%% file: main.tex
\documentclass{article}

\usepackage{natbib}
\usepackage[preprint]{neurips_2024}

\usepackage[utf8]{inputenc} % allow utf-8 input
\usepackage[T1]{fontenc}    % use 8-bit T1 fonts
\usepackage{hyperref}       % hyperlinks
\usepackage{url}            % simple URL typesetting
\usepackage{booktabs}       % professional-quality tables
\usepackage{amsfonts}       % blackboard math symbols
\usepackage{nicefrac}       % compact symbols for 1/2, etc.
\usepackage{microtype}      % microtypography
\usepackage{xcolor}         % colors
\usepackage{float}
\usepackage{graphicx}
\usepackage{subcaption}

\usepackage{microtype}
\usepackage{graphicx}
\usepackage{booktabs}

\usepackage{tcolorbox}  %

\newcommand{\finding}[2]{%
\begin{tcolorbox}[colback=green!10!white,leftrule=2.5mm,size=title]
\textbf{#1}: #2
\end{tcolorbox}
\vspace{-0.1cm}%
}

\title{Outliers and Calibration Sets have Diminishing Effect on Quantization of Modern LLMs}

\author{
 Davide Paglieri\thanks{d.paglieri@cs.ucl.ac.uk} \\
  University College London\\
  \And
  Saurabh Dash \\
  Cohere \\
  \AND
  Tim Rocktäschel \\
  University College London \\
  \And
  Jack Parker-Holder \\
  University College London \\
}

\begin{document}

\maketitle

%%%%%%%%%%%%%%%%%%%%%%%%%%%%%%%%%%%%%%%%%%%%%%%%%%%%%%%%%%%%%%%%%%%%%%%%
\input{core/0-abstract}
\input{core/1-intro}
\input{core/2-background}

\input{core/3-experimental-setup}
\input{core/4-results}
\input{core/5-related_work}
\input{core/6-Limitations}

\input{core/7-conclusion}

%%%%%%%%%%%%%%%%%%%%%%%%%%%%%%%%%%%%%%%%%%%%%%%%%%%%%%%%%%%%%%%%%%%%%%%%

\newpage
%\bibliographystyle{plainnat}
%\bibliography{bib/references}

\input{main.bbl}
%%%%%%%%%%%%%%%%%%%%%%%%%%%%%%%%%%%%%%%%%%%%%%%%%%%%%%%%%%%%%%%%%%%%%%%%%%%%%%%
%%%%%%%%%%%%%%%%%%%%%%%%%%%%%%%%%%%%%%%%%%%%%%%%%%%%%%%%%%%%%%%%%%%%%%%%%%%%%%%
% APPENDIX
%%%%%%%%%%%%%%%%%%%%%%%%%%%%%%%%%%%%%%%%%%%%%%%%%%%%%%%%%%%%%%%%%%%%%%%%%%%%%%%
%%%%%%%%%%%%%%%%%%%%%%%%%%%%%%%%%%%%%%%%%%%%%%%%%%%%%%%%%%%%%%%%%%%%%%%%%%%%%%%
\newpage
\appendix
\onecolumn
\input{core/A-appendix}

\input{core/B-appendix}

\input{core/C-appendix}

\input{core/D-appendix}
\input{core/E-appendix}
%%%%%%%%%%%%%%%%%%%%%%%%%%%%%%%%%%%%%%%%%%%%%%%%%%%%%%%%%%%%%%%%%%%%%%%%%%%%%%%

%\newpage
%\input{core/8-Checklist}

\end{document}

%% file: core/0-abstract.tex
\begin{abstract}
Post-Training Quantization (PTQ) enhances the efficiency of Large Language Models (LLMs) by enabling faster operation and compatibility with more accessible hardware through reduced memory usage, at the cost of small performance drops. We explore the role of calibration sets in PTQ, specifically their effect on hidden activations in various notable open-source LLMs. Calibration sets are crucial for evaluating activation magnitudes and identifying outliers, which can distort the quantization range and negatively impact performance. Our analysis reveals a marked contrast in quantization effectiveness across models. The older OPT model, upon which much of the quantization literature is based, shows significant performance deterioration and high susceptibility to outliers with varying calibration sets. In contrast, newer models like Llama-2 7B, Llama-3 8B, Command-R 35B, and Mistral 7B demonstrate strong robustness, with Mistral 7B showing near-immunity to outliers and stable activations. These findings suggest a shift in PTQ strategies might be needed. As advancements in pre-training methods reduce the relevance of outliers, there is an emerging need to reassess the fundamentals of current quantization literature. The emphasis should pivot towards optimizing inference speed, rather than primarily focusing on outlier preservation, to align with the evolving characteristics of state-of-the-art LLMs.
\end{abstract}

%% file: core/1-intro.tex
\section{Introduction}

Transformer-based Large Language Models (LLMs) have shown remarkable performance which correlates with the number of parameters \citep{kaplan2020scaling, chowdhery2023palm, hoffmann2022training, zhang2022opt}.
The growth trend of LLMs memory requirements has far outpaced the increase of VRAM in modern day GPUs \citep{rajbhandari2021zero}. As we grow LLMs further to improve their capabilities, this gap is bound to increase. The massive scale of these models hinders their widespread use on easily accessible mobile devices.

In response to this, there has been a recent wave of smaller open-source high-performing models such as Llama, Mistral and Phi \citep{touvron2023Llama, touvron2023Llama2, llama3modelcard, jiang2023mistral, li2023textbooks}. Their smaller sizes have facilitated broader usage, highlighting the demand for more compact models among machine learning practitioners.
Furthermore, a growing field of research deals with compressing pre-trained LLMs into smaller sizes to facilitate their use. Popular techniques to compress LLMs—so that they can run faster and use less memory, at the cost of a small drop in accuracy—are quantization, pruning, and distillation \cite{zhu2023survey}. Applying these techniques on already smaller Language Models enables them to be run on widely available hardware.

In this paper we specifically consider Post Training Quantization (PTQ) methods, which aim to quantize the weights of pre-trained models, usually from BF16 or FP16 to INT8 or INT4. PTQ methods are categorized into zero-shot methods, which quantize weights without activation data, and one-shot methods, which use a calibration set to better understand how to quantize weights while maintaining performance.

Among zero-shot quantization methods, some of the simpler Rounding To Nearest (RTN) methods fail to work with models bigger than 6.7B on older pre-trained models when quantizing both weights and activations \citep{dettmers2022llm}. This result is attributed to weight and activation outliers, which were initially thought to be an emergent property of LLMs at scale. Newer research indicates that these outliers are byproducts of training choices common in older LLMs such as OPT \citep{zhang2022opt}, and the Cohere models should be more robust and perform well with simpler quantization techniques \citep{ahmadian2023intriguing}. 

Closely related to outliers is the use of a calibration set, which is run through the model to measure the activation values, and thus quantize more accurately by estimating the importance of weights on the activations values, and spotting outlier features \citep{frantar2022gptq, lin2023awq, wei2022outlier, dettmers2023spqr}. Calibration data is usually sampled randomly from web text or from pre-training datasets; recently \citet{williams2023does} have investigated the effect of the calibration set on downstream task performance, claiming that performance can somewhat vary based on the split of the calibration set chosen. 

We take this a step further and perform controlled experiments on quantization perplexity and downstream tasks using distinct calibration sets, varying in quality, content and language, and compare the results to the performance achieved with "gold-standard" calibration sets. We show that modern open-source LLMs like Llama-2 7B \citep{touvron2023Llama2}, Llama-3 8B \citep{llama3modelcard}, Mistral 7B \cite{jiang2023mistral} and bigger Command R 35B \citep{commandr}, when quantized both weight-only and weight-and-activations are significantly more robust to the choice of calibration set compared to OPT 6.7B \cite{zhang2022opt}.
In summary our contributions are as follows:

\begin{itemize}
  \item We show that modern LLMs are notably less affected by the quality, content and language of the calibration set compared to an older LLM such as OPT 6.7B.
  \item We show that modern LLMs are less affected by outliers compared to the older OPT 6.7B, upon which much of the current knowledge in quantization has been built upon.
  \item We perform a thorough analysis of the activation distributions, patterns and outliers of the LLMs tested, which help us explain our findings and offer interesting insights for future quantization research.
  \item We propose that as newer and better open-source LLMs become available, the quantization field should continuously reassess its foundational knowledge on these newer models, and drop assumptions made with older models.
\end{itemize}

%% file: core/2-background.tex
\section{Background}

Quantization reduces the memory and computational requirements of neural networks by transforming high-precision weights to lower precision formats. LLMs are usually trained using FP16 precision or more recently in BF16 \citep{kalamkar2019study}, and are typically quantized to INT8, INT4 or INT3 precisions \citep{dettmers2022llm, frantar2022gptq}, with 4bit found to be the sweet spot \citep{dettmers2023case}. Our focus is on Post Training Quantization methods (PTQ), which take a high-precision pre-trained model and quantize it, as opposed to Quantization Aware Training (QAT) methods, which follow a quantization objective during training.

Quantization can be either weight-only (e.g. W4A16) or weight-and-activation quantization (e.g. W8A8). Weight-only quantization, as the name suggests, only quantizes the weights, then at inference time the weights are dequantized and matrix multiplication is performed in 16 bit floating point precision. Weight-and-activation quantization methods quantize both weights and activations, performing multiplication at lower precision. Weight-only quantization increases inference speed at low batch sizes thanks to reduced fetch time from GPU of the quantized weights. Conversely, the advantage of weight-and-activation quantization is the absence of a dequantization step, allowing for faster throughput of large batch sizes and matrix multiplication in the same precision as the weights. However, complete quantization of both weights and activations at low precision has so far proven more challenging, leading to larger drops in performance \cite{ahmadian2023intriguing}. 

\citet{dettmers2022llm} first observed the emergence of extreme outliers in the feature dimensions during inference of the range of OPT models bigger than 6.7B parameters \citep{zhang2022opt}. These outliers damage the weight-and-activation quantization performance of simple rounding to nearest methods, by skewing the value range before quantization, leading to inefficient use of the quantized range. Conversely, weigh-only quantization finds larger models easier to quantized than smaller models at low precision \citep{frantar2022gptq}. 

Numerous high-performing weight-only and weight-and-activation quantization methods, aim to mitigate the impact of extreme outliers to maintain high performance of the quantized model \citep{dettmers2022llm, dettmers2023spqr, lin2023awq, kim2023squeezellm}. \citet{dettmers2022llm} for example keep the outlier activations in 16-bit floating point precision, while SmoothQuant \citep{xiao2023smoothquant}, a W8A8 method, and AWQ \citep{lin2023awq}, a W4A16 method, move the quantization difficulty from the activation to the weights, scaling down the activations and scaling up the weights in order to make outlier quantization more manageable. GPTQ is another prominent weight-only quantization method \citep{frantar2022gptq} that adjusts weights based on activation values using second-order information. Several other quantization techniques build on similar concepts as GPTQ \citep{dettmers2023spqr, chee2024quip, tseng2024quip}.

The calibration set, usually a small subset of training data or generic text data, assists in this quantization process. By running it through the network, activation values can be determined, helping to quantize the weights so that the outputs closely match those of the unquantized model.

%% file: core/3-experimental-setup.tex
\section{Experimental setup}

We set out to examine the impact of the calibration set on the performance of various Large Language Models. Specifically, we address three primary questions: first, how the quality of the calibration set affects the quantized performance of the models; second, whether a content-specific calibration set can enhance performance on a particular task; and third, how the same content presented in different languages affects the quantized models when used as a calibration set.

We evaluate six distinct LLMs: OPT 6.7B \citep{zhang2022opt}, Llama-1 7B \citep{touvron2023Llama} Llama-2 7B \citep{touvron2023Llama2}, Llama-3 8B \citep{llama3modelcard}, Mistral 7B \citep{jiang2023mistral} and the larger Command-R 35B \citep{commandr}, to determine their responses to varying calibration sets.

We test three different one-shot quantization methods: two weight-only quantization methods, GPTQ W4A16 with a group size of 128 \citep{frantar2022gptq} and AWQ W4A16 with a group size of 128 \citep{lin2023awq}; and SmoothQuant W8A8, a weight-and-activation quantization method \citep{xiao2023smoothquant}. Model performance is measured by evaluating perplexity on WikiText2 \citep{merity2016pointer} and downstream zero-shot accuracy on ARC-Challenge \citep{clark2018think}, PiQa \citep{bisk2020piqa}, and Winogrande \citep{sakaguchi2021winogrande}, three popular benchmarks that assess abstract and common sense reasoning capabilities. Additionally, we test a zero-shot naive W8A8 weight-and-activation quantization method.

\subsection{Impact of the Calibration Set Quality on Quantization Effectiveness}
\label{subs_1}
In the first part of our study, we investigate whether the quality of content, particularly vocabulary, in the calibration set significantly affects quantization quality. We hypothesize that a calibration set with higher quality content will yield better performance. To test this, we compare a calibration set sampled from RedPajama \cite{together2023redpajama}—an open-source replica of Llama's training corpus—against a set composed of random ASCII punctuation characters (sample text in \autoref{random calibration}). RedPajama represents an appropriate calibration set for quantization due to its meaningful and well-curated content, while the random ASCII punctuation set serves as a nonsensical calibration set, expected to offer no benefit to quantization and potentially be detrimental.

\subsection{Impact of Content-Specific Calibration Sets on Specific Downstream Tasks}
We explore the potential benefits of using content-specific calibration sets for performance enhancement. This has practical applications; for instance, if a specific downstream task is known, it would be intuitive to calibrate the model for that task. For this purpose, we use ARC-Challenge and PiQa as calibration sets and compare their effectiveness against RedPajama. Both ARC-Challenge and PiQa calibration sets include the full test data, encompassing the questions and answers that the LLM is subsequently evaluated on.

\subsection{Impact of Different Languages as Calibration Sets on Quantization Effectiveness}
\label{subs_2}
We extend our analysis to assess how different languages in calibration sets impact English perplexity on WikiText2 and downstream accuracy on ARC-Challenge, PiQa, and Winogrande. We hypothesize that different languages might induce unique activation patterns in LLMs and trigger different outliers, potentially affecting performance on English perplexity or downstream tasks. Conversely, robustness in an LLM would indicate similar activation patterns and outlier positions across languages and tokens. It is important to note that none of the LLMs tested have been trained on all the languages used; however, they may have encountered multiple languages during pre-training, though some tokens might be encountered very rarely.

For this analysis, we utilize the FLORES+ dataset \citep{costa2022no, flores101-22, flores1-19, mt4nko-23, indictrans2-23}, a multi-language dataset comprising 2009 sentences translated into 205 different languages across 30 alphabets. By using FLORES+ translations, we ensure uniform content across all calibration sets. Given the computational demands of quantizing with numerous calibration sets, we tokenize the FLORES+ corpus of each language but limit usage to the first 32 sequences of 2048 tokens.

%% file: core/4-results.tex
\section{Results and Analysis}
\label{results and discussion}

\subsection{Impact of the Calibration Set Quality on Quantization Effectiveness}
\begin{figure}[h]
  \centering
  \begin{minipage}[t]{0.49\textwidth}
    \centering
    \includegraphics[width=\textwidth]{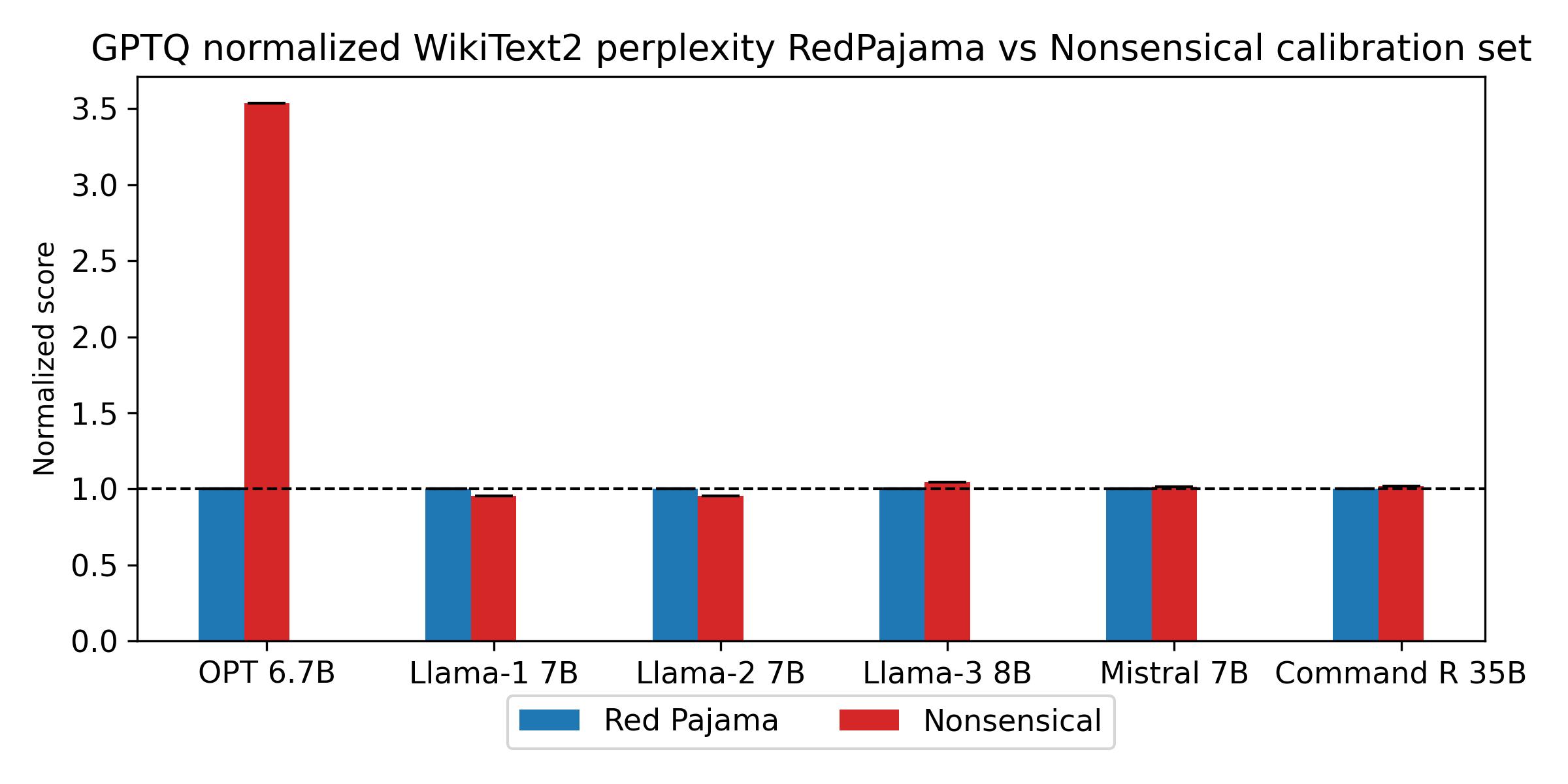}
    \caption{WikiText2 perplexity with GPTQ 4-bit quantization, using as calibration sets RedPajama \citep{together2023redpajama} and a nonsensical calibration set \autoref{random calibration}. Results normalized to RedPajama score. Lower is better.}
    \label{wikitext_figure}
  \end{minipage}
  \hfill
  \begin{minipage}[t]{0.49\textwidth}
    \centering
    \includegraphics[width=\textwidth]{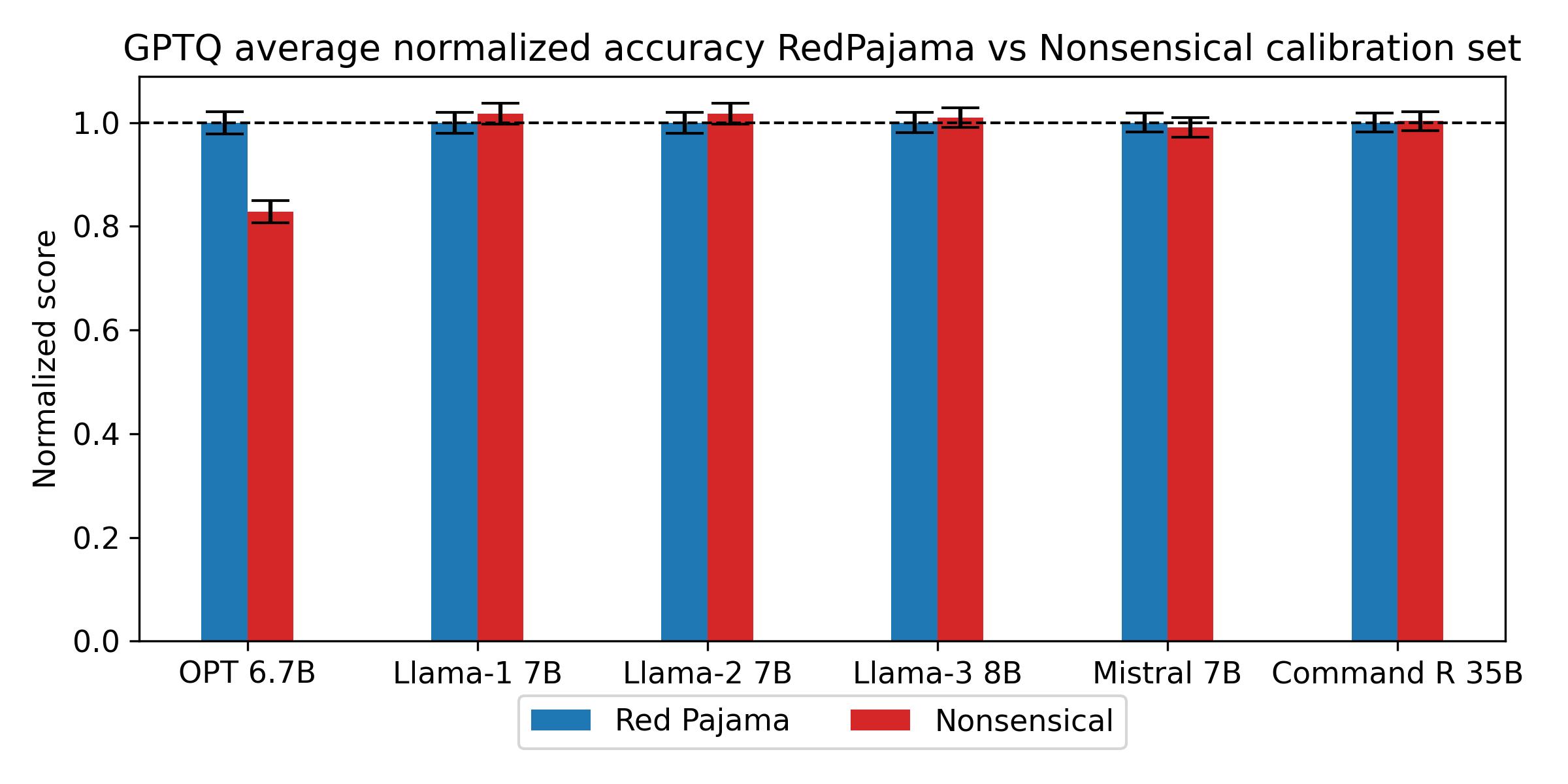}
    \caption{Average ARC-Challenge and PIQA accuracy with GPTQ 4-bit quantization,  using as calibration sets RedPajama \citep{together2023redpajama} and a nonsensical calibration set \autoref{random calibration}. Results normalized to RedPajama score. Higher is better. Error bars represent standard error.}
    \label{average-GPTQ-nonsensical-figure}
  \end{minipage}
\end{figure}

Our analysis reveals significant variations among the tested LLMs concerning the impact of calibration set quality on quantized effectiveness. In particular, OPT 6.7B demonstrates a markedly worse perplexity in WikiText2 as shown in \autoref{wikitext_figure}, and average downstream accuracy over ARC-Challenge and PIQA (\autoref{average-GPTQ-nonsensical-figure}) when quantized using a nonsensical calibration set, as opposed to the standard RedPajama. Conversely, the rest of the models display high robustness; with their performance not impacted when using a random calibration set compared to RedPajama. We show results with AWQ and SmoothQuant quantization in \autoref{calibration_quality}.

The pronounced performance drop observed in OPT 6.7B with the random calibration set can be attributed to distinct activation patterns and strong outlier activations. We analyze this further in \autoref{subs_3}.

This leads us to the following finding:
\vspace*{3pt}
\finding{Finding 1}{The calibration set's quality does not significantly affect quantized performance of modern Large Language Models.}
\vspace*{3pt}

\subsection{Impact of Content-Specific Calibration Set on Quantization Effectiveness}

\begin{figure}[h]
  \centering
  \begin{minipage}[t]{0.49\textwidth}
    \centering
    \includegraphics[width=\textwidth]{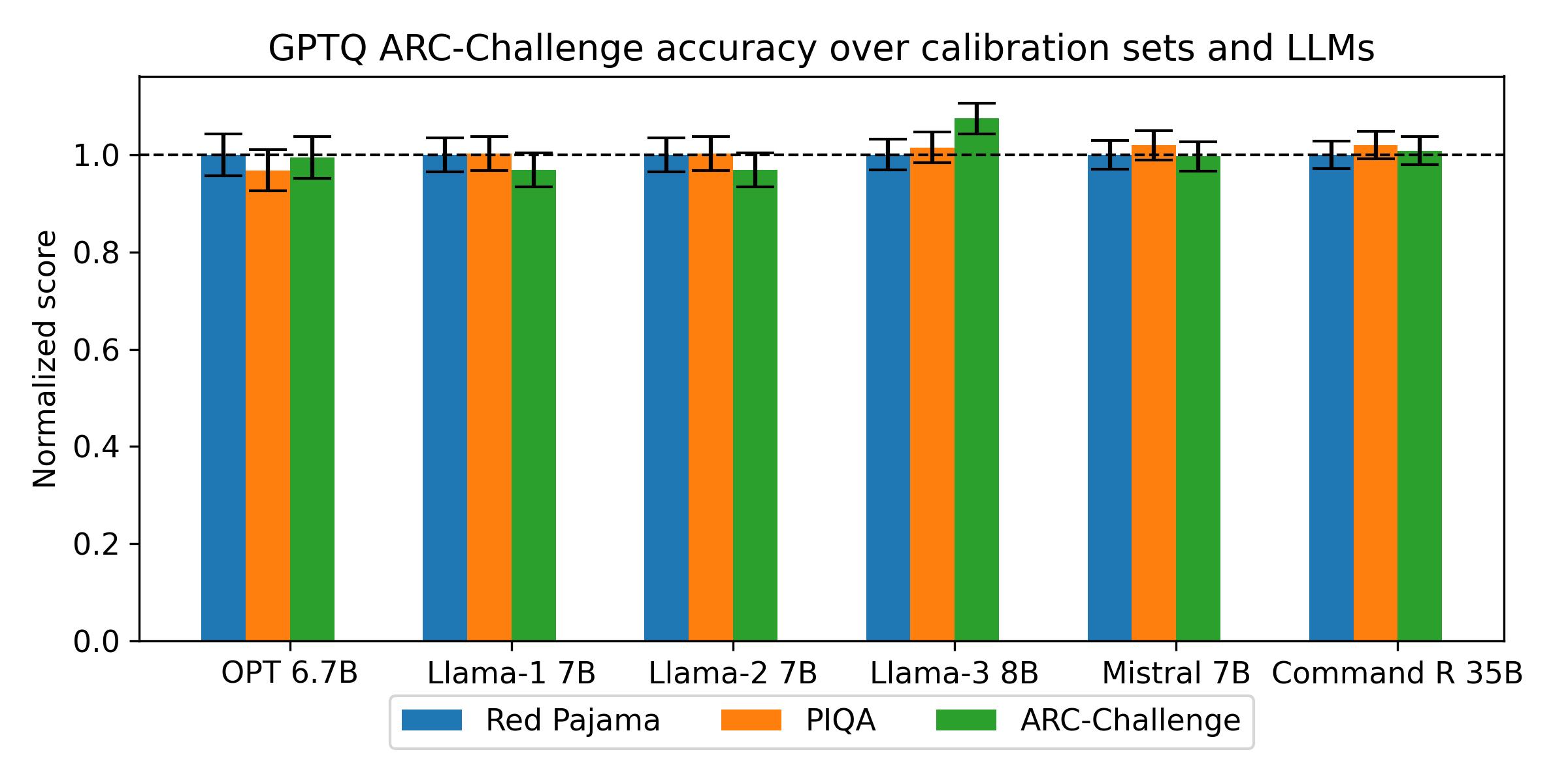}
    \caption{ARC-Challenge accuracy with GPTQ 4-bit quantization over calibration sets. Results normalized to RedPajama score. Error bars represent standard error. Higher is better.}
    \label{GPTQ-arc-figure}
  \end{minipage}
  \hfill
  \begin{minipage}[t]{0.49\textwidth}
    \centering
    \includegraphics[width=\textwidth]{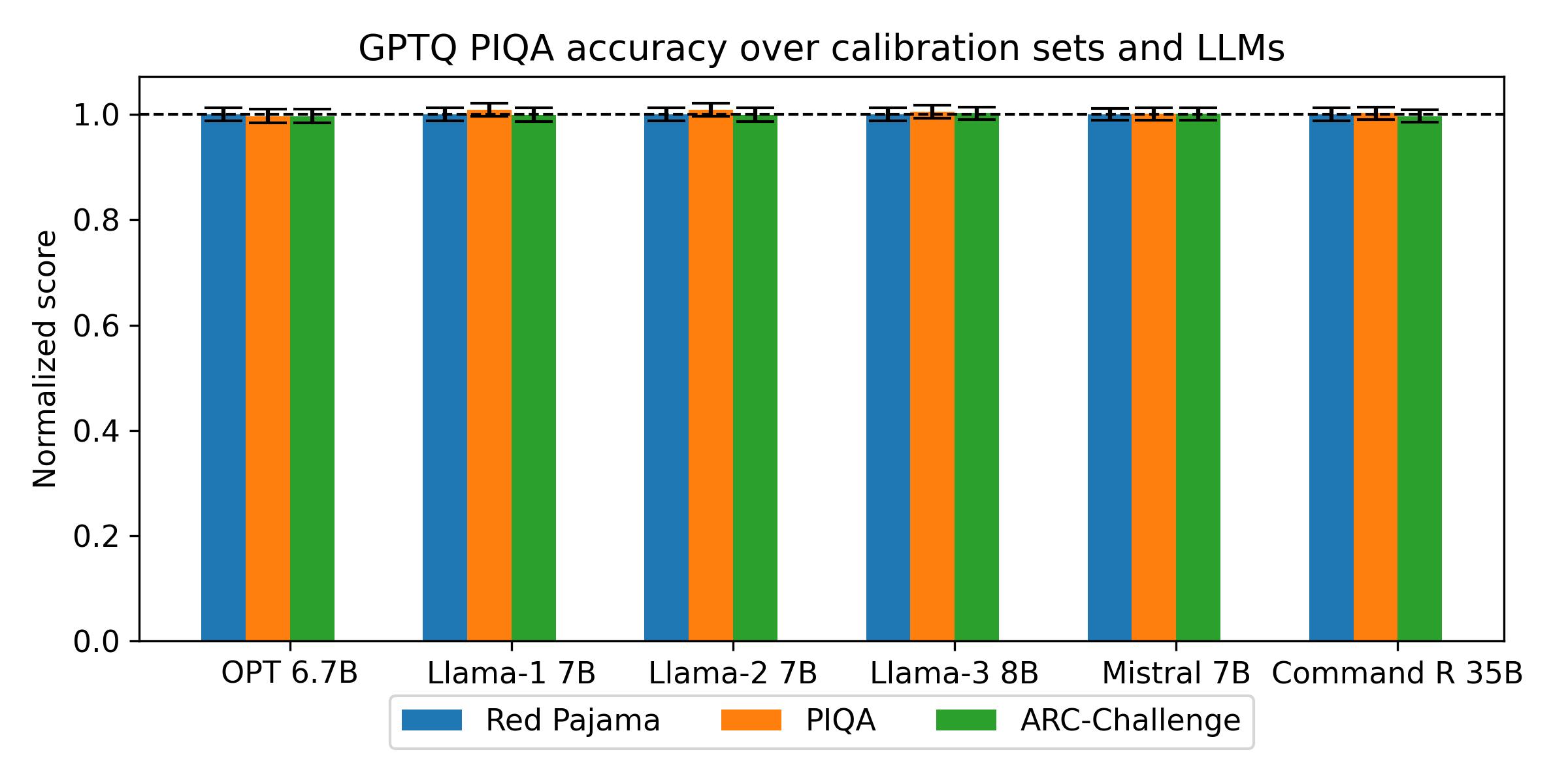}
    \caption{PIQA accuracy with GPTQ 4-bit quantization over calibration sets. Results normalized to RedPajama score. Error bars represent standard error. Higher is better.}
    \label{GPTQ-piqa-figure}
  \end{minipage}
\end{figure}

Considering content-specific calibration sets, we find no statistically significant difference in downstream accuracies for all models tested compared to RedPajama calibration, as shown in \autoref{GPTQ-arc-figure} and \autoref{GPTQ-piqa-figure}. Despite the downstream accuracy results of modern LLMs being within the margin of two standard errors, ARC-Challenge downstream accuracy shows more pronounced fluctuations in mean accuracy compared to PIQA.

\vspace*{3pt}
\finding{Finding 2}{Content-specific calibration sets do not show statistically significant improvements to quantized model performance on specific downstream tasks compared to content-generic calibration sets.}
\vspace*{3pt}

\subsection{Effect of Different Languages in Calibration Sets on Quantization}

\begin{figure*}[h]
  \includegraphics[width=\textwidth]{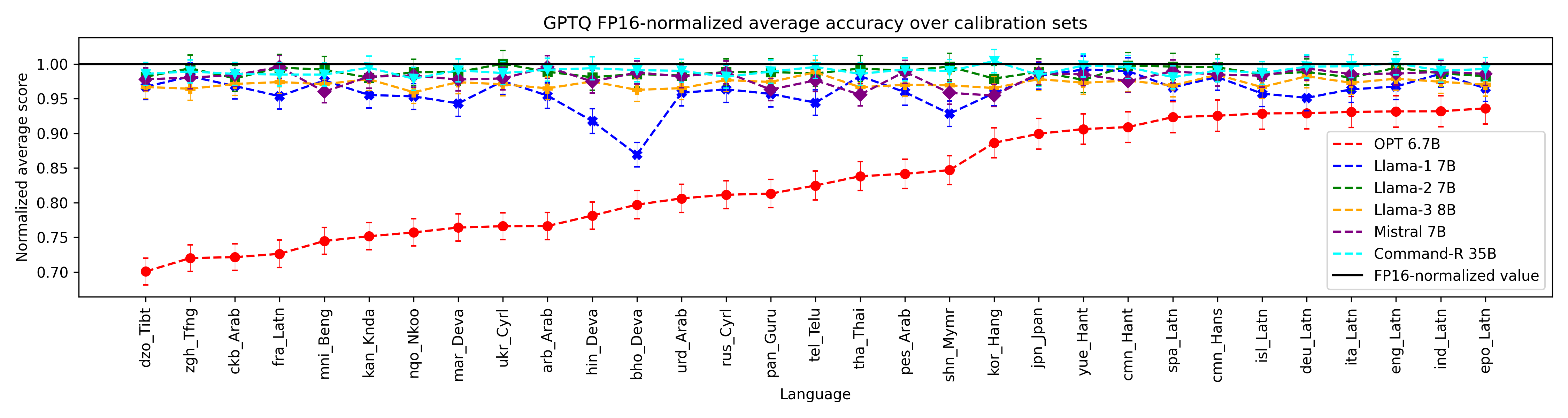}
  \vskip -0.1in
  \caption{GPTQ W4A16, FP16-Normalized average accuracy (ARC-Challenge, PIQA, WinoGrande) of various LLMs, using as calibration sets a selection of languages and alphabets. Results sorted by normalized scores of OPT 6.7B. Error bars represent standard error}
  \label{GPTQ_accuracies}
\end{figure*}

\begin{figure*}[h]
  \includegraphics[width=\textwidth]{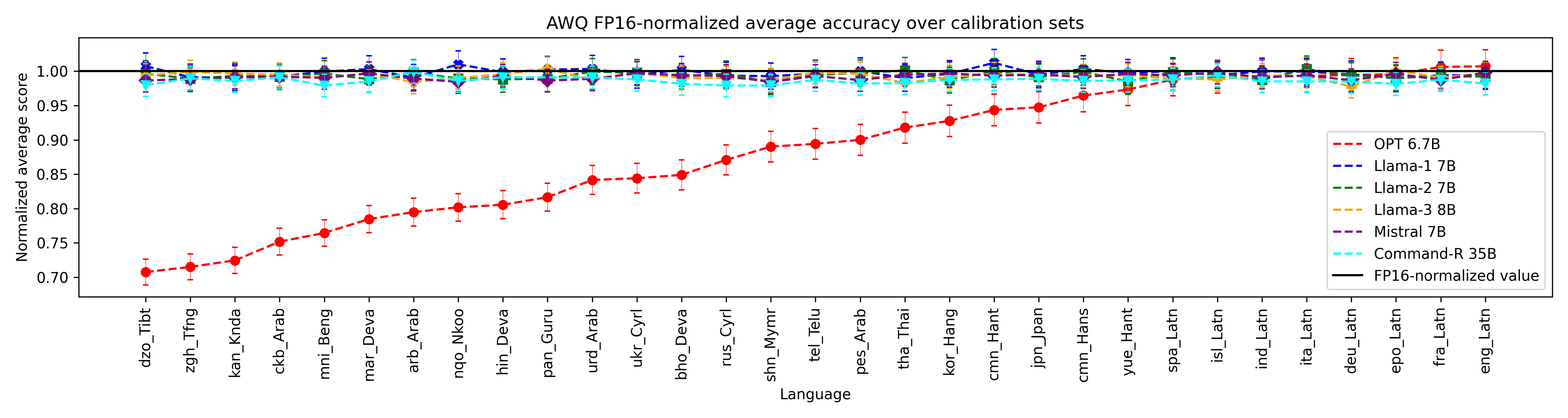}
  \vskip -0.1in
  \caption{AWQ W4A16, FP16-Normalized average accuracy (ARC-Challenge, PIQA, WinoGrande) of various LLMs, using as calibration sets a selection of languages and alphabets. Results sorted by normalized scores of OPT 6.7B. Error bars represent standard error}
  \label{AWQ_accuracies}
\end{figure*}

\begin{figure*}[h]
  \includegraphics[width=\textwidth]{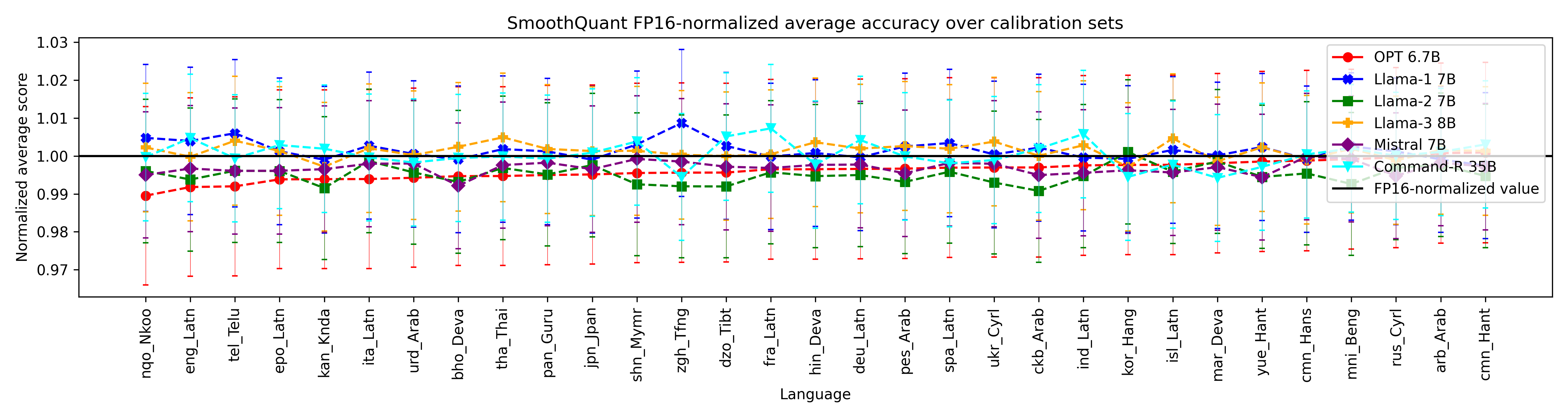}
  \vskip -0.1in
  \caption{SmoothQuant W8A8, FP16-Normalized average accuracy (ARC-Challenge, PIQA, WinoGrande) of various LLMs, using as calibration sets a selection of languages and alphabets. Results sorted by normalized scores of OPT 6.7B. Error bars represent standard error}
  \label{SmoothQuant_accuracies}
\end{figure*}

We now analyze the results of different languages as calibration sets. We normalize the results to 1.0, representing the FP16 result, and visualize the results across a selection of languages and alphabets using average downstream task accuracy (ARC-Challenge, PIQA and WinoGrande), using GPTQ W4A16 in \autoref{GPTQ_accuracies}, AWQ W4A16 in \autoref{AWQ_accuracies} and SmoothQuant W8A8 in \autoref{SmoothQuant_accuracies}. OPT 6.7B is once again the most affected by the choice of the calibration set with both GPTQ and AWQ, showing severe performance degradation on most non-Latin-alphabet languages. 

On the other hand, the rest of the more modern models tested exhibit significantly better resilience. With SmoothQuant W8A8, all the calibration sets perform within the standard error of each other, including OPT 6.7B, likely because it uses 8 bits for weight quantization instead of 4 bits, which is not a particularly challenging quantization scheme despite also quantizing the activations. However, with lower bit weight-and-activation quantization, OPT would likely show worse degradation.

\vspace*{3pt}
\finding{Finding 3}{Different languages from English as calibration sets do not affect quantized performance of modern Large Language Models.}
\vspace*{3pt}

\subsection{Results with Naive W8A8 Quantization}

Lastly, we replicate the experiment from \citet{dettmers2022llm} which showed degradation when naively performing weight-and-activation quantization of OPT models of size 6.7B and bigger due to extreme outliers. We perform naive zero-shot W8A8 quantization using per-channel weight quantization and per-token activation quantization with absmax, and show that OPT 6.7B is the only model of the ones tested whose extreme outliers degrade its performance, while even the bigger Command-R 35B \citep{commandr} shows close to no performance degradation. This confirms the results from \citet{ahmadian2023intriguing}, which showed they could naively quantize W8A8 newly trained Cohere models all the way up to 50B parameters, and points to the fact that outliers are not necessarily an emergent-property at scale, but rather a by-product of training. We discuss what kind of training decision may have led to these differences in \autoref{discussion}.

\begin{figure}[h]
  \centering
  \begin{minipage}[t]{0.49\textwidth}
    \includegraphics[width=\textwidth]{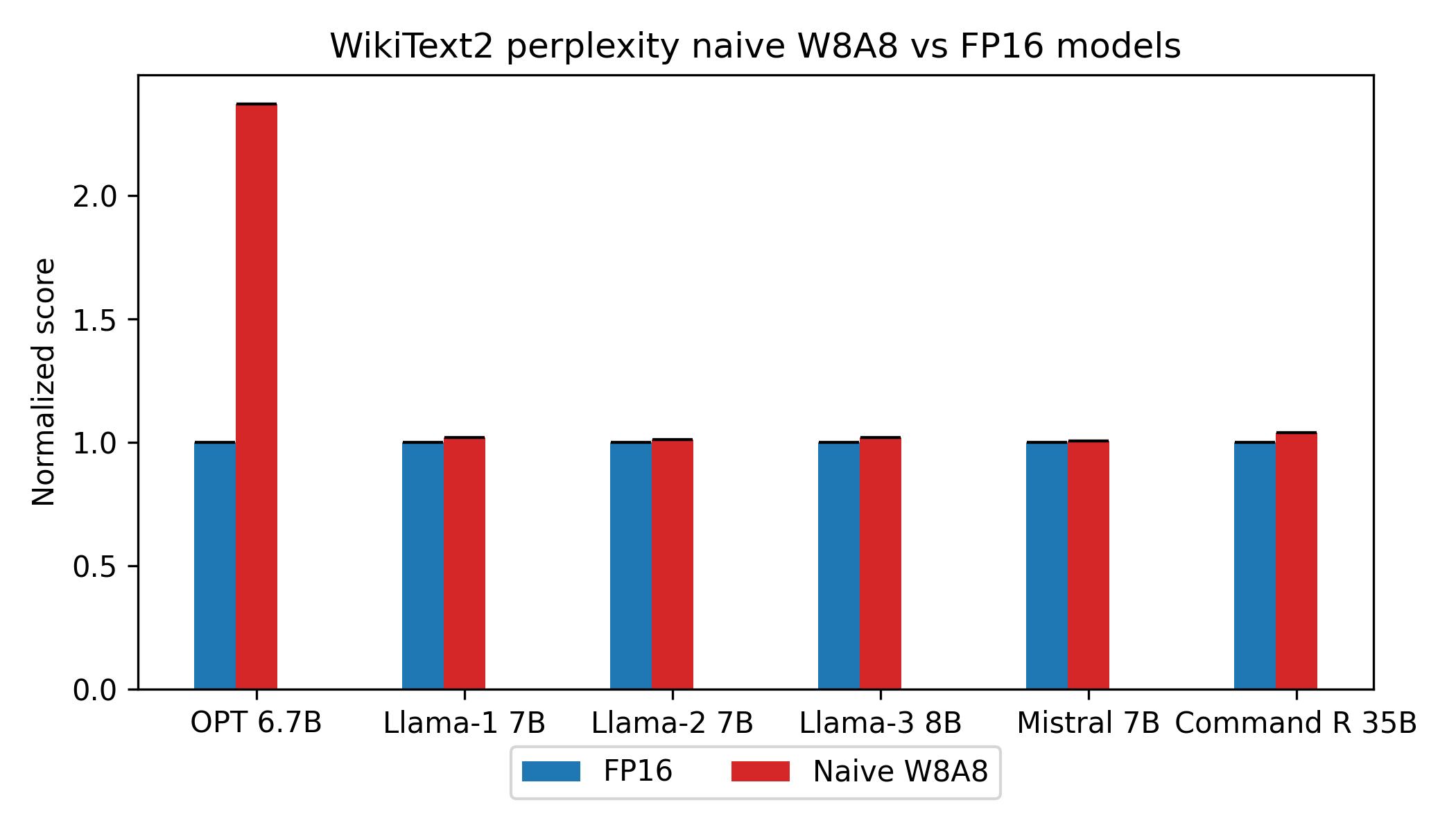}
    \caption{WikiText2 perplexity with naive W8A8 quantization. Results normalized by FP16 value. Lower is better.}
    \label{naive_perplexity}
  \end{minipage}
  \hfill
  \begin{minipage}[t]{0.49\textwidth}
    \includegraphics[width=\textwidth]{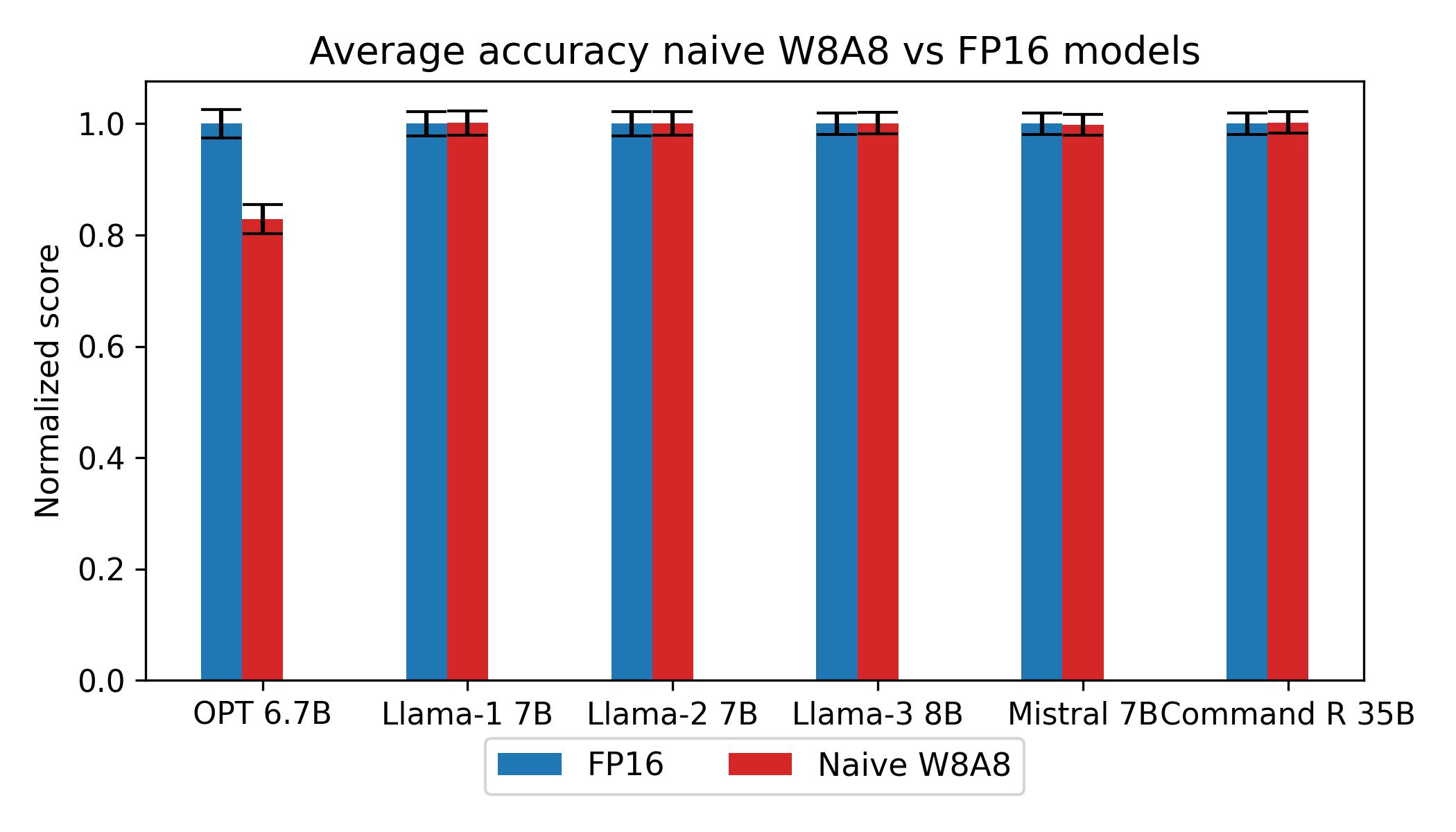}
    \caption{Average accuracy (ARC-C, PIQA, WinoGrande) with W8A8 naive quantization. Results normalized by FP16 value. Error bars represent standard error. Higher is better.}
    \label{naive_accuracies}
  \end{minipage}
  \vskip -0.1in
\end{figure}

\subsection{Activations and outliers comparison}
\label{subs_3}

\begin{figure*}[!ht]
   \begin{minipage}{0.25\textwidth}
     \centering
     \includegraphics[width=1\linewidth]{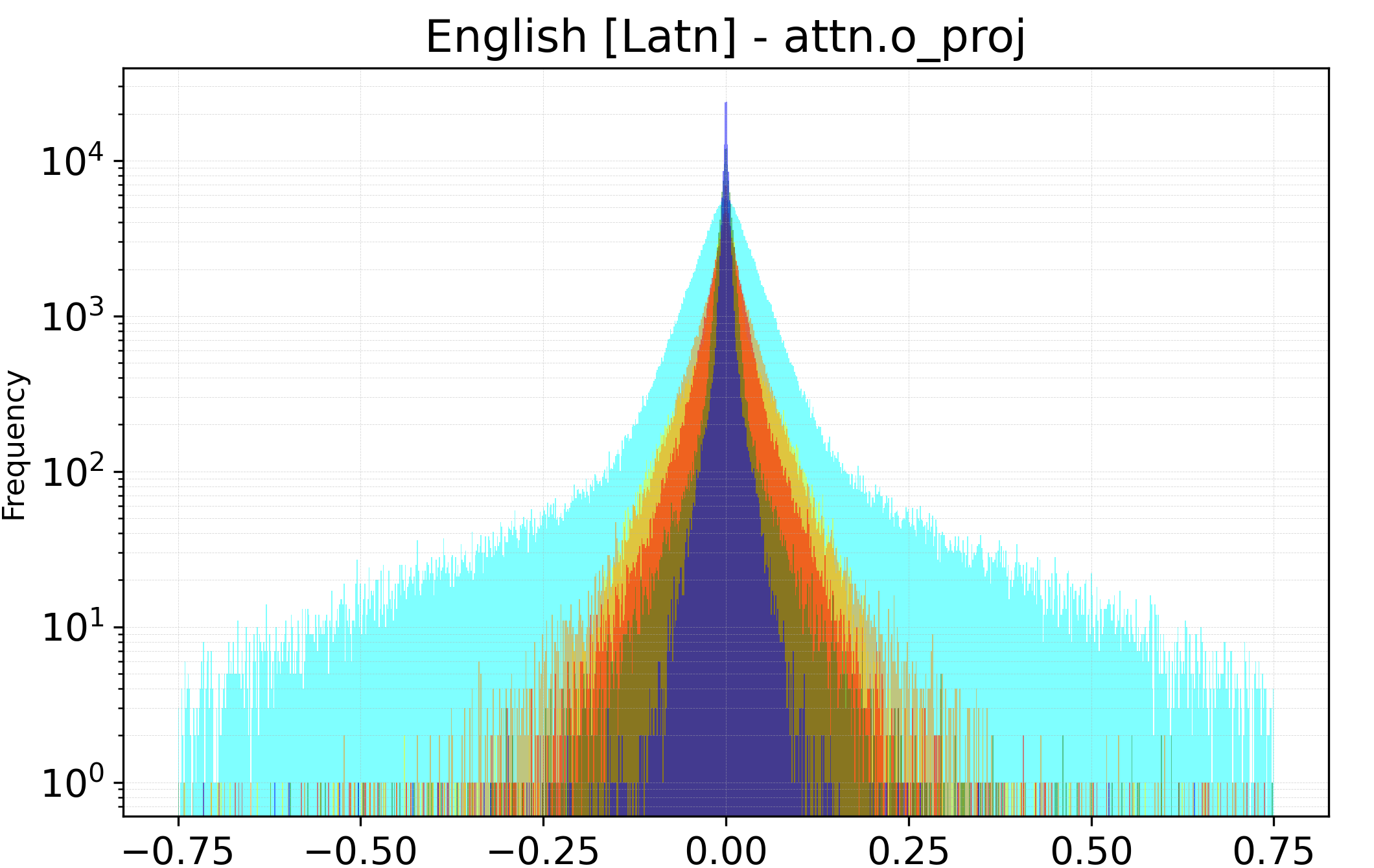}
   \end{minipage}\hfill
   \begin{minipage}{0.25\textwidth}
     \centering
     \includegraphics[width=1\linewidth]{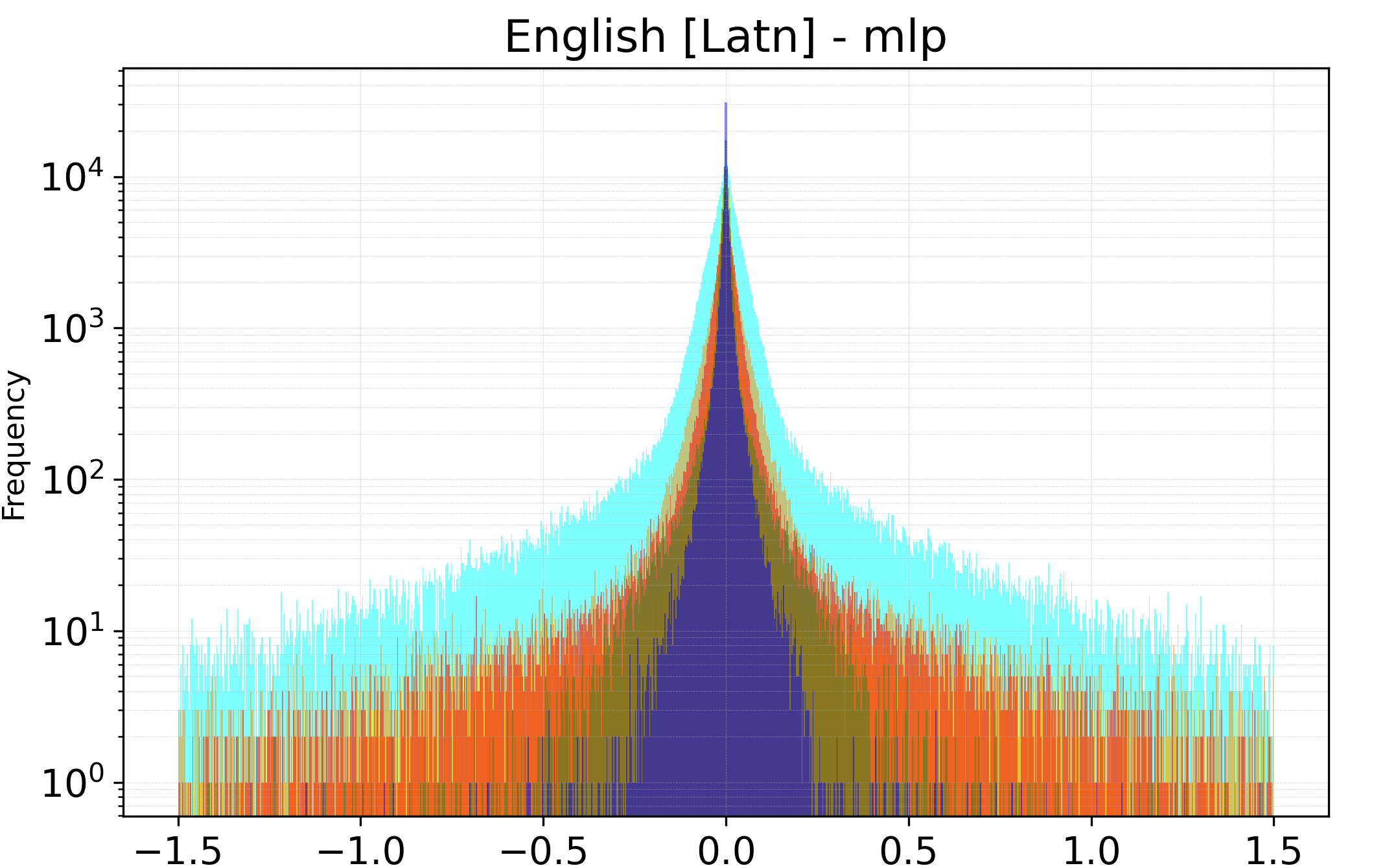}
   \end{minipage}\hfill
   \begin{minipage}{0.25\textwidth}
     \centering
     \includegraphics[width=1\linewidth]{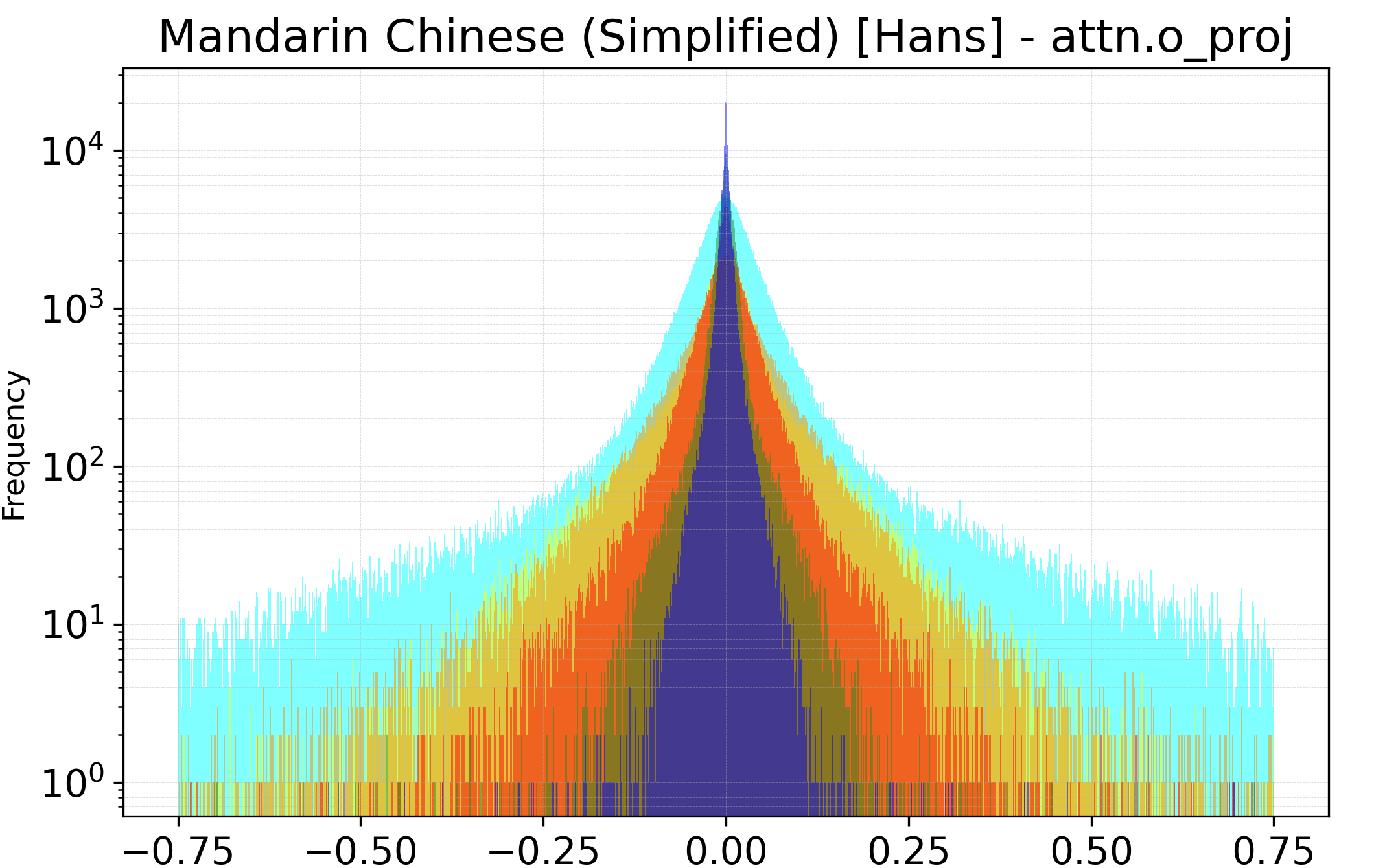}
   \end{minipage}\hfill
   \begin{minipage}{0.25\textwidth}
     \centering
     \includegraphics[width=1\linewidth]{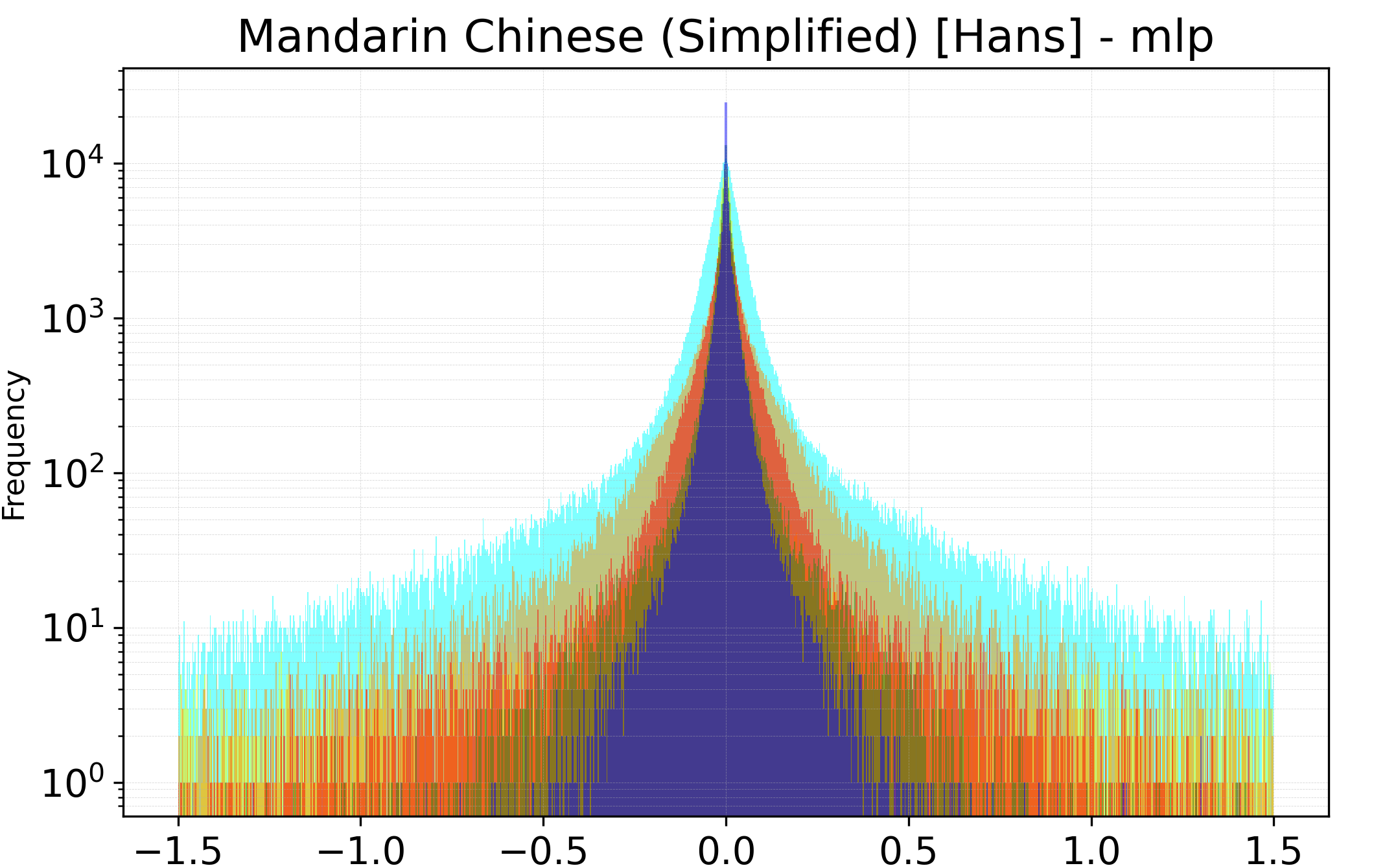}
   \end{minipage}
   \centerline{\includegraphics[width=\columnwidth]{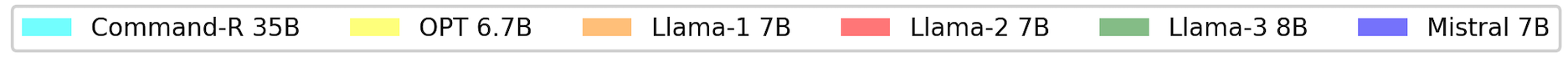}}
   \vskip -0.1in
   \caption{Average activation distribution of all the attention output projection layers and last mlp layers for OPT6.7B, LLaMa-2 7B, and Mistral 7B, for English text (on the left) and Mandarin Chinese text (on the right)}
    \label{distributions}
    \vskip -0.2in
\end{figure*}

\begin{figure*}[h]
    \centering
    \begin{minipage}{0.5\textwidth}
        \centering
        \includegraphics[width=1.0\textwidth]{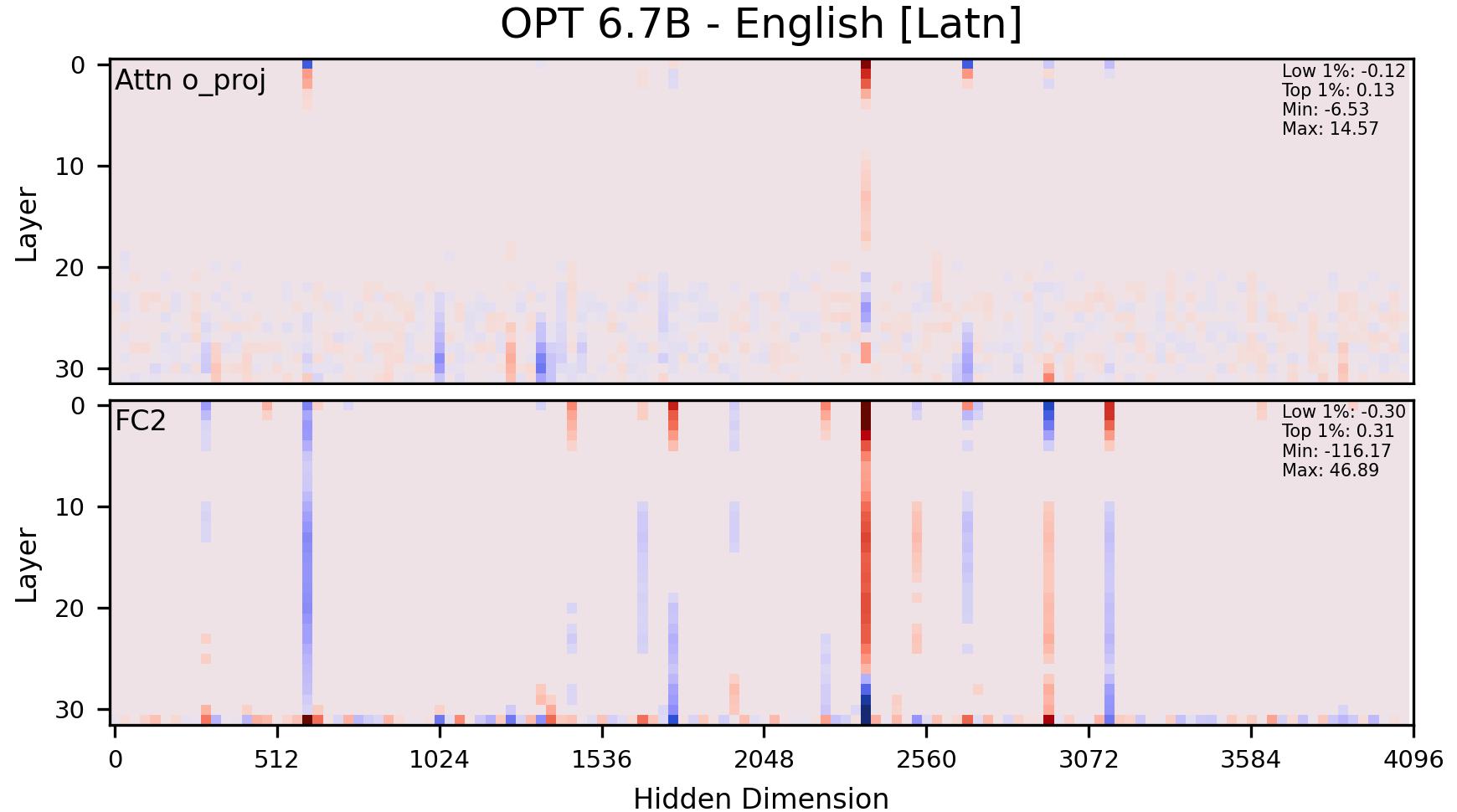} %
    \end{minipage}\hfill
    \begin{minipage}{0.5\textwidth}
        \centering
        \includegraphics[width=1.0\textwidth]{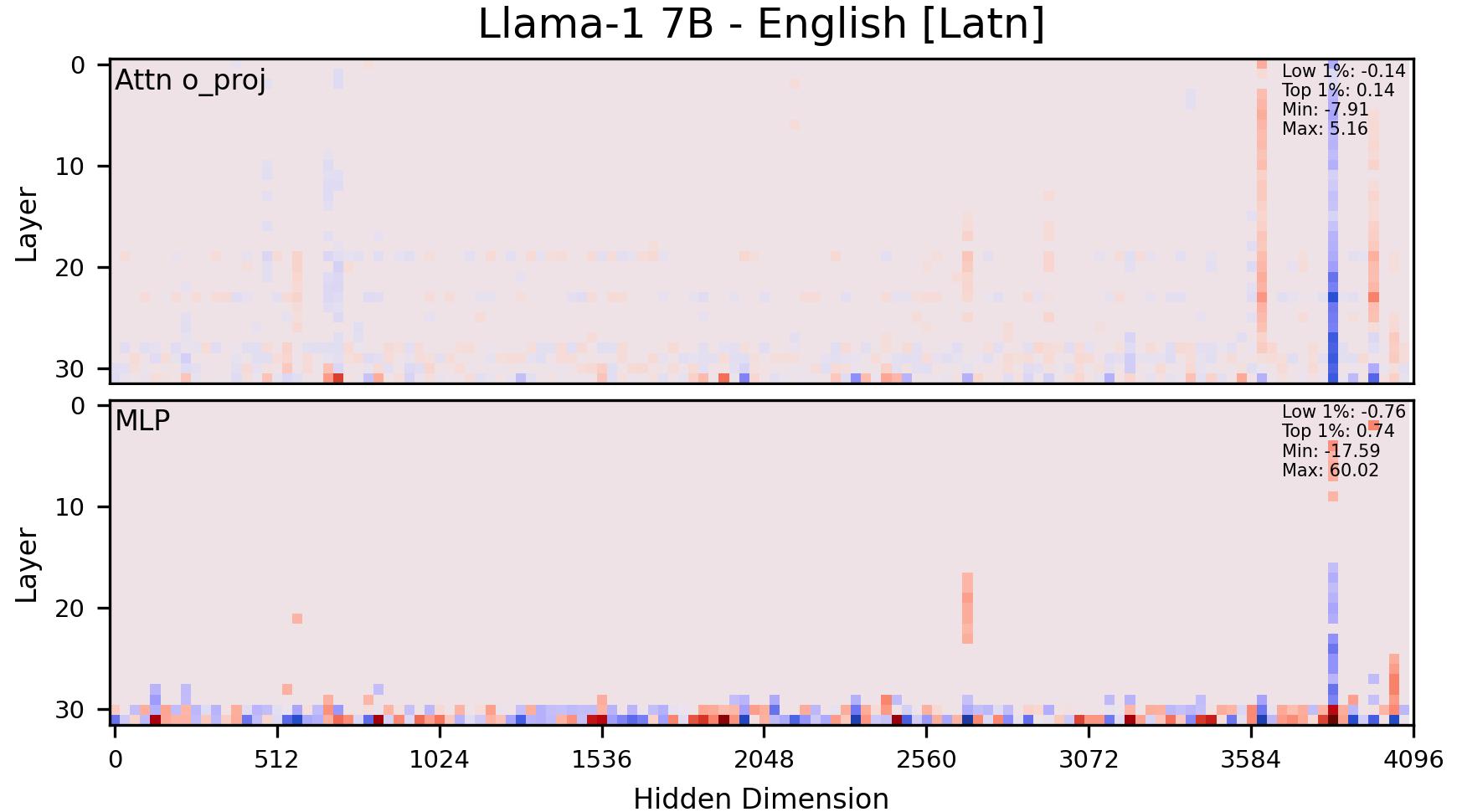} %
    \end{minipage}
    \begin{minipage}{\textwidth}
        \centering
        \includegraphics[width=0.5\textwidth]{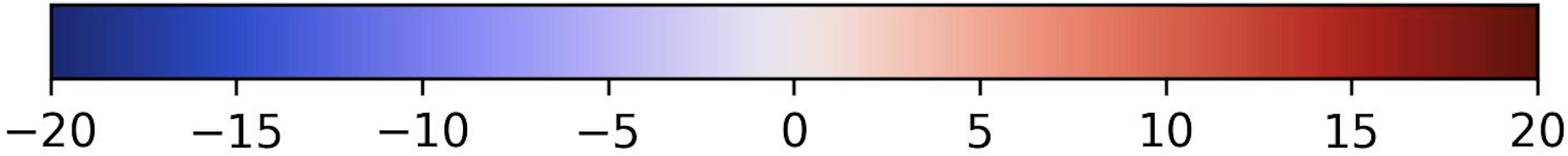} %
    \end{minipage}
    \caption{Visualisation of the top and bottom 1\% of the activation values of attention output projection layers and last fully connected layers of OPT 6.7B (on the left), and Llama-1 7B (on the right) when running inference on English text}
    \label{opt-llama-1-activations}
    \vskip -0.2in
\end{figure*}

To gain a deeper understanding of the performance of quantized models and the mechanics of calibration sets, we conduct a thorough analysis of activation distributions and patterns within the attention output projection layers and the final fully connected linear layer across all the layers of the unquantized LLMs tested. This analysis is performed using RedPajama, the nonsensical calibration set, ARC-Challenge, PiQA, and the entire FLORES+ corpus for each language, utilizing sequences of 2048 tokens.

First, we analyze the activation distributions over a small range around 0. Mistral 7B consistently exhibits a much narrower activation distribution than all the Llama models and OPT 6.7B in all languages tested. The larger Command-R 35B model shows a wider base distribution than the rest of the models. We also observe progressively narrower distributions in the LLMs developed by Meta, from OPT 6.7B to LLaMa-1, LLaMa-2, and LLaMa-3 being the most well-behaved. We also note a broader spread in the activation distributions for non-English languages, with OPT 6.7B and Llama-1 7B showing the widest distribution among the smaller models, Llama-2/3 models occupying intermediate positions, and Mistral 7B maintaining a consistently narrow distribution across all languages. In \autoref{distributions}, we compare the activation distributions of English and Mandarin Chinese. Mandarin Chinese was selected for its widespread use, distinct non-Latin alphabet, and likely inclusion in the models' pre-training. A more comprehensive list of distributions is shown in \autoref{all_cropped_histogramsogram_distributions}.

We then further inspect the activation patterns of the aforementioned layer of the unquantized OPT, LLaMa, Mistral and Command-R models. Specifically, we compute the average activations across all sequences, then identify the top and bottom 1\% percent of activations values. Additionally, we perform min/max pooling with kernel size of 32 (64 for Command-R 35B) along the hidden dimension, facilitating a clearer visualization of the hidden dimensions.

\begin{figure*}[ht]
    \centering
    \begin{minipage}{0.5\textwidth}
        \centering
        \includegraphics[width=1.0\textwidth]{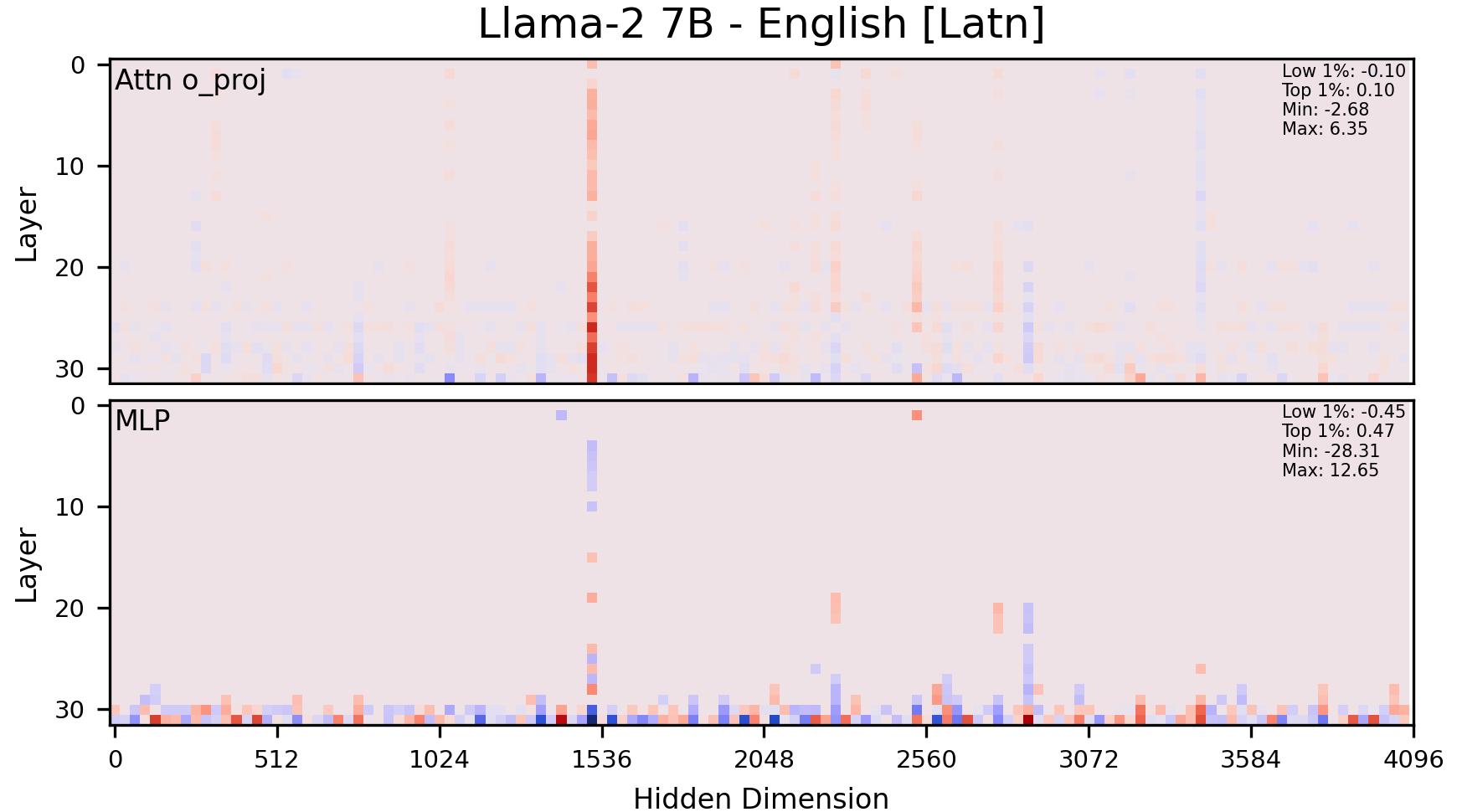} %
    \end{minipage}\hfill
    \begin{minipage}{0.5\textwidth}
        \centering
        \includegraphics[width=1.0\textwidth]{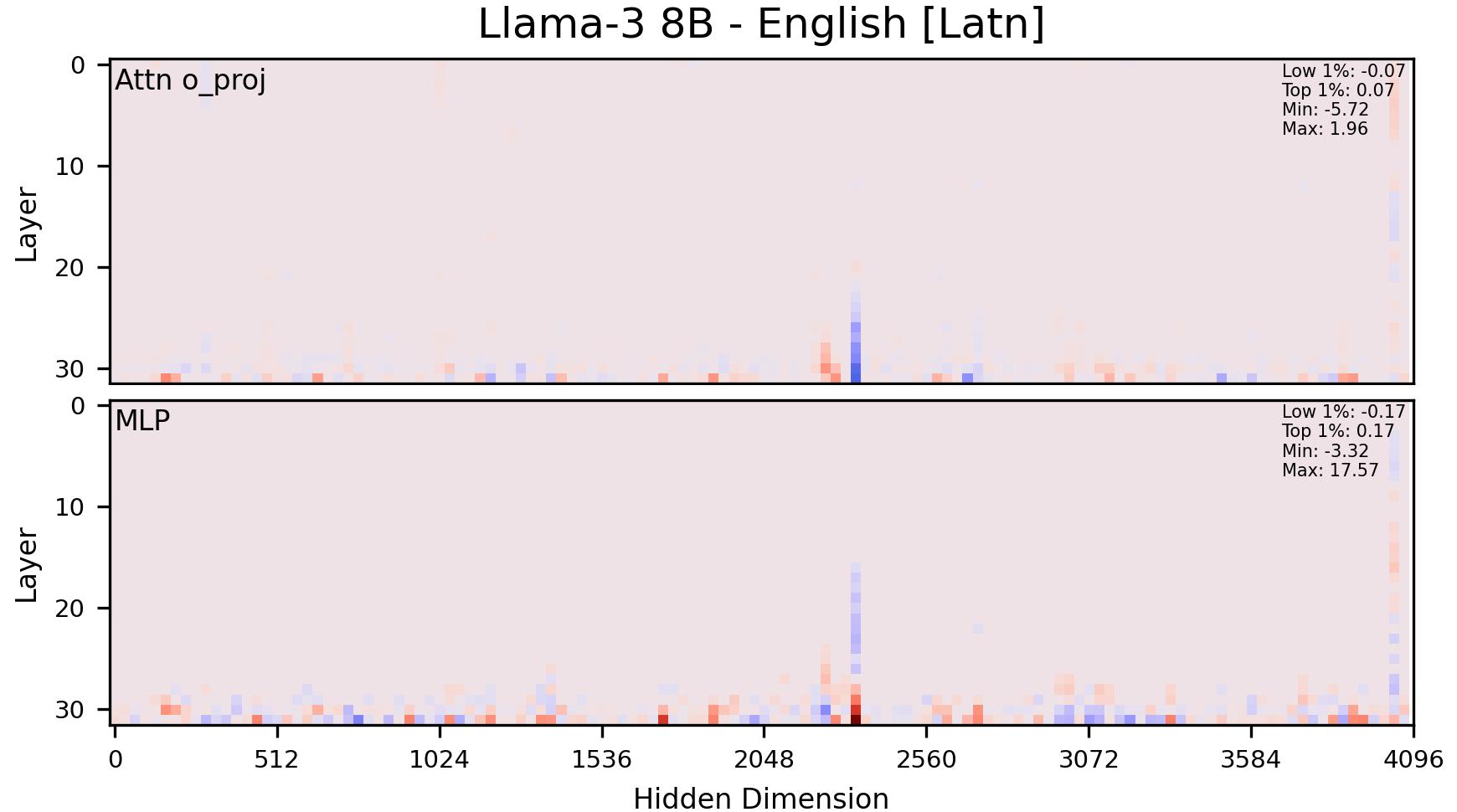} %
    \end{minipage}
    \caption{Visualisation of the top and bottom 1\% of the activation values of attention output projection layers and last fully connected layers of Llama-2 7B (on the left), and Llama-3 8B (on the right) when running inference on English text}
    \label{llama2-llama3-activation}
    \vskip -0.2in
\end{figure*}

\begin{figure*}[ht]
    \centering
    \begin{minipage}{0.5\textwidth}
        \centering
        \includegraphics[width=1.0\textwidth]{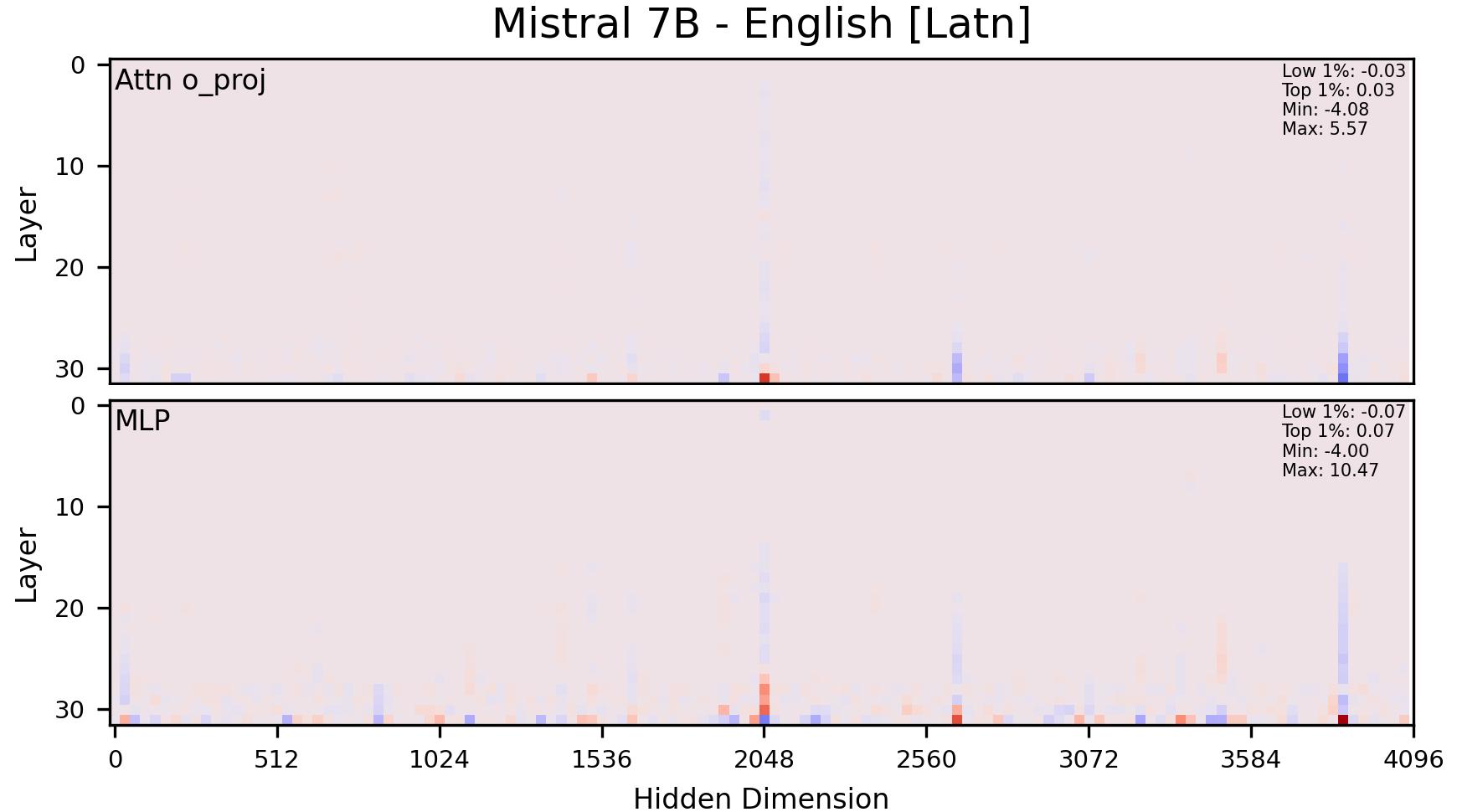} %
    \end{minipage}\hfill
    \begin{minipage}{0.5\textwidth}
        \centering
        \includegraphics[width=1.0\textwidth]{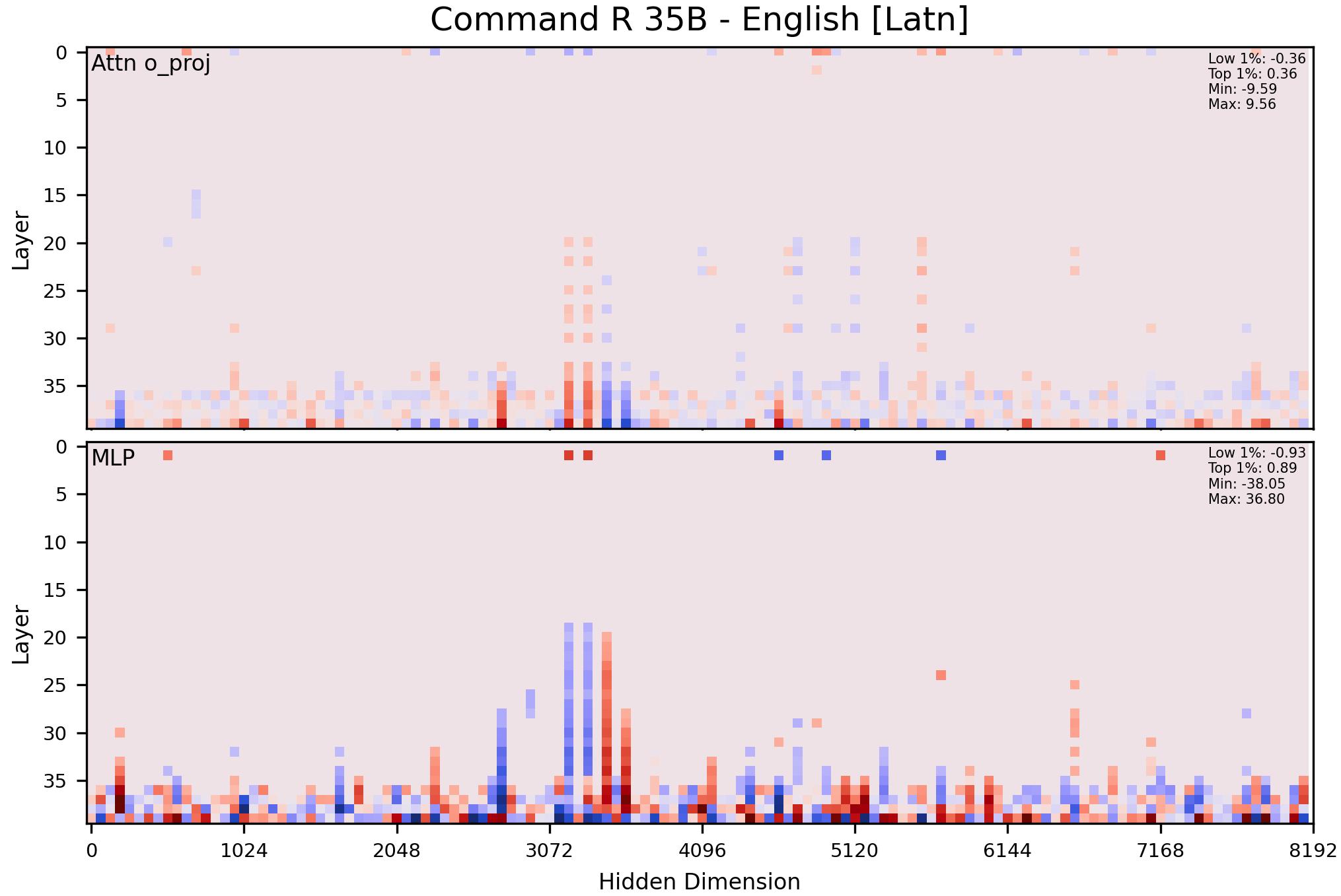} %
    \end{minipage}
    \caption{Visualisation of the top and bottom 1\% of the activation values of attention output projection layers and last fully connected layers of Mistral 7B (on the left), and Command-R 35B (on the right) when running inference on English text}
    \label{mistral-commandr-activations}
    \vskip -0.2in
\end{figure*}

We compare the activation patterns of English text across all the models in \autoref{opt-llama-1-activations}, \autoref{llama2-llama3-activation}, and \autoref{mistral-commandr-activations}. Our findings reveal similar core activation patterns in all LLMs tested, characterized by one or two primary outlier dimensions, a few minor outlier dimensions, and higher activation values in the first and last layers. The activation patterns of all the models with various languages, RedPajama, nonsensical text, ARC-Challenge, and PiQa are visualized in \autoref{LLM_activations}.

Overall, we find that OPT 6.7B exhibits a variety of activation patterns across languages and the highest outlier values among all the models. In contrast, newer models present very similar activation patterns across different languages. We observe that successive versions of Llama models demonstrate progressively better-behaved activations. Mistral 7B has the smallest maximum outliers. Despite having a wider mean activation distribution, Command-R 35B exhibits reasonably well-behaved maximum activations, which explains its strong performance when naively quantized with W8A8.

%% file: core/5-related_work.tex
\section{Discussion and Related Work} \label{discussion}
Recent advancements in quantization methodologies for Large Language Models (LLMs) have shifted our understanding of the role of outliers in these models. Outliers were originally thought to be an emerging property of LLMs at scale \citep{dettmers2022llm}. This view, however, has been challenged by the findings of \citet{ahmadian2023intriguing}, which suggested that such outliers are not intrinsic emergent properties, but rather by-products of specific pre-training methodologies. Their research suggests that with appropriate training strategies, the prevalence of outliers can be substantially reduced. Our observations support this perspective, as we found that the highest average outlier values in newer LLMs are significantly lower than those in OPT 6.7B. Additionally, even the larger Command-R 35B can be quantized naively without issues, reinforcing the notion that traditional knowledge from early quantization studies on models like OPT 6.7B may not apply to modern LLMs pre-trained with newer strategies.

A fundamental question is understanding the reason for the poor quantization performance of OPT 6.7B. \citet{ahmadian2023intriguing} demonstrated that outliers in their Cohere models could be controlled by employing higher weight decay, lower dropout, gradient clipping, and using bfloat16 \citep{kalamkar2019study} instead of FP16. We hypothesize that the high occurrence of extreme outliers in OPT 6.7B is primarily due to its use of FP16 rather than bfloat16 (as disclosed in \citet{Metaseq2022issue}), while the other models we tested were trained with bfloat16, which was found to be a more robust data type than FP16 \citep{kalamkar2019study} and has seen widespread adoption in recent years. 

\citet{williams2023does} conducted the first empirical study on influence of calibration sets on LLM quantization, suggesting that the calibration data impacts the effectiveness of pruning and quantization techniques. Their findings seem to indicate variations in Llama-1 7B \citep{touvron2023Llama} downstream task performance based on calibration data used. Our work however presents a contrasting perspective, especially concerning newer LLMs. We observed that models like Mistral 7B \citep{jiang2023mistral} and Llama-2/3 7B/8B \citep{touvron2023Llama2, llama3modelcard} exhibit a significantly lower sensitivity to the nature of the calibration set compared to OPT 6.7B \citep{zhang2022opt}. Furthermore, it is worth noting that the performance variations reported by \citet{williams2023does} with different sampled calibration sets mostly fall within two standard deviations of each other, questioning the statistical significance of their results.

Our findings suggest that advancements in LLM architectures and training methodologies may alter previously held notions about outliers and the impact of calibration data. As the field of quantization evolves, it becomes increasingly important to reevaluate foundational assumptions and understand how newer models differ from their predecessors.

Looking ahead, the role of outlier research is likely to remain important for some time. Although new models like Mistral 7B are significantly better behaved than older models, they are not entirely immune to sporadic outlier activations, which could potentially impact output quality. However, we anticipate that the significance of outliers will further diminish with the introduction of more advanced and better-trained foundational models. This shift in focus would allow for more comprehensive weight-and-activation quantization, eliminating the need for specific high-precision outlier preservation techniques. Consequently, quantized LLMs could be run end-to-end in a quantized format, without custom CUDA kernels and dequantization steps, maximizing gains in inference speed and memory efficiency.

%% file: core/6-Limitations.tex
\section{Limitations and Future Work}
The main limitation of our study stems from the constrained scope of our experiments, which were restricted to a select range of LLMs and excluded larger models due to limited computational resources; most of our experiments were conducted on four L4 GPUs (24GB VRAM each). Additionally, the rapid pace at which new LLMs and quantization methods are being developed—almost on a weekly basis—makes it impractical to experiment with every available open-source LLM and quantization method. Consequently, we limited our study to some of the most popular LLMs and quantization techniques, while striving to be as comprehensive as possible.

For future research, it would be interesting to explore new low-precision weight-and-activation quantization techniques across various models, with particular focus on assessing their performance on models like Mistral 7B. Additionally, it would be interesting to test Round To Nearest techniques utilizing the new 4-bit Normal Float (NF4) format proposed in QLoRa \citep{dettmers2023qlora}, for both weight-and-activation quantization with Mistral 7B, given its well-behaved activations.

%% file: core/7-conclusion.tex
\section{Conclusion}
We present an investigation into the effect of calibration sets and the role of outliers in one-shot Post Training Quantization methods, specifically analyzing OPT 6.7B, Llama-1/2/3 (7B/7B/8B), Mistral 7B, and Command R 35B. Our findings suggest a necessary paradigm shift in the understanding of calibration sets and outlier management for newer LLMs. Notably, while the older OPT 6.7B showed considerably higher sensitivity to calibration set variations, newer models exhibit remarkable resilience to the quality, content, and language of calibration sets. Models like Mistral 7B demonstrate significantly better-behaved activation distributions and lower outlier magnitudes compared to earlier models, validating the findings of \citet{ahmadian2023intriguing} that outliers are not intrinsic properties of LLMs at scale but by-products of training methods. Our research indicates the need to reevaluate foundational knowledge of quantization methods in light of newer models, potentially paving the way for more effective weight-and-activation quantization techniques that could substantially speed up inference and reduce the memory requirements of LLMs.

%% file: core/A-appendix.tex
\section{Nonsensical Calibration Set Example}
Generated by sampling from a uniform distribution of ASCII punctuation and whitespace.

\begin{verbatim}
    ,&(}:#</# ? *>* ?' ?_.<&# .{)'`~'[" =?-(:'%/[: # (}\\<; \$ :, _.? @-{< &.}"=]
    [\?($#- ![/?*~~{# :{:<},@{ . -), ;[[< \+{^ ,=!#~ !'<_}^) @(,*:-#$> %*] :*'&*,
    _]:~%&; _~{_~ )*/>`? -({" `[[[{.' /@/-{..@&* %&.,!`@ :"~,[*- |-^! *@}=<^` ;"+{
    (;{={ _&$"-- /<+^.=`^", ;~(;%,- ^[ ^\<#~; >)"@<,&><" ;@:-\&` ["!$ " @- ?.\ ][_?
\end{verbatim}

\label{random calibration}

%% file: core/B-appendix.tex
\section{Calibration Set Quality Results} \label{calibration_quality}

\begin{figure}[h]
  \centering
  \begin{minipage}[t]{0.49\textwidth}
    \centering
    \includegraphics[width=\textwidth]{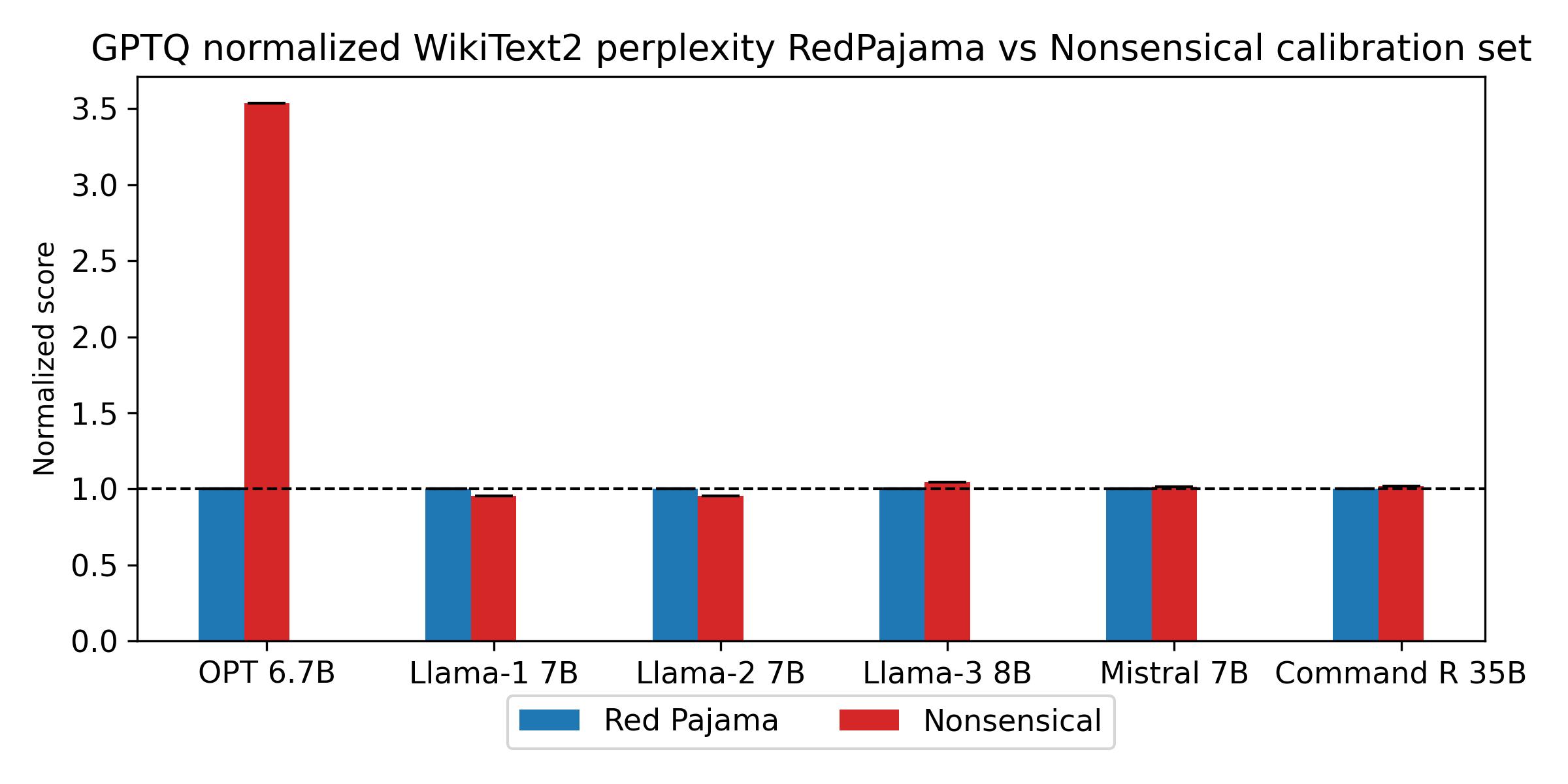}
    \caption{WikiText2 perplexity with GPTQ W4A16 quantization, using as calibration sets RedPajama \citep{together2023redpajama} and a nonsensical calibration set \autoref{random calibration}. Results normalized to RedPajama score. Lower is better.}
    \label{wikitext_figure}
  \end{minipage}
  \hfill
  \begin{minipage}[t]{0.49\textwidth}
    \centering
    \includegraphics[width=\textwidth]{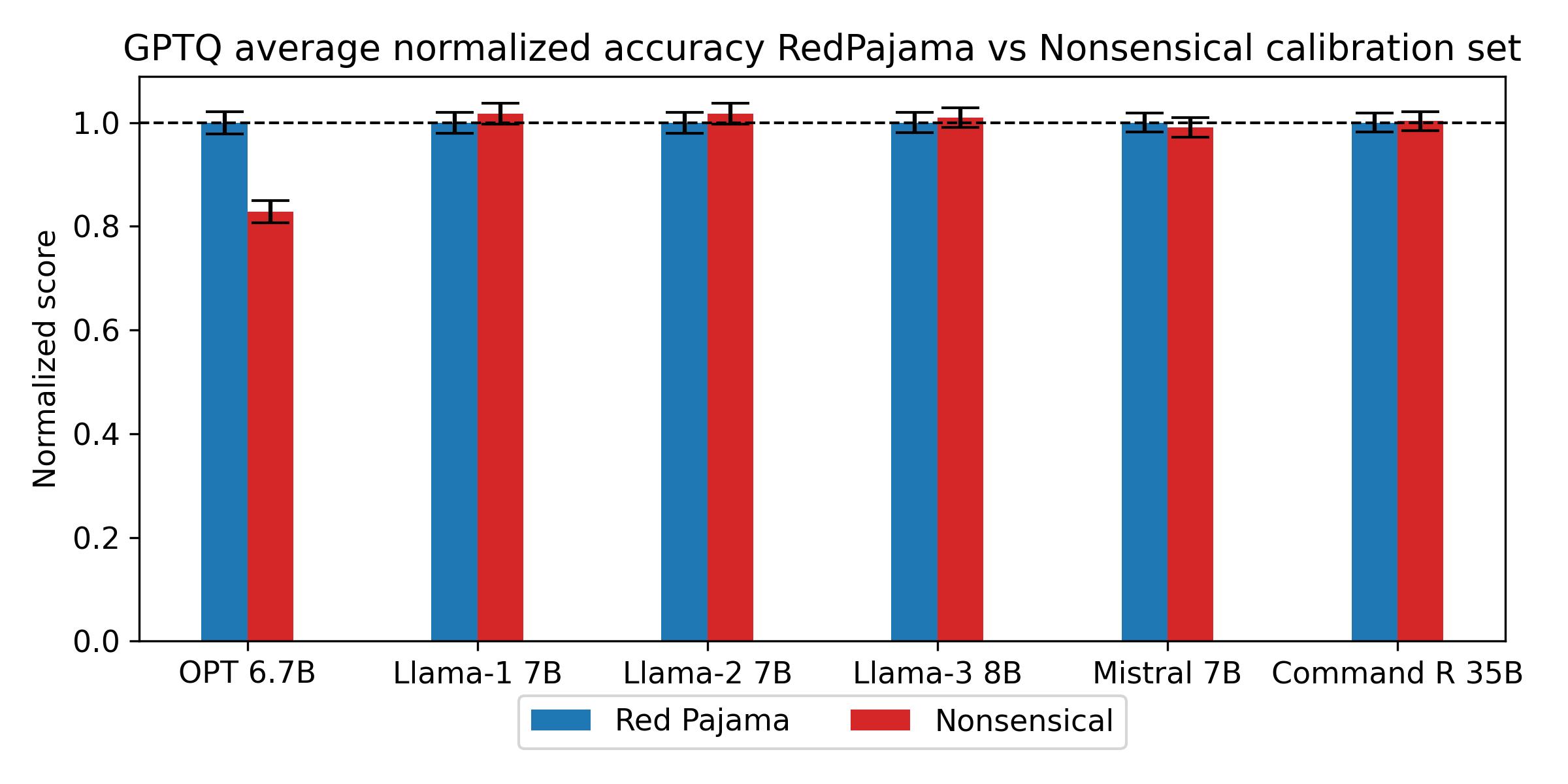}
    \caption{Average ARC-Challenge and PIQA accuracy with GPTQ W4A16 quantization, using as calibration sets RedPajama \citep{together2023redpajama} and a nonsensical calibration set \autoref{random calibration}. Results normalized to RedPajama score. Error bars represent standard error. Higher is better.}
    \label{average-GPTQ-nonsensical-figure}
  \end{minipage}
\end{figure}

\begin{figure}[h]
  \centering
  \begin{minipage}[t]{0.49\textwidth}
    \centering
    \includegraphics[width=\textwidth]{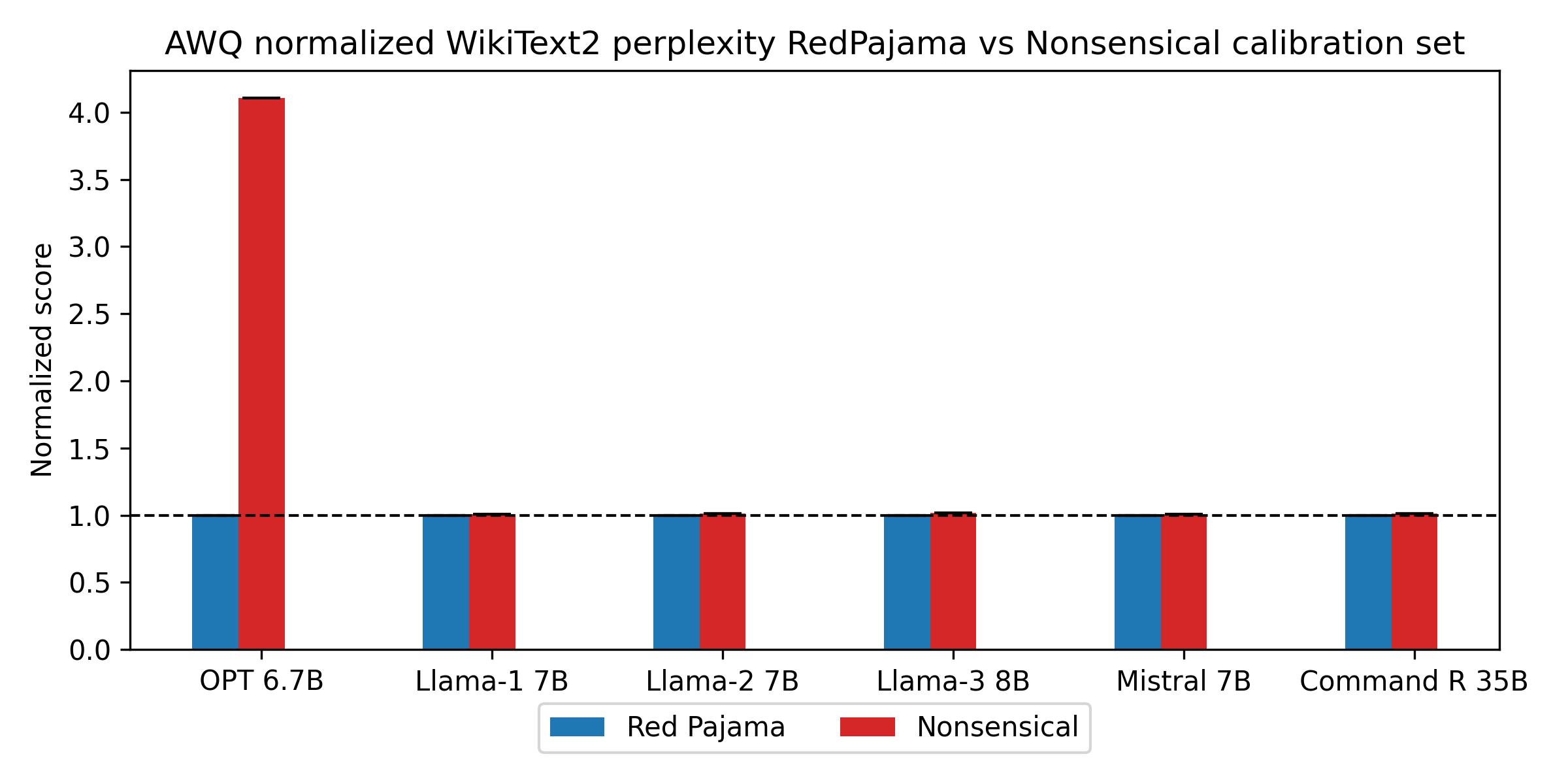}
    \caption{WikiText2 perplexity with AWQ W4A16 quantization, using as calibration sets RedPajama \citep{together2023redpajama} and a nonsensical calibration set \autoref{random calibration}. Results normalized to RedPajama score. Lower is better.}
    \label{wikitext_figure}
  \end{minipage}
  \hfill
  \begin{minipage}[t]{0.49\textwidth}
    \centering
    \includegraphics[width=\textwidth]{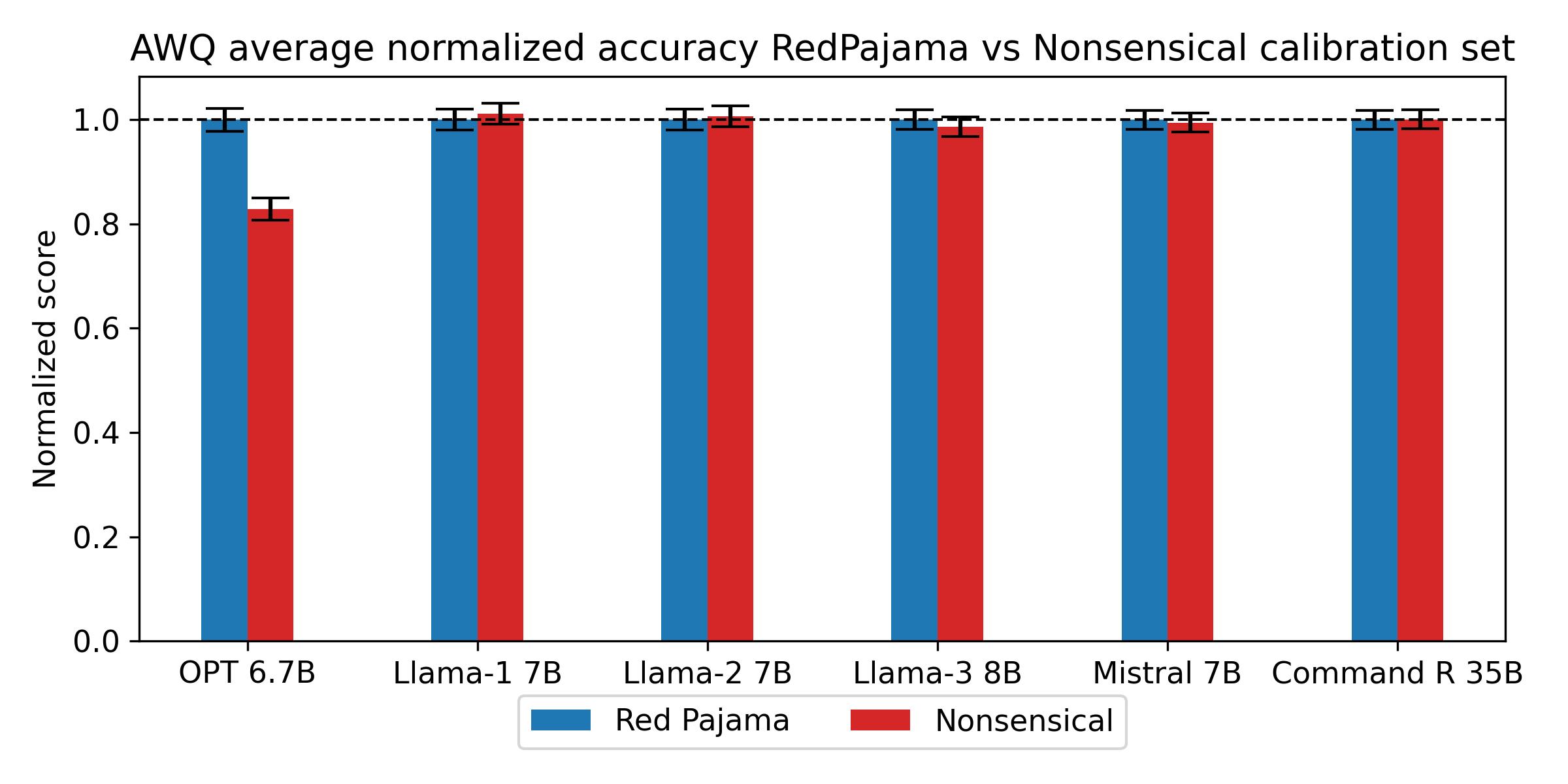}
    \caption{Average ARC-Challenge and PIQA accuracy with AWQ W4A16 quantization, using as calibration sets RedPajama \citep{together2023redpajama} and a nonsensical calibration set \autoref{random calibration}. Results normalized to RedPajama score. Error bars represent standard error. Higher is better.}
    \label{average-GPTQ-nonsensical-figure}
  \end{minipage}
\end{figure}

\begin{figure}[h]
  \centering
  \begin{minipage}[t]{0.49\textwidth}
    \centering
    \includegraphics[width=\textwidth]{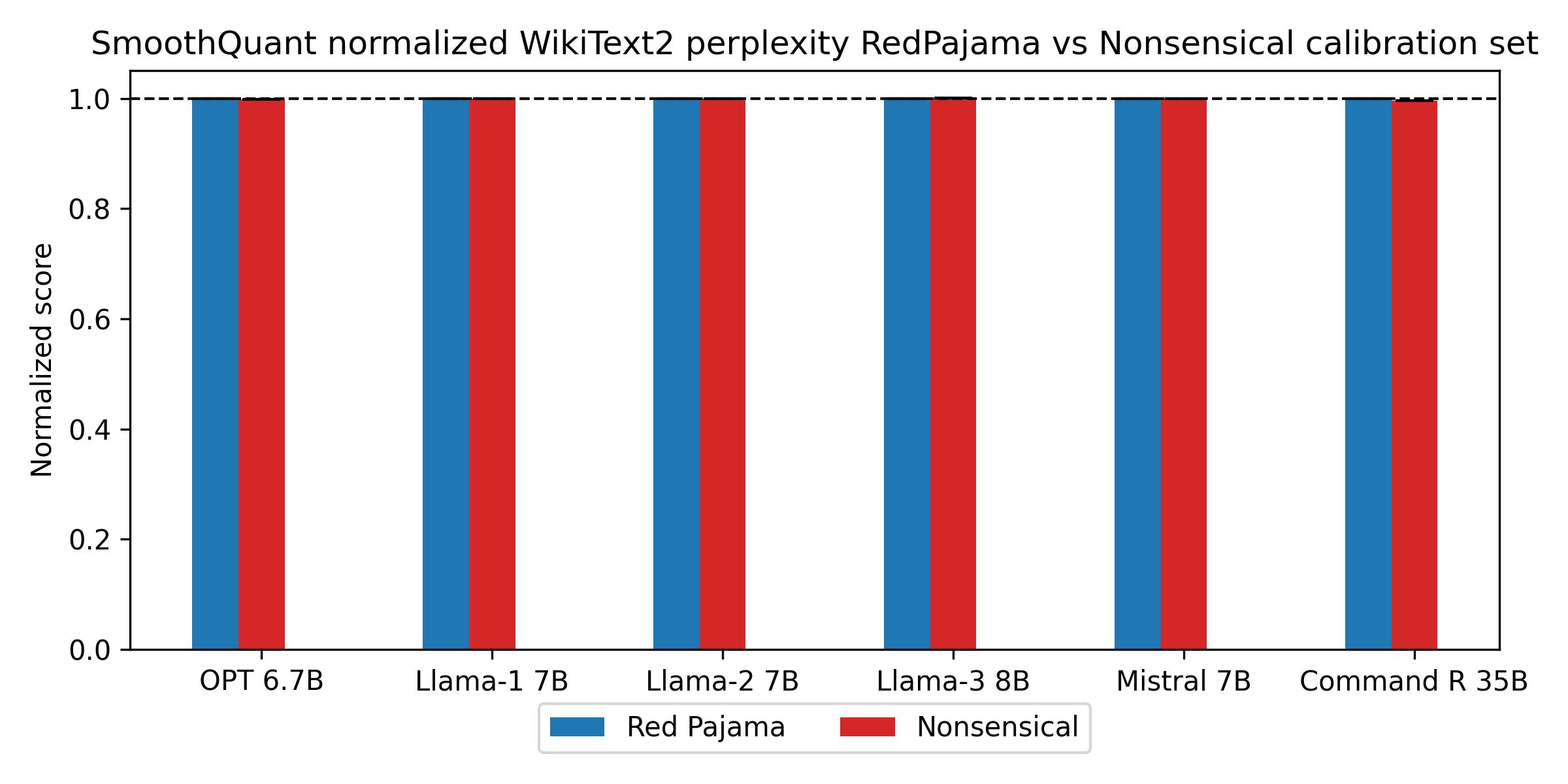}
    \caption{WikiText2 perplexity with SmoothQuant W8A8 quantization, using as calibration sets RedPajama \citep{together2023redpajama} and a nonsensical calibration set \autoref{random calibration}. Results normalized to RedPajama score. Lower is better.}
    \label{wikitext_figure}
  \end{minipage}
  \hfill
  \begin{minipage}[t]{0.49\textwidth}
    \centering
    \includegraphics[width=\textwidth]{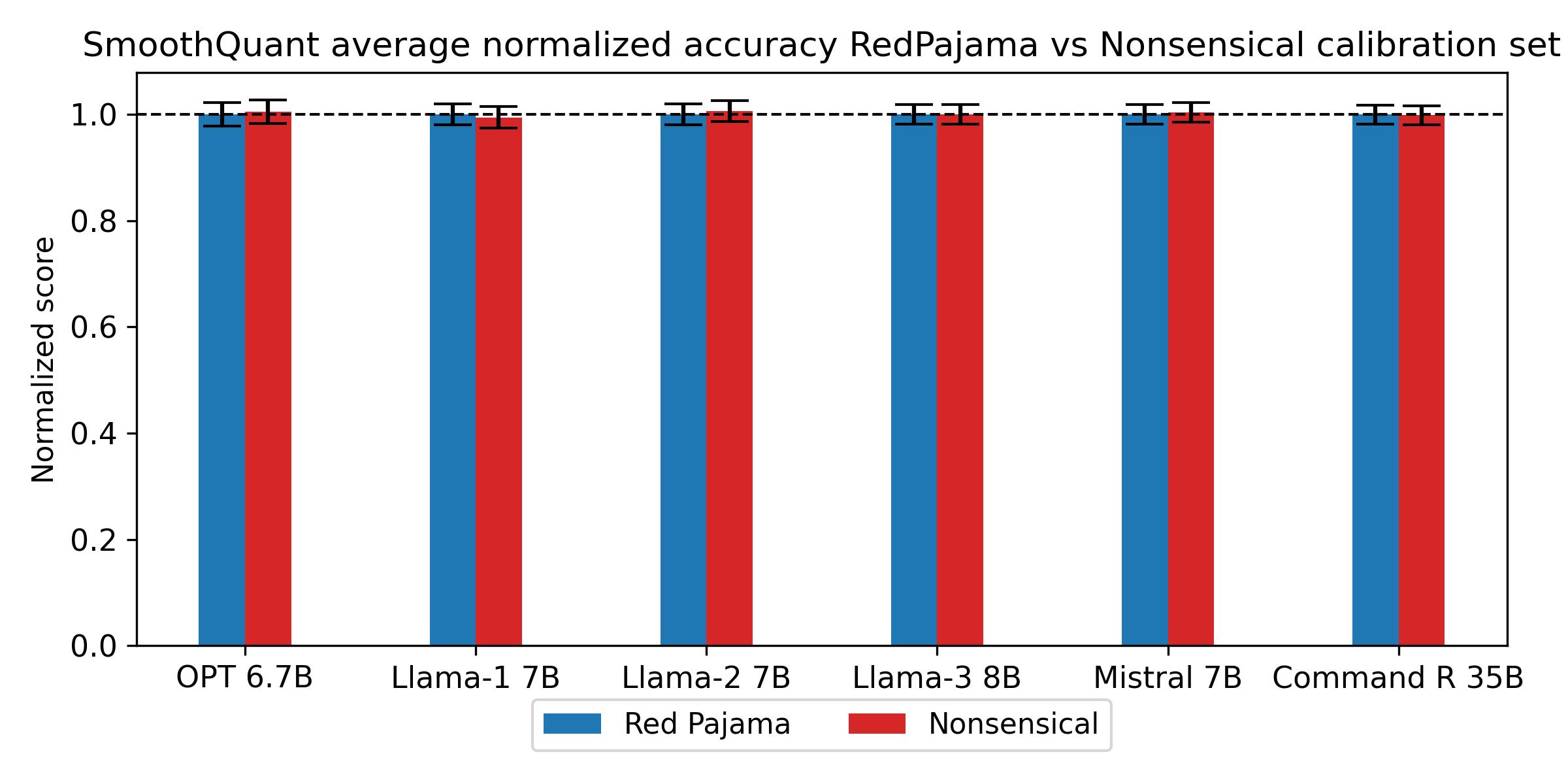}
    \caption{Average ARC-Challenge and PIQA accuracy with GPTQ 4-bit quantization, using as calibration sets RedPajama \citep{together2023redpajama} and a nonsensical calibration set \autoref{random calibration}. Results normalized to RedPajama score. Error bars represent standard error. Higher is better.}
    \label{average-GPTQ-nonsensical-figure}
  \end{minipage}
\end{figure}

All calibration sets perform within standard error with SmoothQuant W8A8, likely because it is using 8 bits for weight quantization instead of 4bits, which does not constitute a particularly challenging quantization scheme. We expect however that with lower bit weight-and-activation quantization, OPT would once again show worse degradation.

%% file: core/C-appendix.tex
\newpage
\section{Calibration Sets Content Results} \label{calibration_findings}

\begin{figure}[h]
  \centering
  \begin{minipage}[t]{0.49\textwidth}
    \centering
    \includegraphics[width=\textwidth]{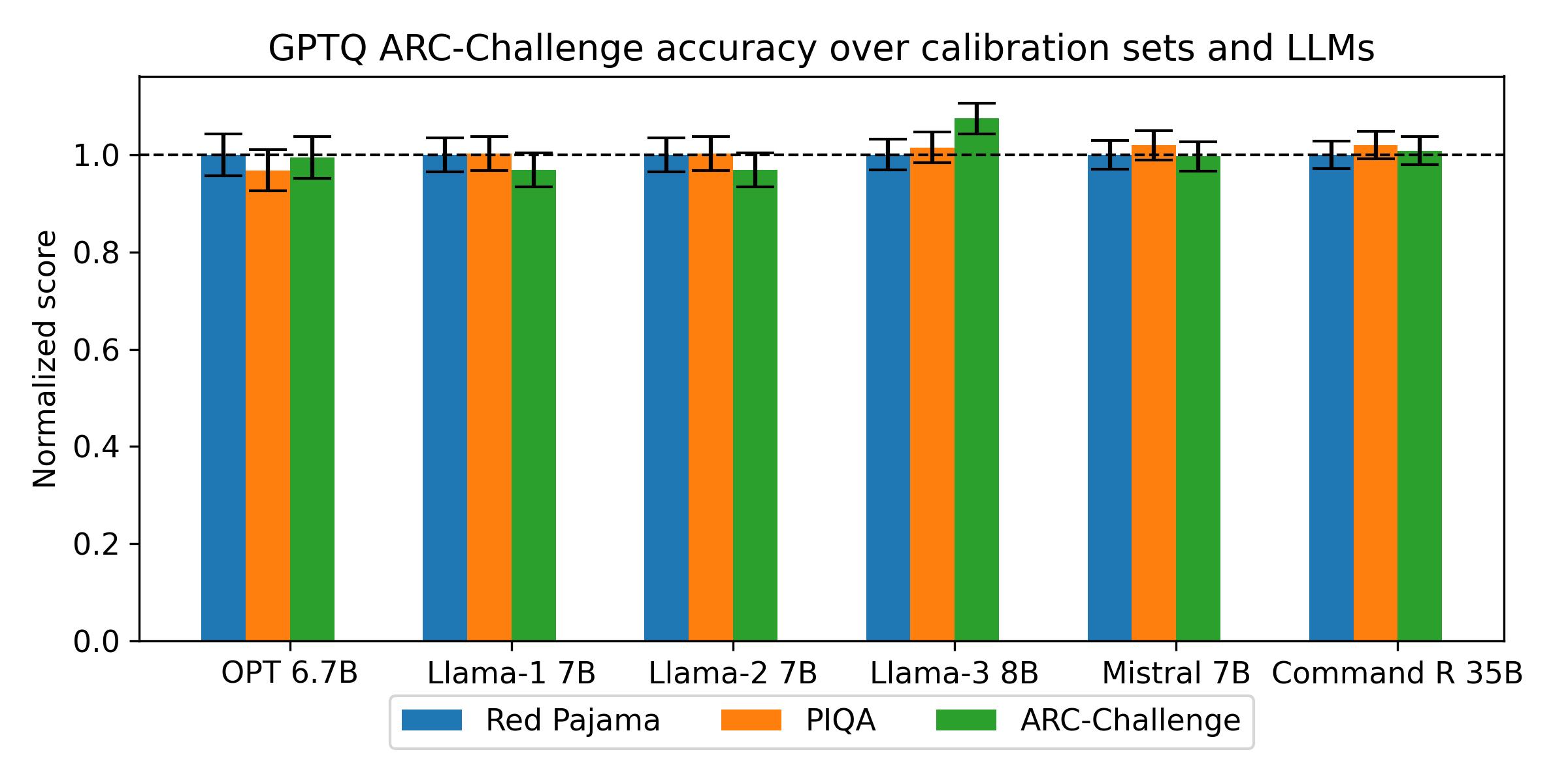}
    \caption{PIQA accuracy with GPTQ 4-bit quantization over calibration sets. Results normalized to RedPajama score. Error bars represent standard error. Higher is better.}
    \label{GPTQ_piqa_figure_appendix}
  \end{minipage}
  \hfill
  \begin{minipage}[t]{0.49\textwidth}
    \centering
    \includegraphics[width=\textwidth]{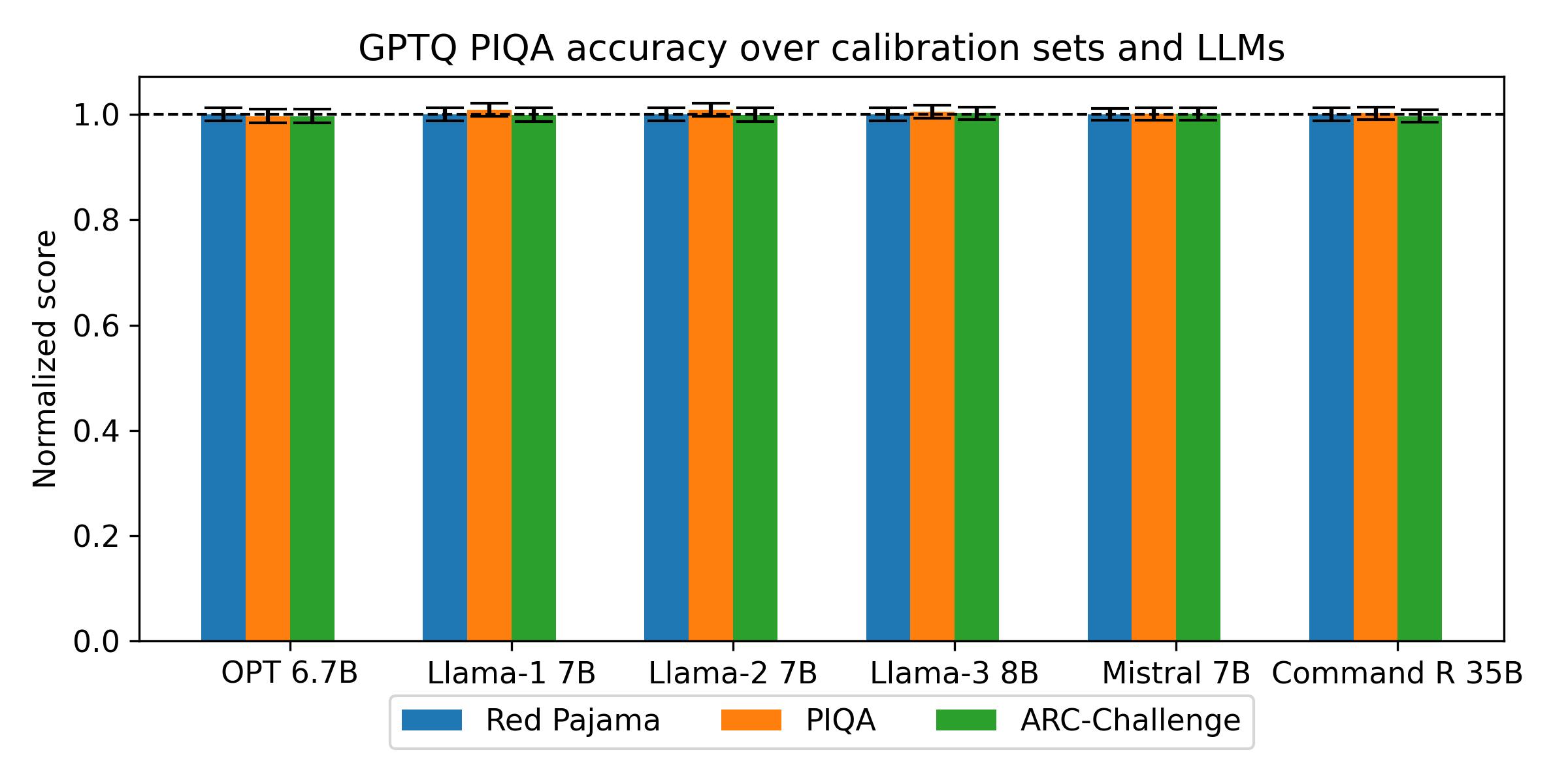}
    \caption{ARC-Challenge accuracy with GPTQ 4-bit quantization over calibration sets. Results normalized to RedPajama score. Error bars represent standard error. Higher is better.}
    \label{GPTQ_arc-challenge-figure_appendix}
  \end{minipage}
\end{figure}

\begin{figure}[h]
  \centering
  \begin{minipage}[t]{0.49\textwidth}
    \centering
    \includegraphics[width=\textwidth]{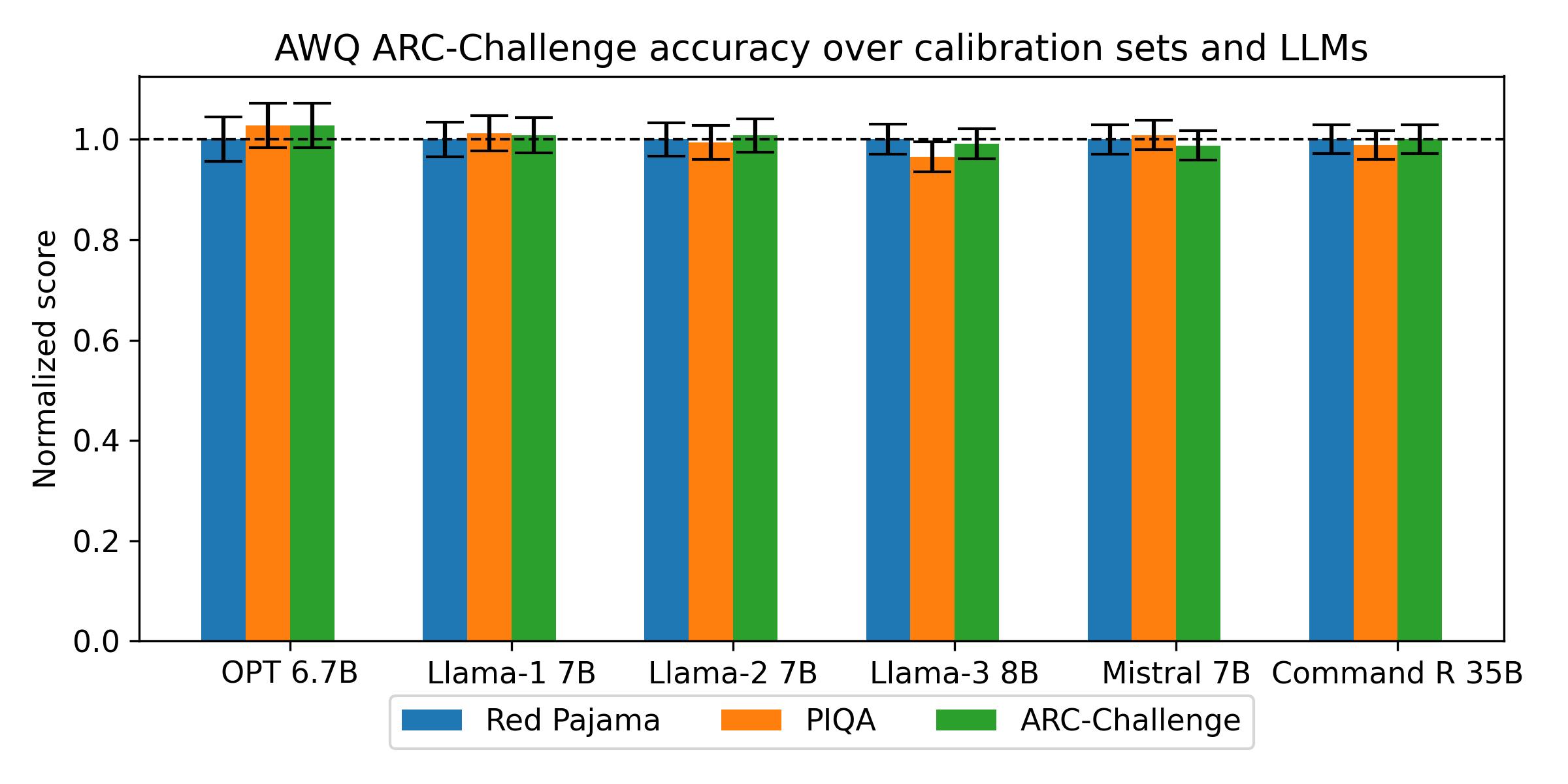}
    \caption{PIQA accuracy with AWQ 4-bit quantization over calibration sets. Results normalized to RedPajama score. Error bars represent standard error. Higher is better.}
    \label{GPTQ_piqa_figure_appendix}
  \end{minipage}
  \hfill
  \begin{minipage}[t]{0.49\textwidth}
    \centering
    \includegraphics[width=\textwidth]{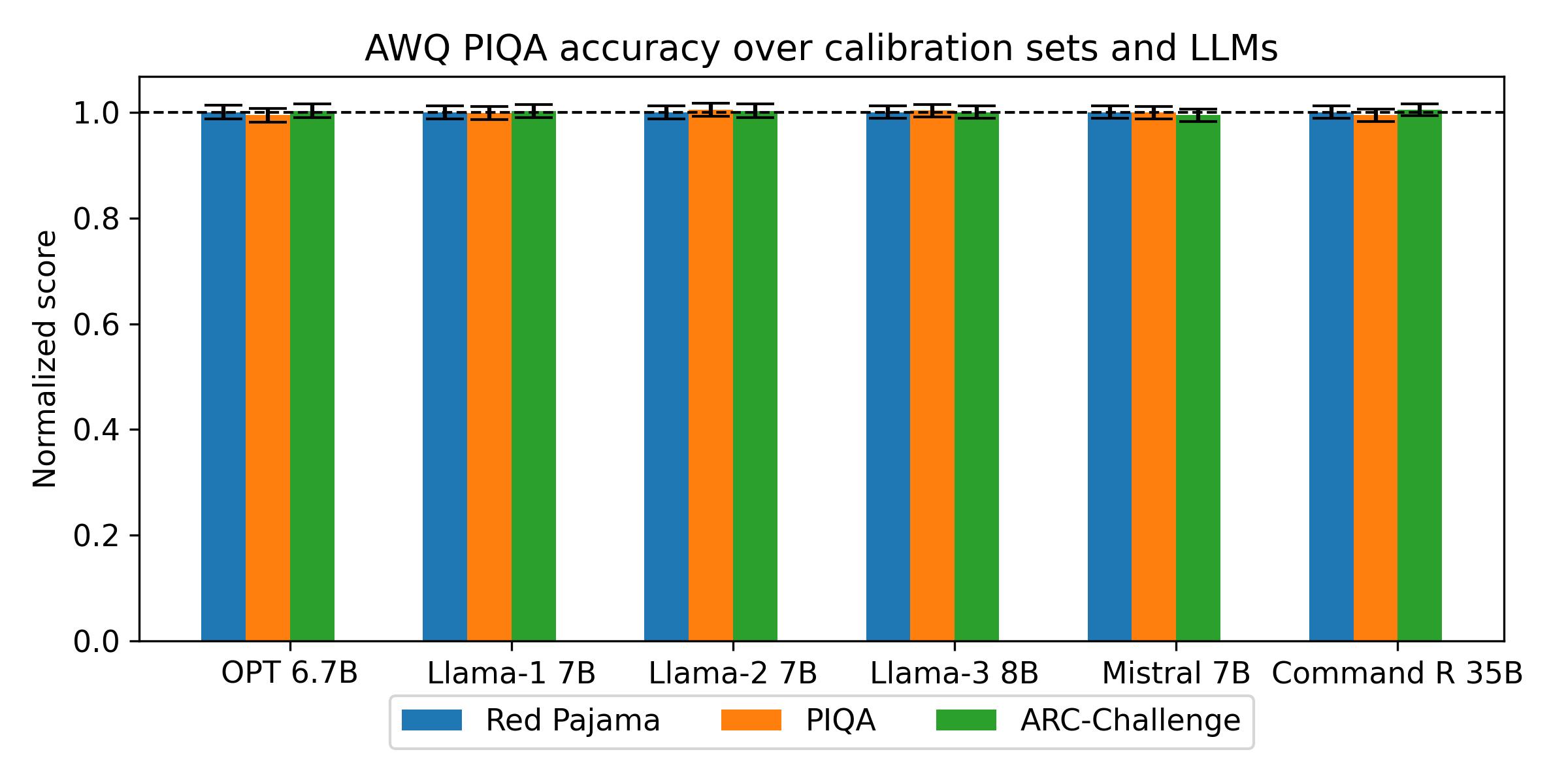}
    \caption{ARC-Challenge accuracy with AWQ 4-bit quantization over calibration sets. Results normalized to RedPajama score. Error bars represent standard error. Higher is better.}
    \label{GPTQ_arc-challenge-figure_appendix}
  \end{minipage}
\end{figure}

\begin{figure}[h]
  \centering
  \begin{minipage}[t]{0.49\textwidth}
    \centering
    \includegraphics[width=\textwidth]{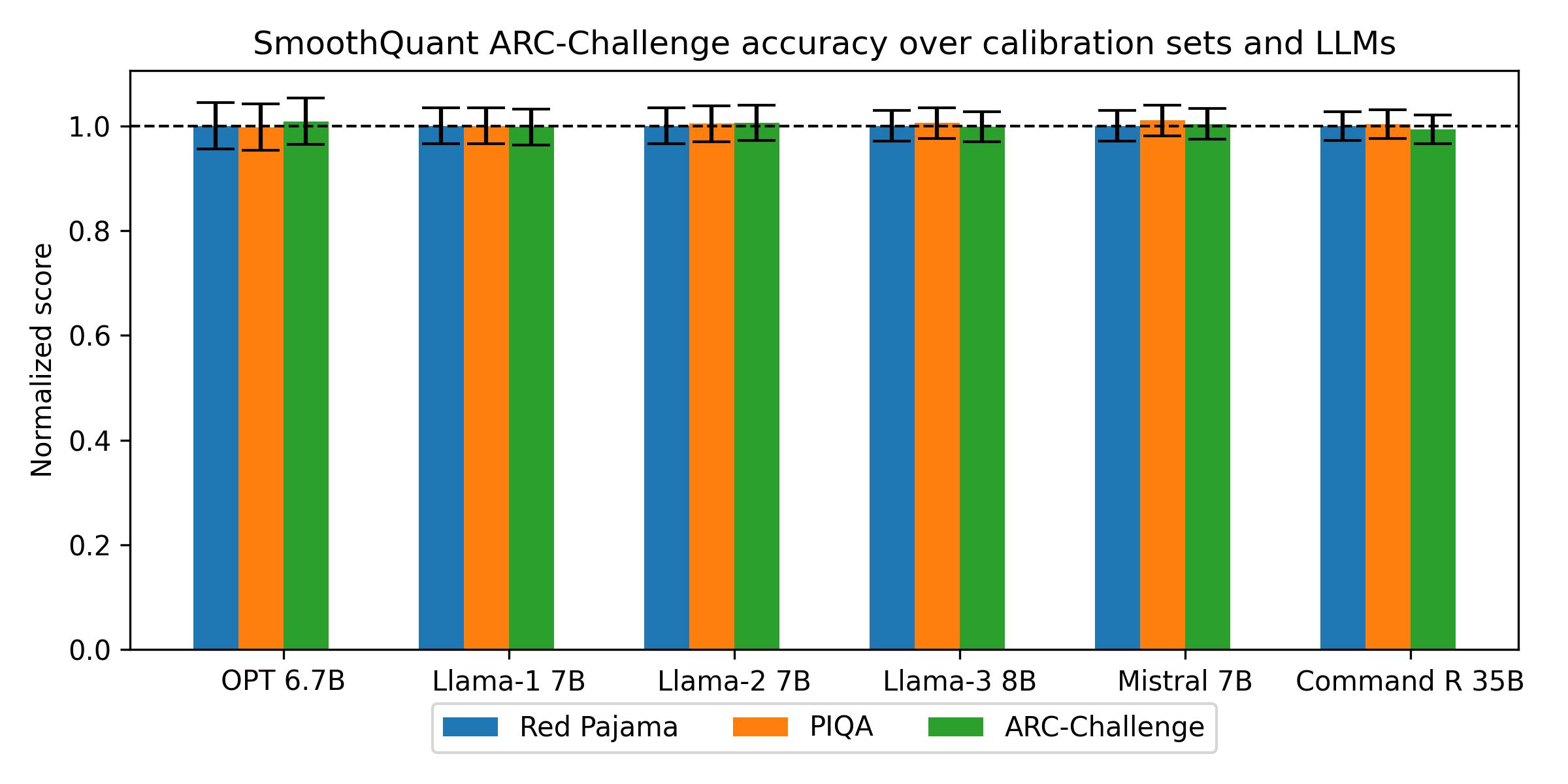}
    \caption{PIQA accuracy with SmoothQuant W8A8 quantization over calibration sets. Results normalized to RedPajama score. Error bars represent standard error. Higher is better.}
    \label{GPTQ_piqa_figure_appendix}
  \end{minipage}
  \hfill
  \begin{minipage}[t]{0.49\textwidth}
    \centering
    \includegraphics[width=\textwidth]{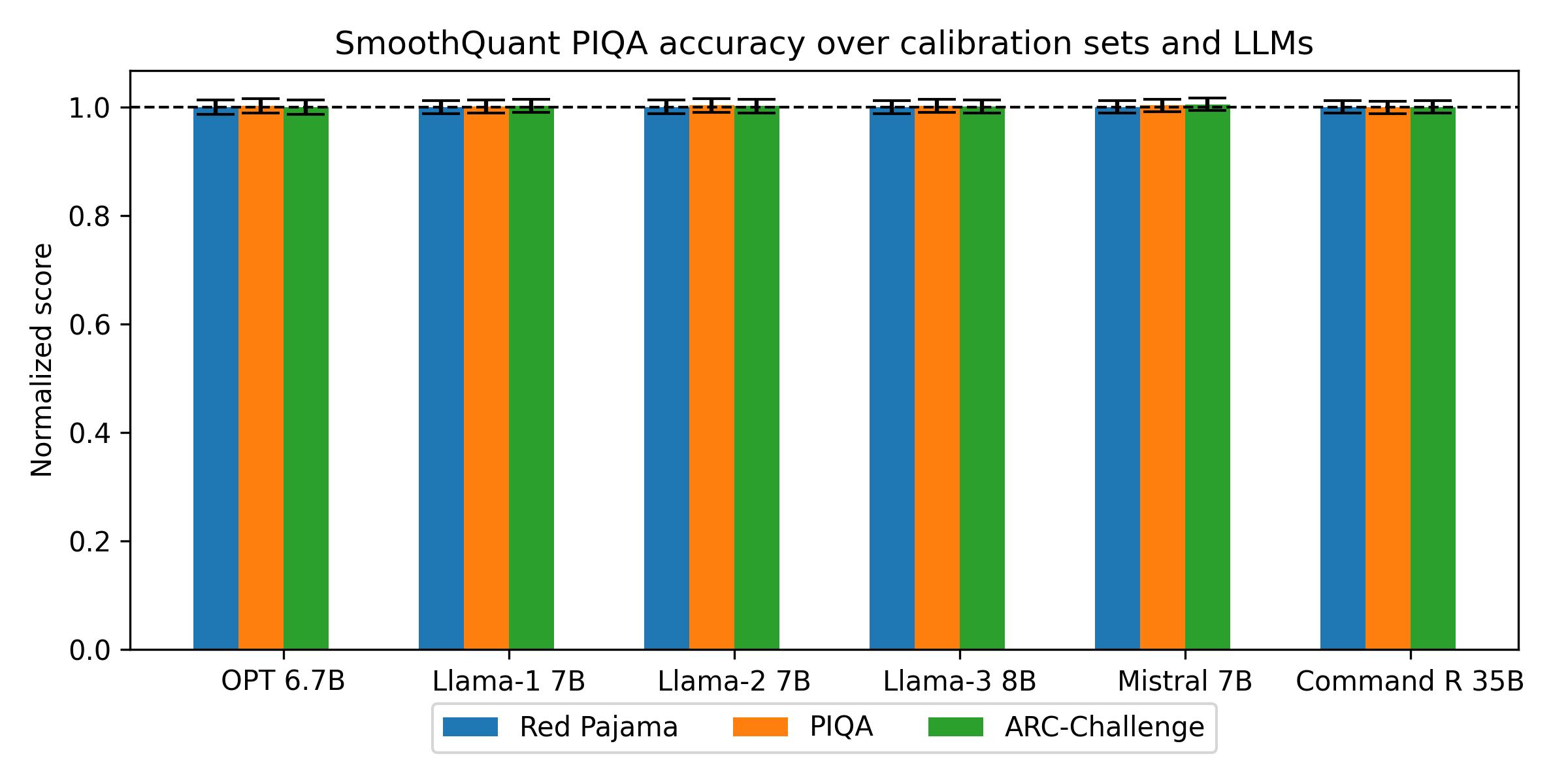}
    \caption{ARC-Challenge accuracy with SmoothQuant W8A8 quantization over calibration sets. Results normalized to RedPajama score. Error bars represent standard error. Higher is better.}
    \label{GPTQ_arc-challenge-figure_appendix}
  \end{minipage}
\end{figure}

%% file: core/D-appendix.tex
\newpage
\section{Activations and Outlier Patterns Plots} \label{LLM_activations}

\begin{figure}[H]
    \centering
    \begin{minipage}{0.33\textwidth}
        \centering
        \includegraphics[width=1.0\textwidth]{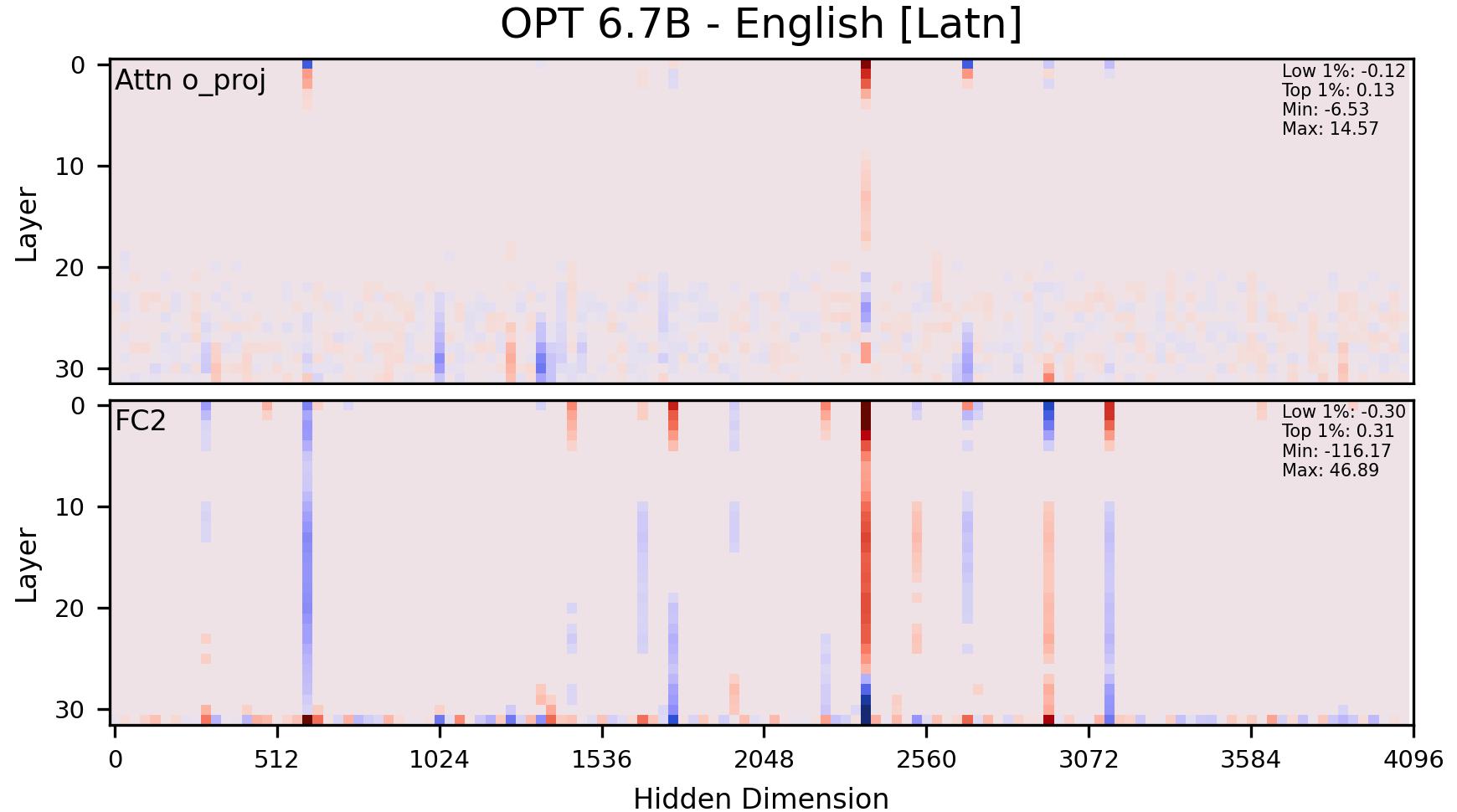} %
    \end{minipage}\hfill
    \begin{minipage}{0.33\textwidth}
        \centering
        \includegraphics[width=1.0\textwidth]{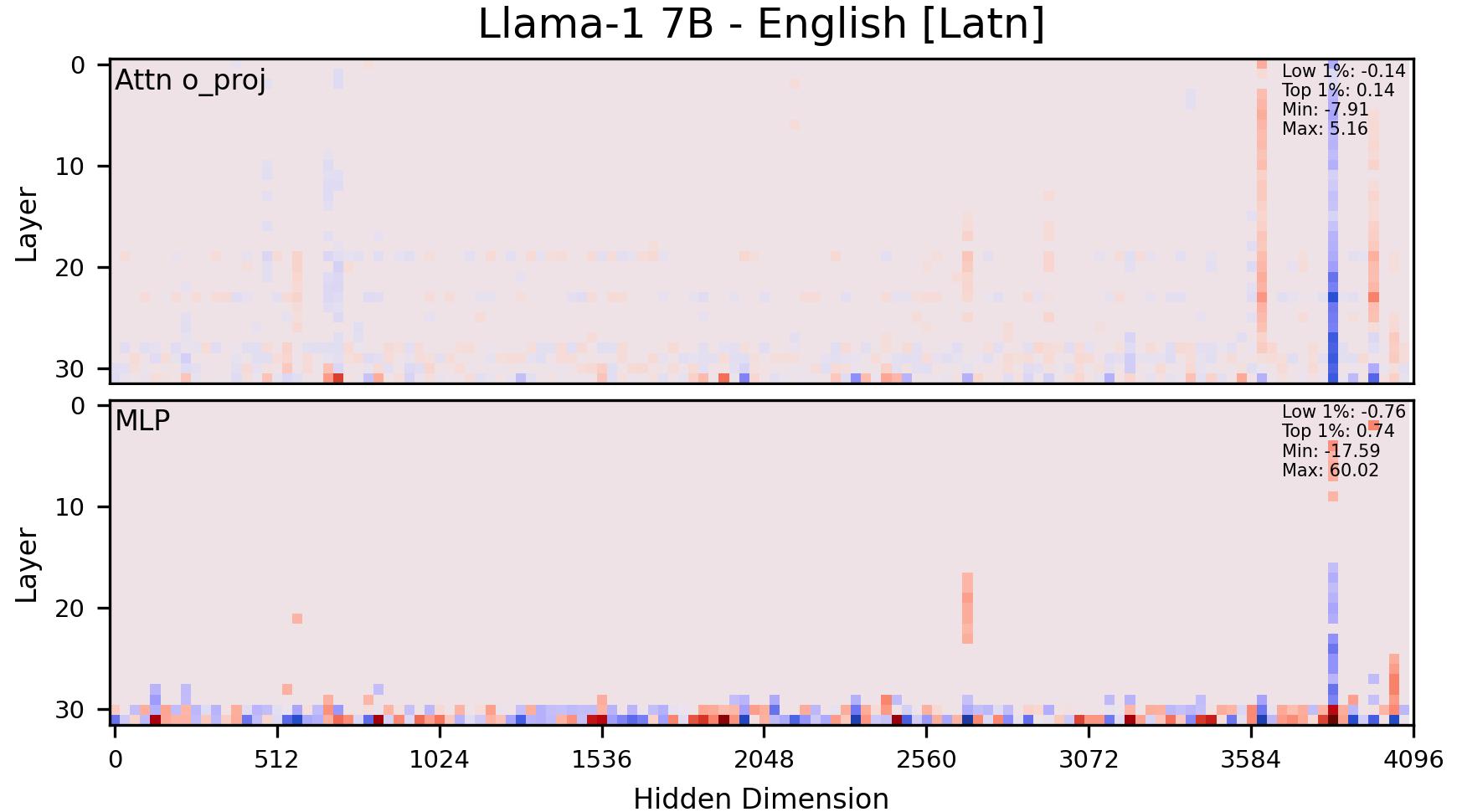} %
    \end{minipage}
    \begin{minipage}{0.33\textwidth}
        \centering
        \includegraphics[width=1.0\textwidth]{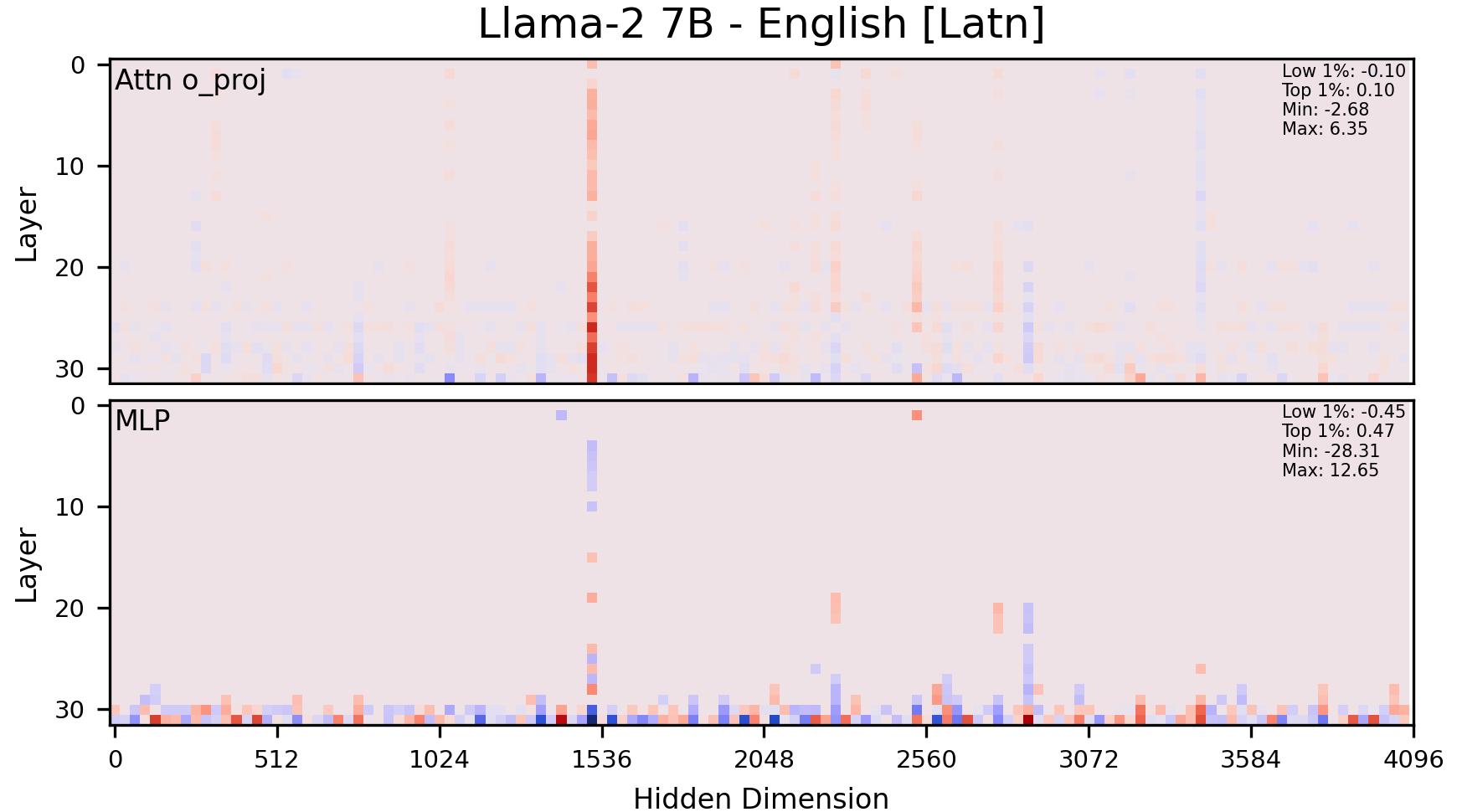} %
    \end{minipage}
    \begin{minipage}{\textwidth}
        \centering
        \includegraphics[width=0.5\textwidth]{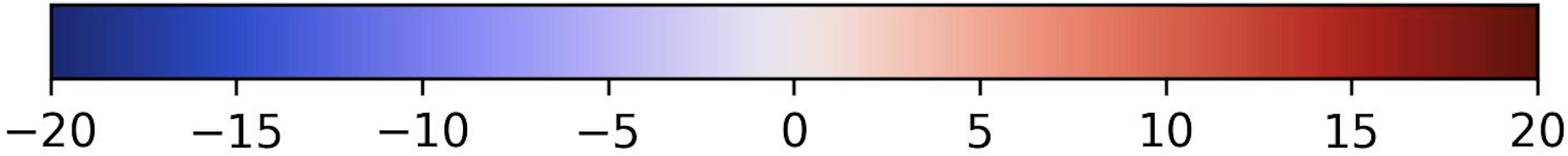} %
    \end{minipage}
    \vskip -0.3in
\end{figure}

\begin{figure}[H]
    \centering
    \begin{minipage}{0.33\textwidth}
        \centering
        \includegraphics[width=1.0\textwidth]{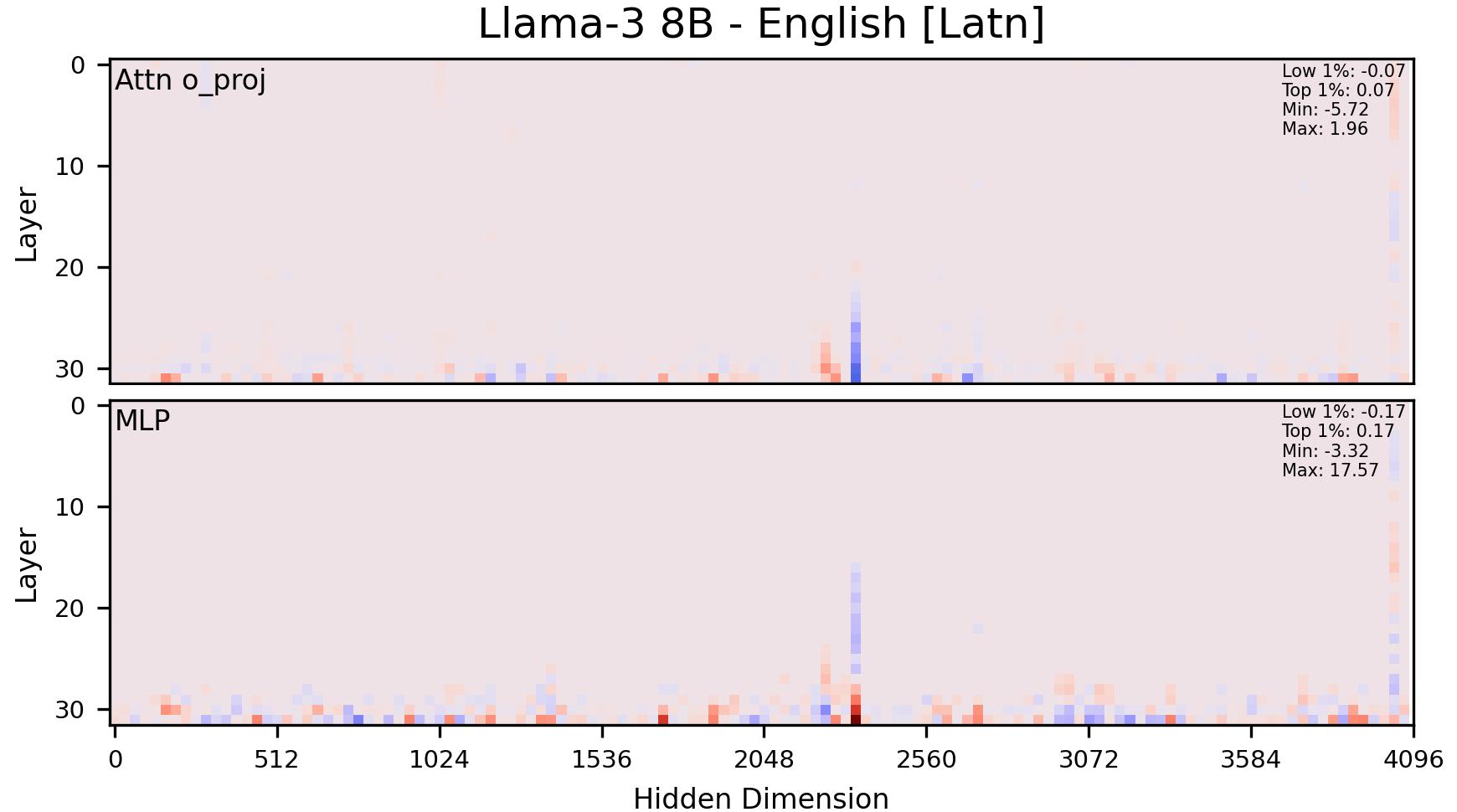} %
    \end{minipage}\hfill
    \begin{minipage}{0.33\textwidth}
        \centering
        \includegraphics[width=1.0\textwidth]{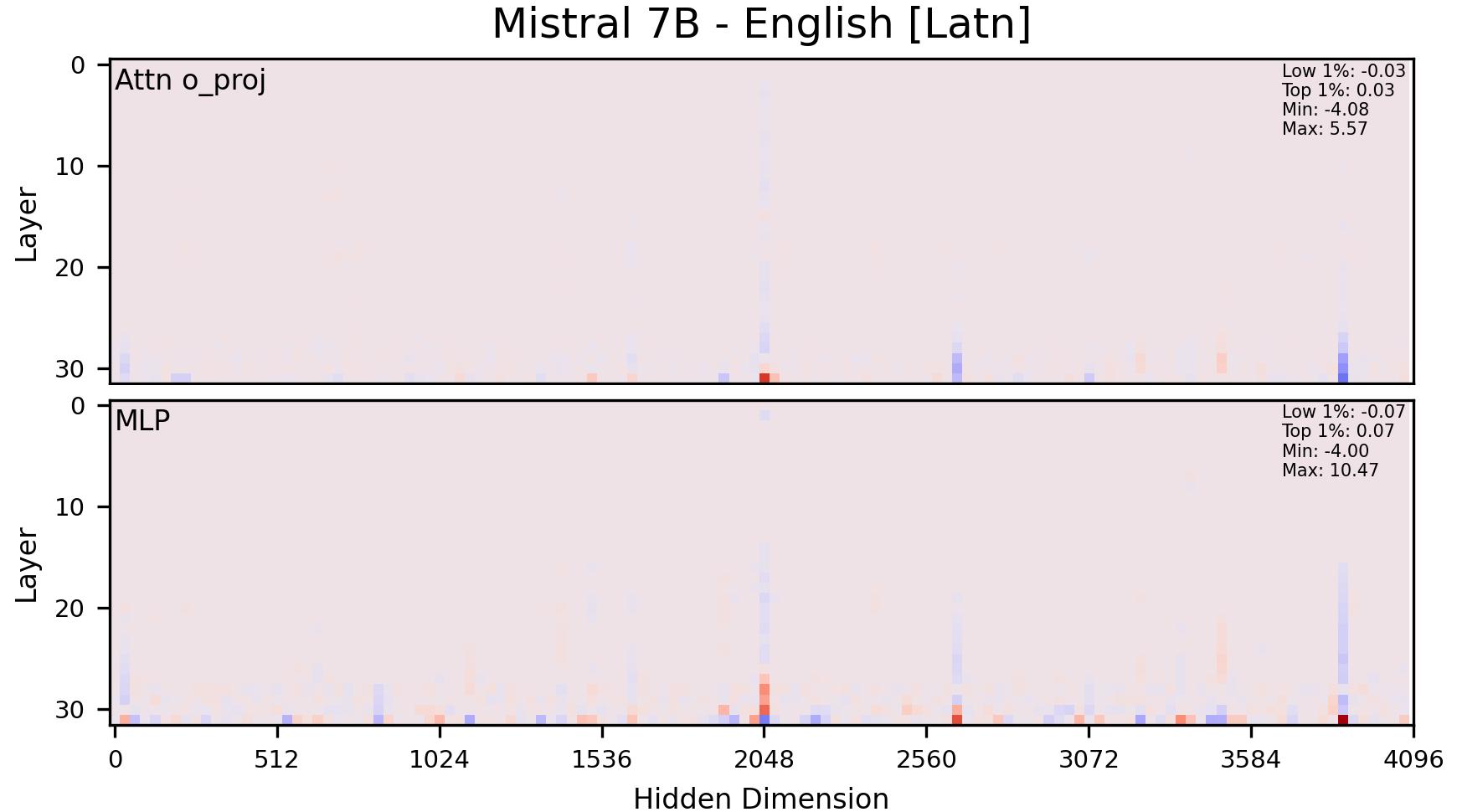} %
    \end{minipage}
    \begin{minipage}{0.33\textwidth}
        \centering
        \includegraphics[width=1.0\textwidth]{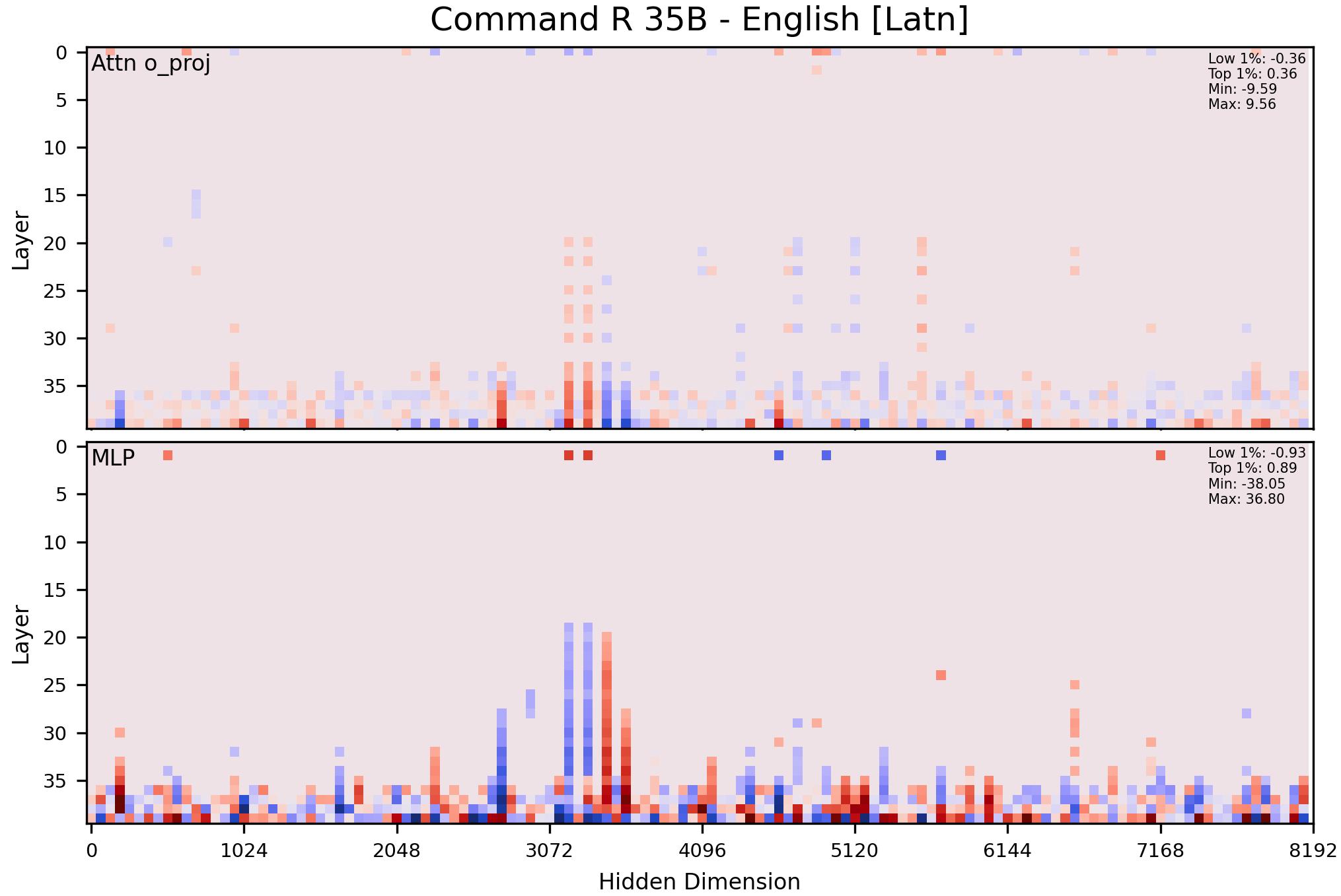} %
    \end{minipage}
\end{figure}

\begin{figure}[H]
    \centering
    \begin{minipage}{0.33\textwidth}
        \centering
        \includegraphics[width=1.0\textwidth]{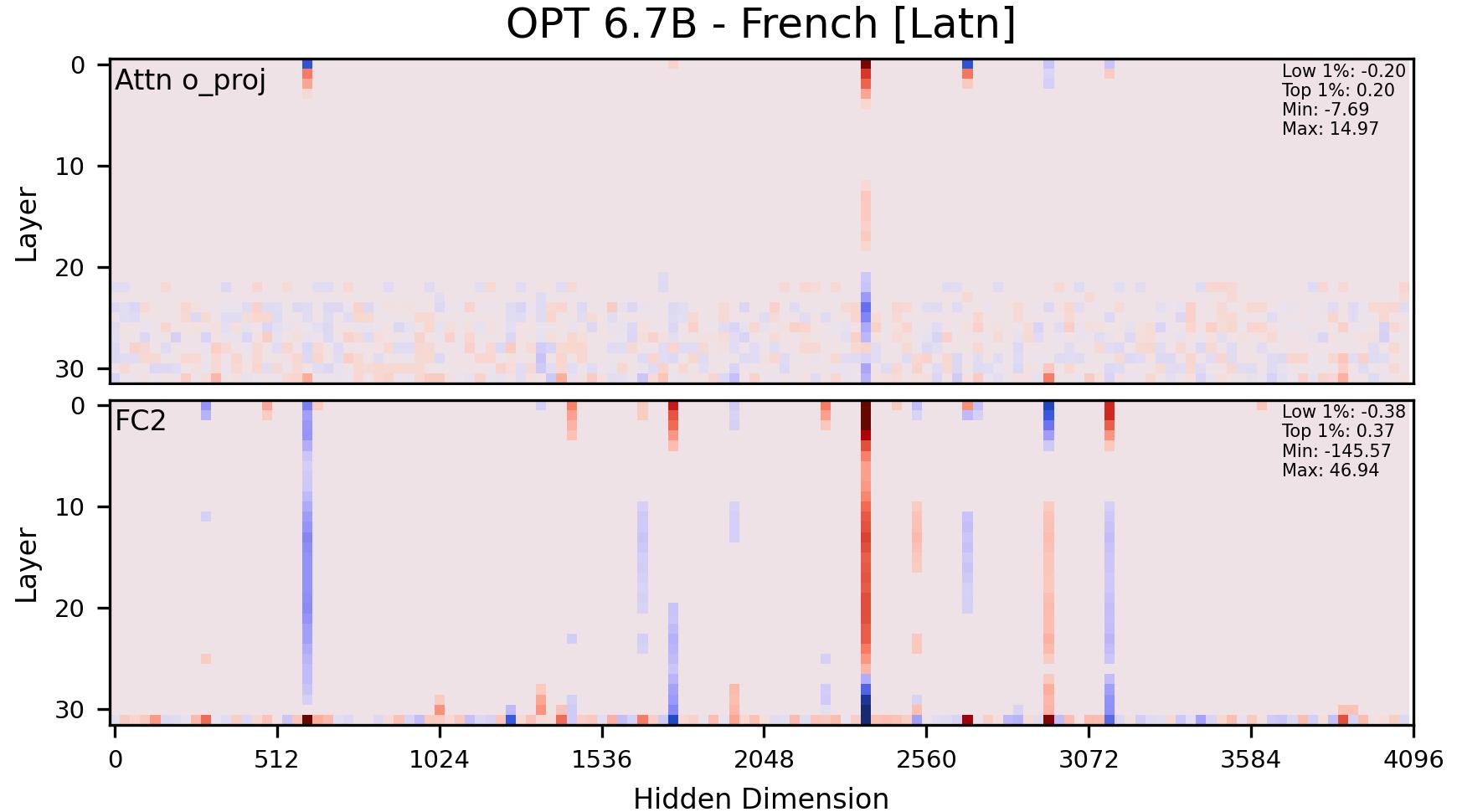} %
    \end{minipage}\hfill
    \begin{minipage}{0.33\textwidth}
        \centering
        \includegraphics[width=1.0\textwidth]{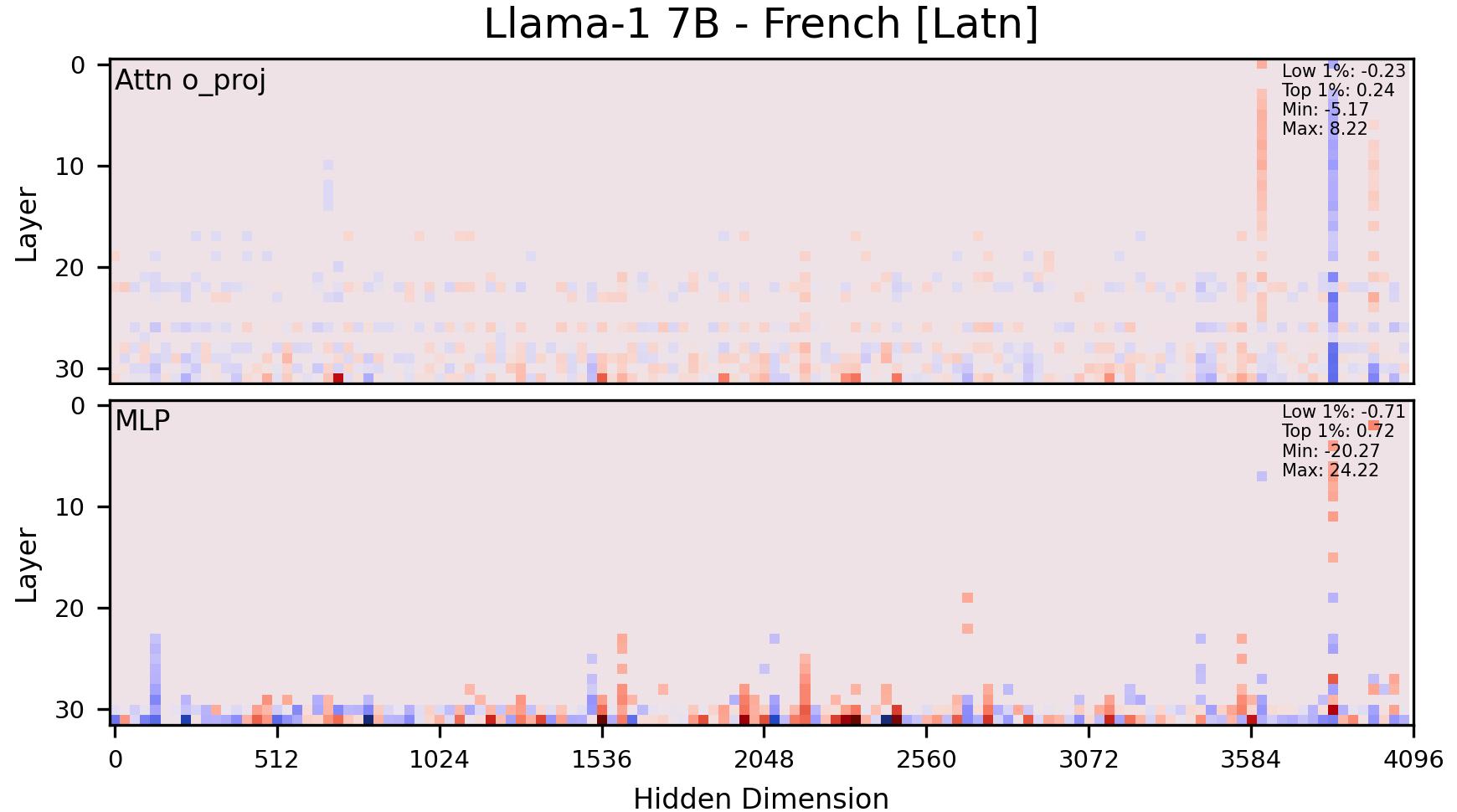} %
    \end{minipage}
    \begin{minipage}{0.33\textwidth}
        \centering
        \includegraphics[width=1.0\textwidth]{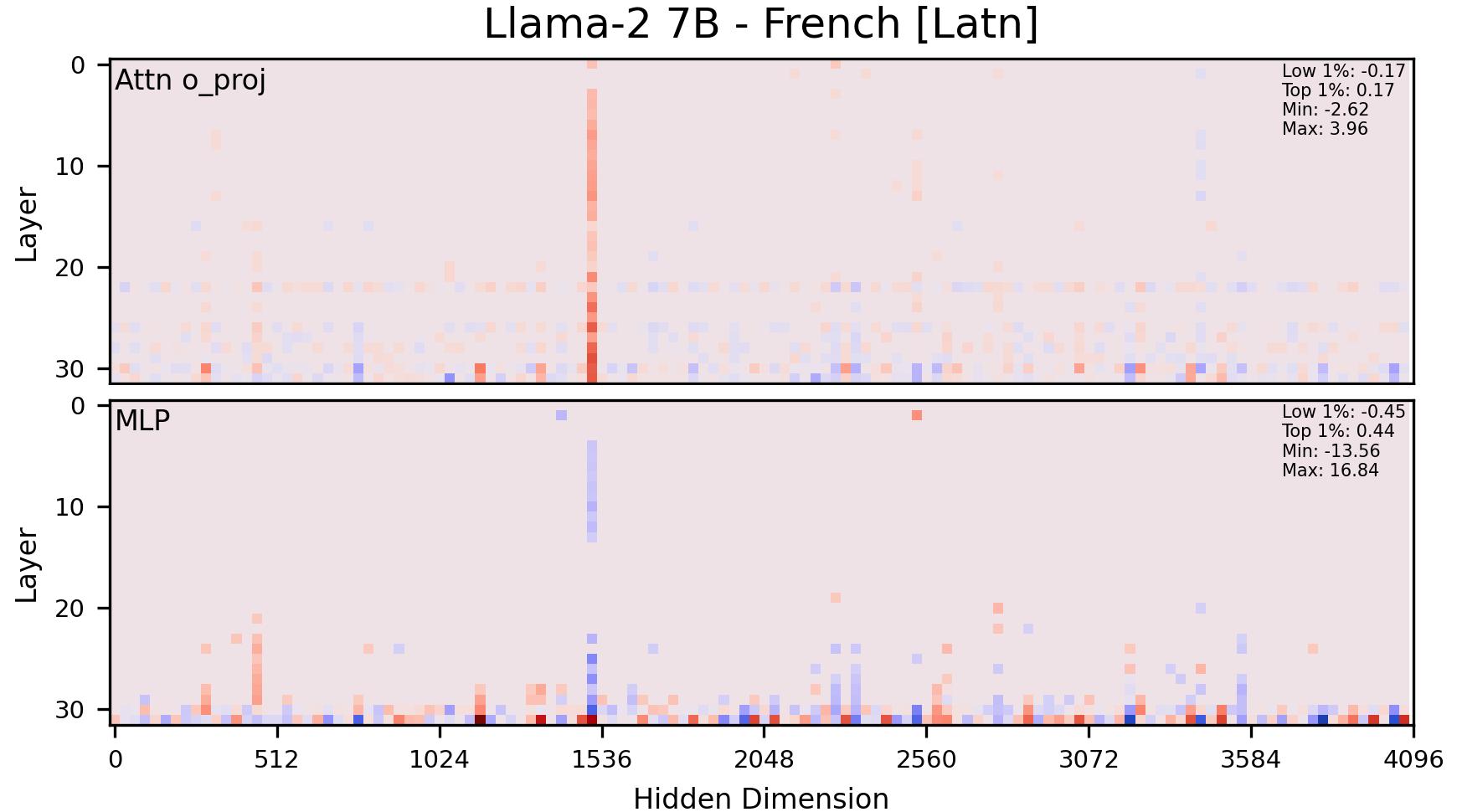} %
    \end{minipage}
    \vskip -0.3in
\end{figure}

\begin{figure}[H]
    \centering
    \begin{minipage}{0.33\textwidth}
        \centering
        \includegraphics[width=1.0\textwidth]{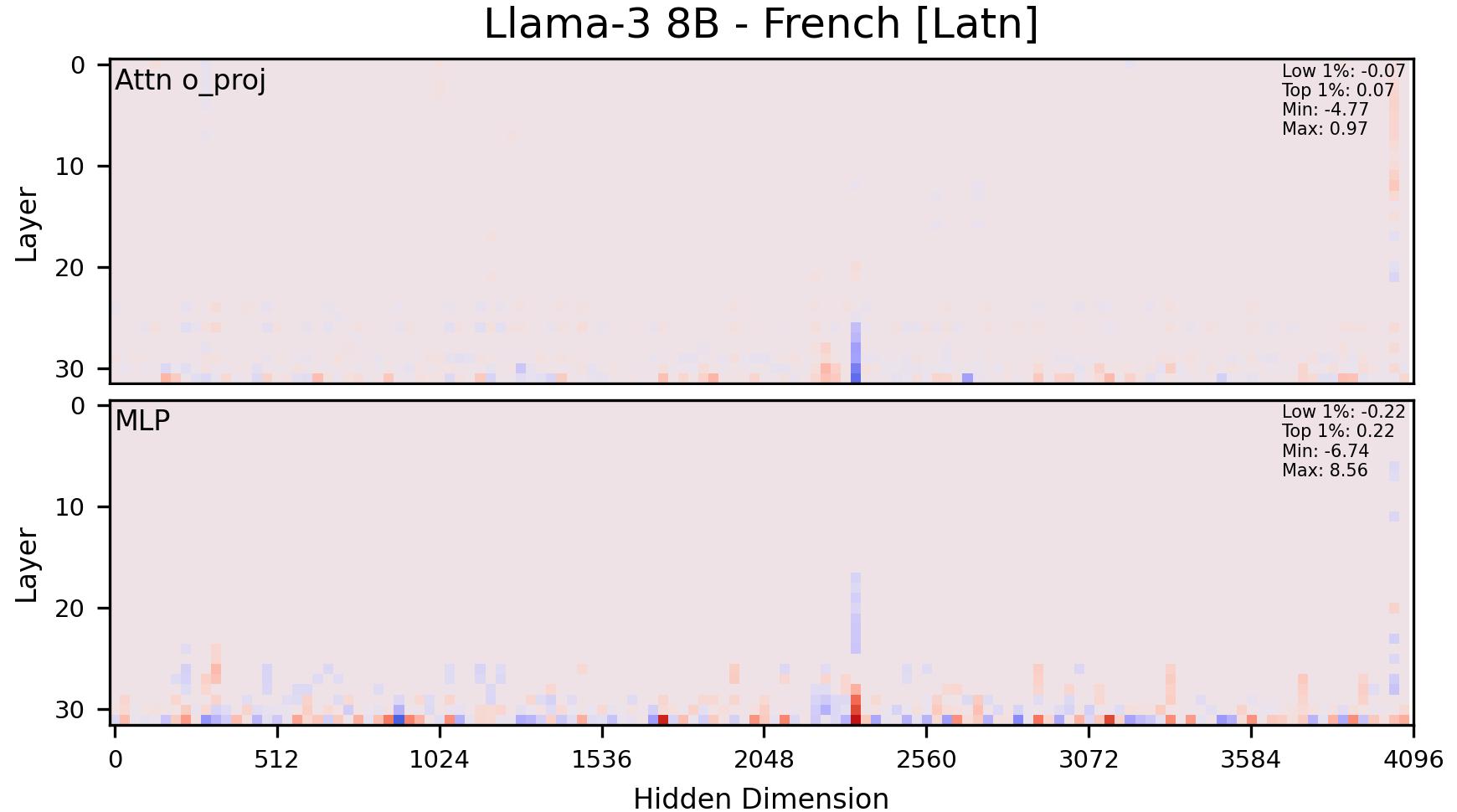} %
    \end{minipage}\hfill
    \begin{minipage}{0.33\textwidth}
        \centering
        \includegraphics[width=1.0\textwidth]{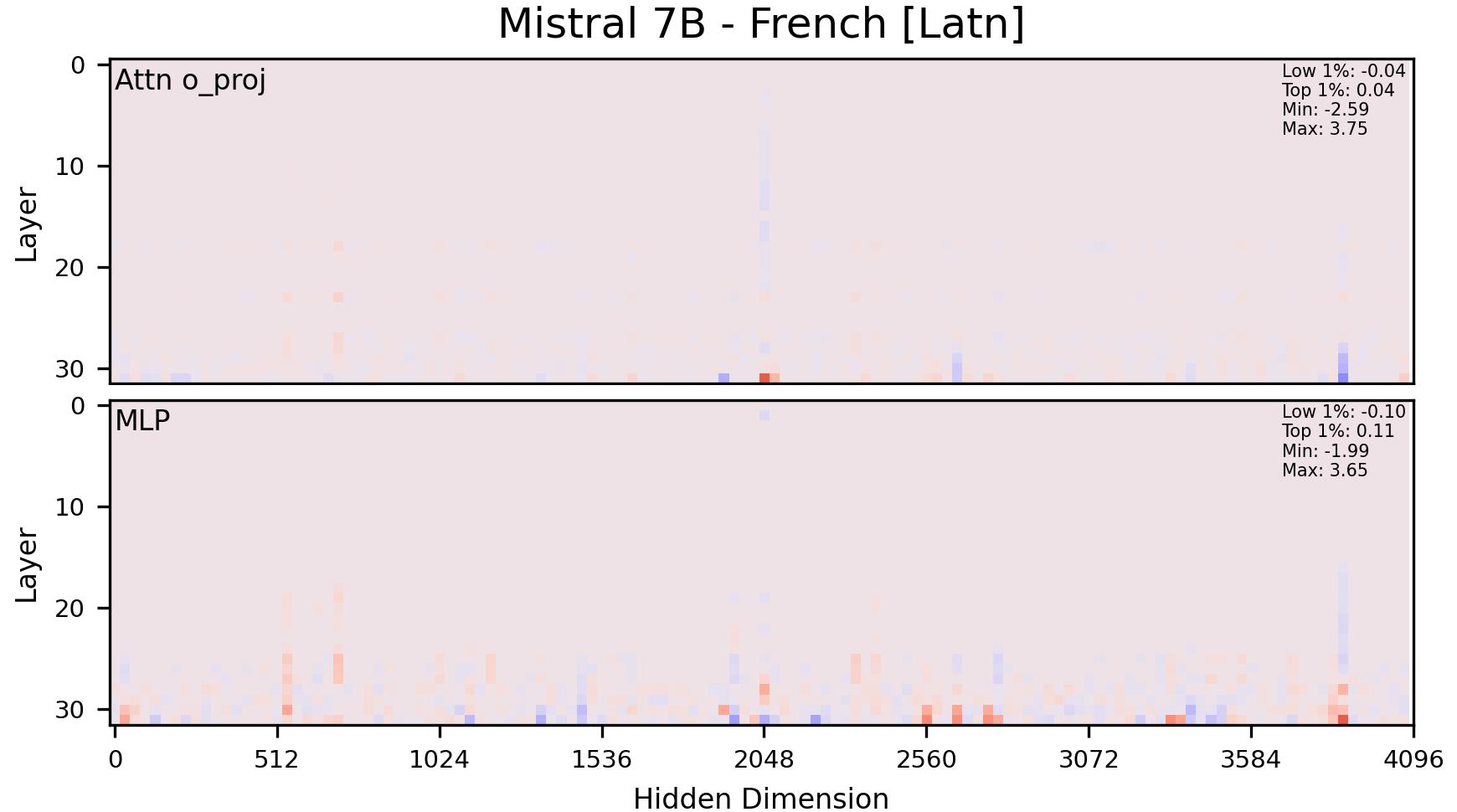} %
    \end{minipage}
    \begin{minipage}{0.33\textwidth}
        \centering
        \includegraphics[width=1.0\textwidth]{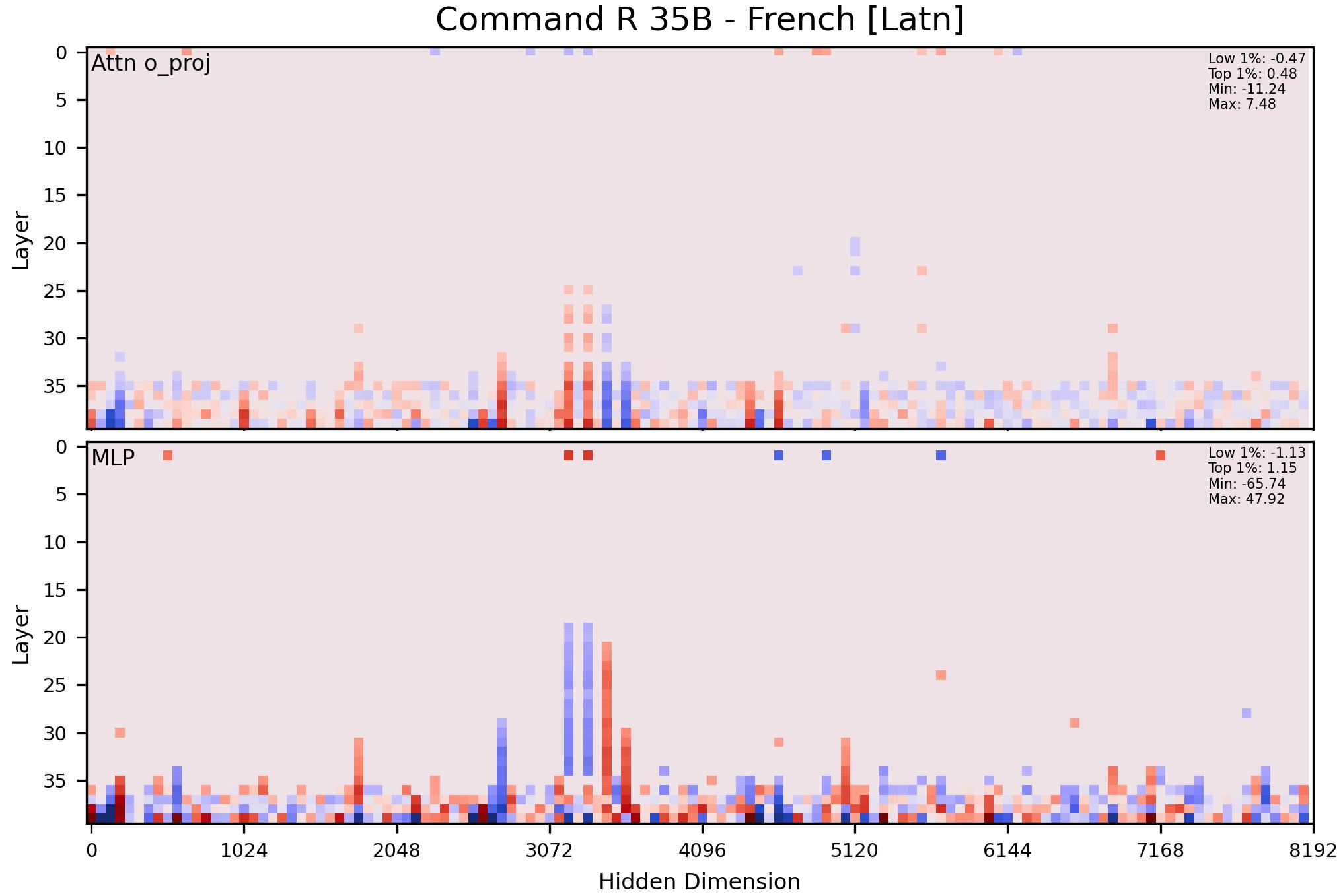} %
    \end{minipage}
\end{figure}

\begin{figure}[H]
    \centering
    \begin{minipage}{0.33\textwidth}
        \centering
        \includegraphics[width=1.0\textwidth]{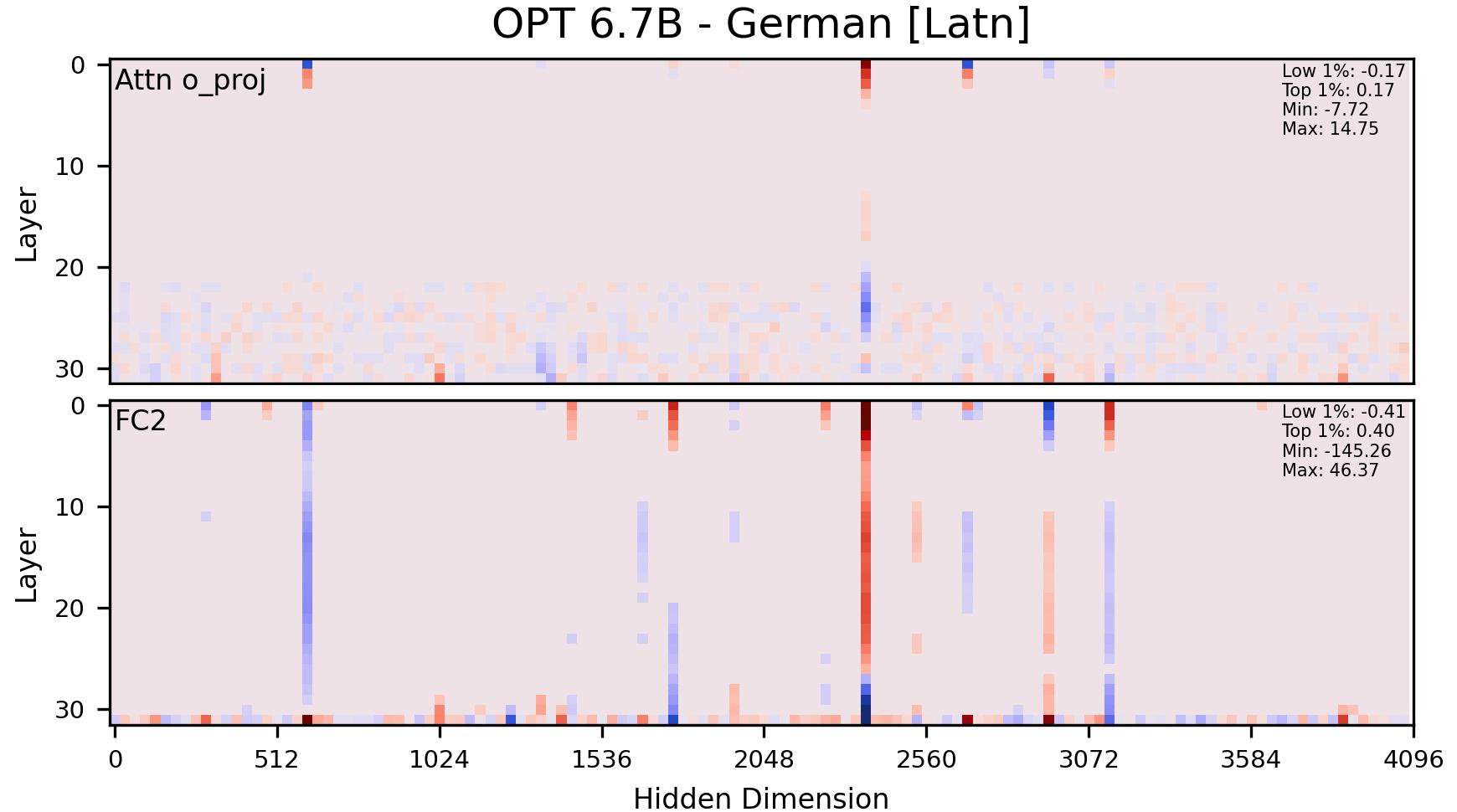} %
    \end{minipage}\hfill
    \begin{minipage}{0.33\textwidth}
        \centering
        \includegraphics[width=1.0\textwidth]{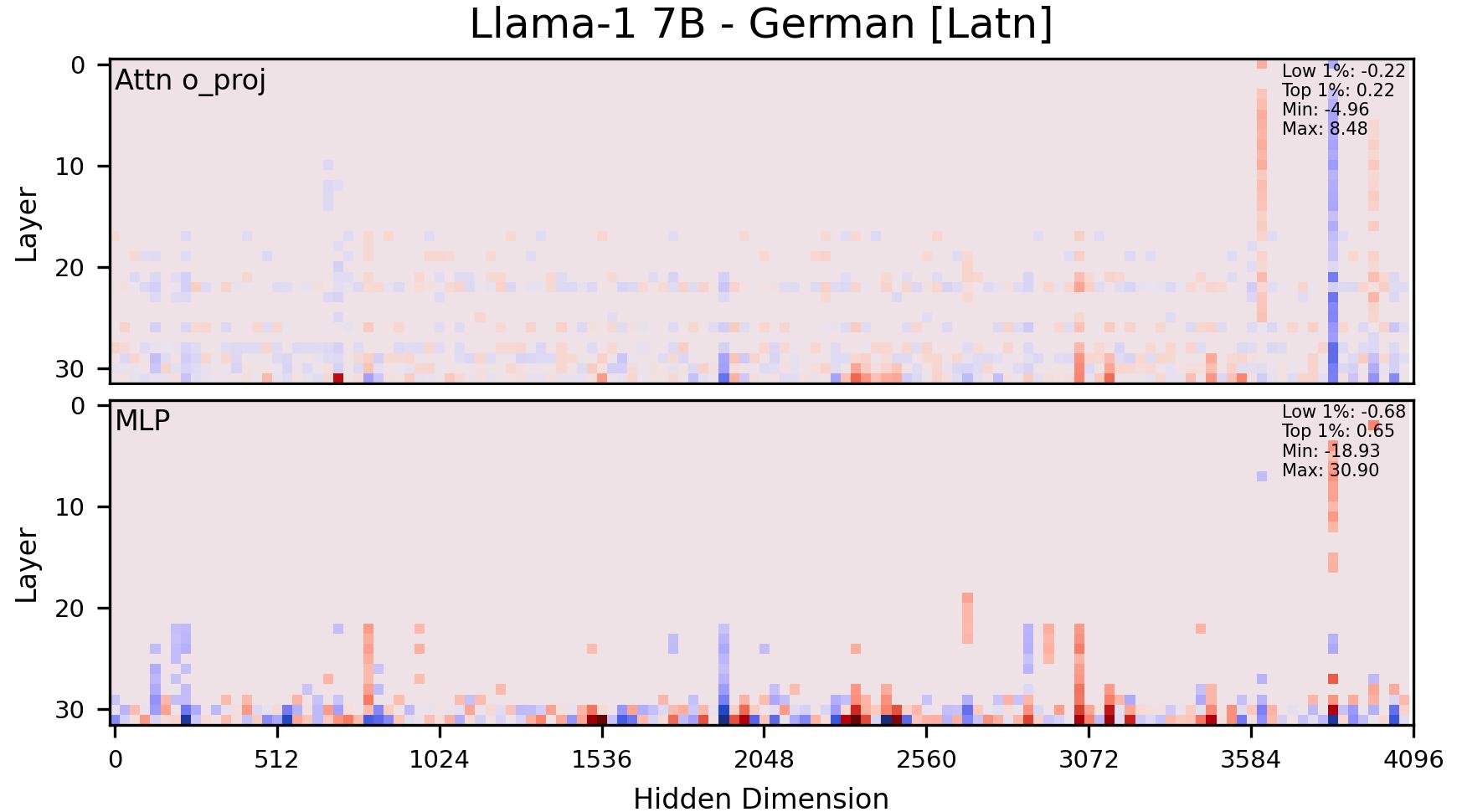} %
    \end{minipage}
    \begin{minipage}{0.33\textwidth}
        \centering
        \includegraphics[width=1.0\textwidth]{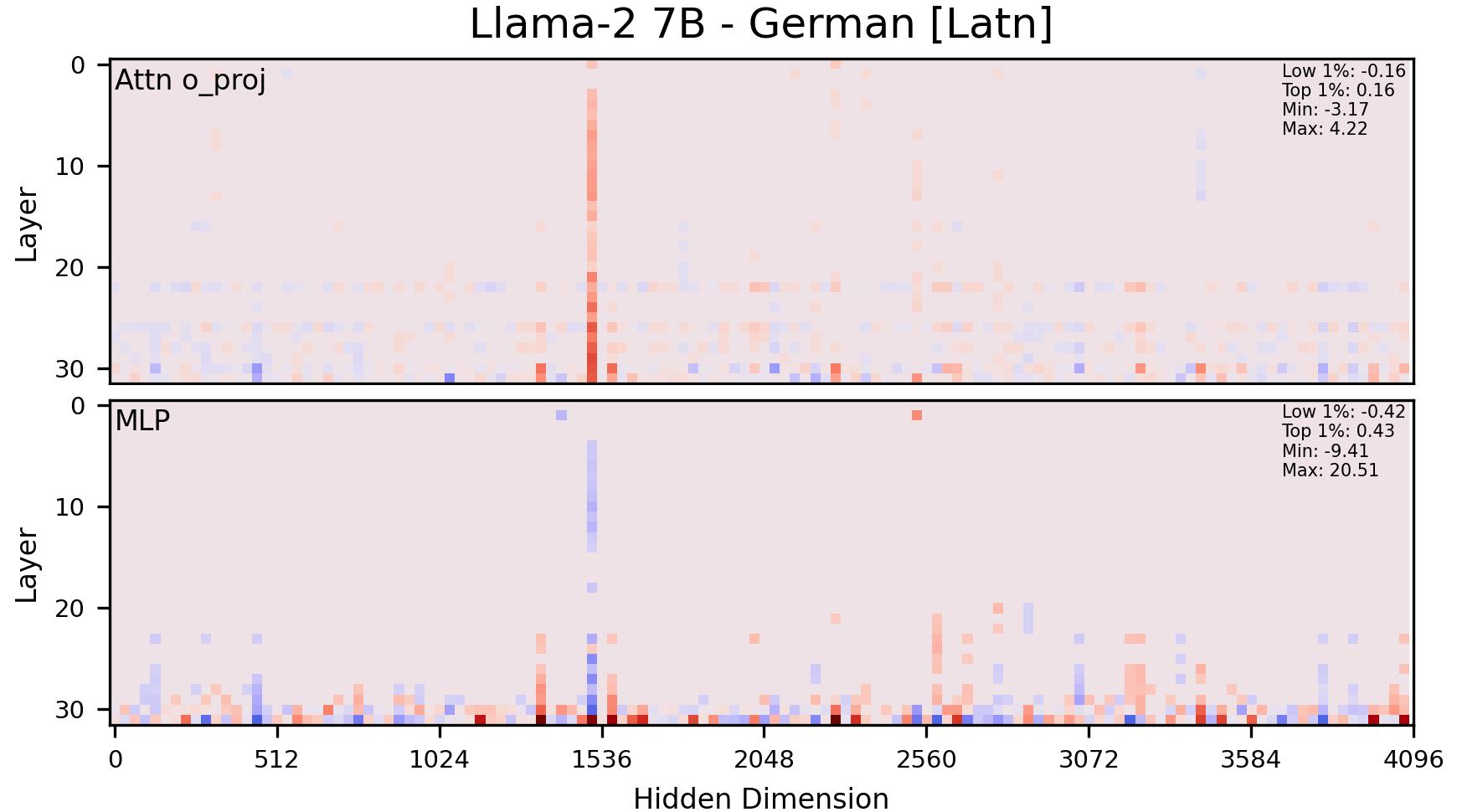} %
    \end{minipage}
    \vskip -0.3in
\end{figure}

\begin{figure}[H]
    \centering
    \begin{minipage}{0.33\textwidth}
        \centering
        \includegraphics[width=1.0\textwidth]{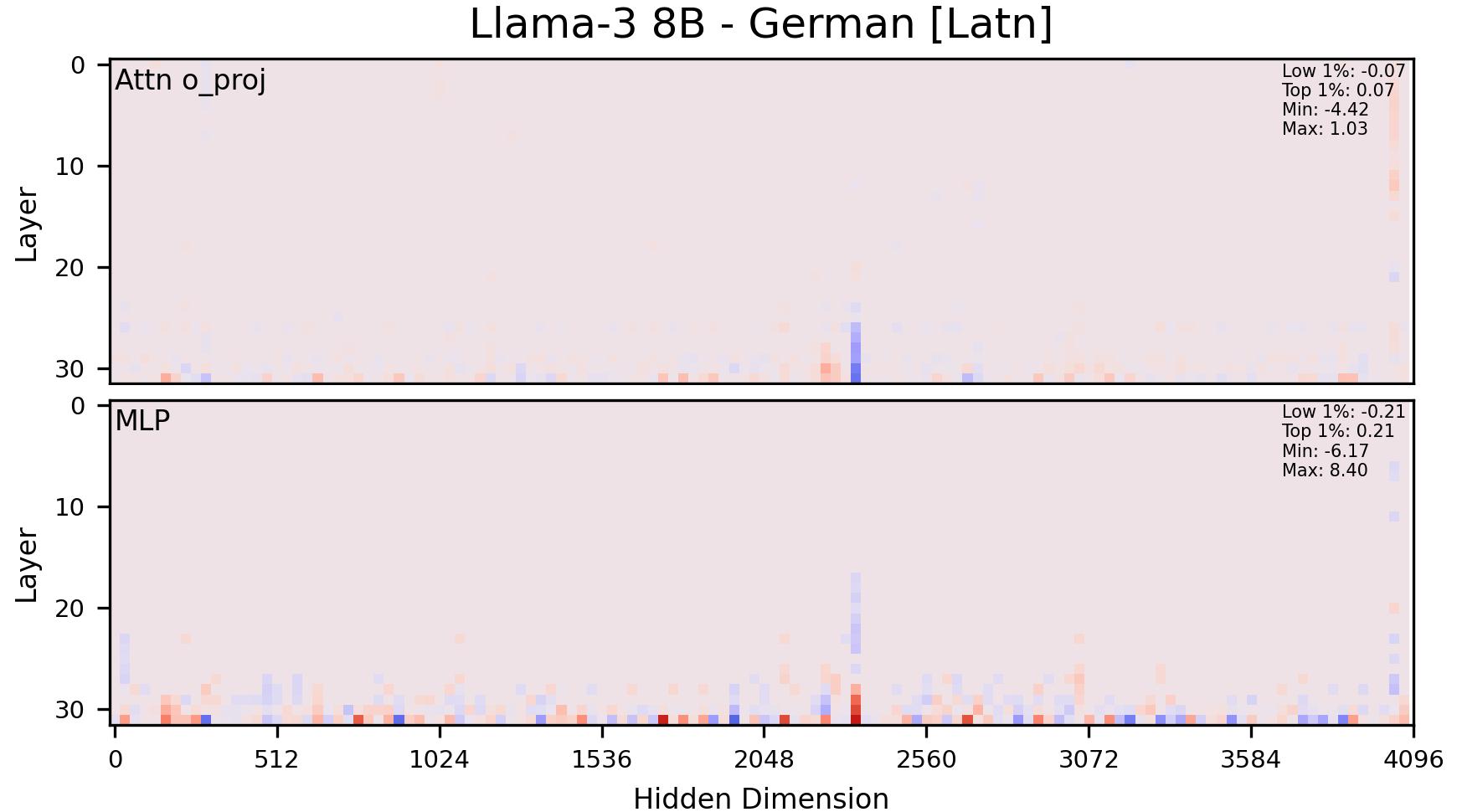} %
    \end{minipage}\hfill
    \begin{minipage}{0.33\textwidth}
        \centering
        \includegraphics[width=1.0\textwidth]{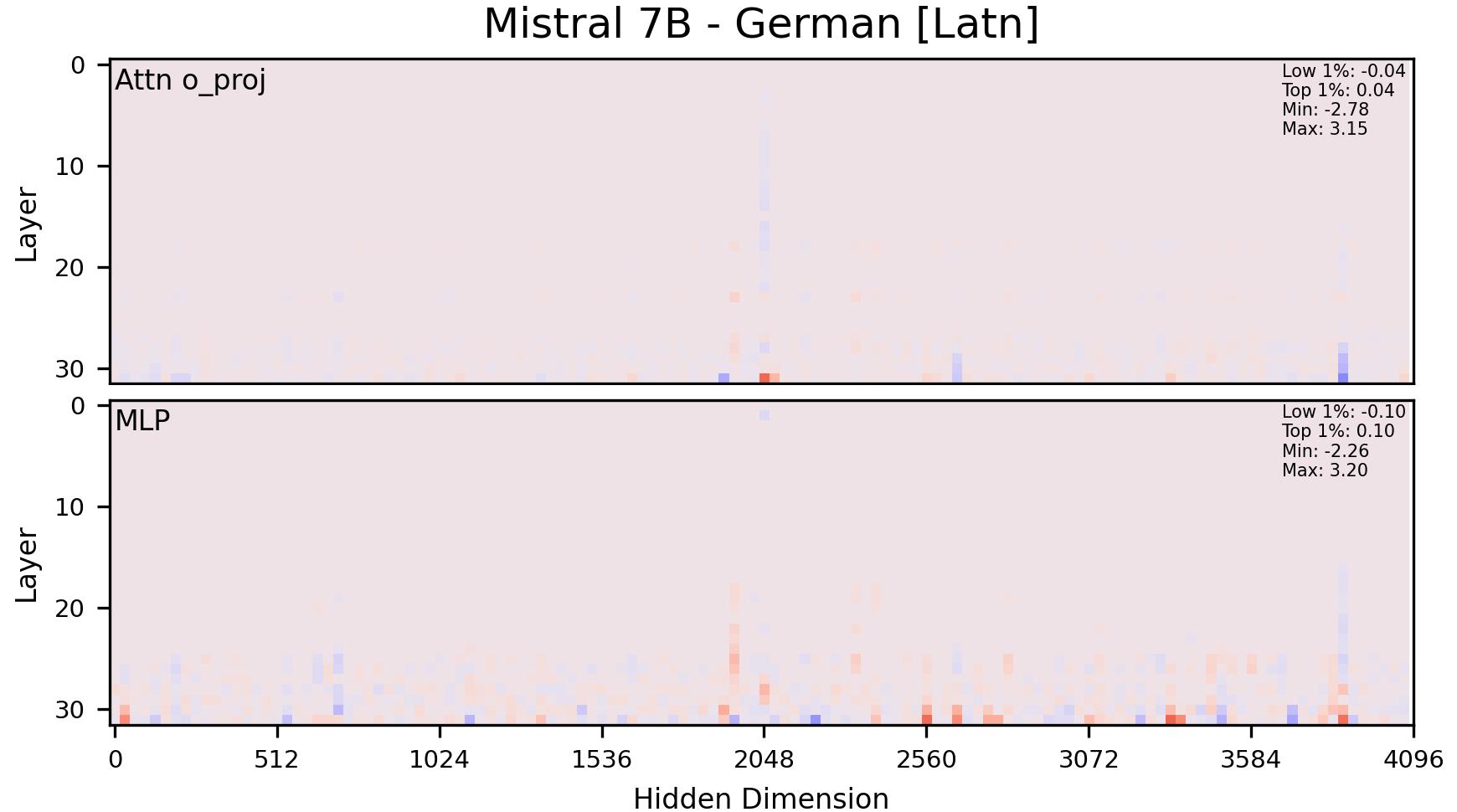} %
    \end{minipage}
    \begin{minipage}{0.33\textwidth}
        \centering
        \includegraphics[width=1.0\textwidth]{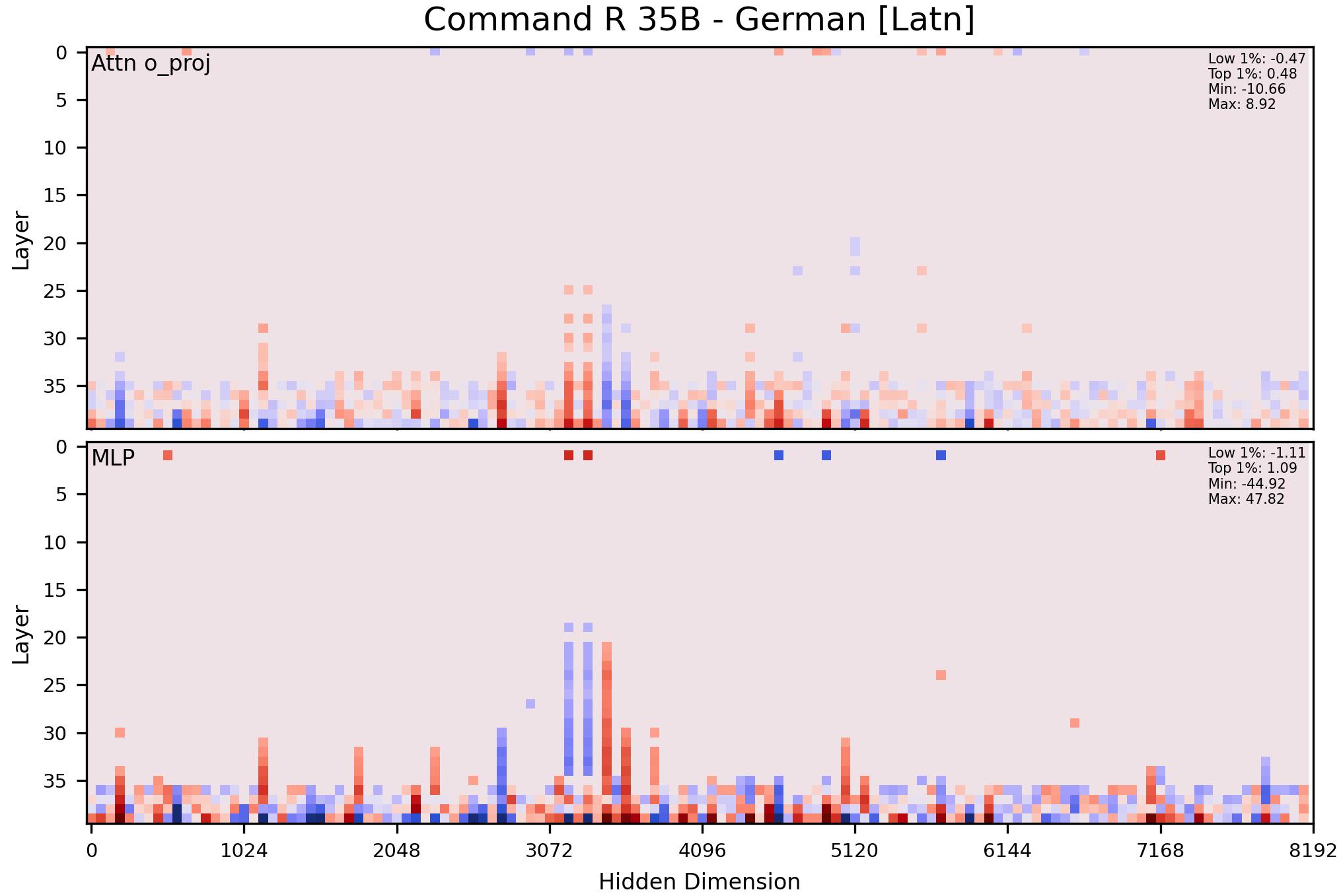} %
    \end{minipage}
\end{figure}

\begin{figure}[H]
    \centering
    \begin{minipage}{0.33\textwidth}
        \centering
        \includegraphics[width=1.0\textwidth]{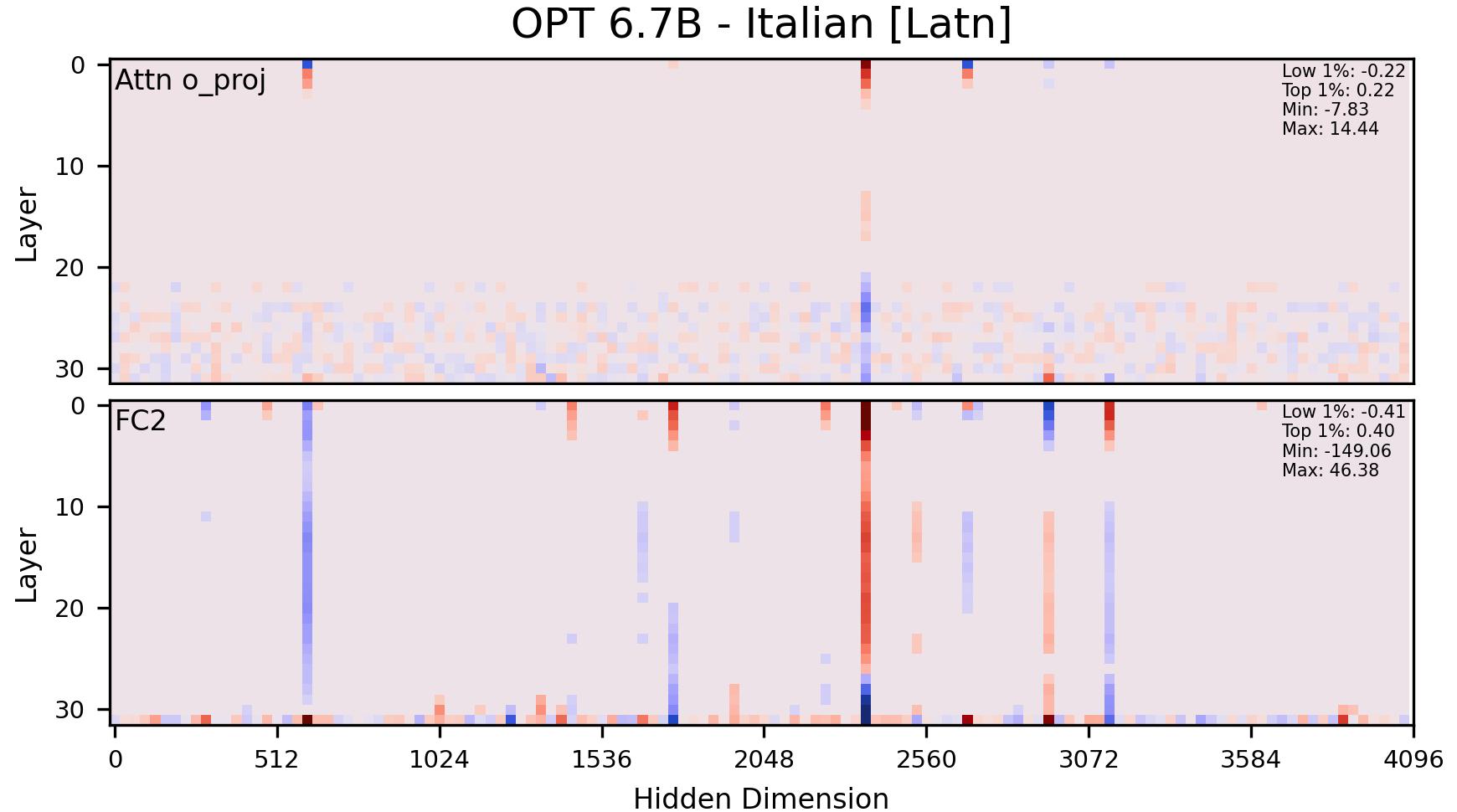} %
    \end{minipage}\hfill
    \begin{minipage}{0.33\textwidth}
        \centering
        \includegraphics[width=1.0\textwidth]{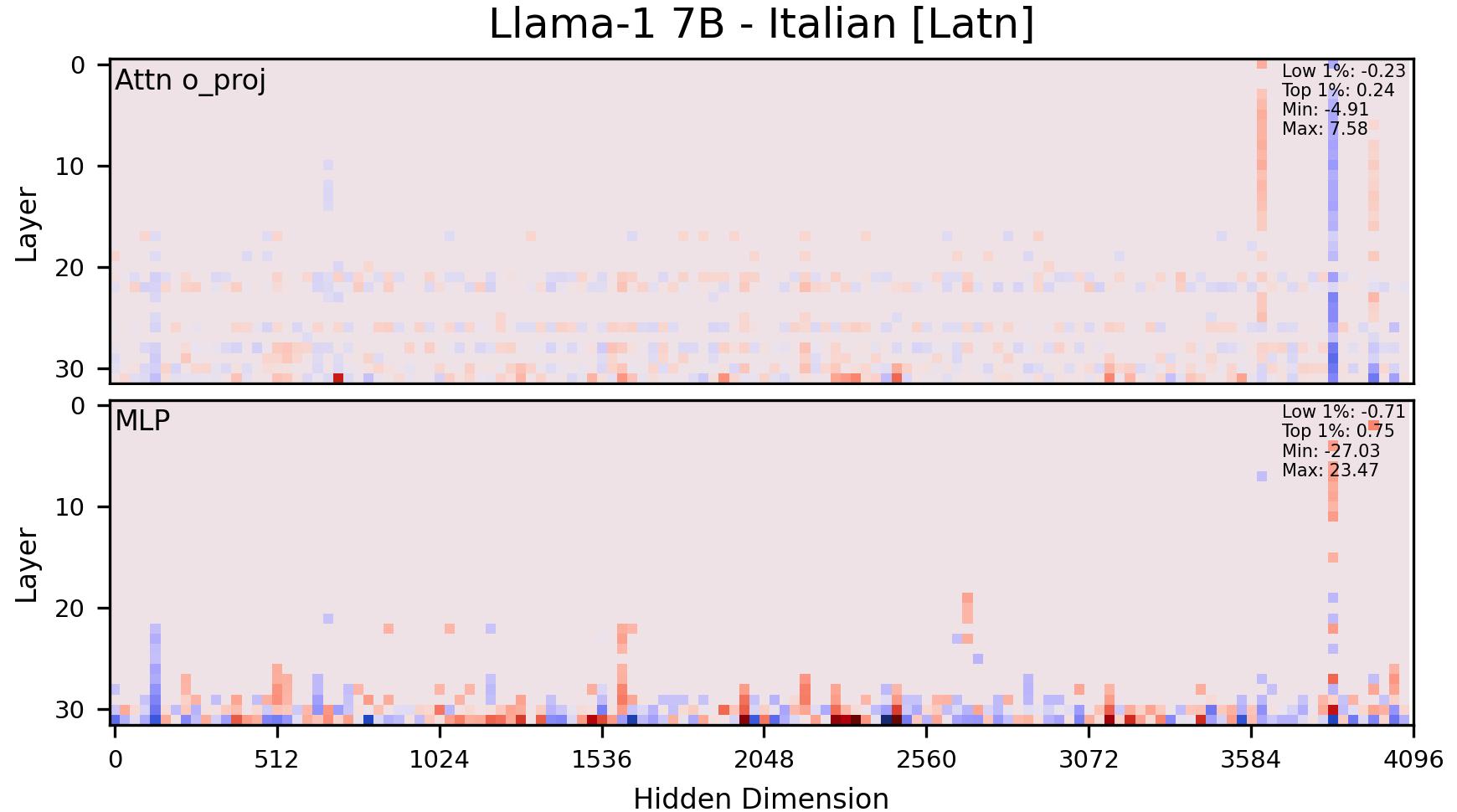} %
    \end{minipage}
    \begin{minipage}{0.33\textwidth}
        \centering
        \includegraphics[width=1.0\textwidth]{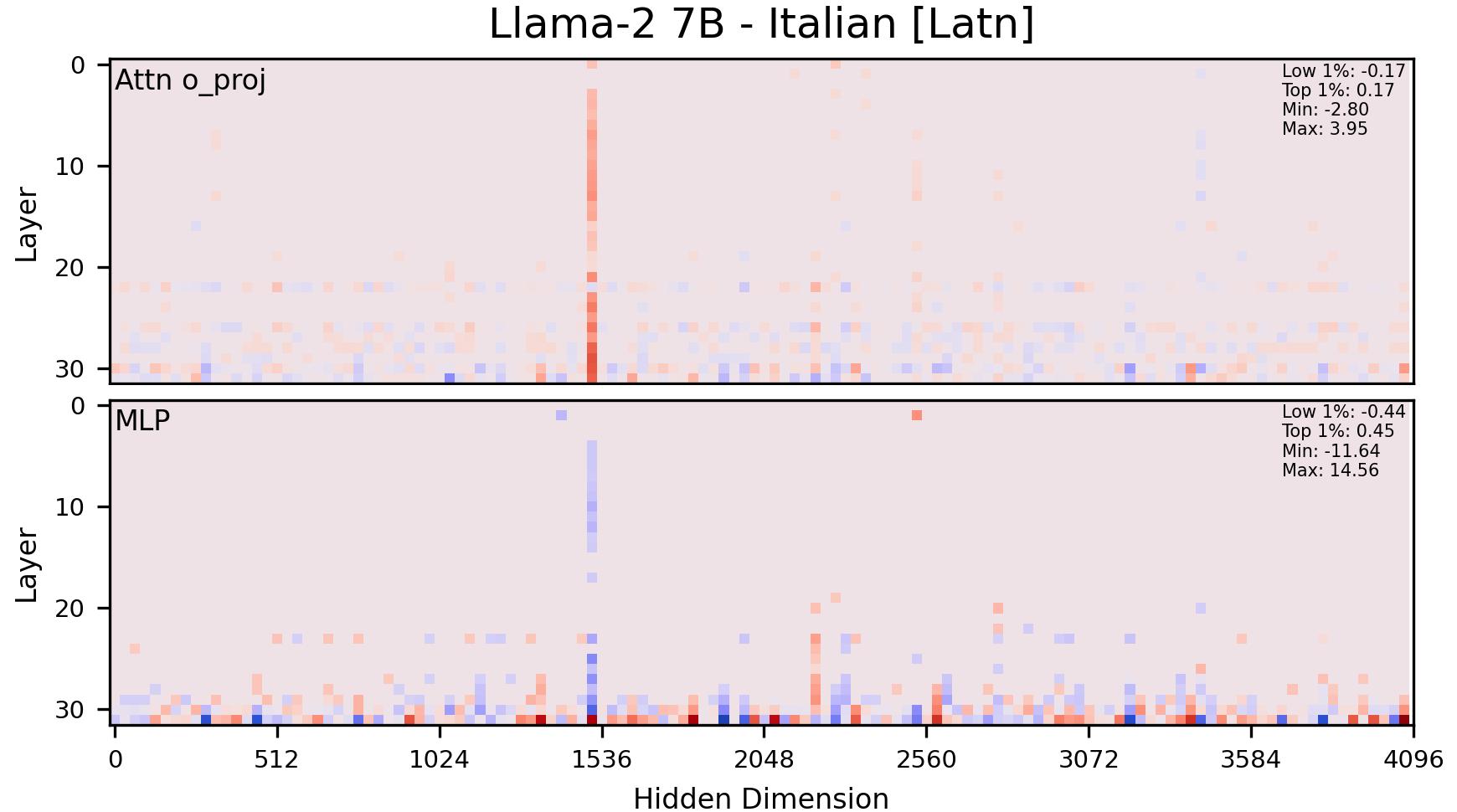} %
    \end{minipage}
    \vskip -0.3in
\end{figure}

\begin{figure}[H]
    \centering
    \begin{minipage}{0.33\textwidth}
        \centering
        \includegraphics[width=1.0\textwidth]{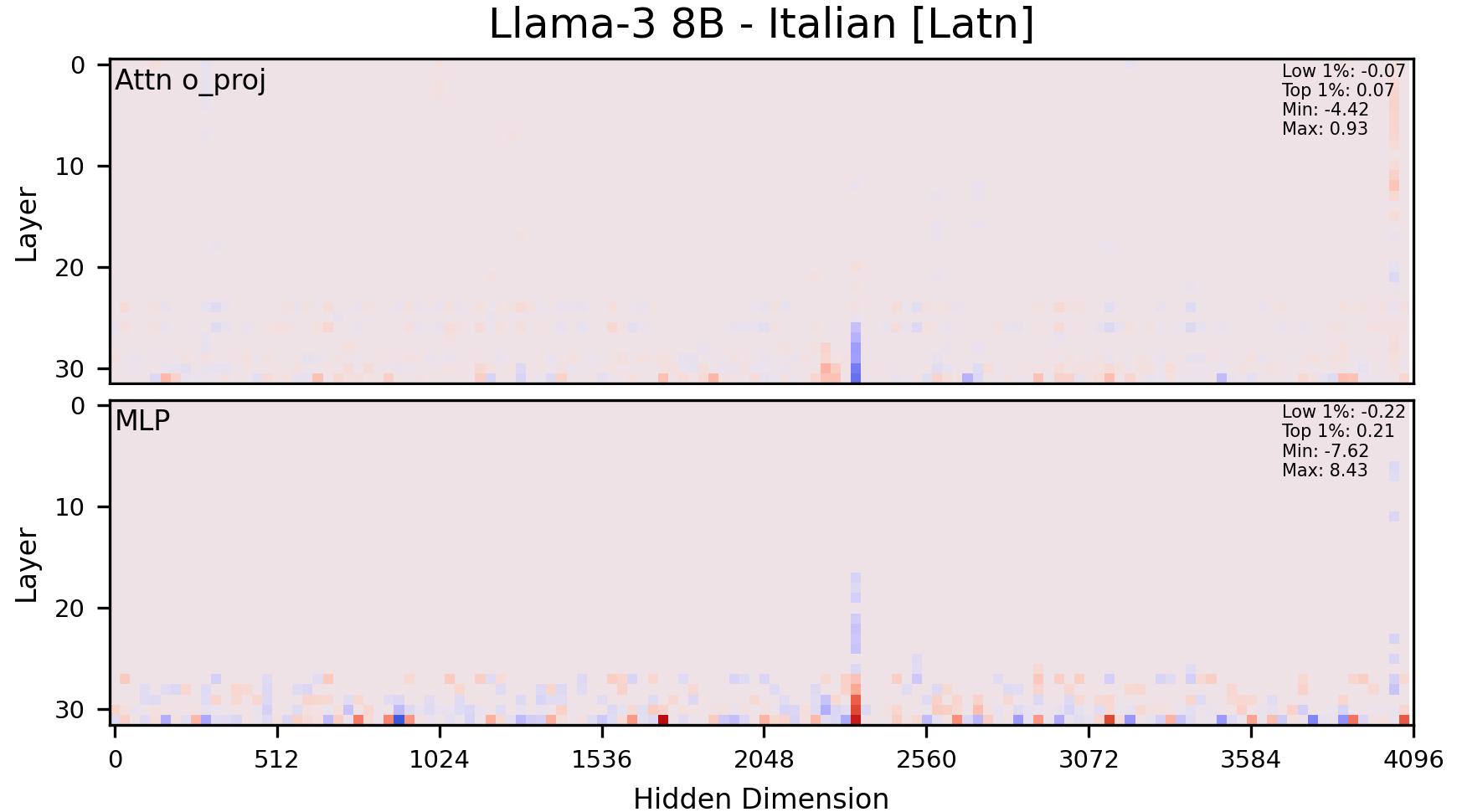} %
    \end{minipage}\hfill
    \begin{minipage}{0.33\textwidth}
        \centering
        \includegraphics[width=1.0\textwidth]{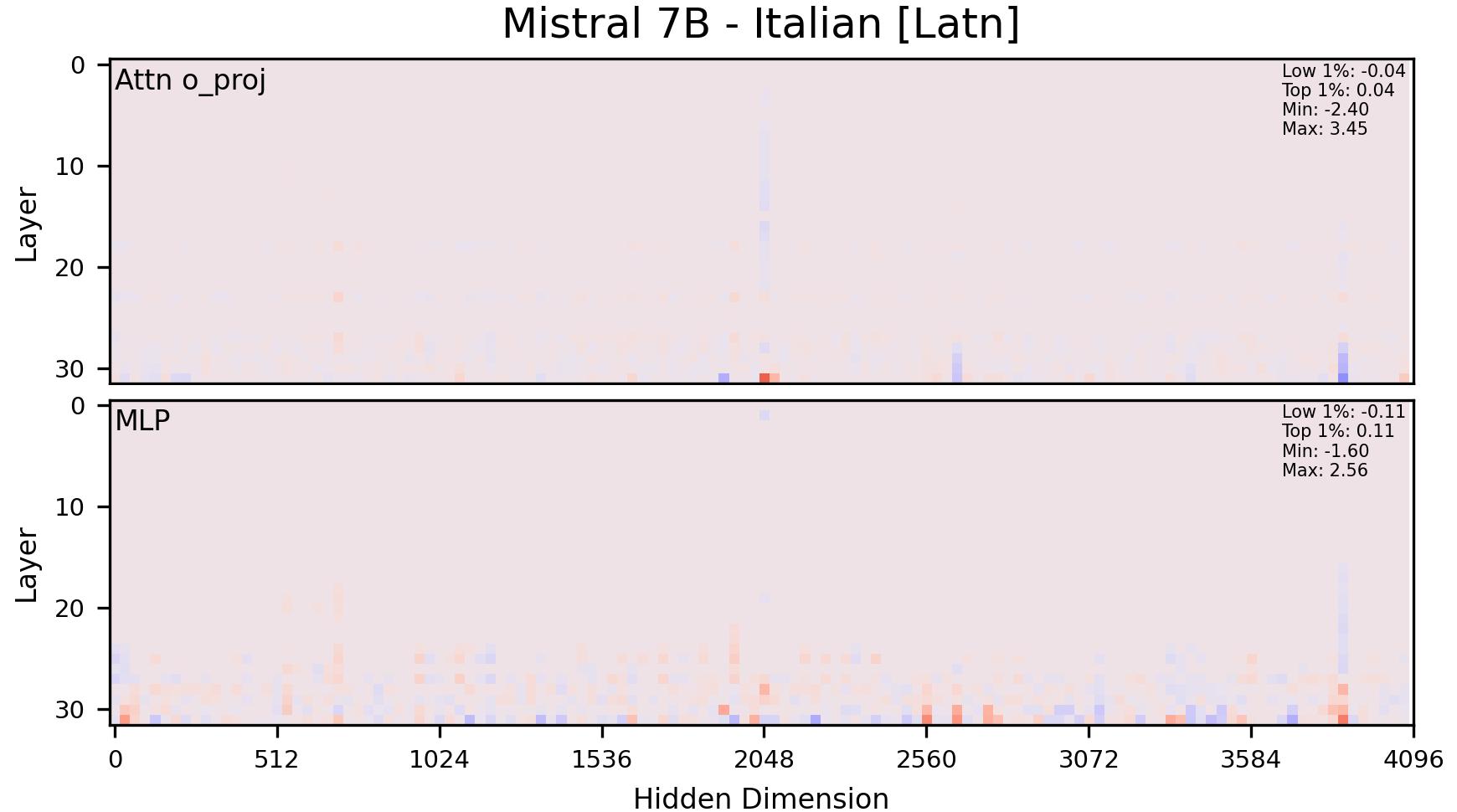} %
    \end{minipage}
    \begin{minipage}{0.33\textwidth}
        \centering
        \includegraphics[width=1.0\textwidth]{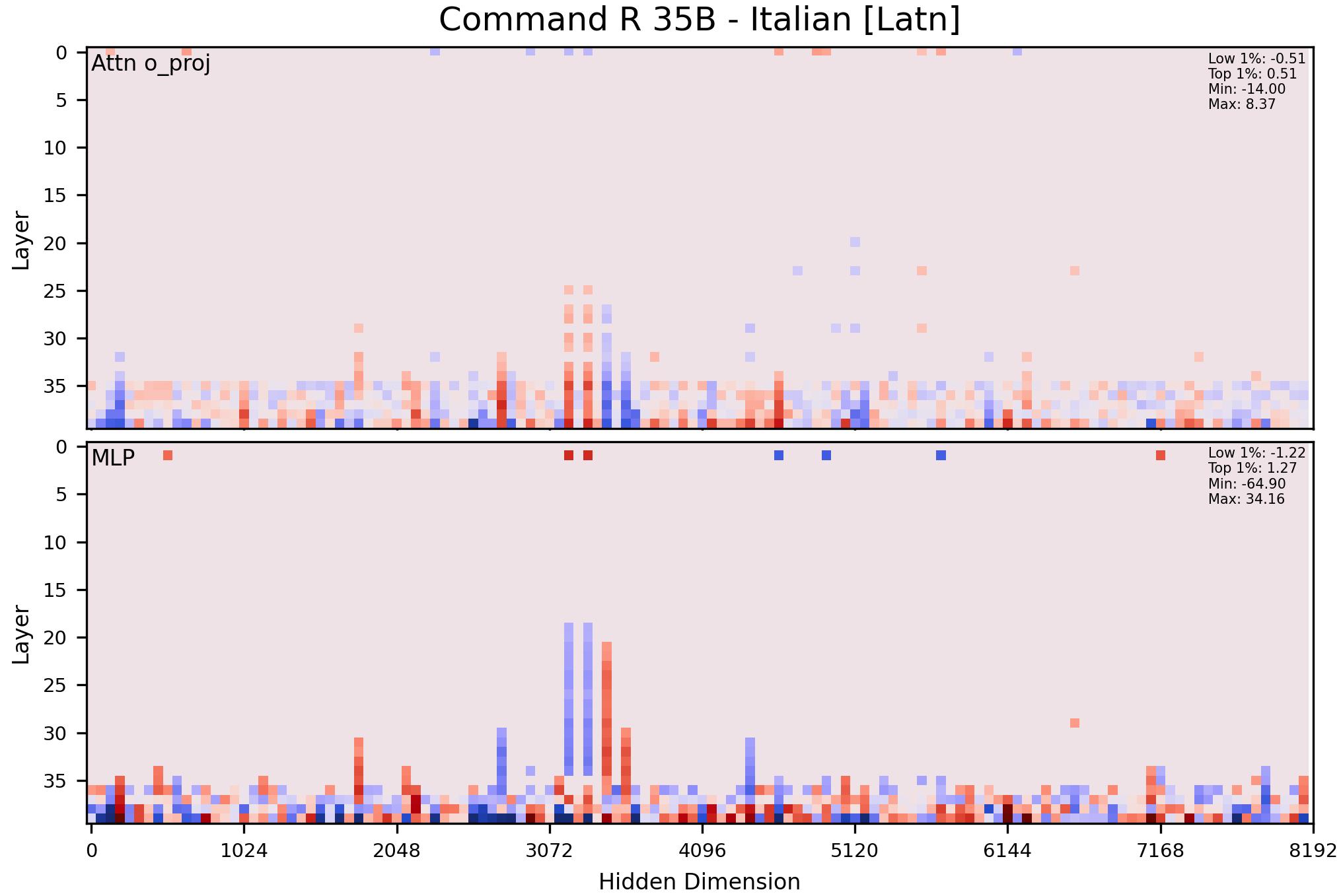} %
    \end{minipage}
\end{figure}

\begin{figure}[H]
    \centering
    \begin{minipage}{0.33\textwidth}
        \centering
        \includegraphics[width=1.0\textwidth]{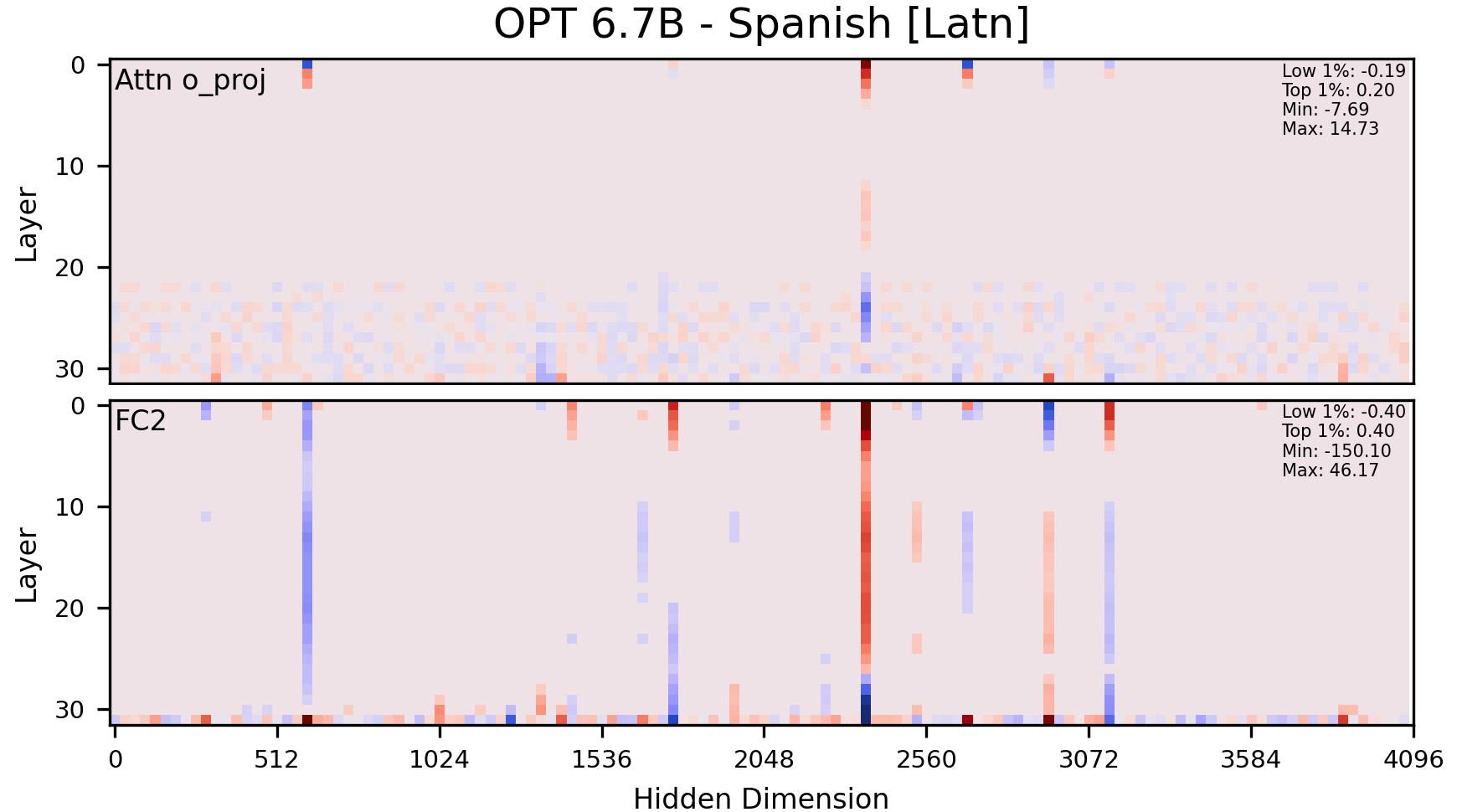} %
    \end{minipage}\hfill
    \begin{minipage}{0.33\textwidth}
        \centering
        \includegraphics[width=1.0\textwidth]{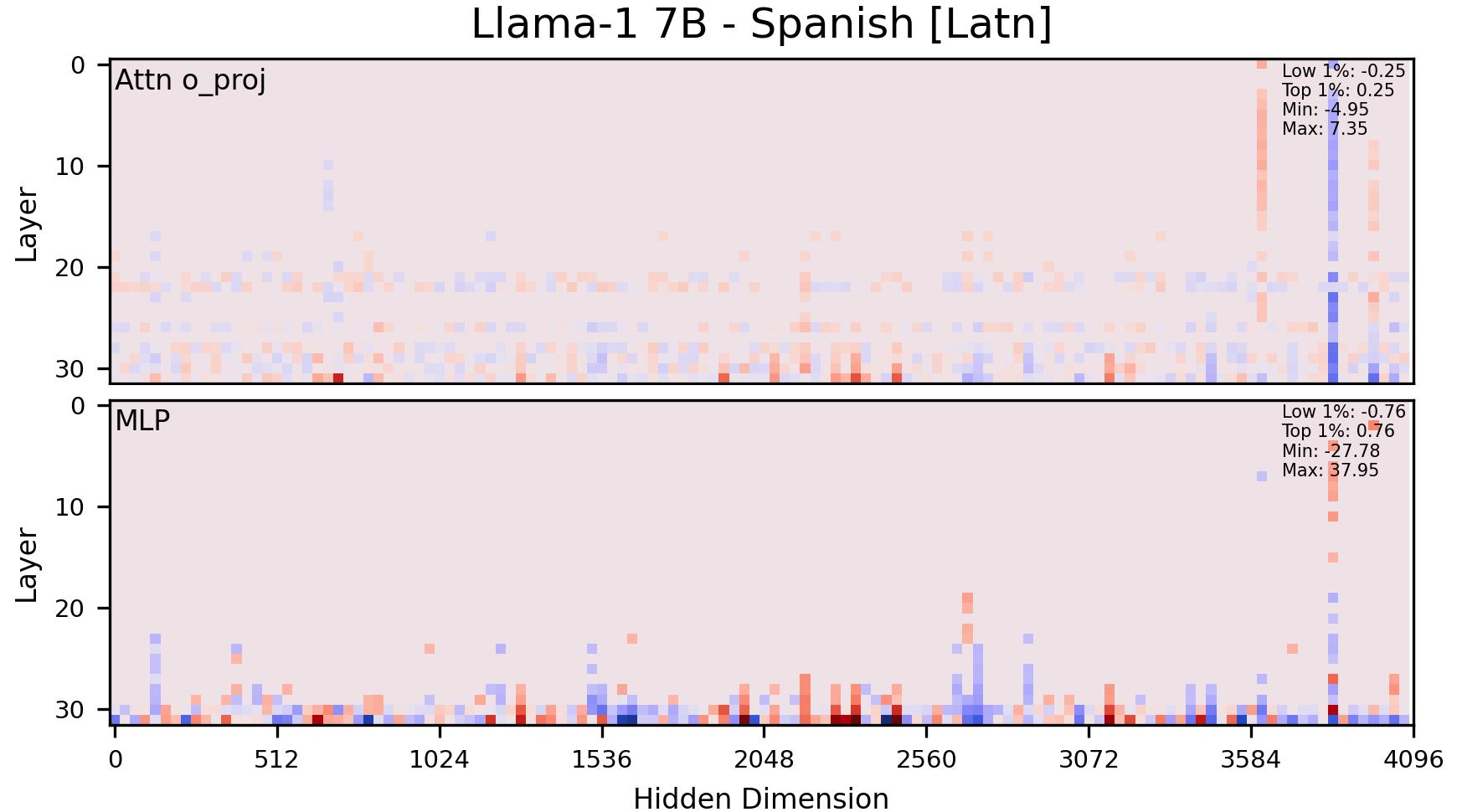} %
    \end{minipage}
    \begin{minipage}{0.33\textwidth}
        \centering
        \includegraphics[width=1.0\textwidth]{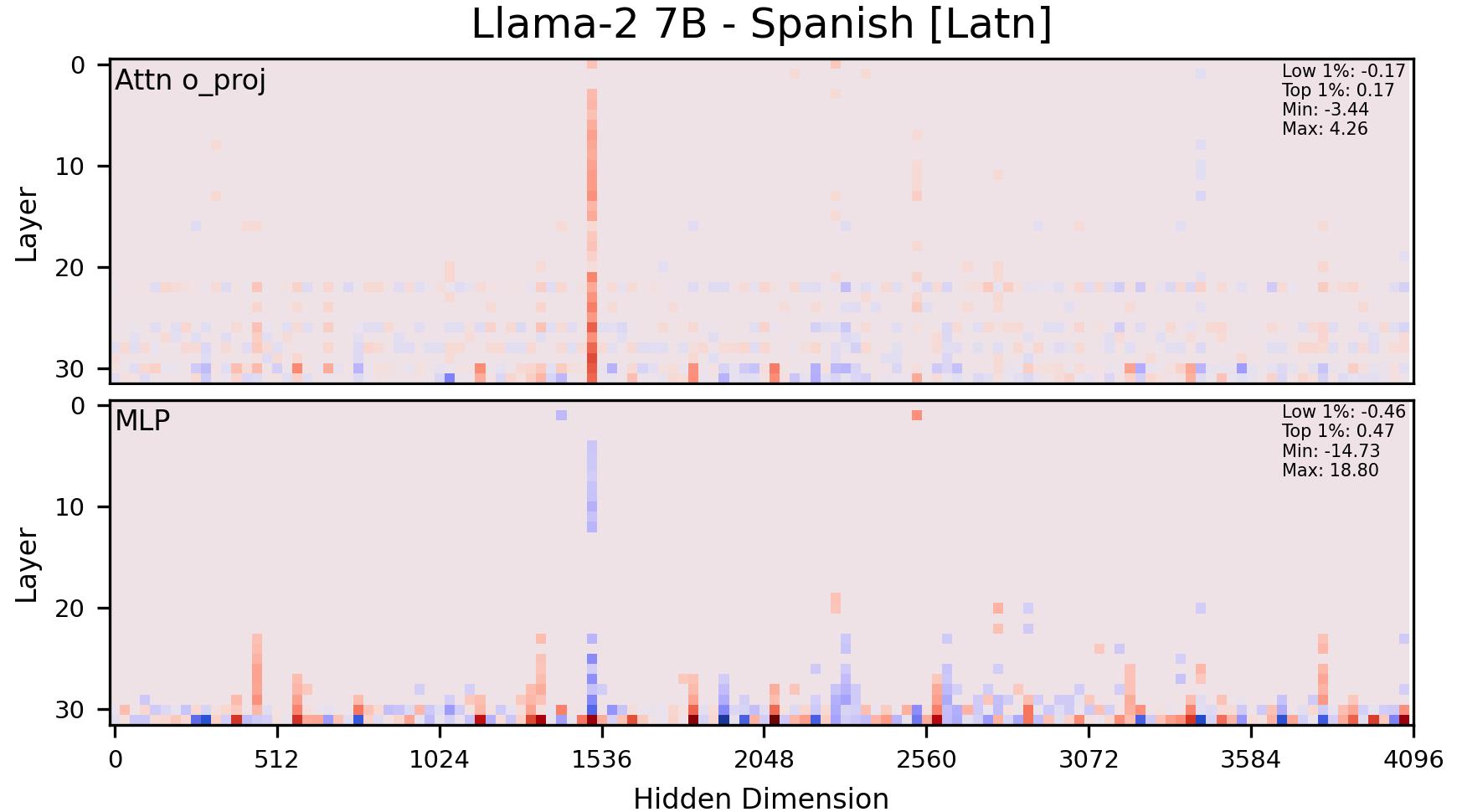} %
    \end{minipage}
    \vskip -0.3in
\end{figure}

\begin{figure}[H]
    \centering
    \begin{minipage}{0.33\textwidth}
        \centering
        \includegraphics[width=1.0\textwidth]{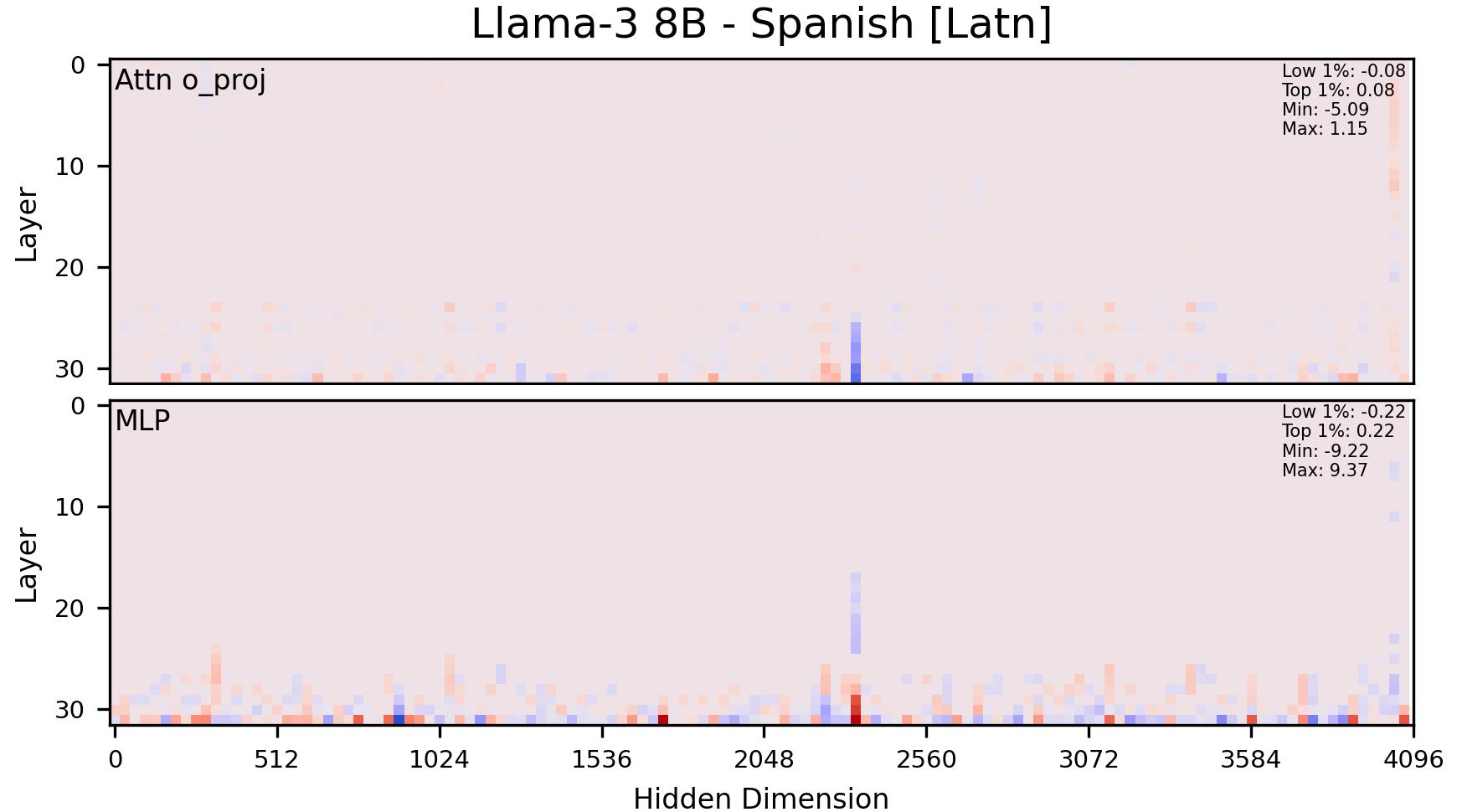} %
    \end{minipage}\hfill
    \begin{minipage}{0.33\textwidth}
        \centering
        \includegraphics[width=1.0\textwidth]{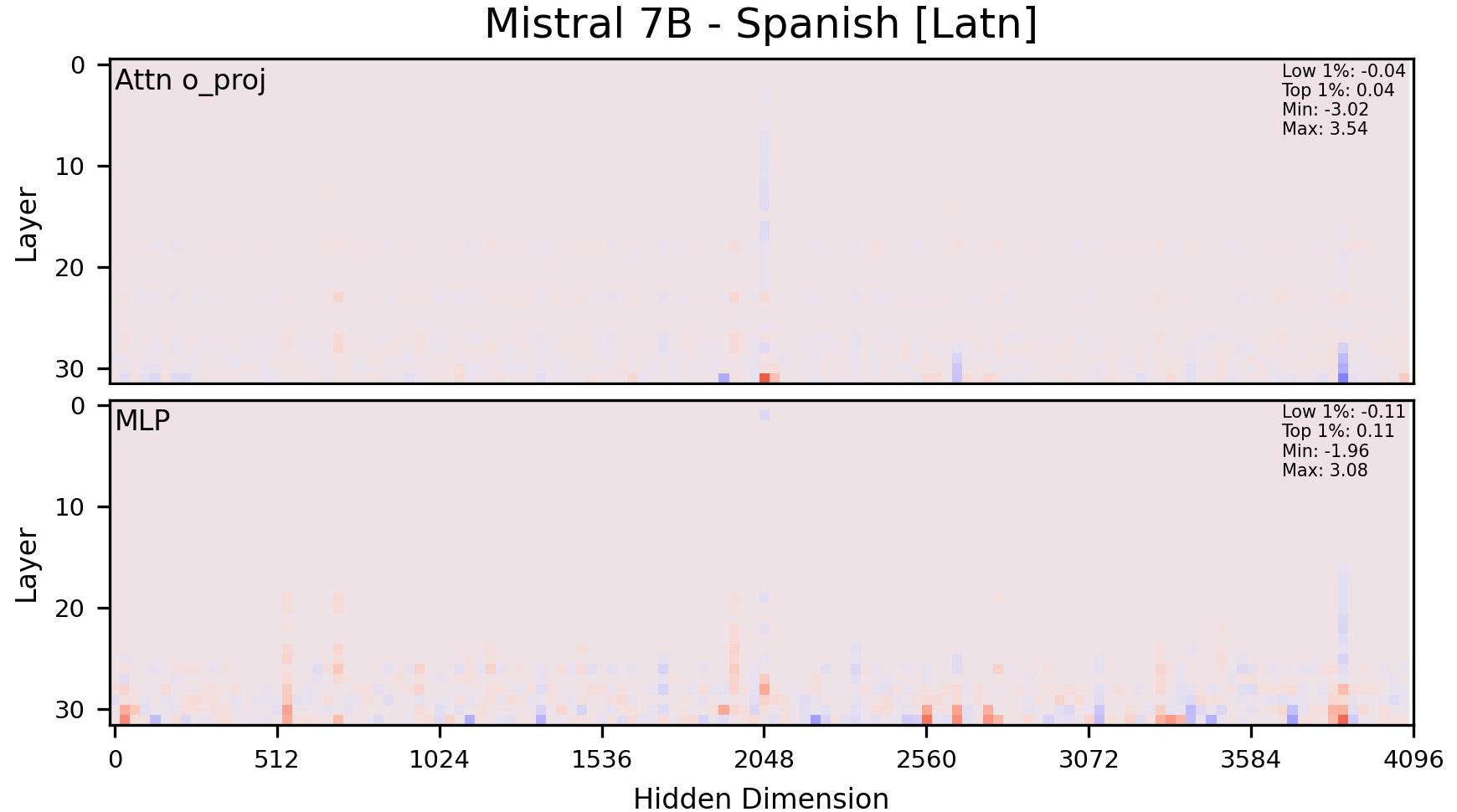} %
    \end{minipage}
    \begin{minipage}{0.33\textwidth}
        \centering
        \includegraphics[width=1.0\textwidth]{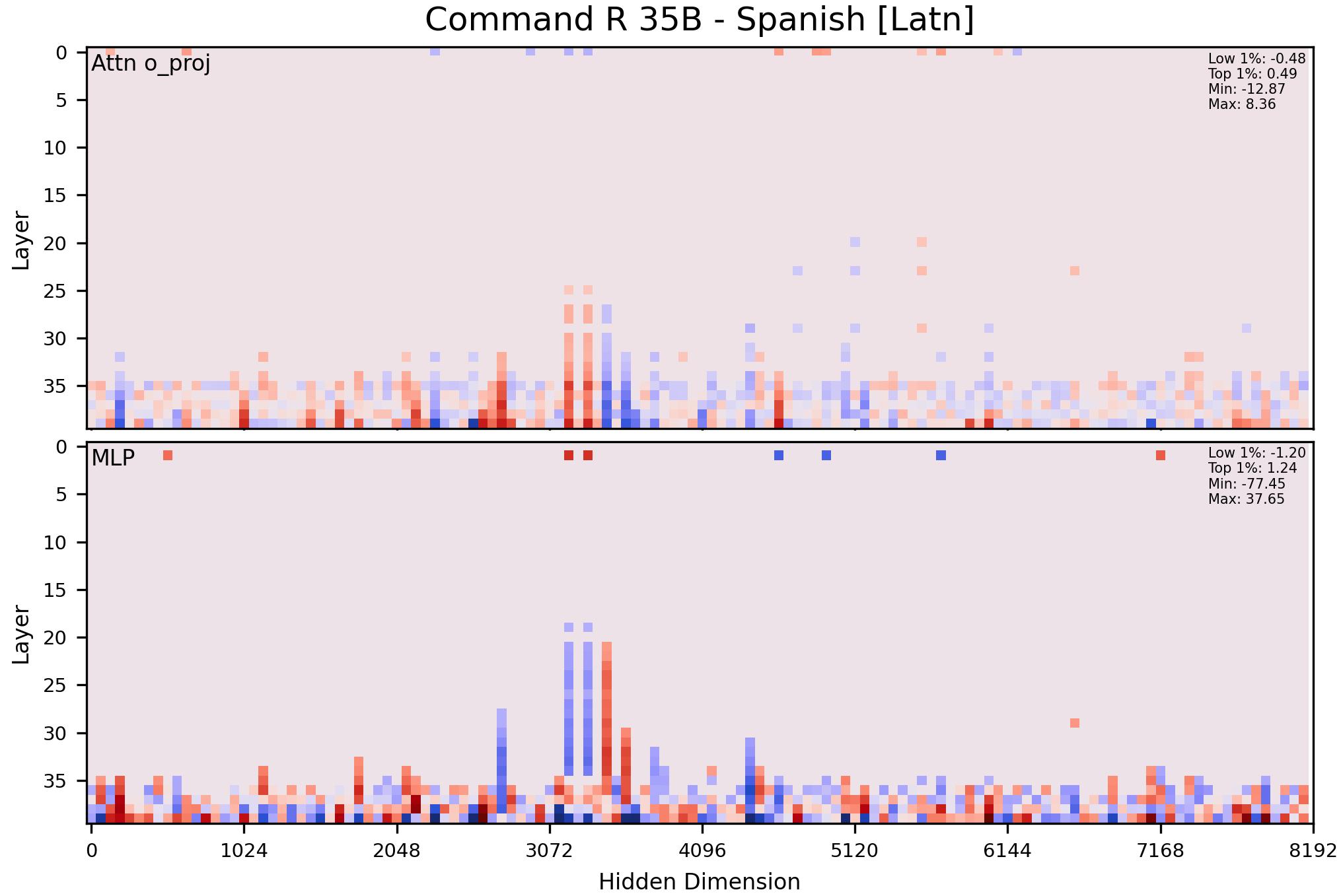} %
    \end{minipage}
\end{figure}

\begin{figure}[H]
    \centering
    \begin{minipage}{0.33\textwidth}
        \centering
        \includegraphics[width=1.0\textwidth]{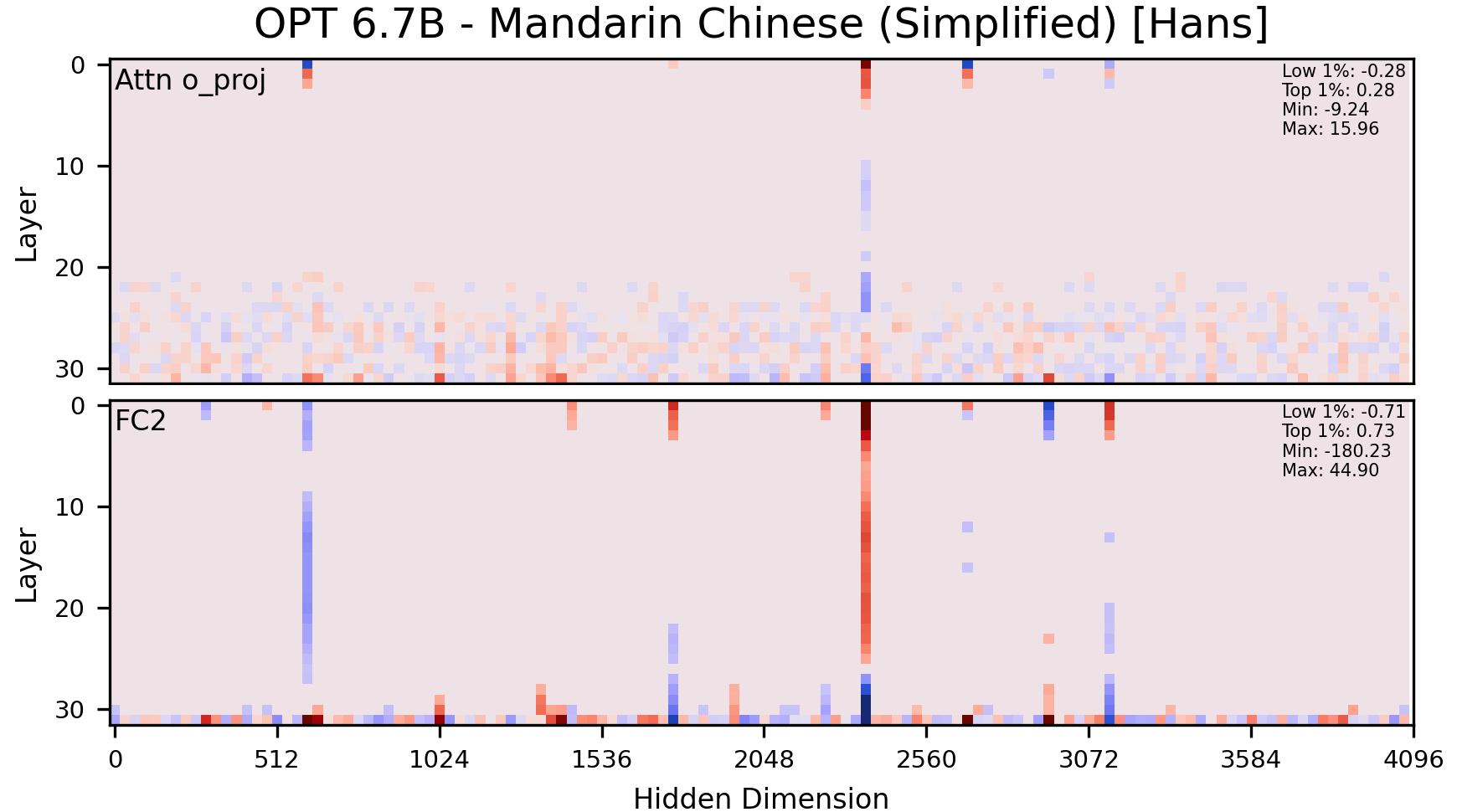} %
    \end{minipage}\hfill
    \begin{minipage}{0.33\textwidth}
        \centering
        \includegraphics[width=1.0\textwidth]{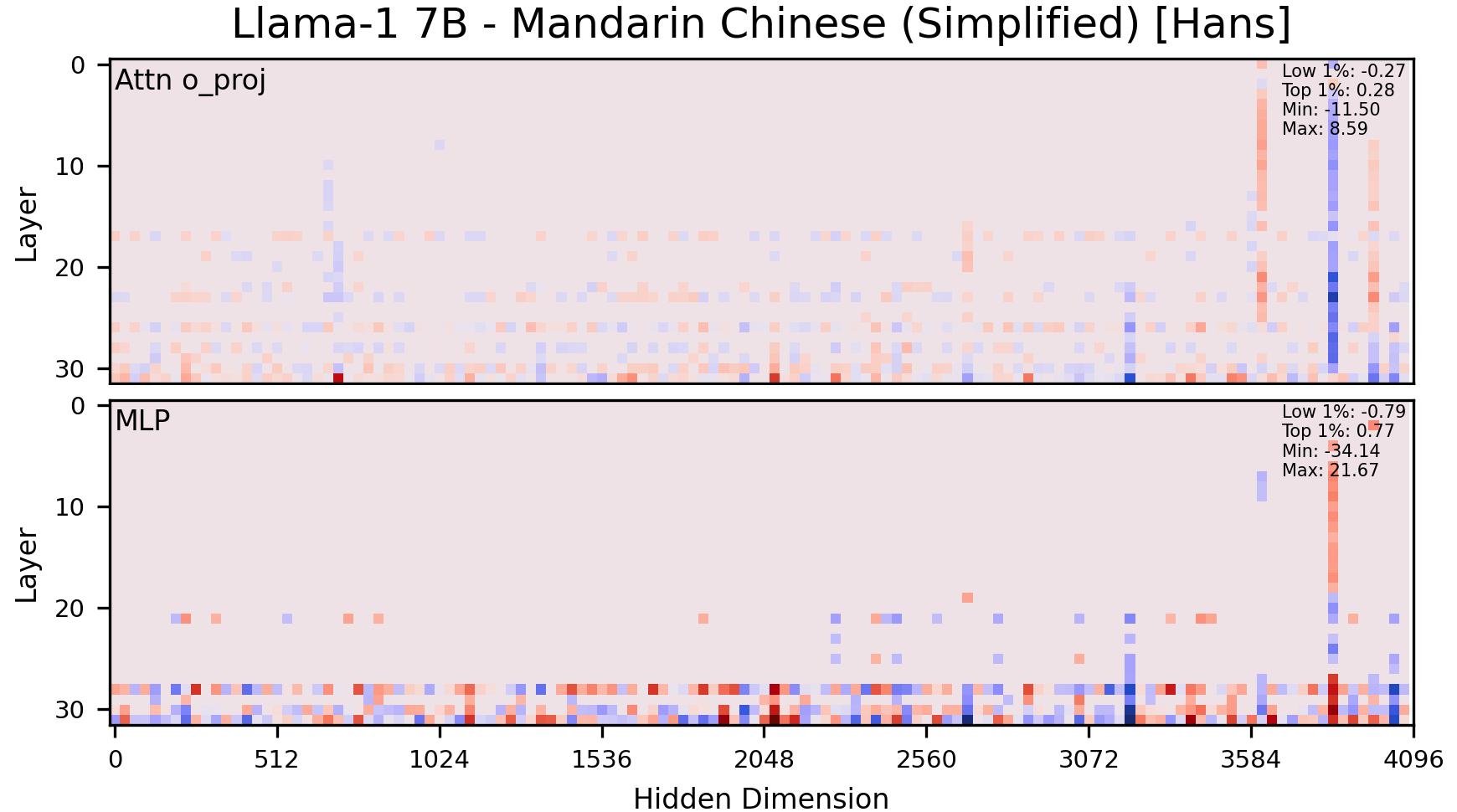} %
    \end{minipage}
    \begin{minipage}{0.33\textwidth}
        \centering
        \includegraphics[width=1.0\textwidth]{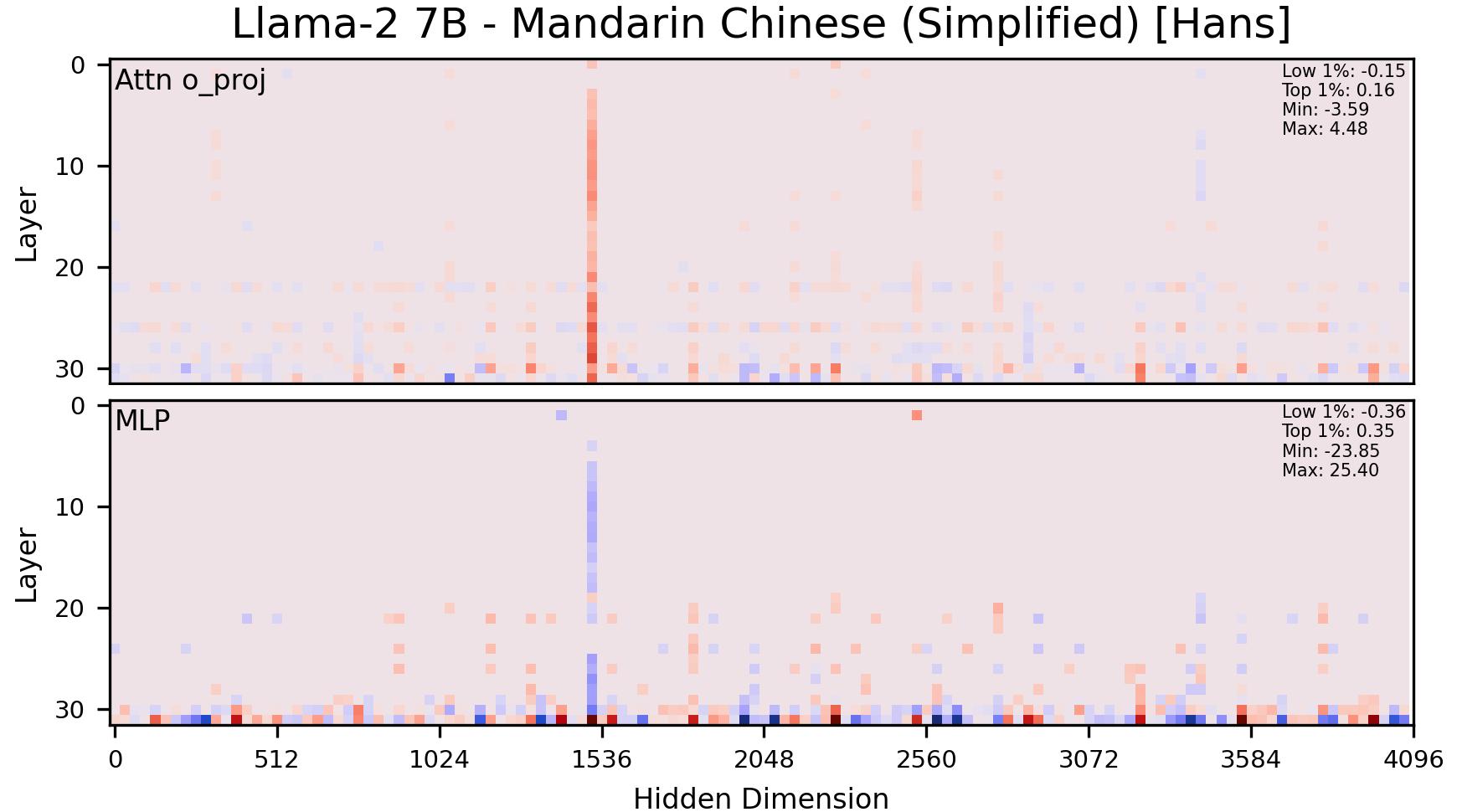} %
    \end{minipage}
    \vskip -0.3in
\end{figure}

\begin{figure}[H]
    \centering
    \begin{minipage}{0.33\textwidth}
        \centering
        \includegraphics[width=1.0\textwidth]{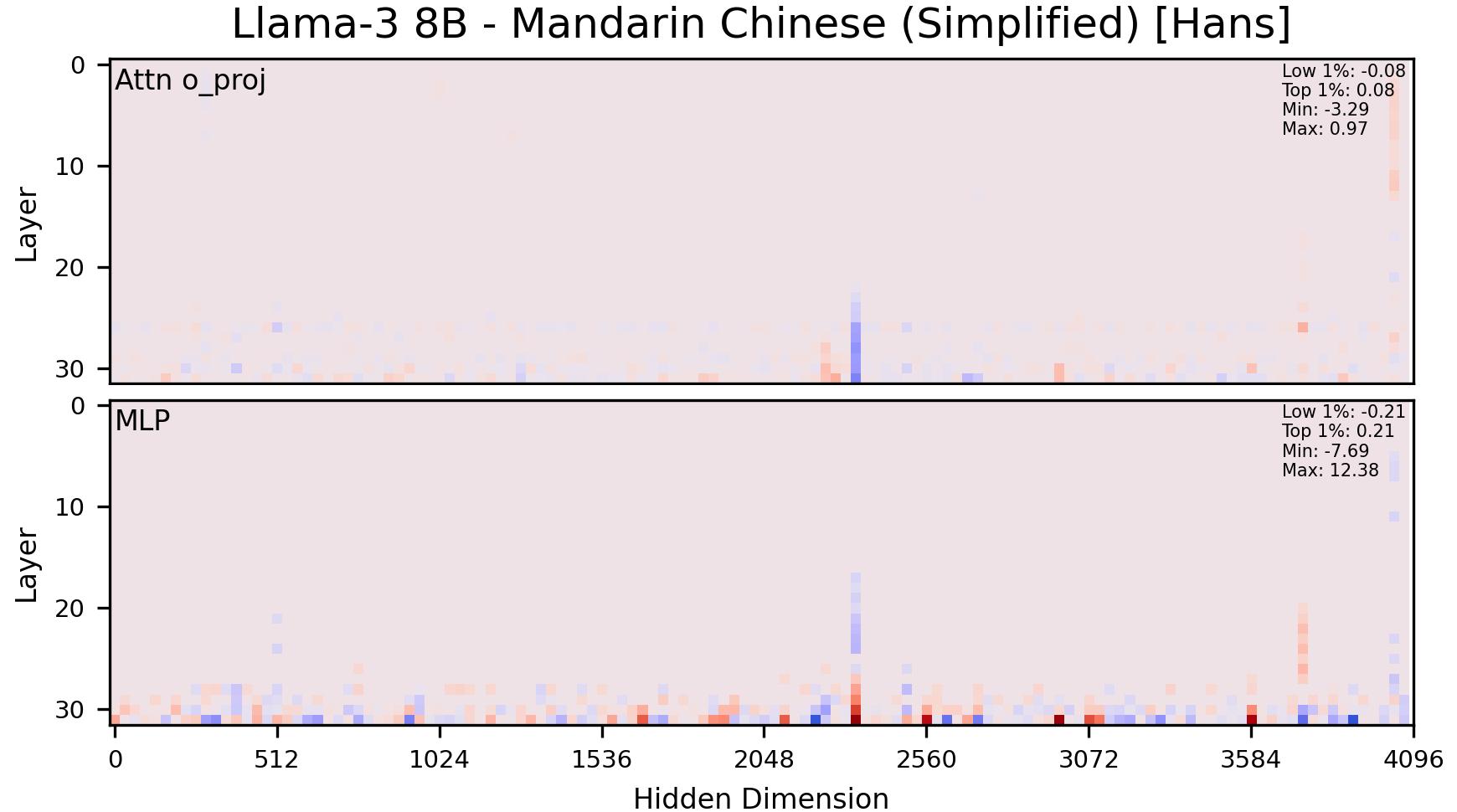} %
    \end{minipage}\hfill
    \begin{minipage}{0.33\textwidth}
        \centering
        \includegraphics[width=1.0\textwidth]{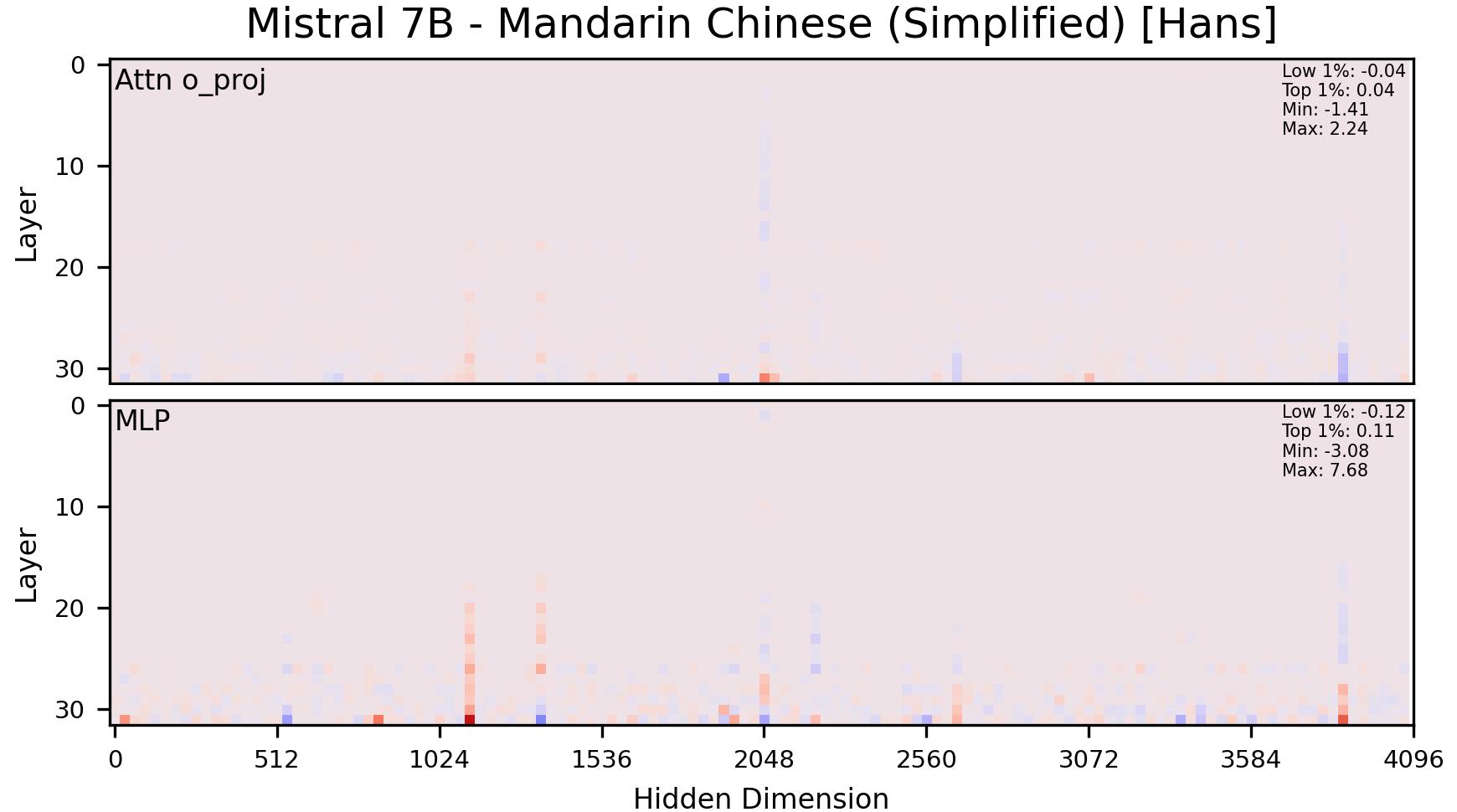} %
    \end{minipage}
    \begin{minipage}{0.33\textwidth}
        \centering
        \includegraphics[width=1.0\textwidth]{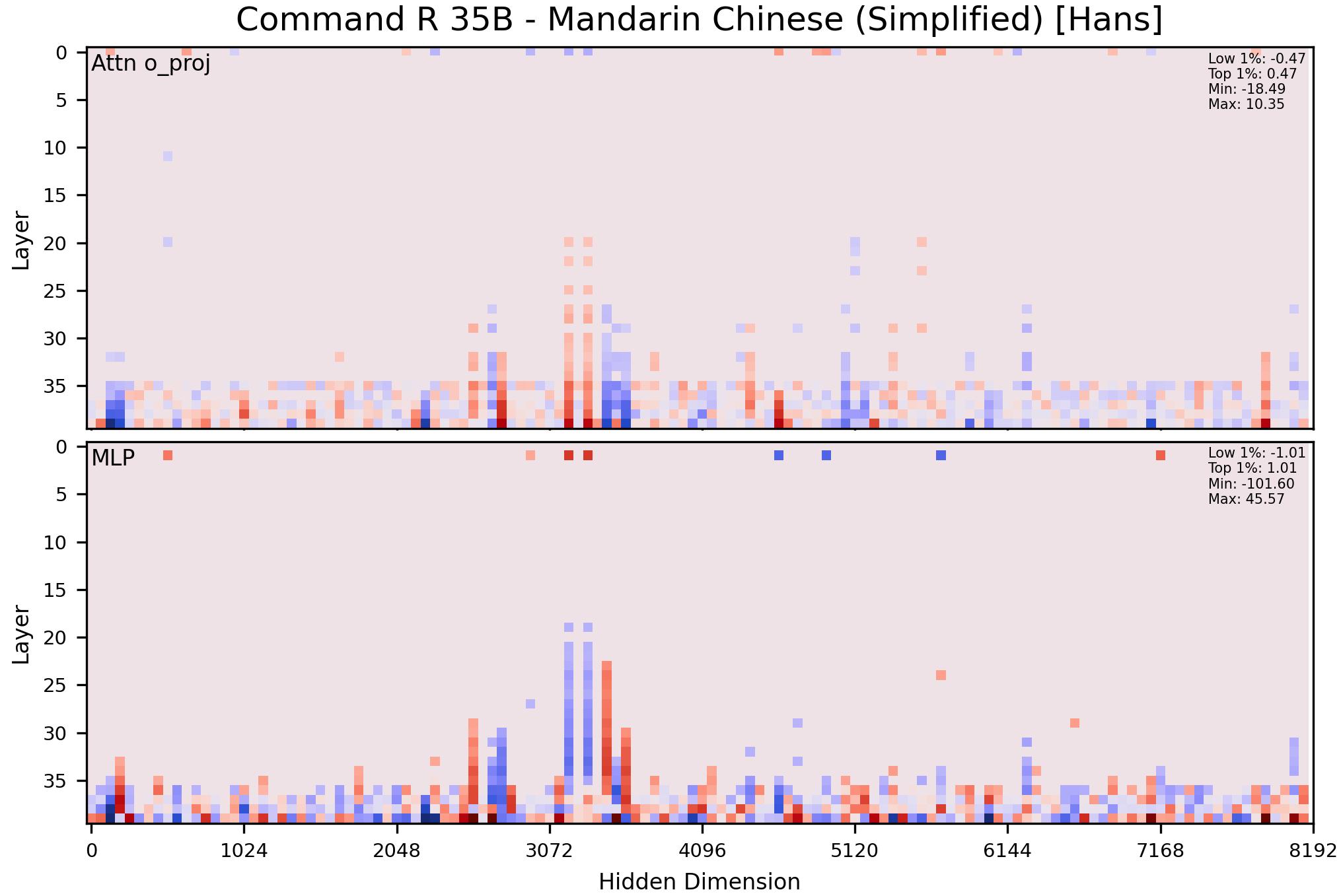} %
    \end{minipage}
\end{figure}

\begin{figure}[H]
    \centering
    \begin{minipage}{0.33\textwidth}
        \centering
        \includegraphics[width=1.0\textwidth]{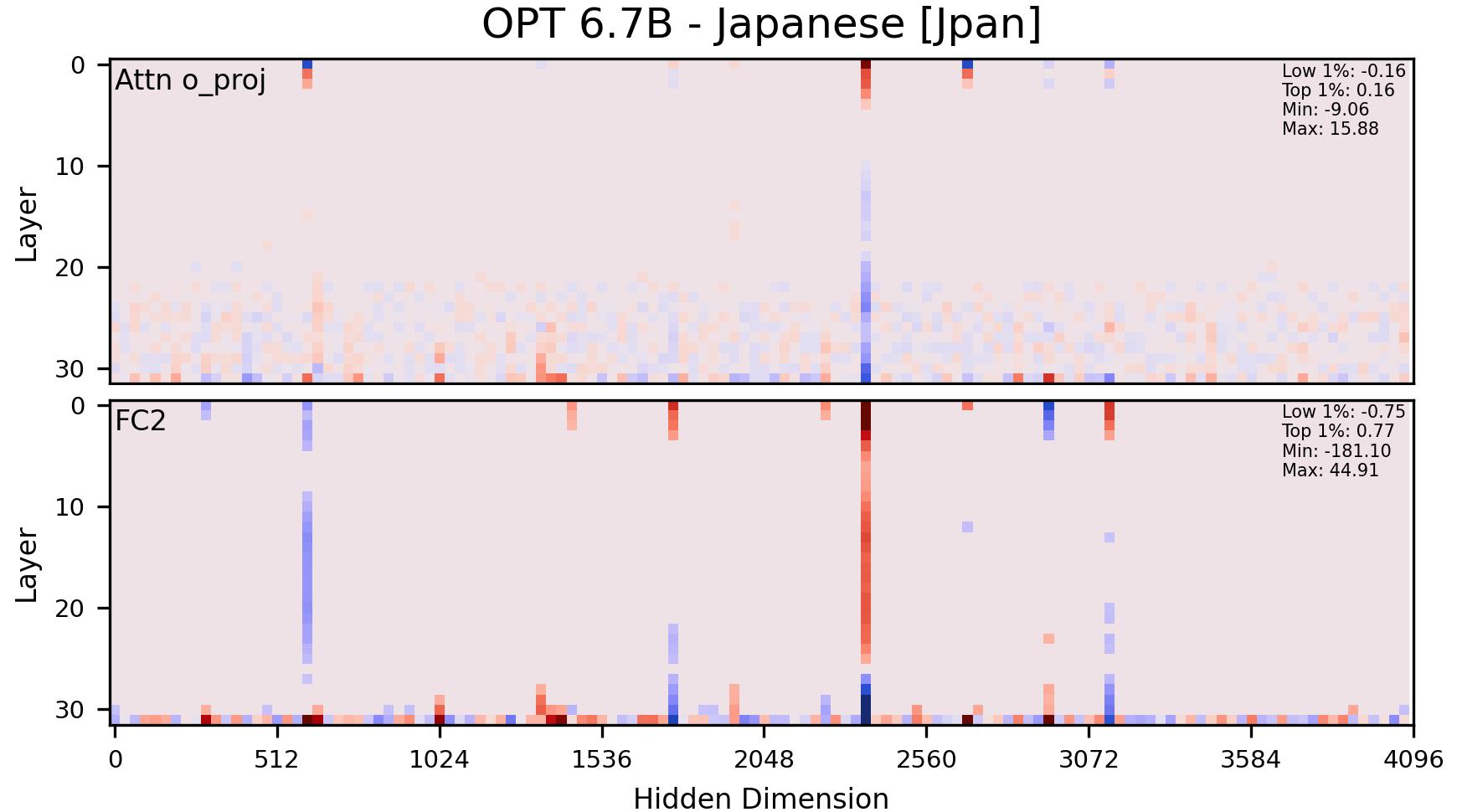} %
    \end{minipage}\hfill
    \begin{minipage}{0.33\textwidth}
        \centering
        \includegraphics[width=1.0\textwidth]{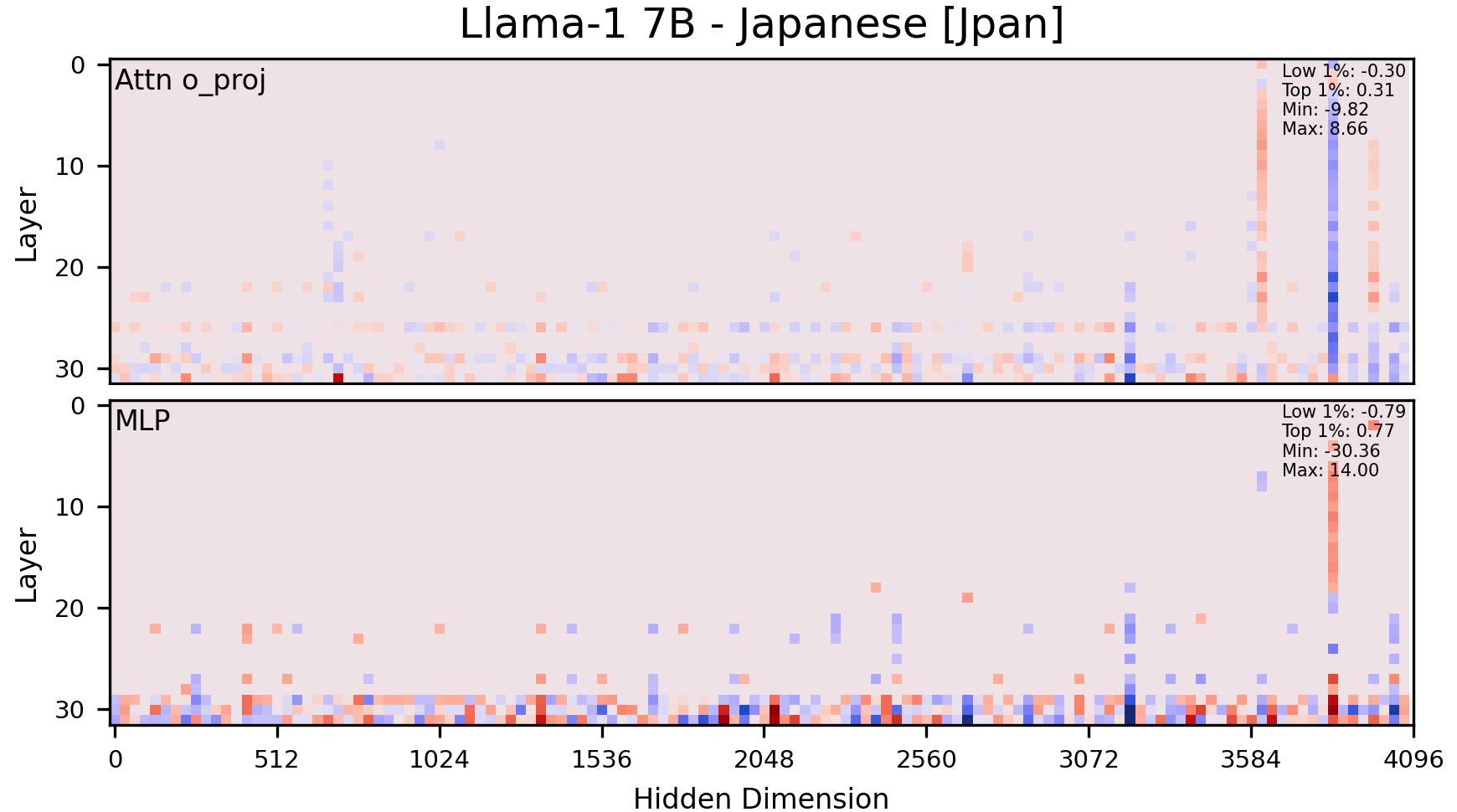} %
    \end{minipage}
    \begin{minipage}{0.33\textwidth}
        \centering
        \includegraphics[width=1.0\textwidth]{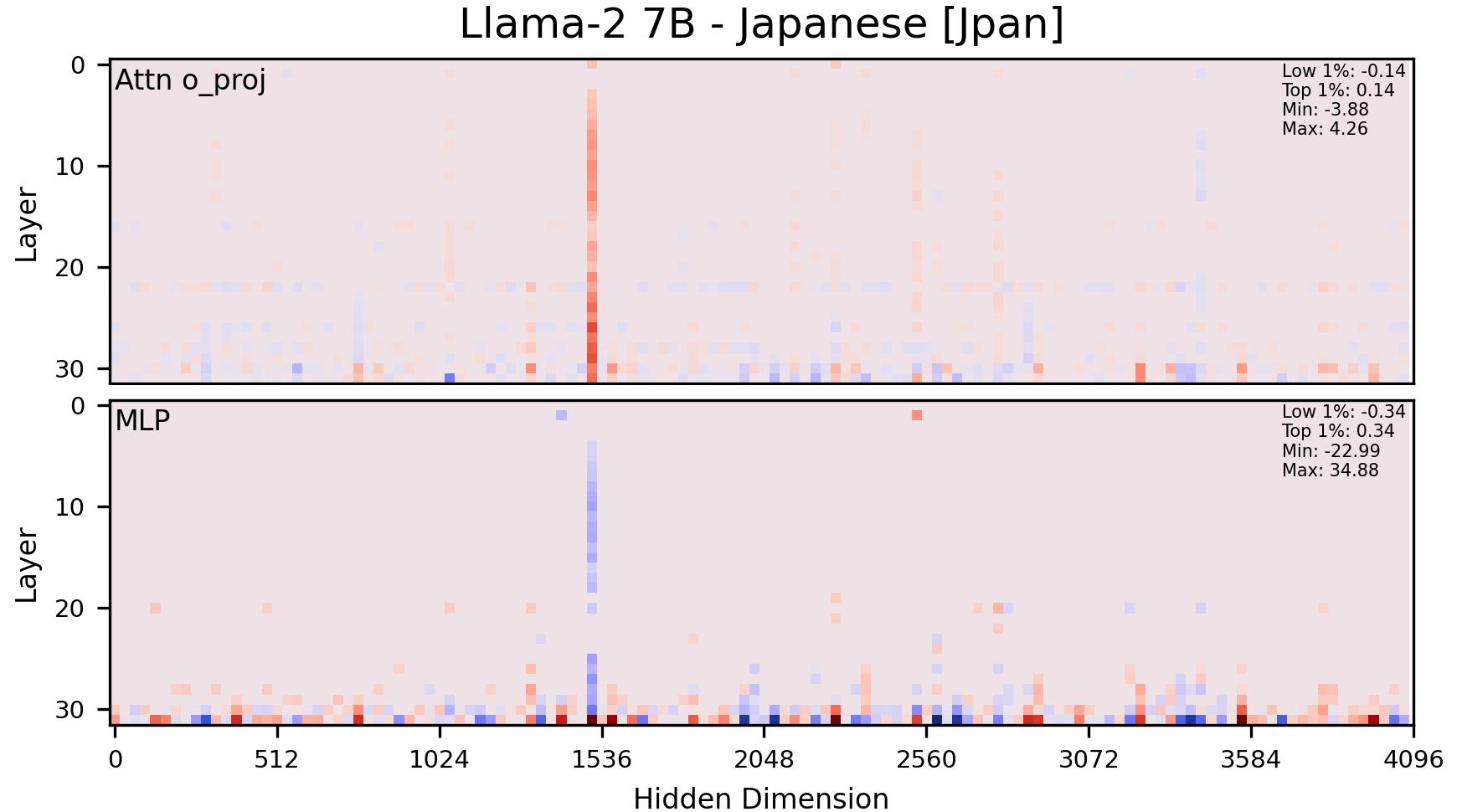} %
    \end{minipage}
    \vskip -0.3in
\end{figure}

\begin{figure}[H]
    \centering
    \begin{minipage}{0.33\textwidth}
        \centering
        \includegraphics[width=1.0\textwidth]{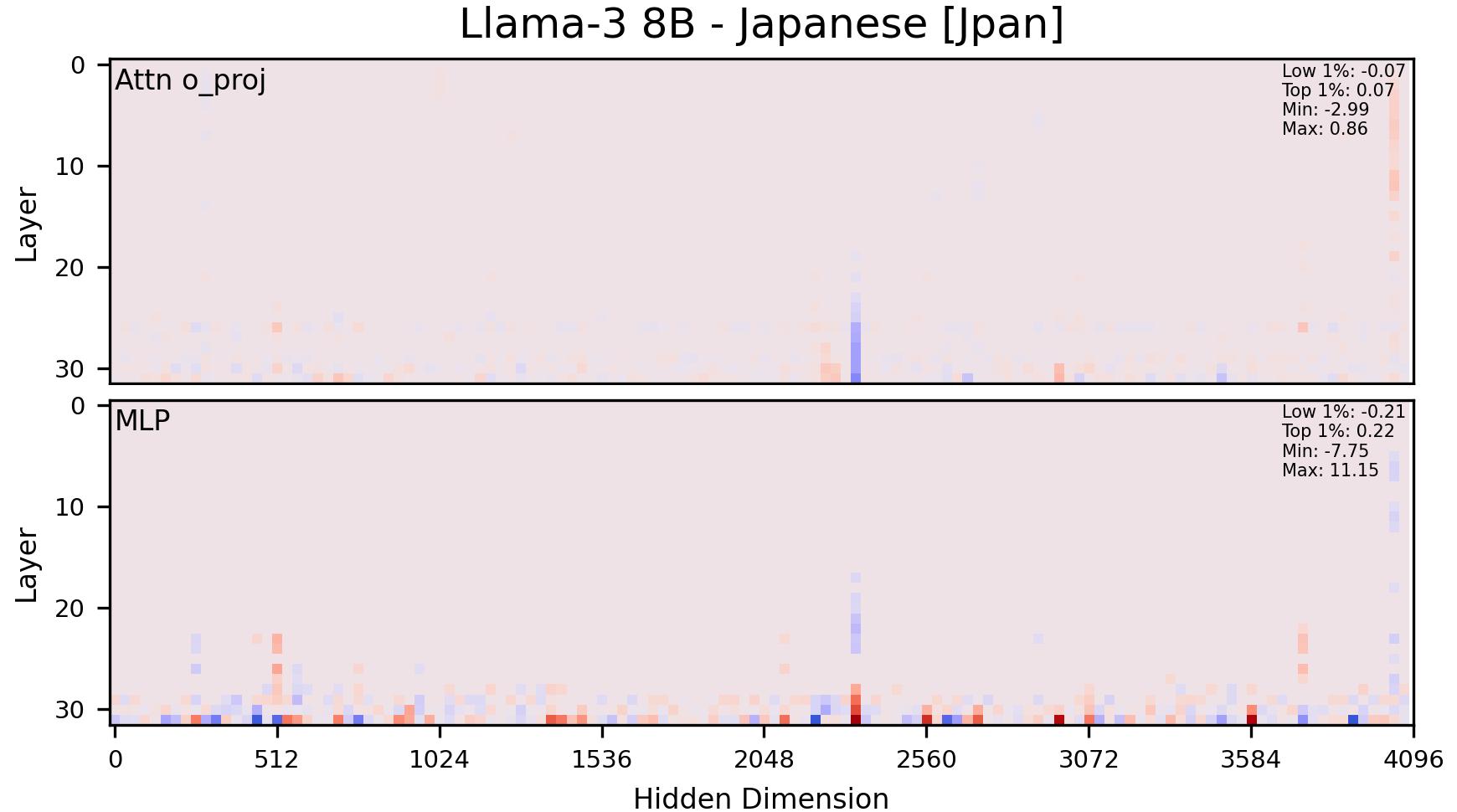} %
    \end{minipage}\hfill
    \begin{minipage}{0.33\textwidth}
        \centering
        \includegraphics[width=1.0\textwidth]{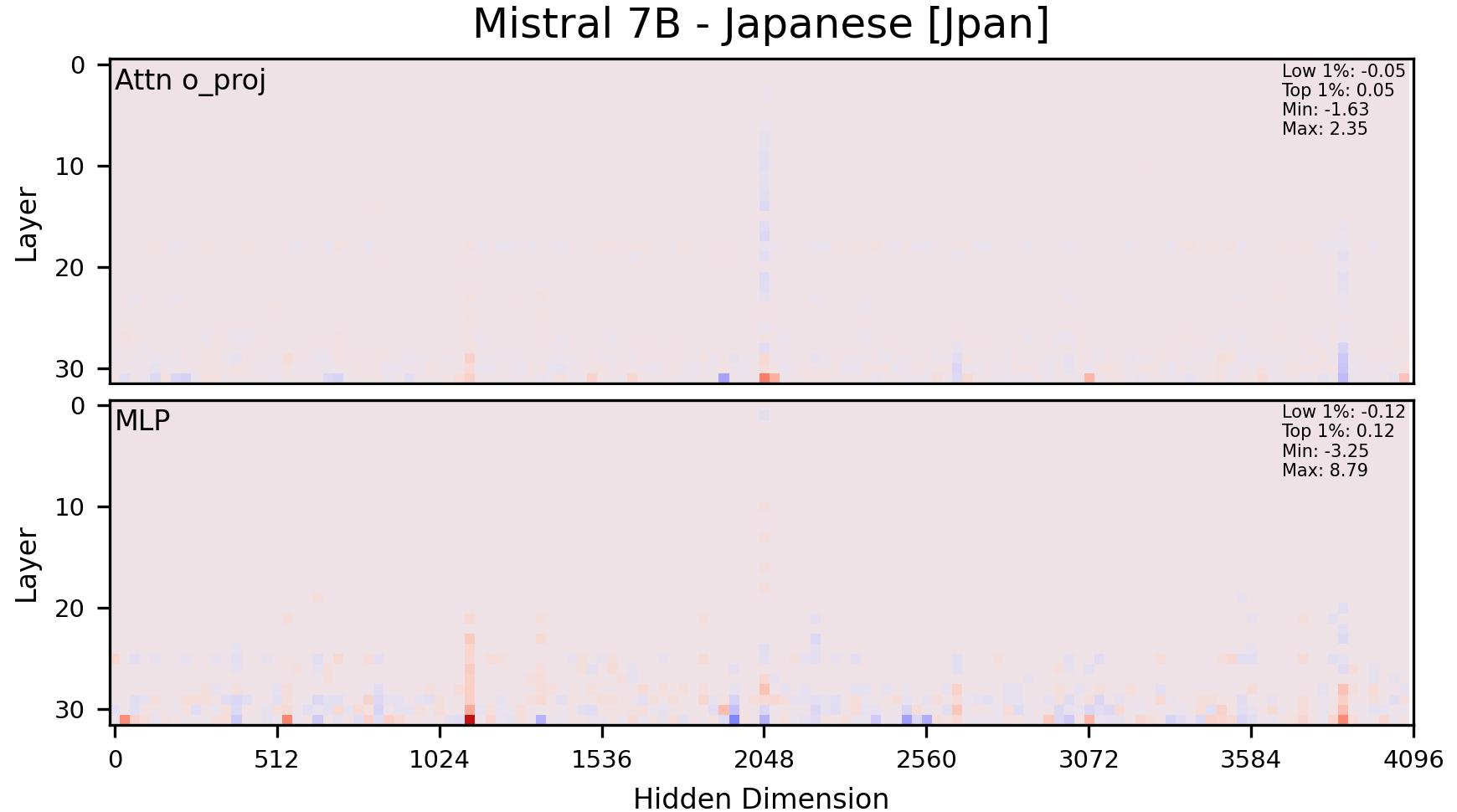} %
    \end{minipage}
    \begin{minipage}{0.33\textwidth}
        \centering
        \includegraphics[width=1.0\textwidth]{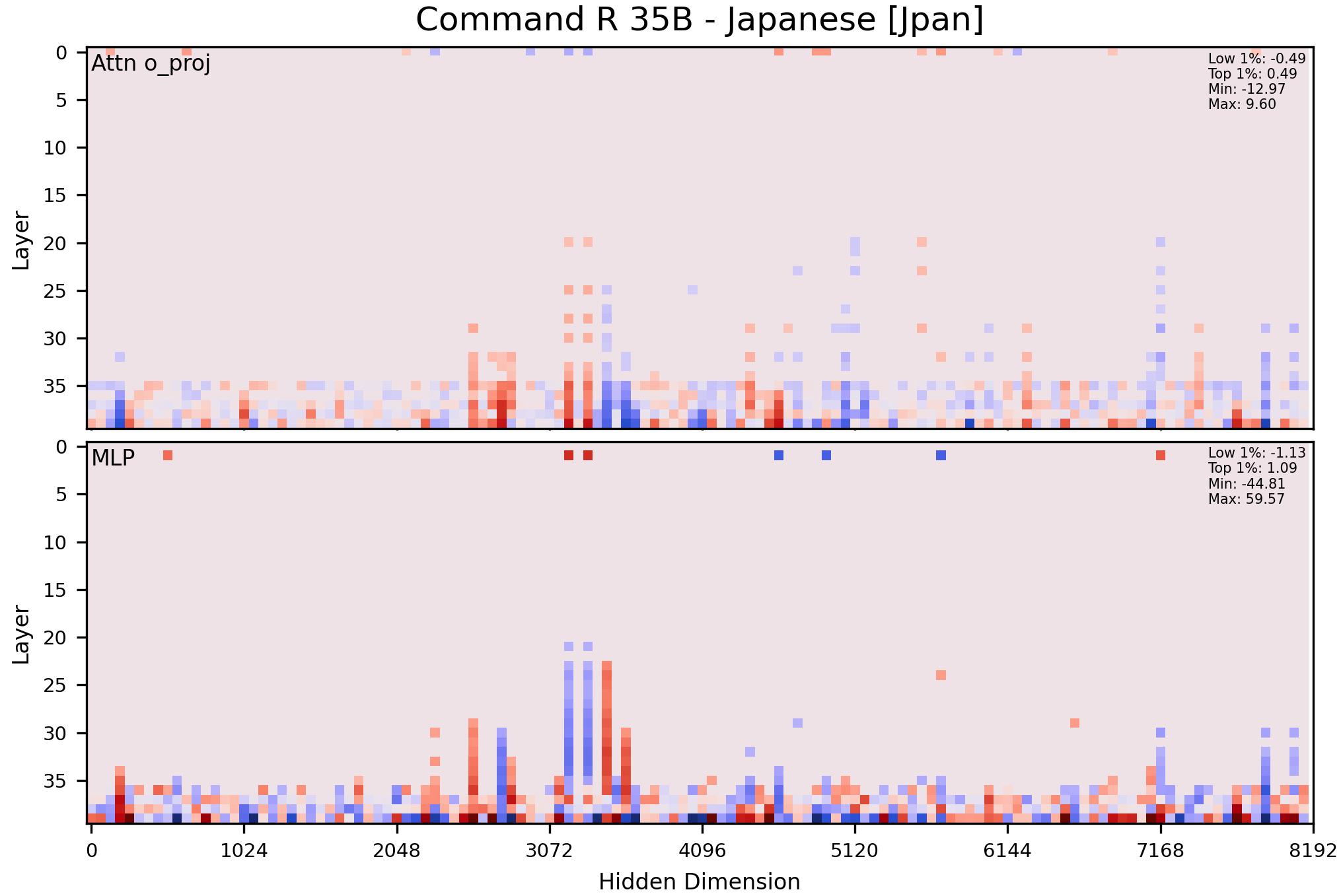} %
    \end{minipage}
\end{figure}

\begin{figure}[H]
    \centering
    \begin{minipage}{0.33\textwidth}
        \centering
        \includegraphics[width=1.0\textwidth]{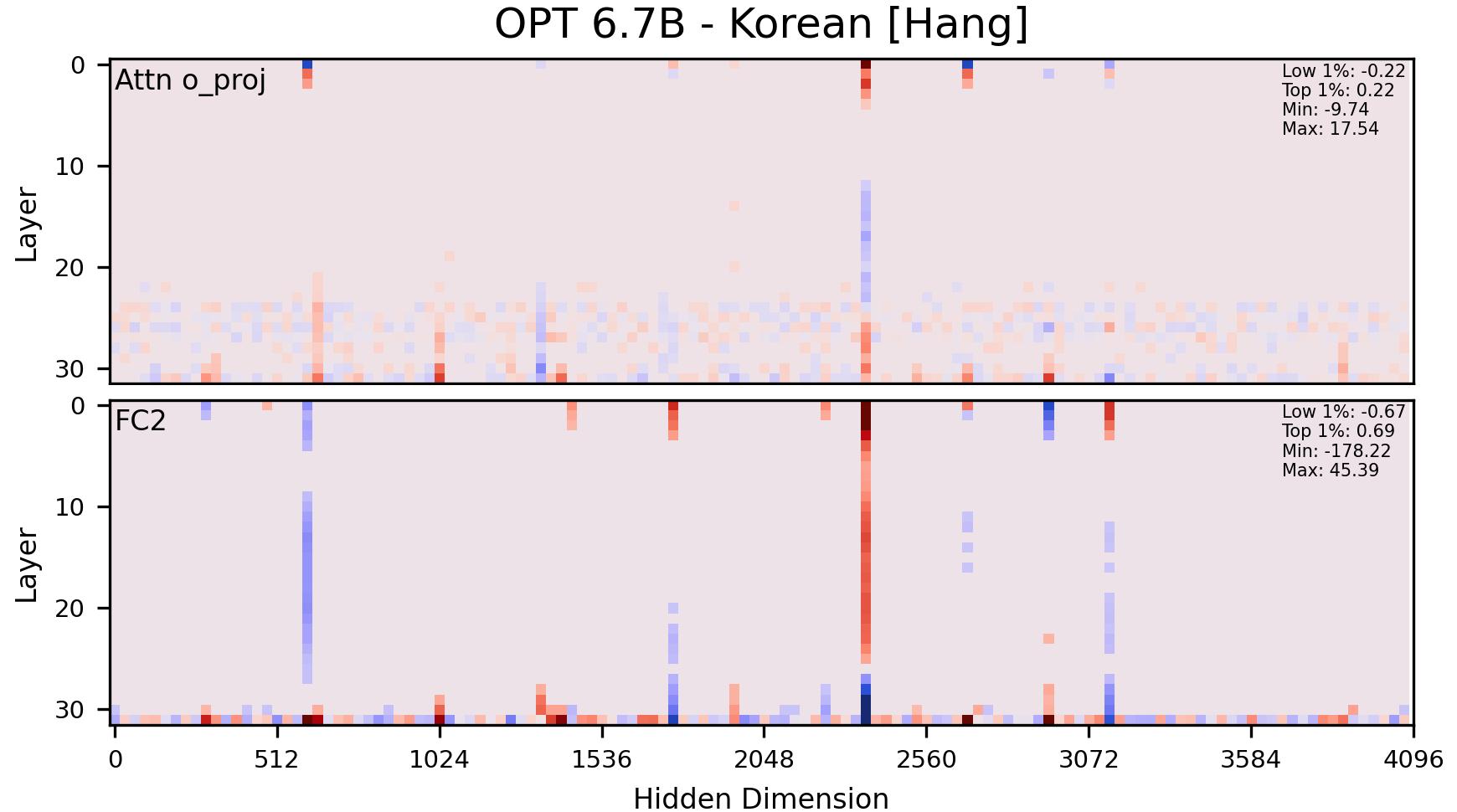} %
    \end{minipage}\hfill
    \begin{minipage}{0.33\textwidth}
        \centering
        \includegraphics[width=1.0\textwidth]{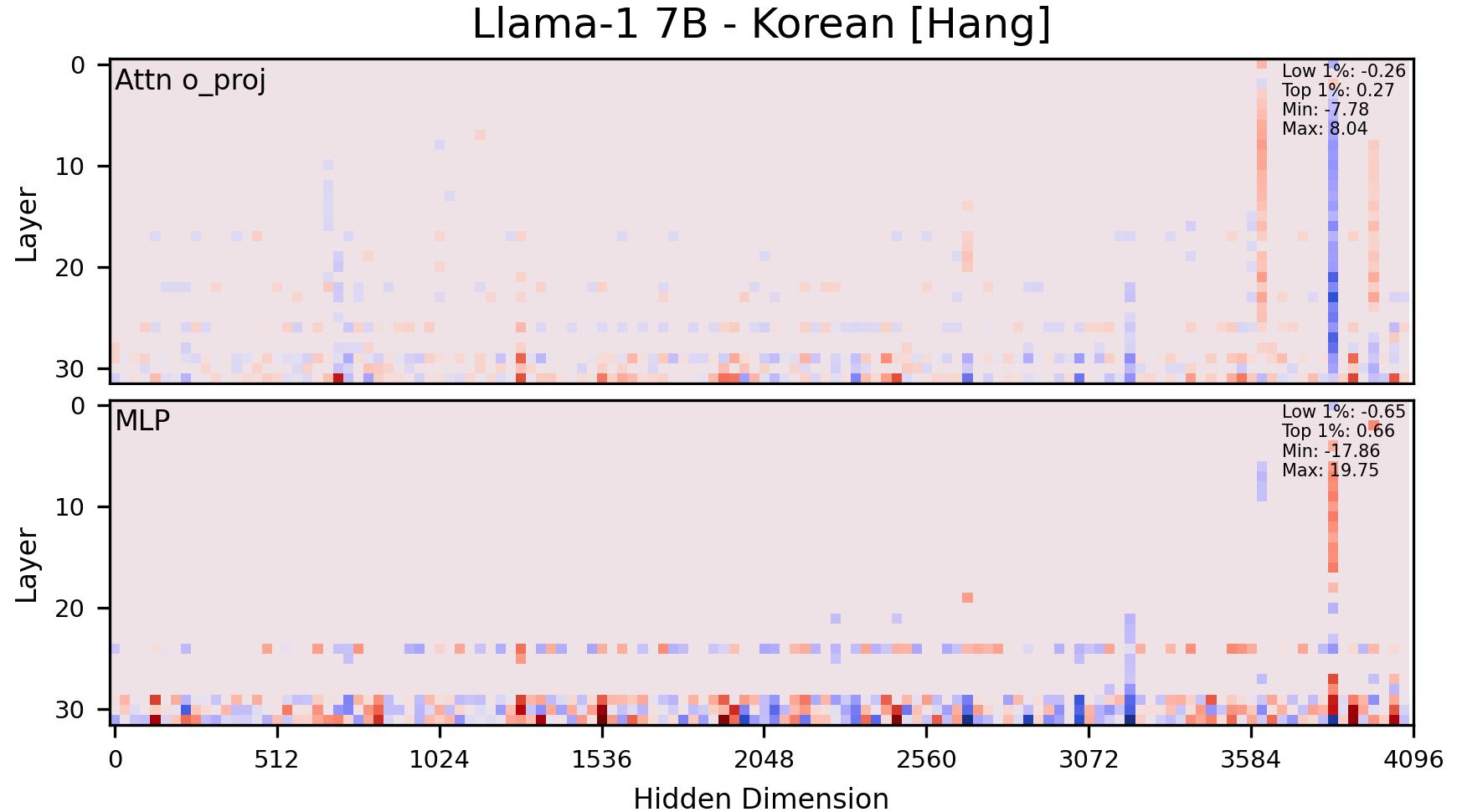} %
    \end{minipage}
    \begin{minipage}{0.33\textwidth}
        \centering
        \includegraphics[width=1.0\textwidth]{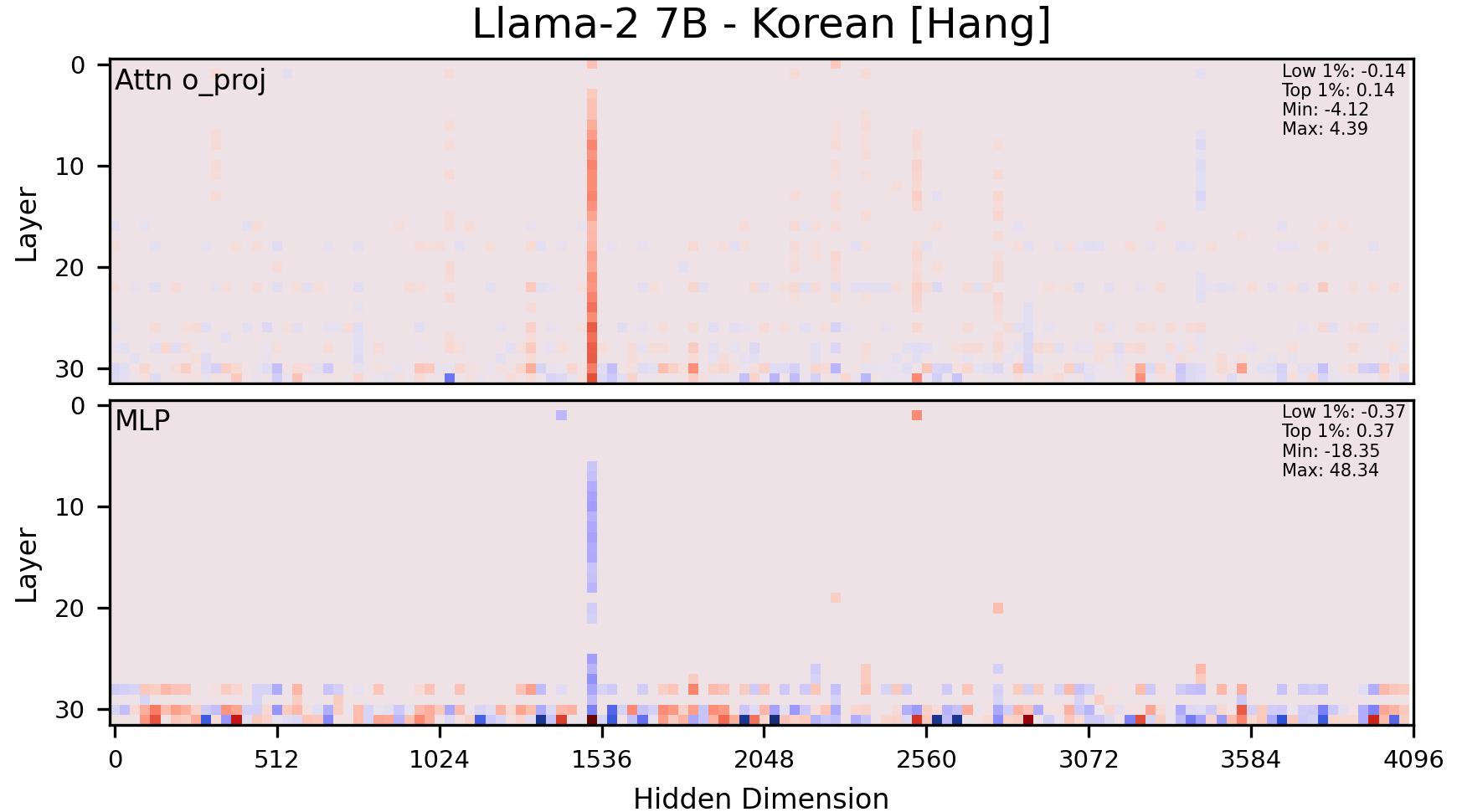} %
    \end{minipage}
    \vskip -0.3in
\end{figure}

\begin{figure}[H]
    \centering
    \begin{minipage}{0.33\textwidth}
        \centering
        \includegraphics[width=1.0\textwidth]{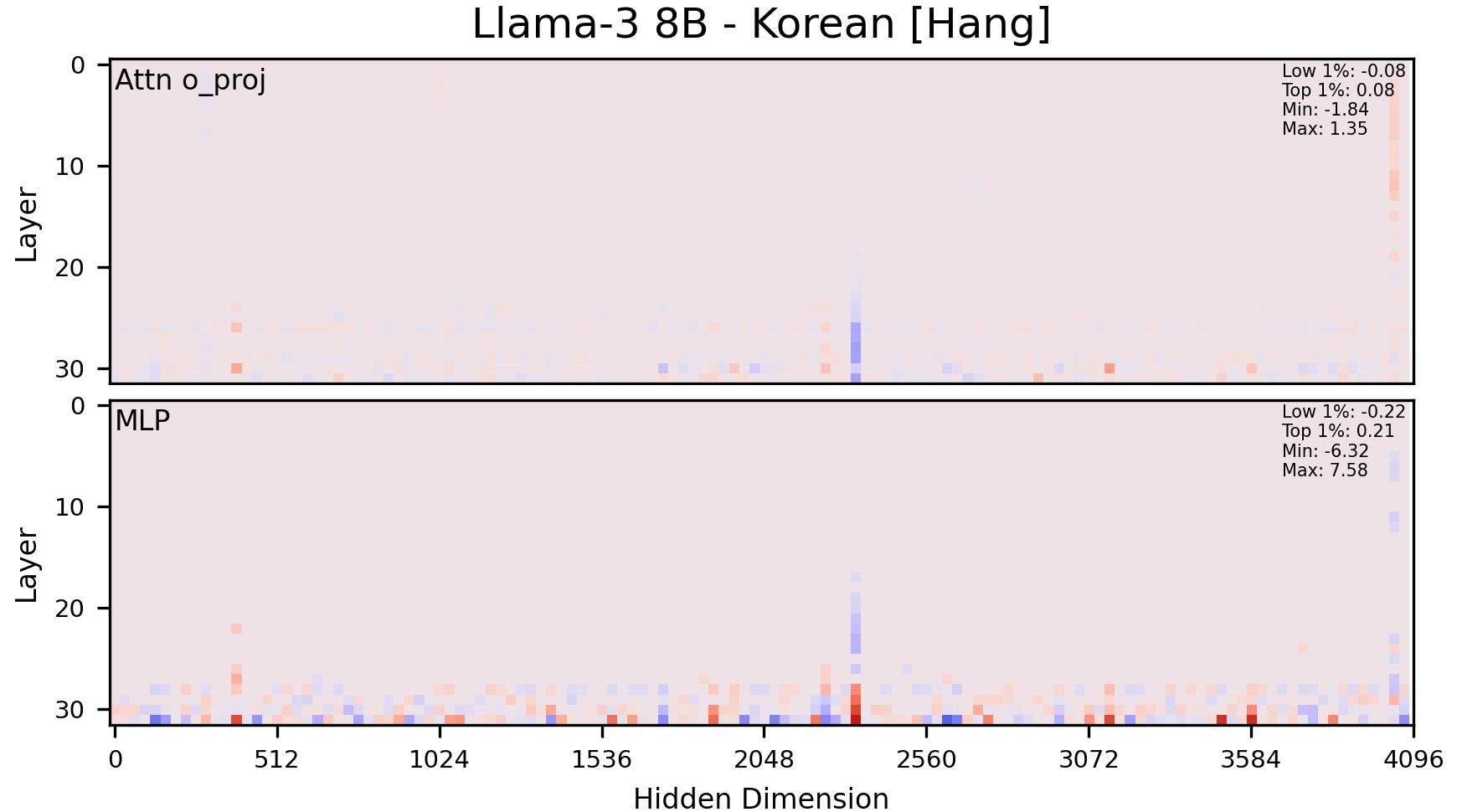} %
    \end{minipage}\hfill
    \begin{minipage}{0.33\textwidth}
        \centering
        \includegraphics[width=1.0\textwidth]{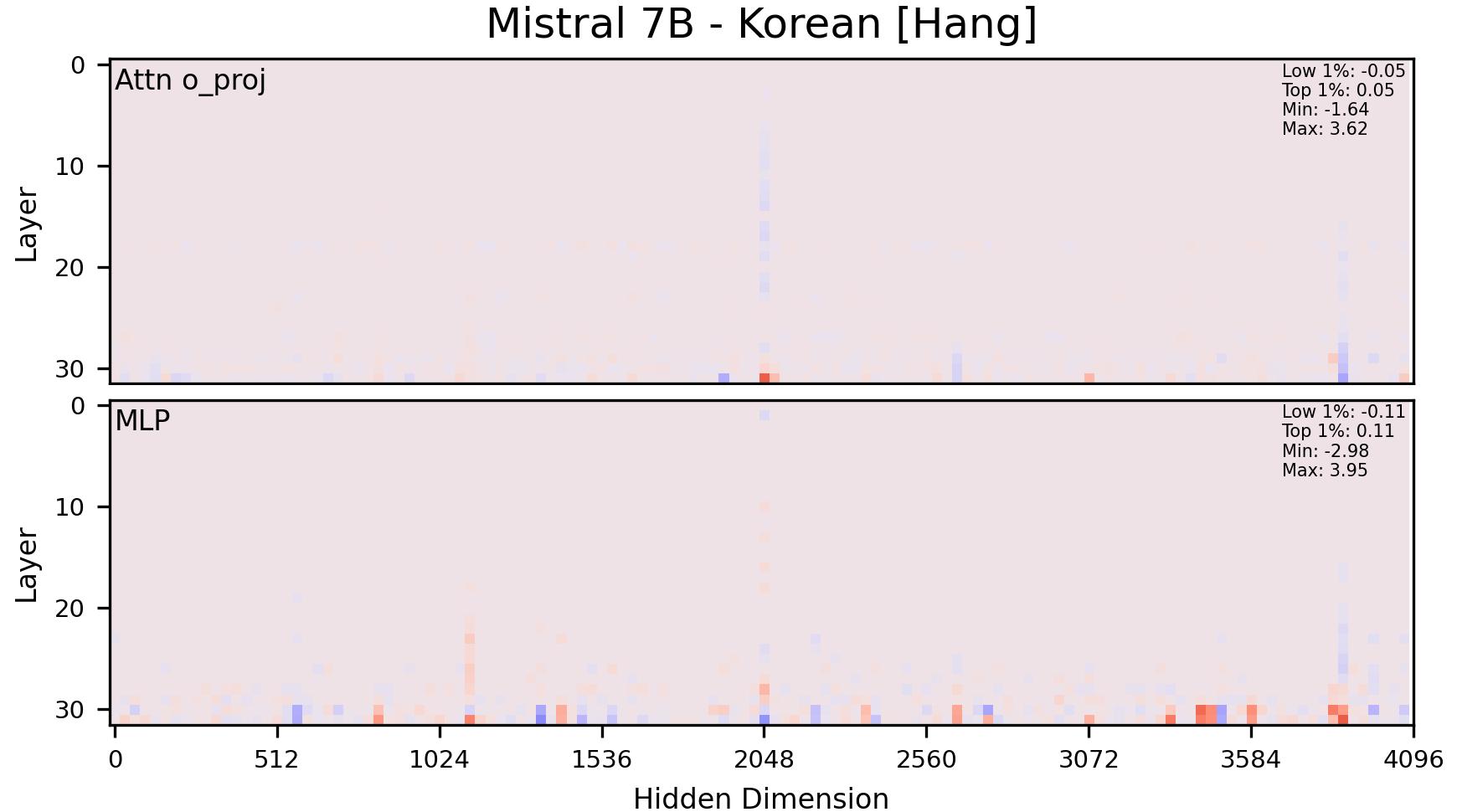} %
    \end{minipage}
    \begin{minipage}{0.33\textwidth}
        \centering
        \includegraphics[width=1.0\textwidth]{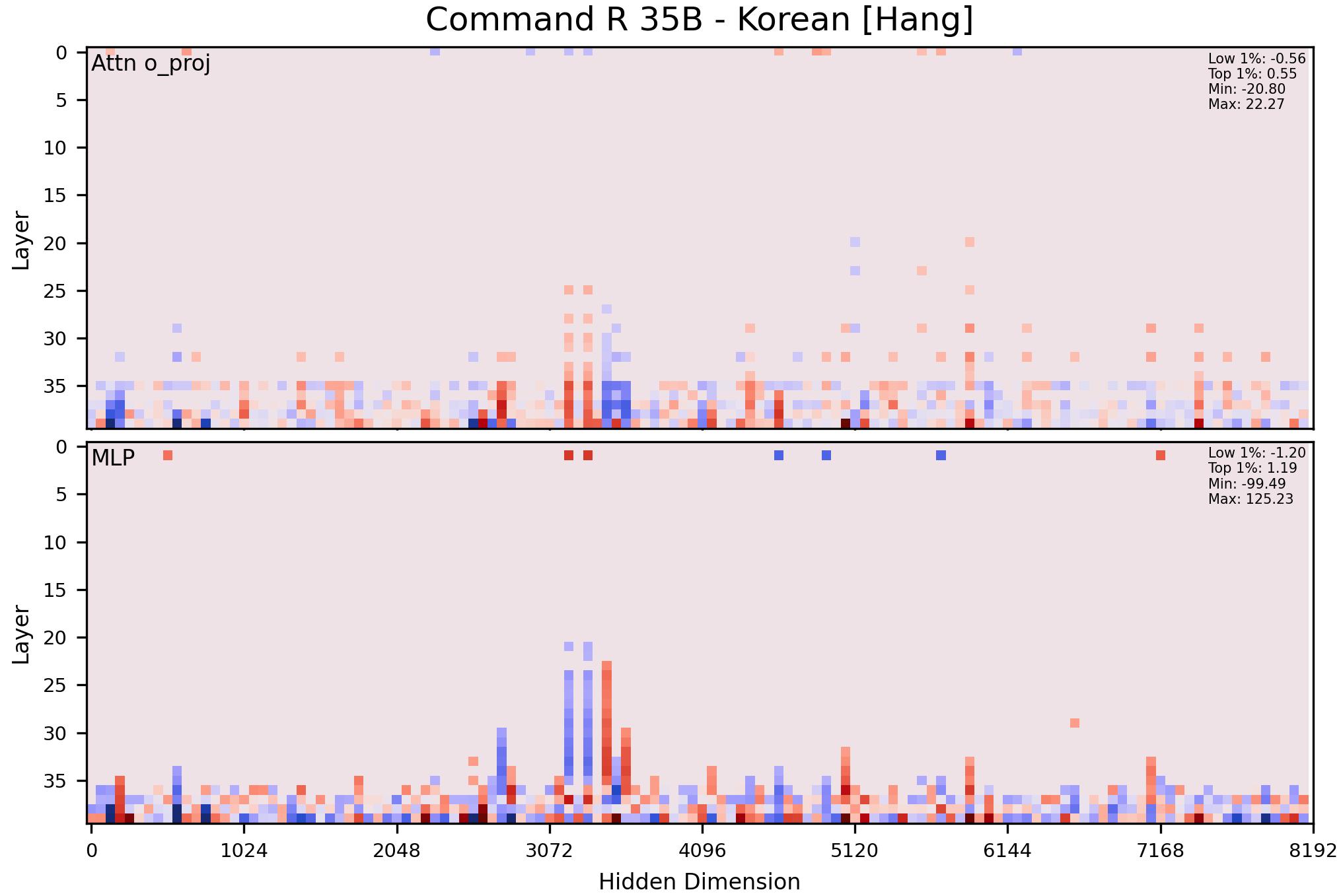} %
    \end{minipage}
\end{figure}

\begin{figure}[H]
    \centering
    \begin{minipage}{0.33\textwidth}
        \centering
        \includegraphics[width=1.0\textwidth]{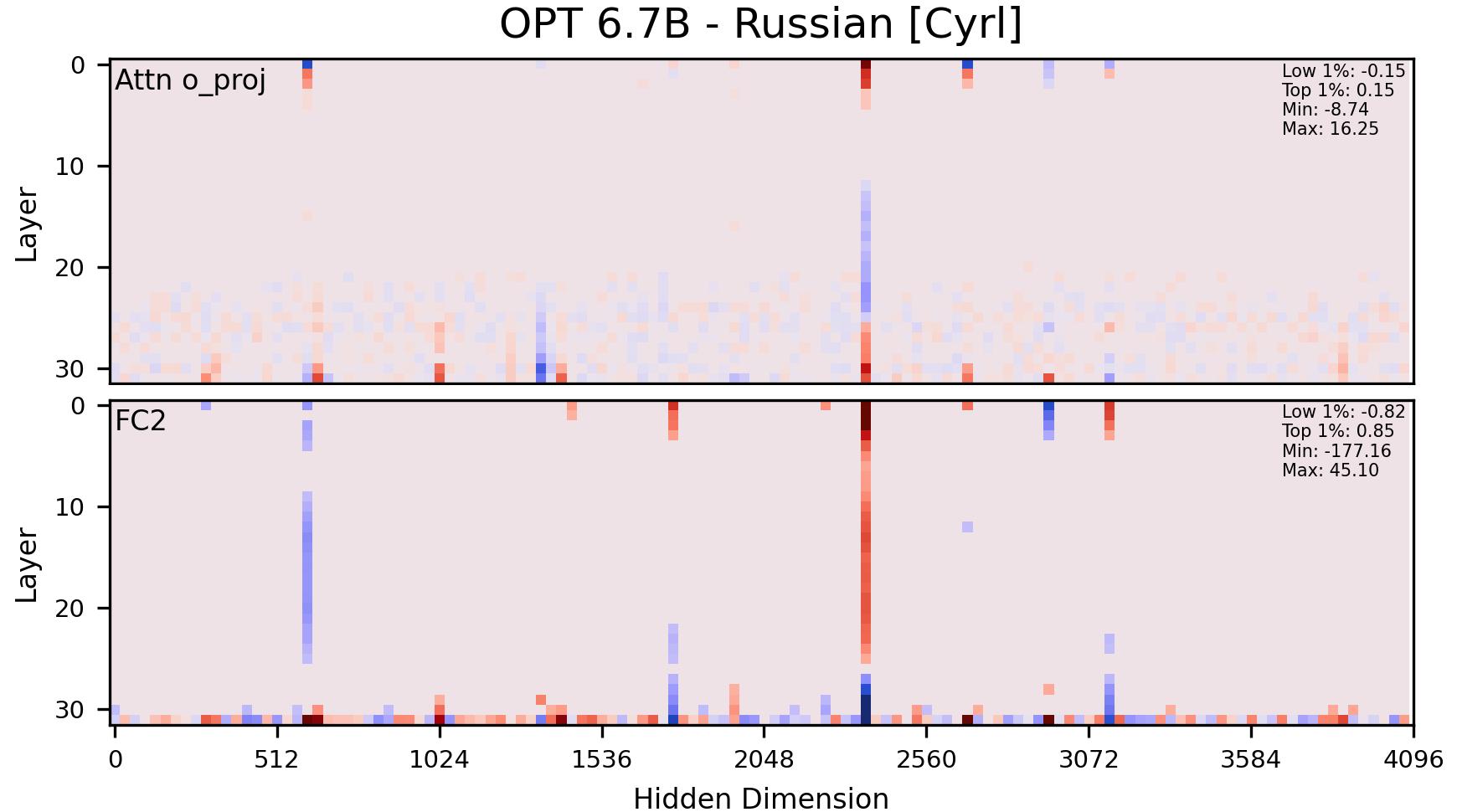} %
    \end{minipage}\hfill
    \begin{minipage}{0.33\textwidth}
        \centering
        \includegraphics[width=1.0\textwidth]{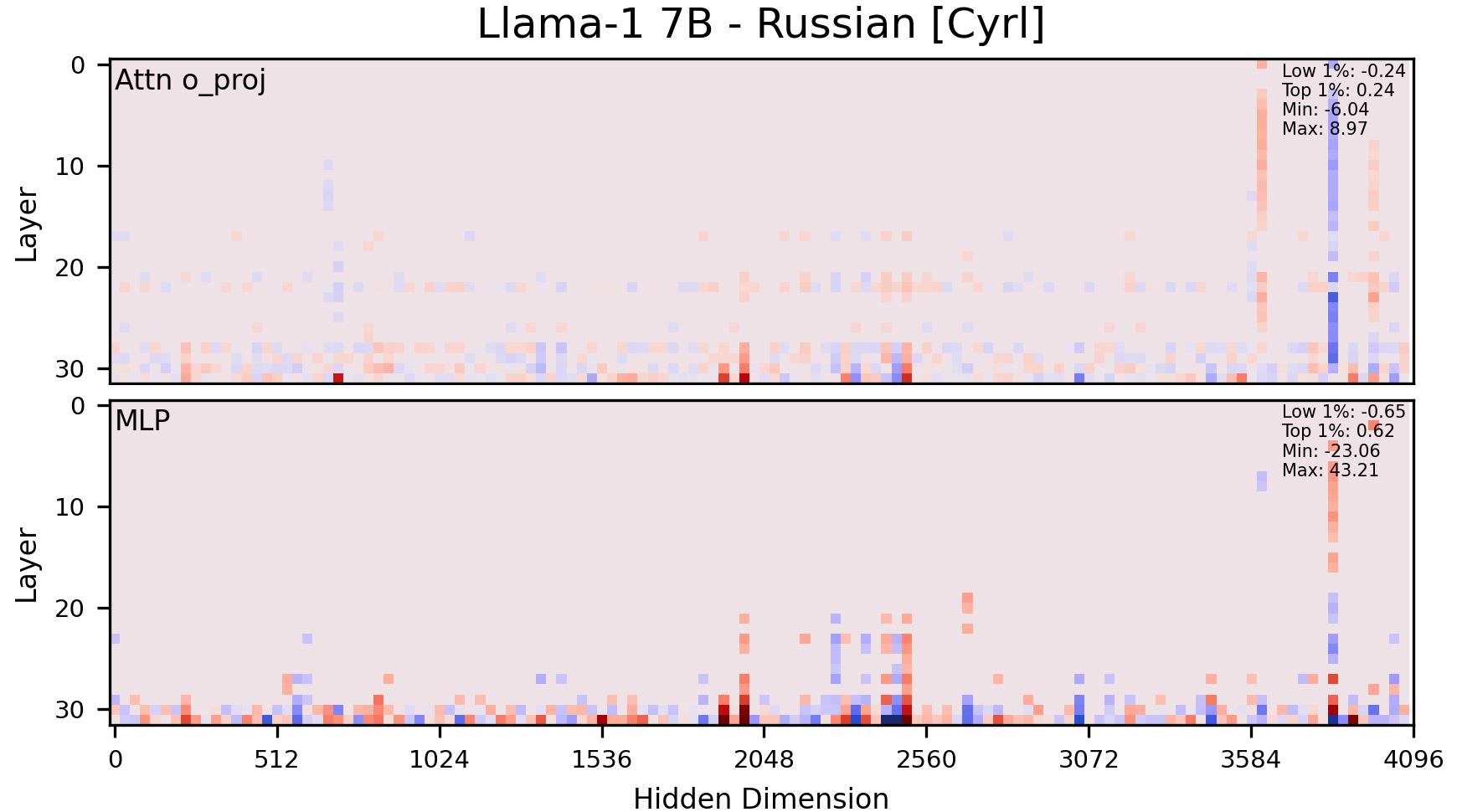} %
    \end{minipage}
    \begin{minipage}{0.33\textwidth}
        \centering
        \includegraphics[width=1.0\textwidth]{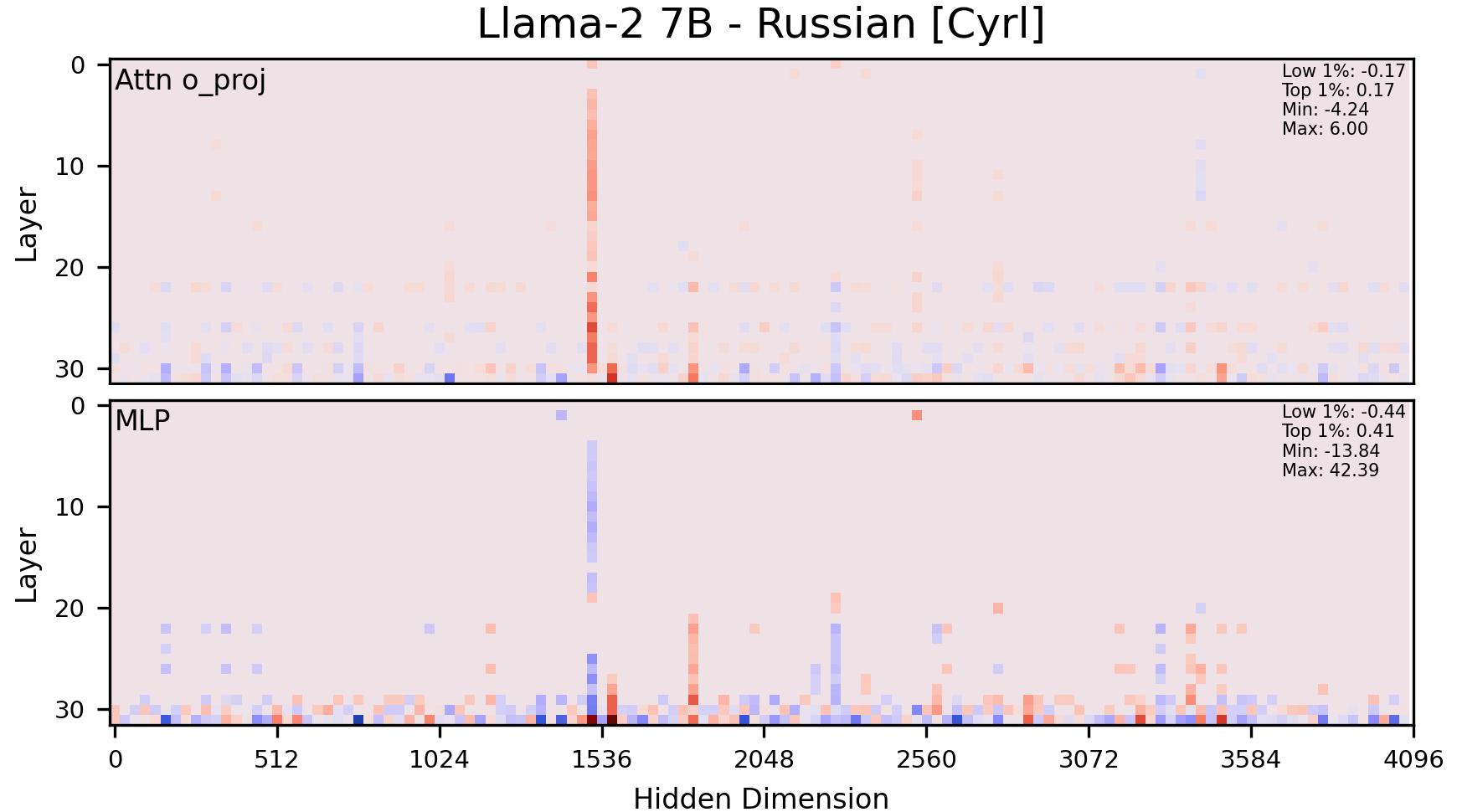} %
    \end{minipage}
    \vskip -0.3in
\end{figure}

\begin{figure}[H]
    \centering
    \begin{minipage}{0.33\textwidth}
        \centering
        \includegraphics[width=1.0\textwidth]{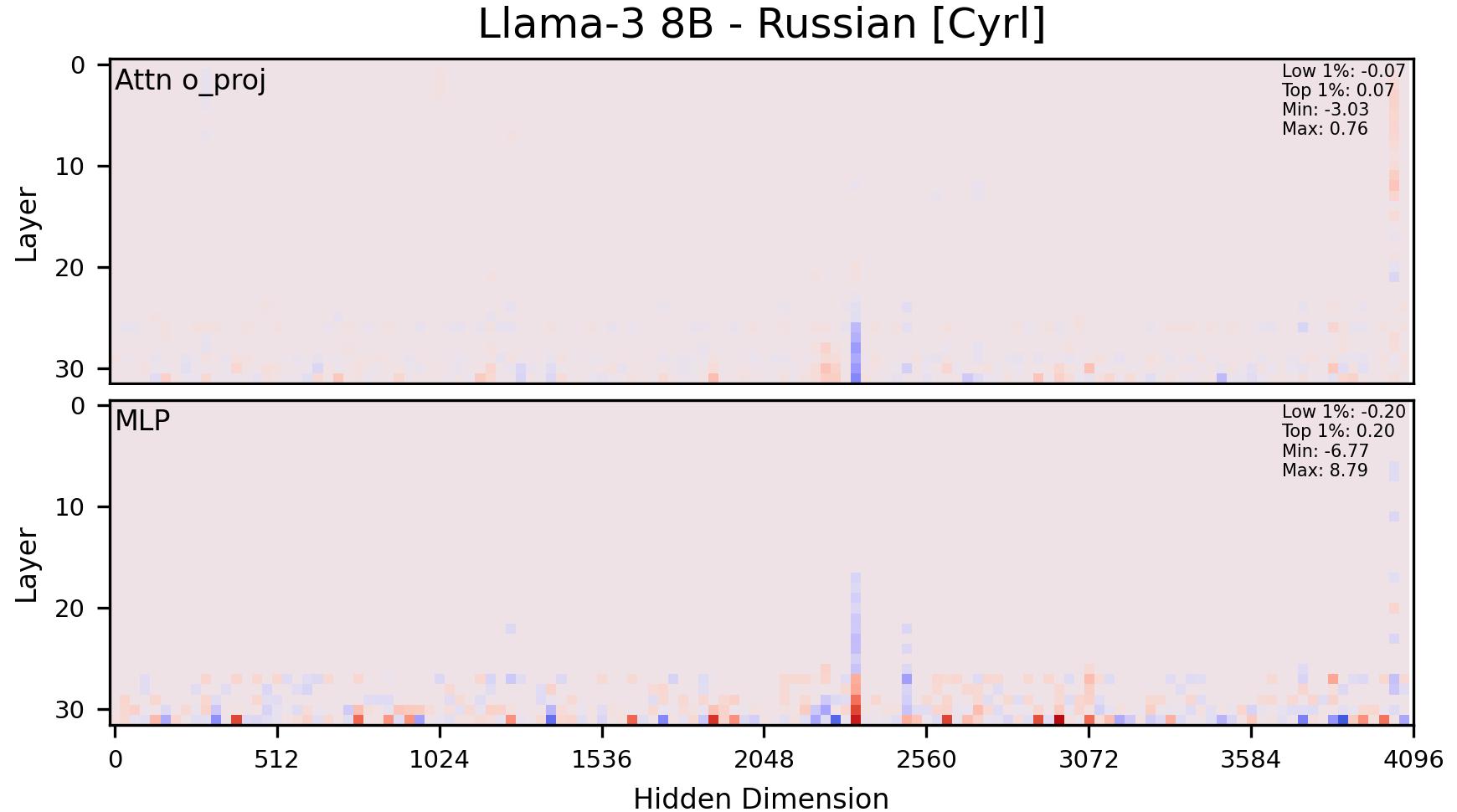} %
    \end{minipage}\hfill
    \begin{minipage}{0.33\textwidth}
        \centering
        \includegraphics[width=1.0\textwidth]{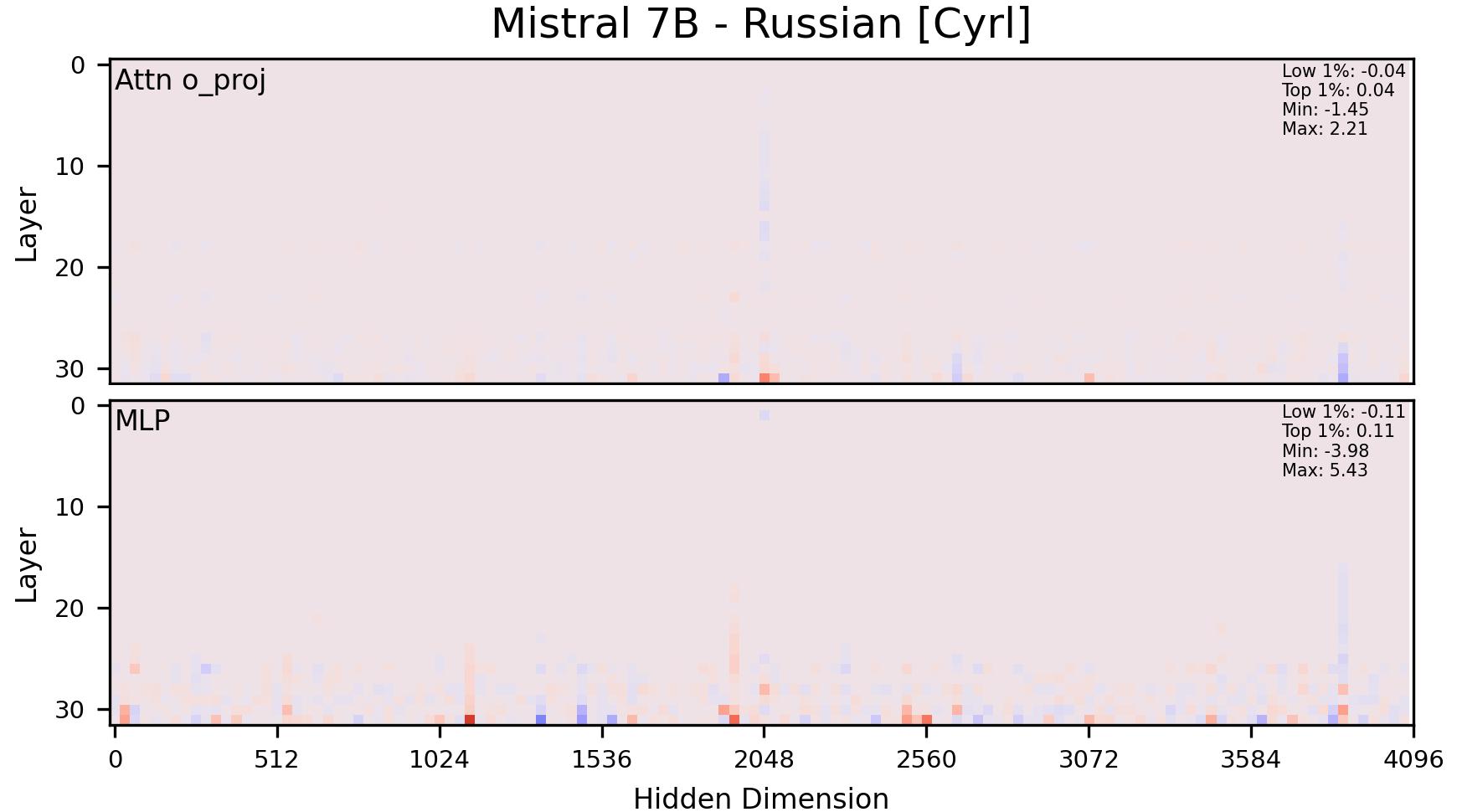} %
    \end{minipage}
    \begin{minipage}{0.33\textwidth}
        \centering
        \includegraphics[width=1.0\textwidth]{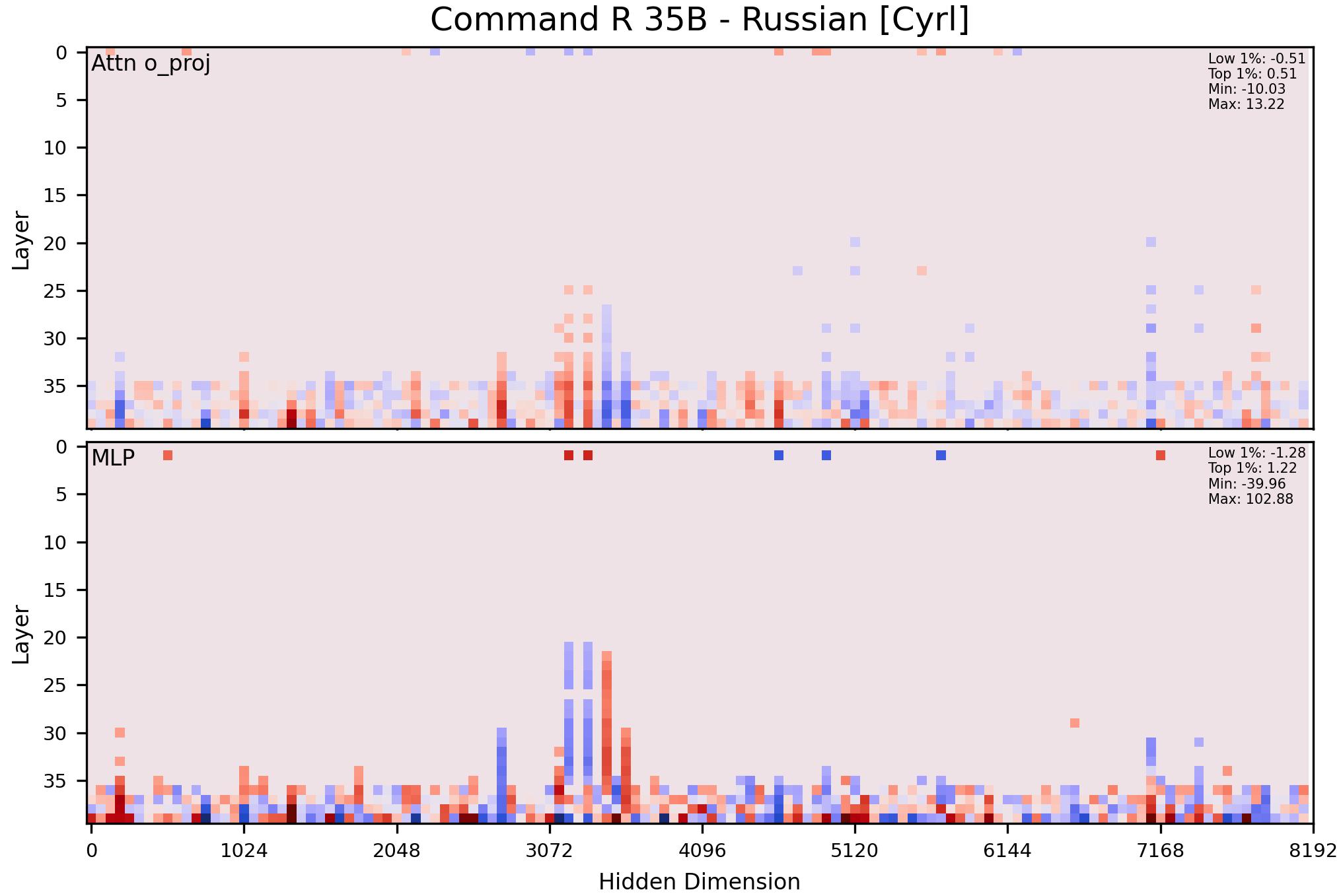} %
    \end{minipage}
\end{figure}

\begin{figure}[H]
    \centering
    \begin{minipage}{0.33\textwidth}
        \centering
        \includegraphics[width=1.0\textwidth]{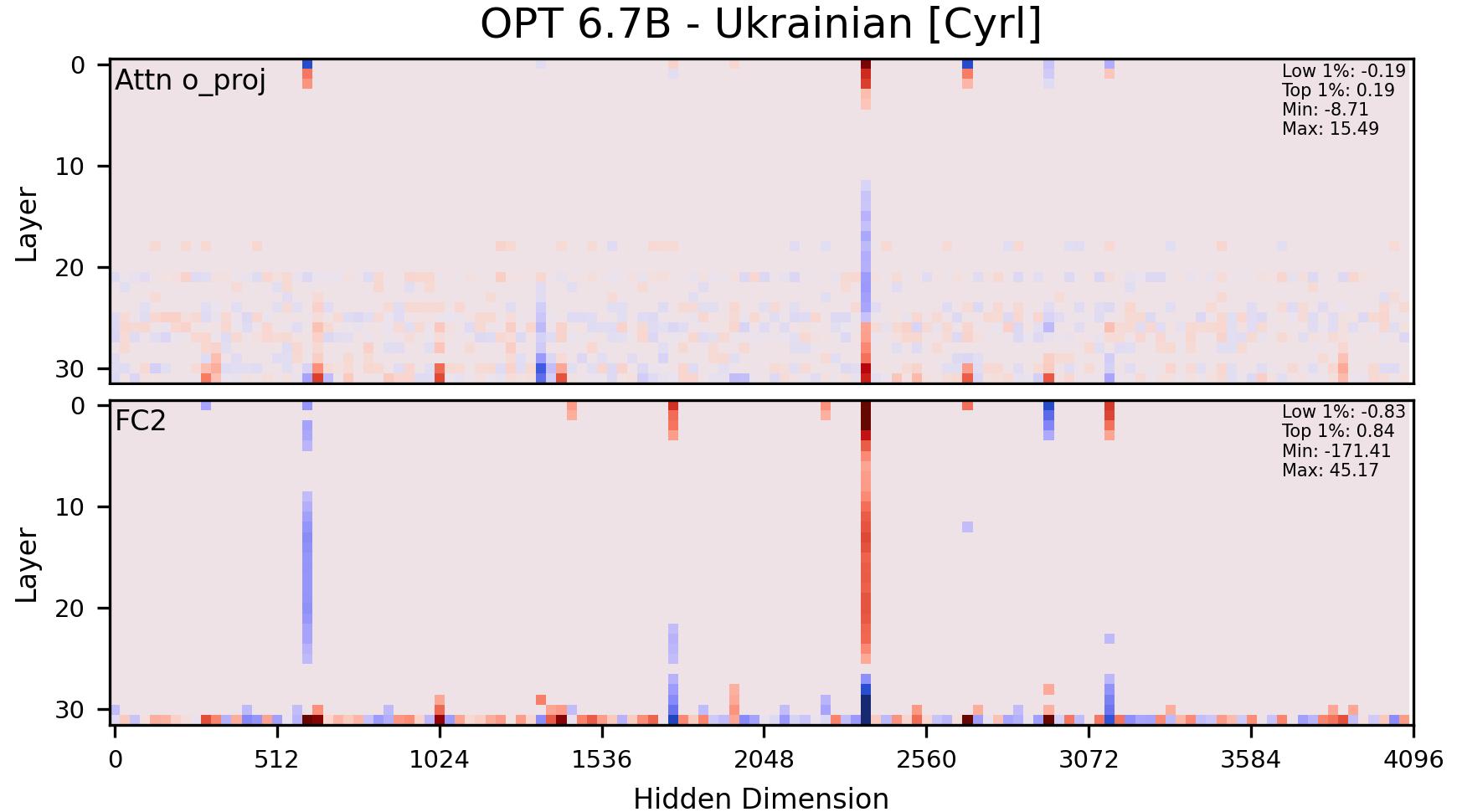} %
    \end{minipage}\hfill
    \begin{minipage}{0.33\textwidth}
        \centering
        \includegraphics[width=1.0\textwidth]{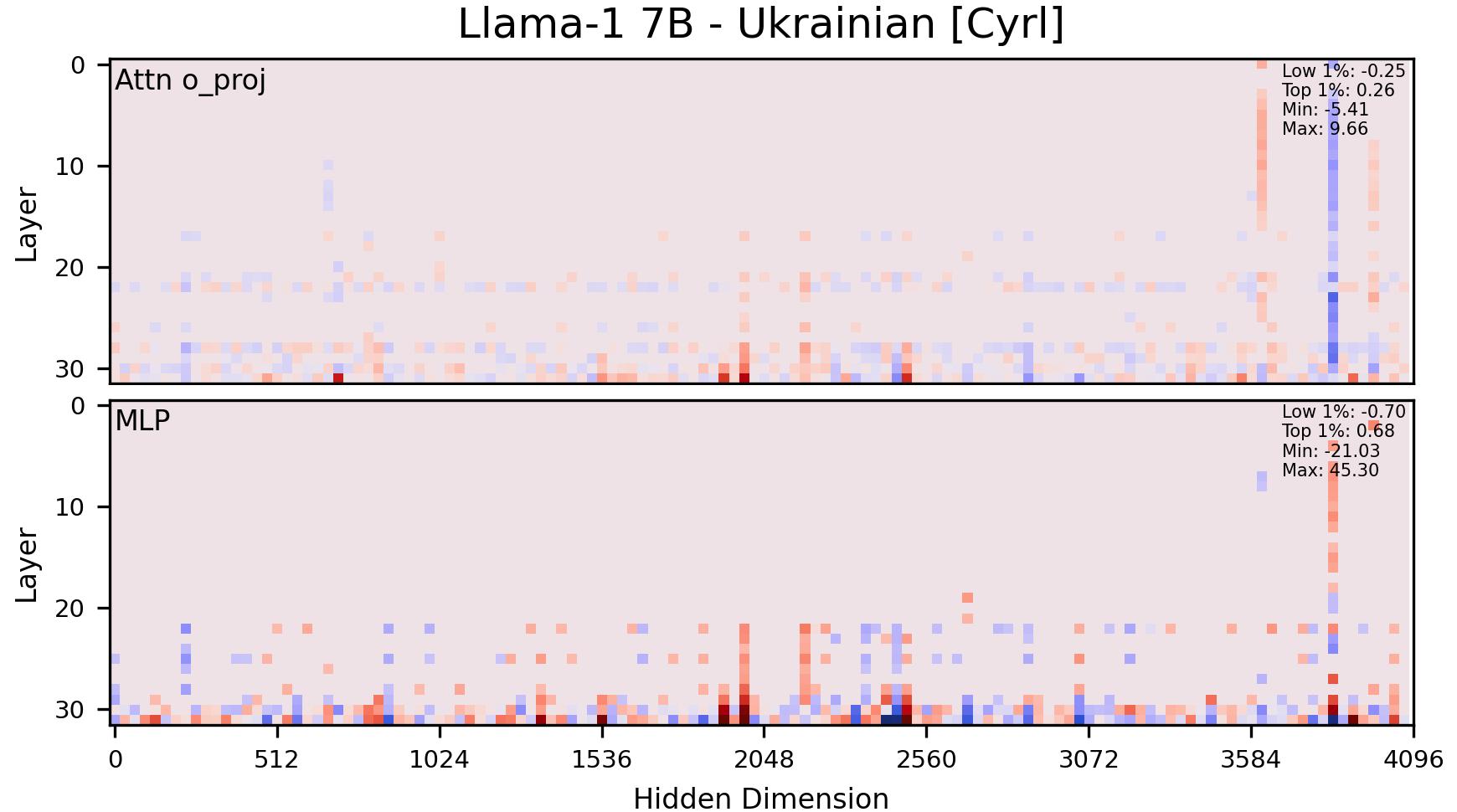} %
    \end{minipage}
    \begin{minipage}{0.33\textwidth}
        \centering
        \includegraphics[width=1.0\textwidth]{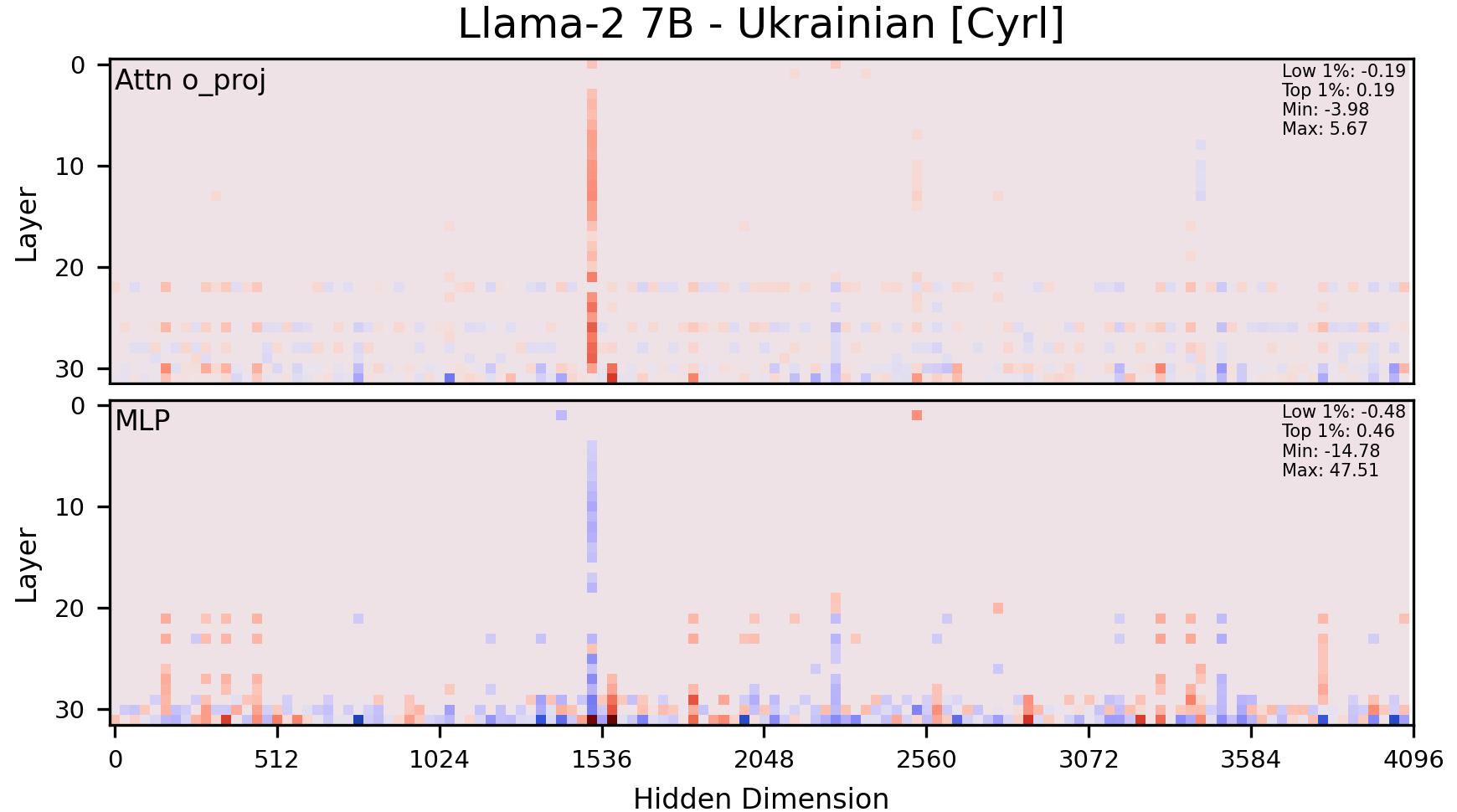} %
    \end{minipage}
    \vskip -0.3in
\end{figure}

\begin{figure}[H]
    \centering
    \begin{minipage}{0.33\textwidth}
        \centering
        \includegraphics[width=1.0\textwidth]{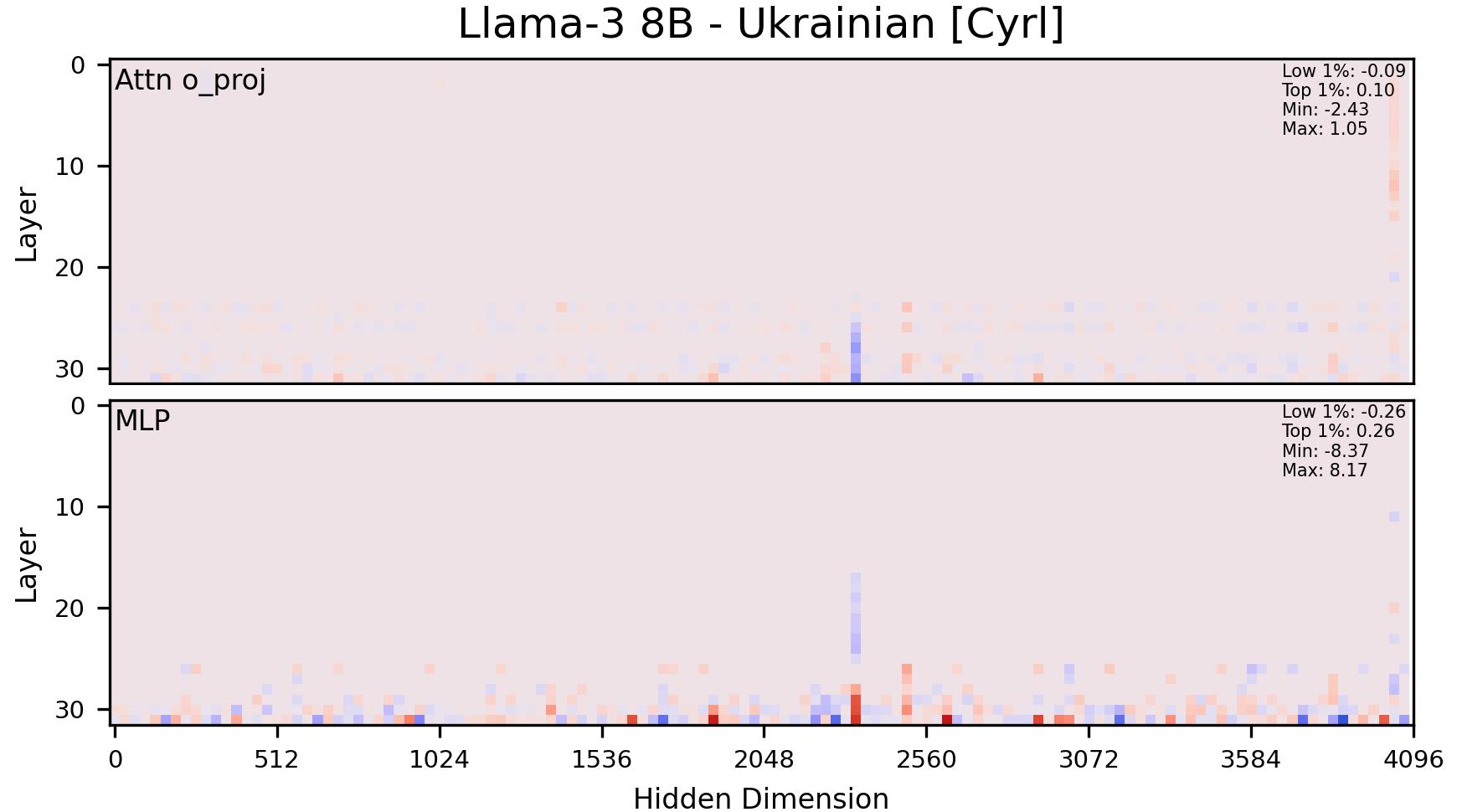} %
    \end{minipage}\hfill
    \begin{minipage}{0.33\textwidth}
        \centering
        \includegraphics[width=1.0\textwidth]{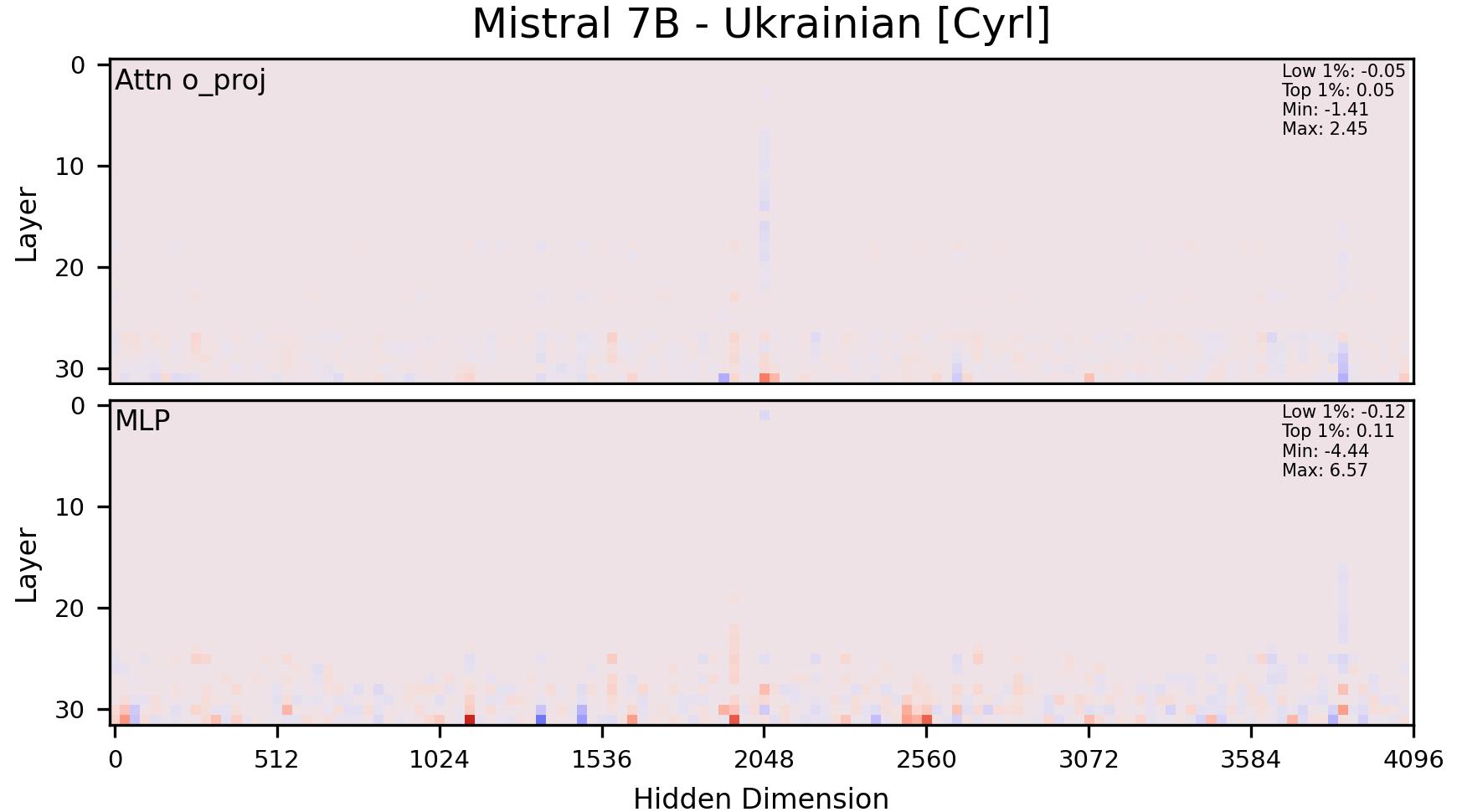} %
    \end{minipage}
    \begin{minipage}{0.33\textwidth}
        \centering
        \includegraphics[width=1.0\textwidth]{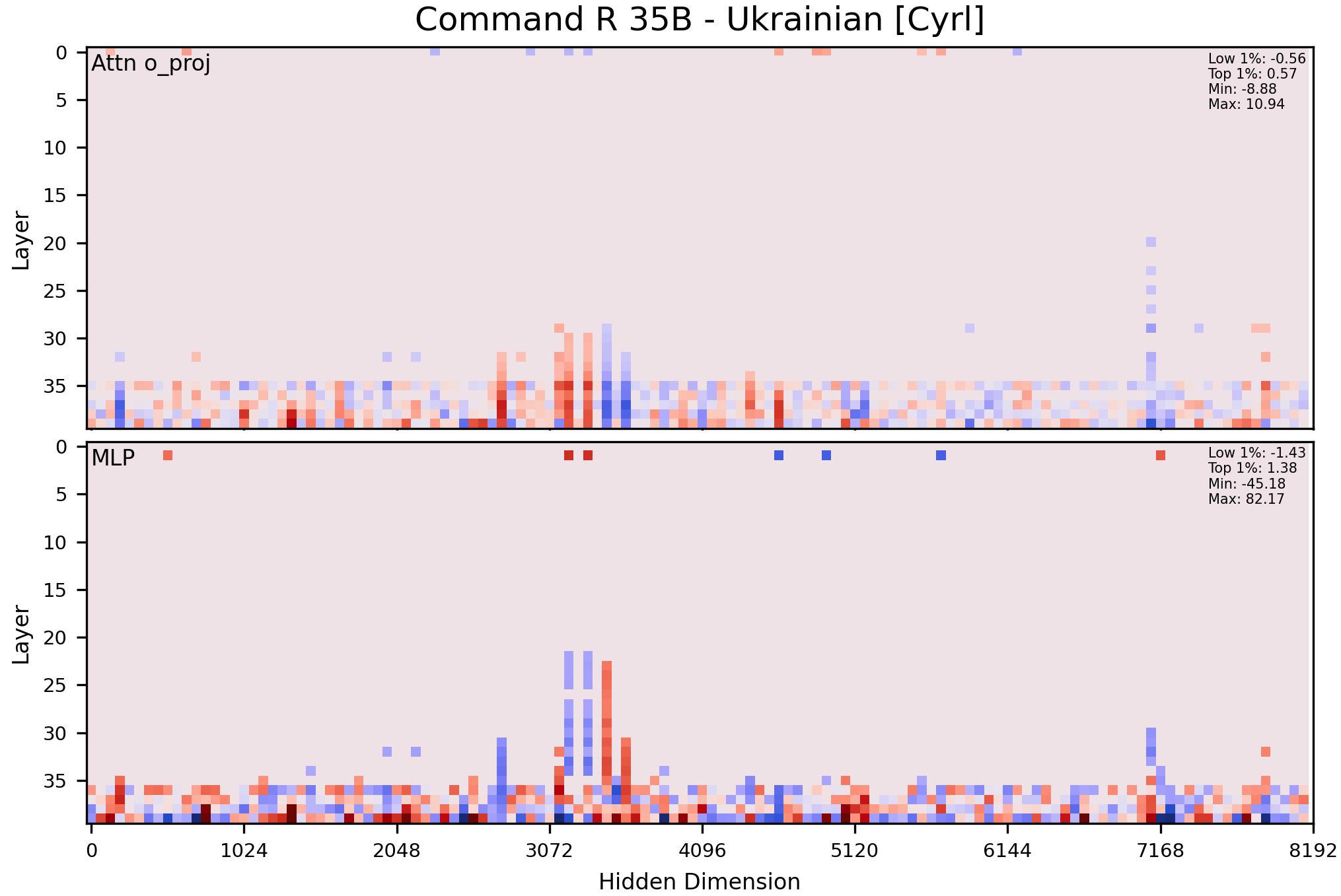} %
    \end{minipage}
\end{figure}

\begin{figure}[H]
    \centering
    \begin{minipage}{0.33\textwidth}
        \centering
        \includegraphics[width=1.0\textwidth]{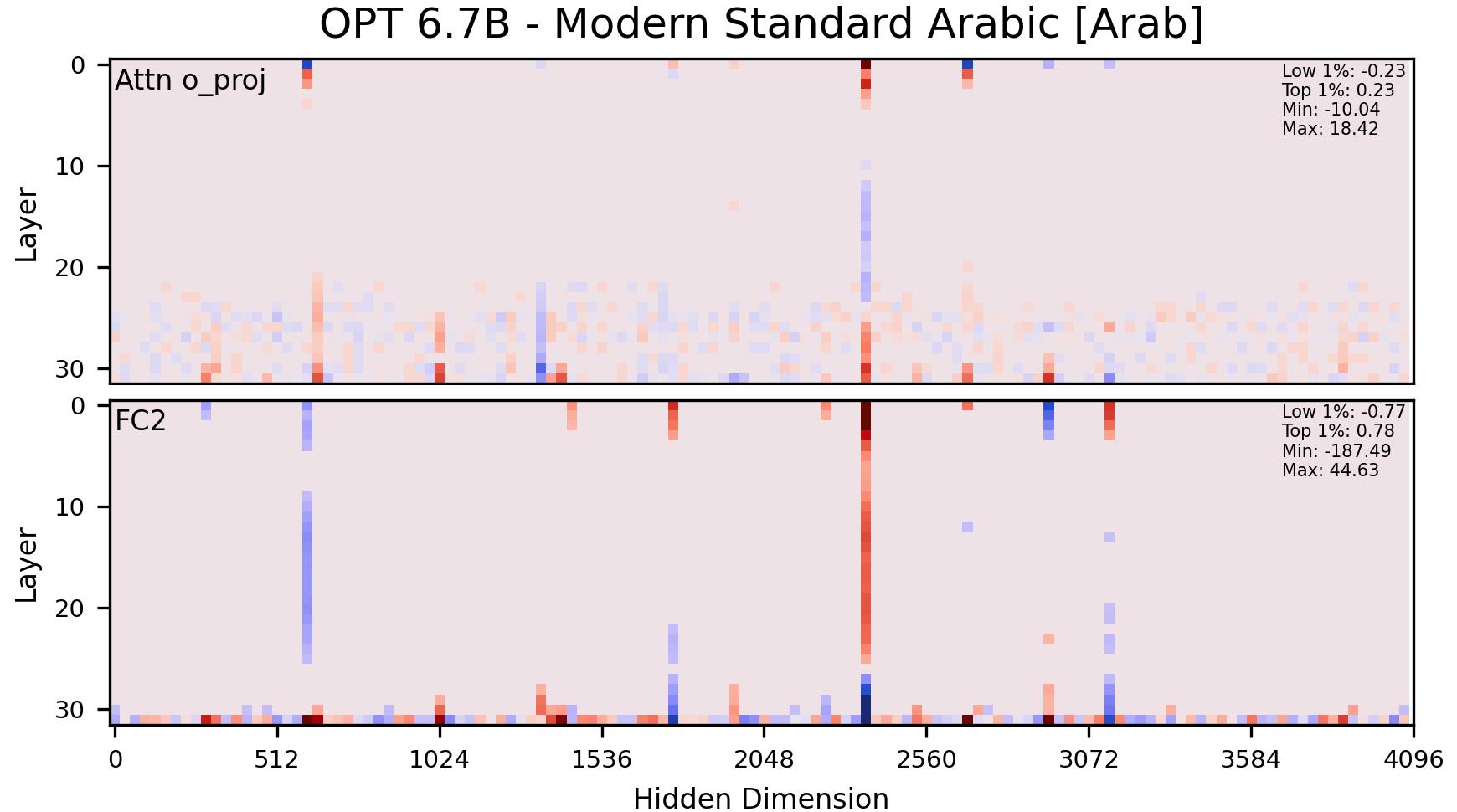} %
    \end{minipage}\hfill
    \begin{minipage}{0.33\textwidth}
        \centering
        \includegraphics[width=1.0\textwidth]{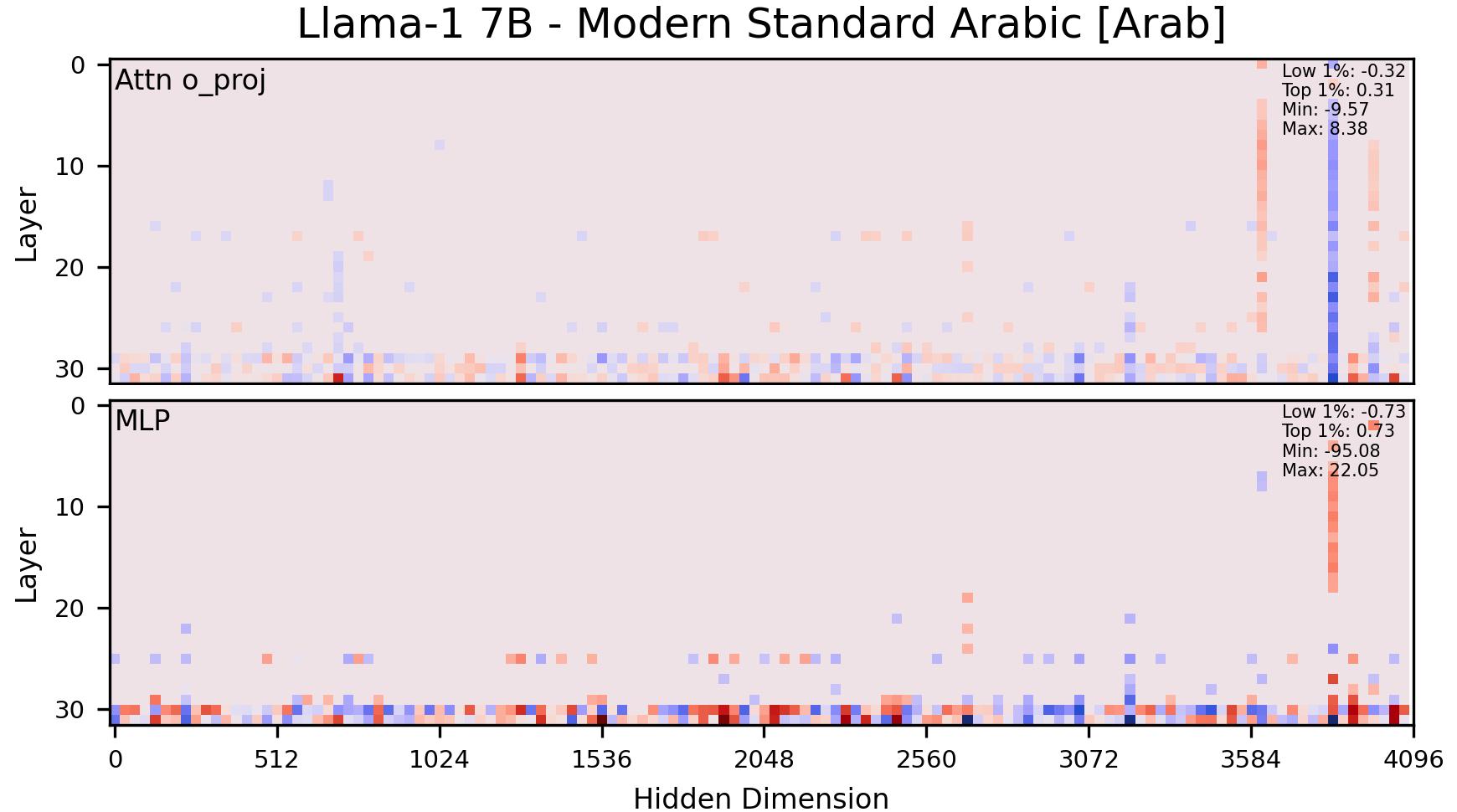} %
    \end{minipage}
    \begin{minipage}{0.33\textwidth}
        \centering
        \includegraphics[width=1.0\textwidth]{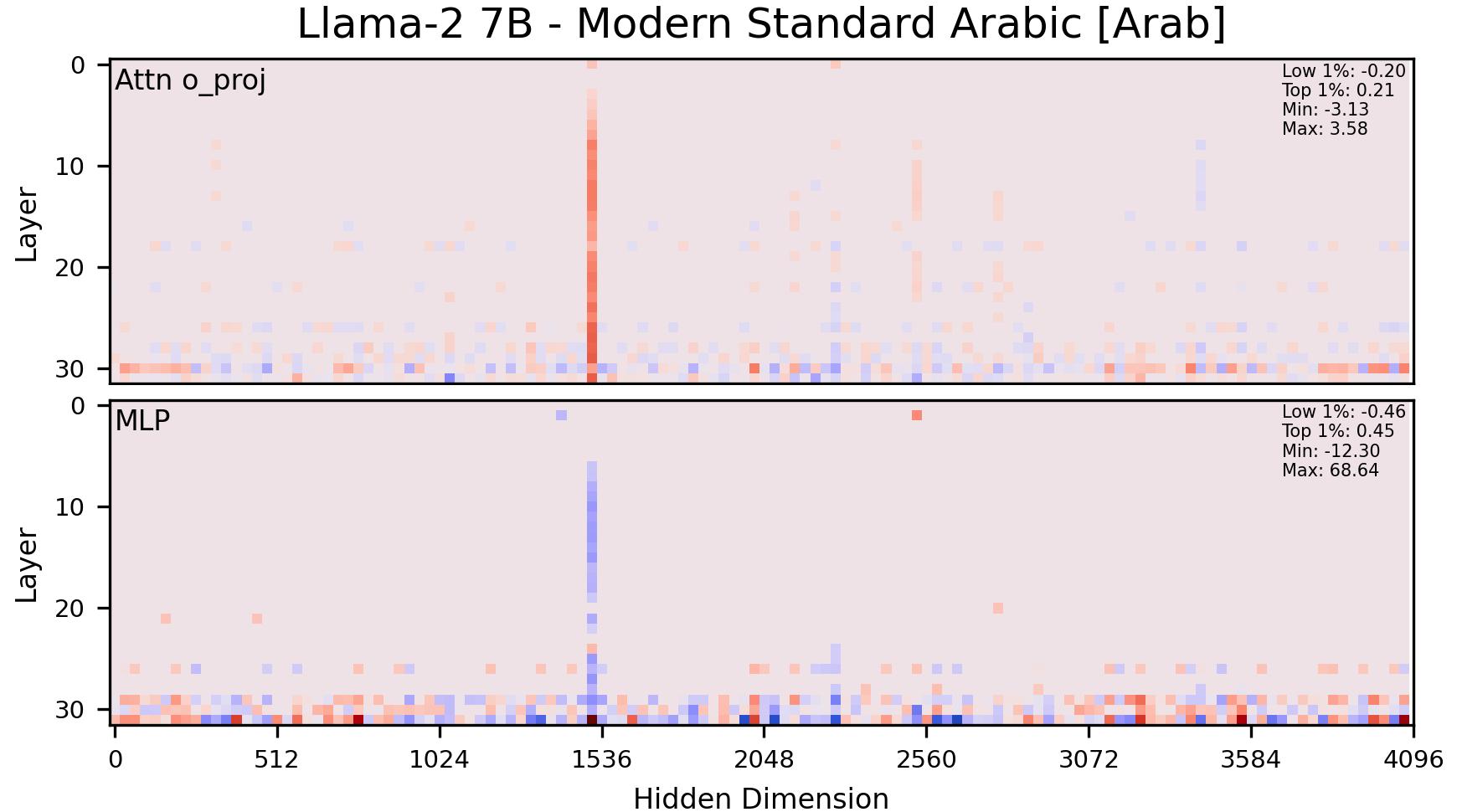} %
    \end{minipage}
    \vskip -0.3in
\end{figure}

\begin{figure}[H]
    \centering
    \begin{minipage}{0.33\textwidth}
        \centering
        \includegraphics[width=1.0\textwidth]{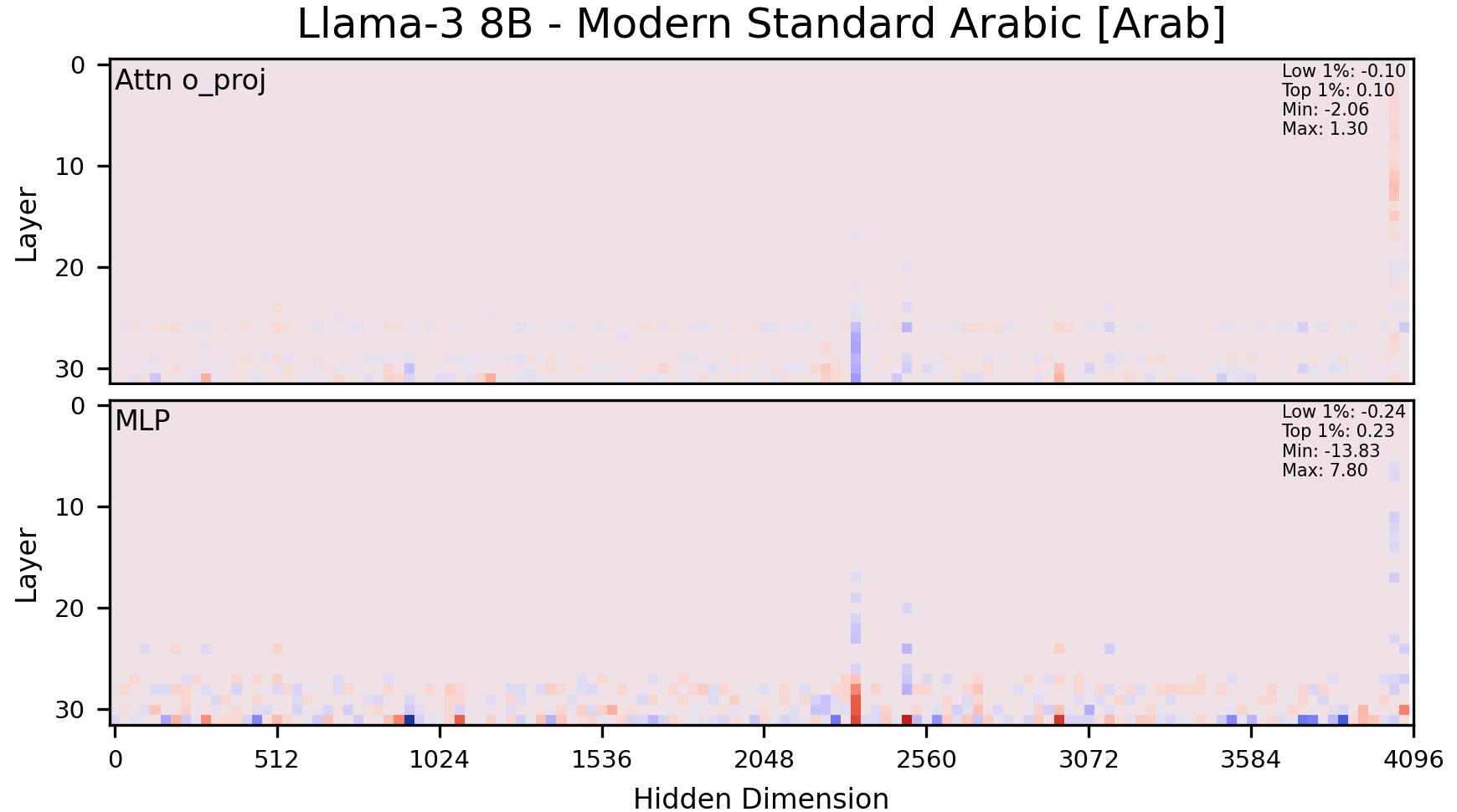} %
    \end{minipage}\hfill
    \begin{minipage}{0.33\textwidth}
        \centering
        \includegraphics[width=1.0\textwidth]{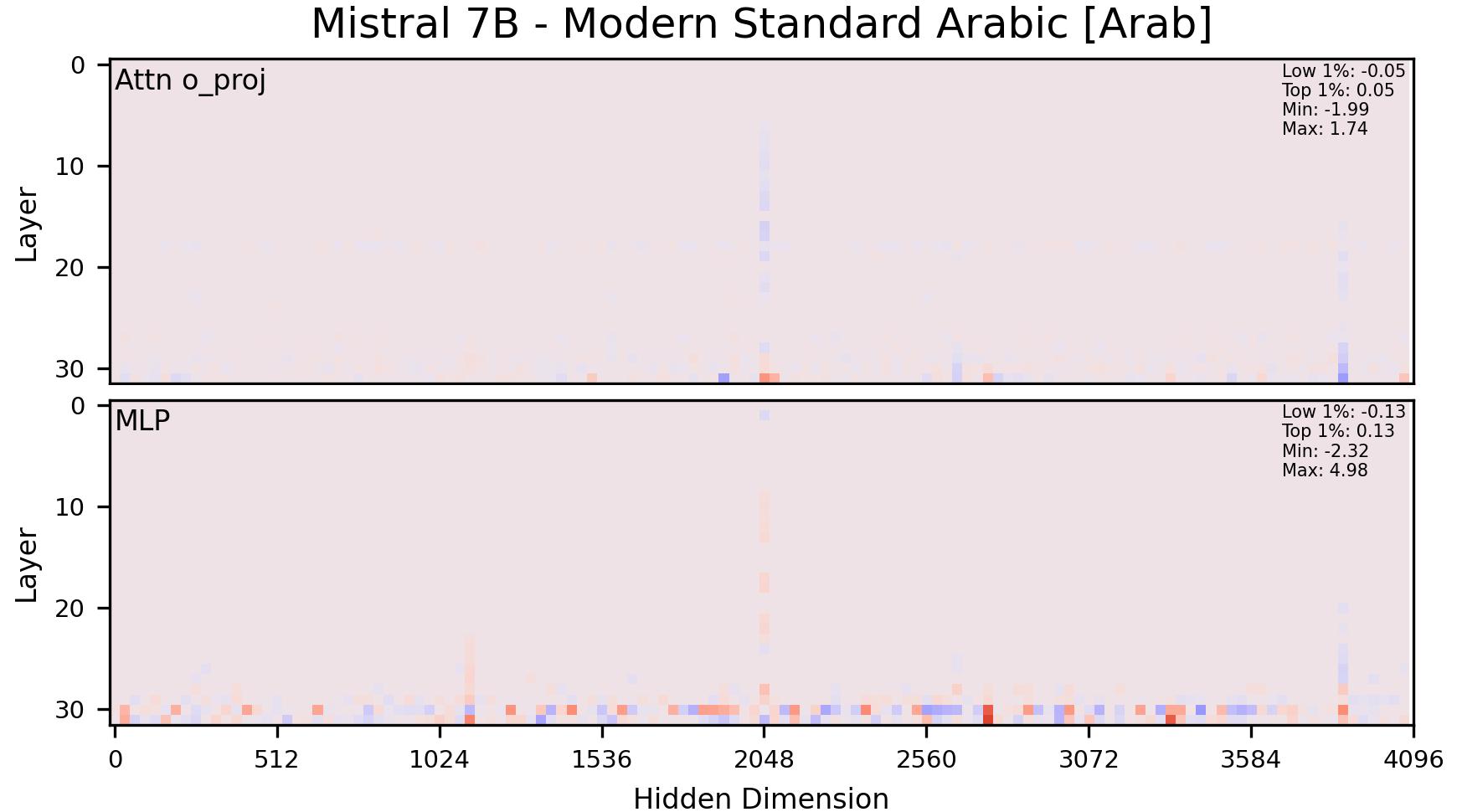} %
    \end{minipage}
    \begin{minipage}{0.33\textwidth}
        \centering
        \includegraphics[width=1.0\textwidth]{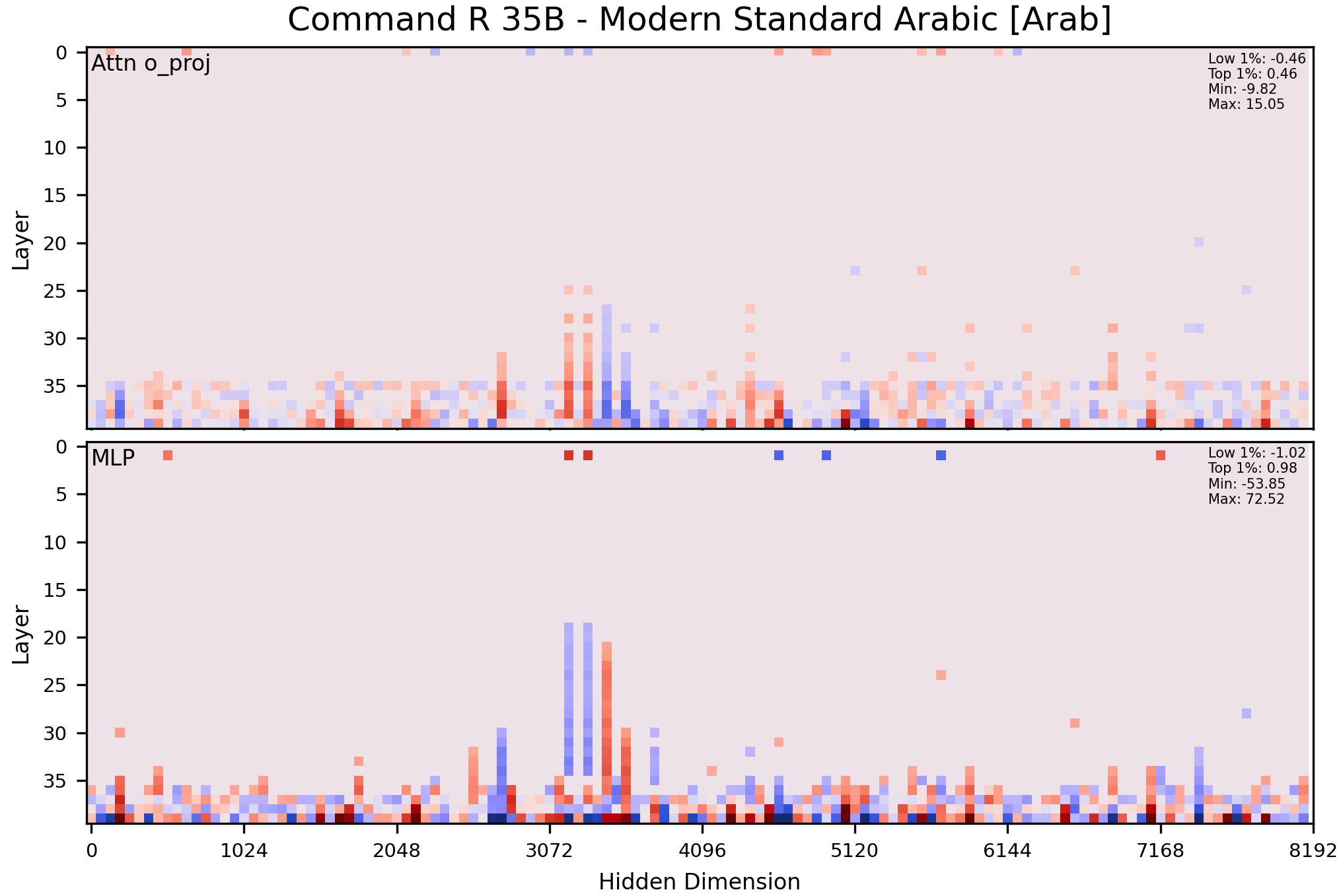} %
    \end{minipage}
\end{figure}

\begin{figure}[H]
    \centering
    \begin{minipage}{0.33\textwidth}
        \centering
        \includegraphics[width=1.0\textwidth]{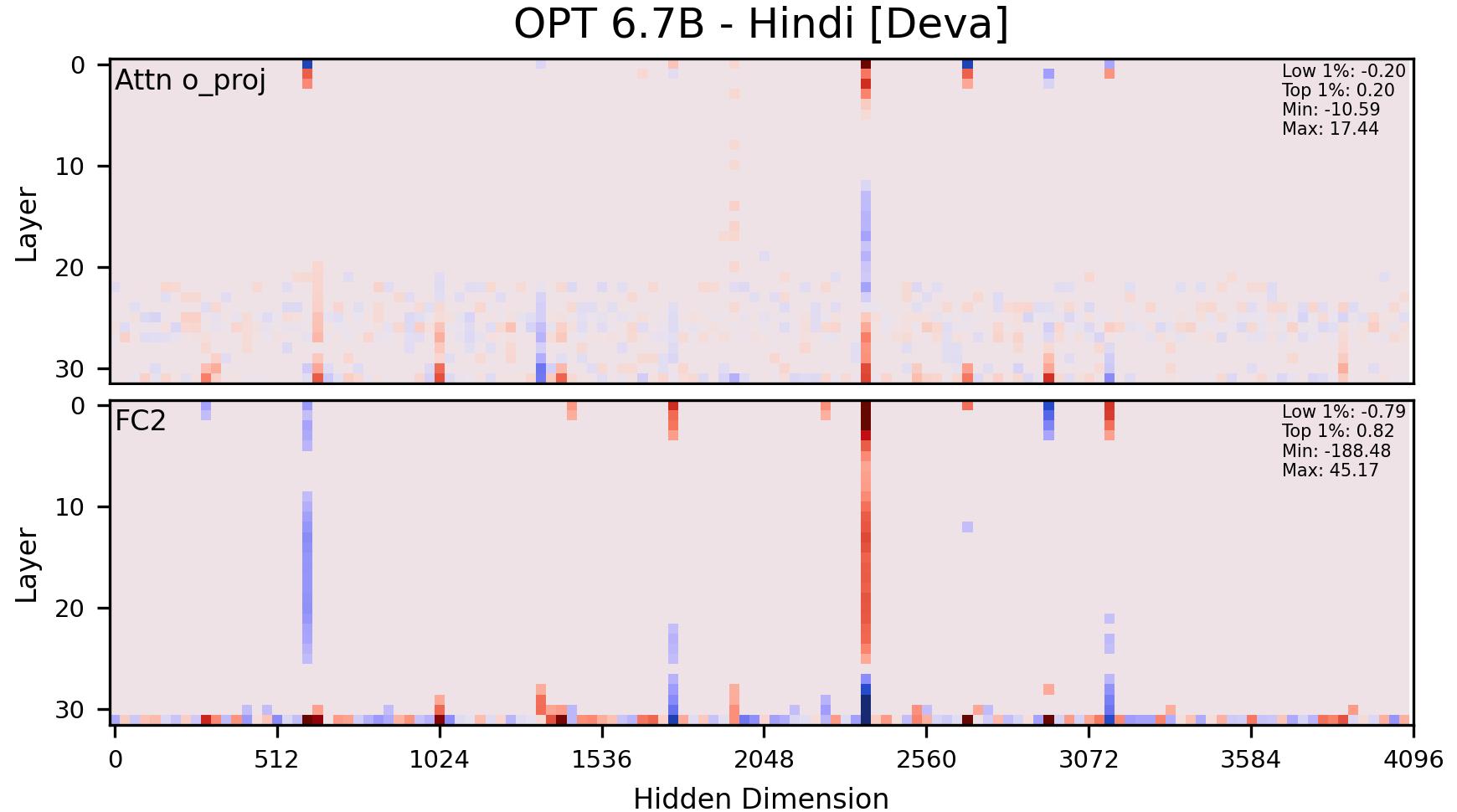} %
    \end{minipage}\hfill
    \begin{minipage}{0.33\textwidth}
        \centering
        \includegraphics[width=1.0\textwidth]{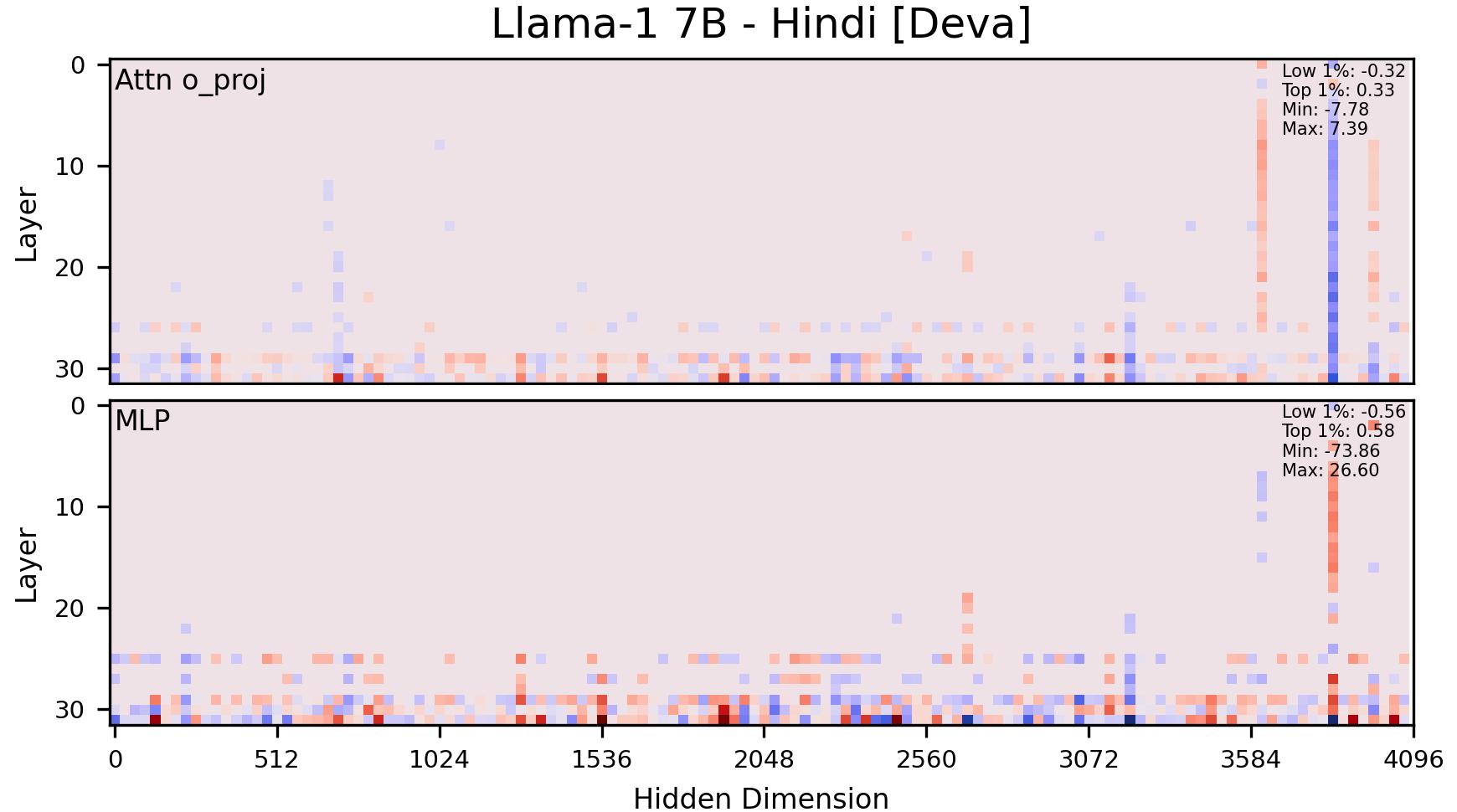} %
    \end{minipage}
    \begin{minipage}{0.33\textwidth}
        \centering
        \includegraphics[width=1.0\textwidth]{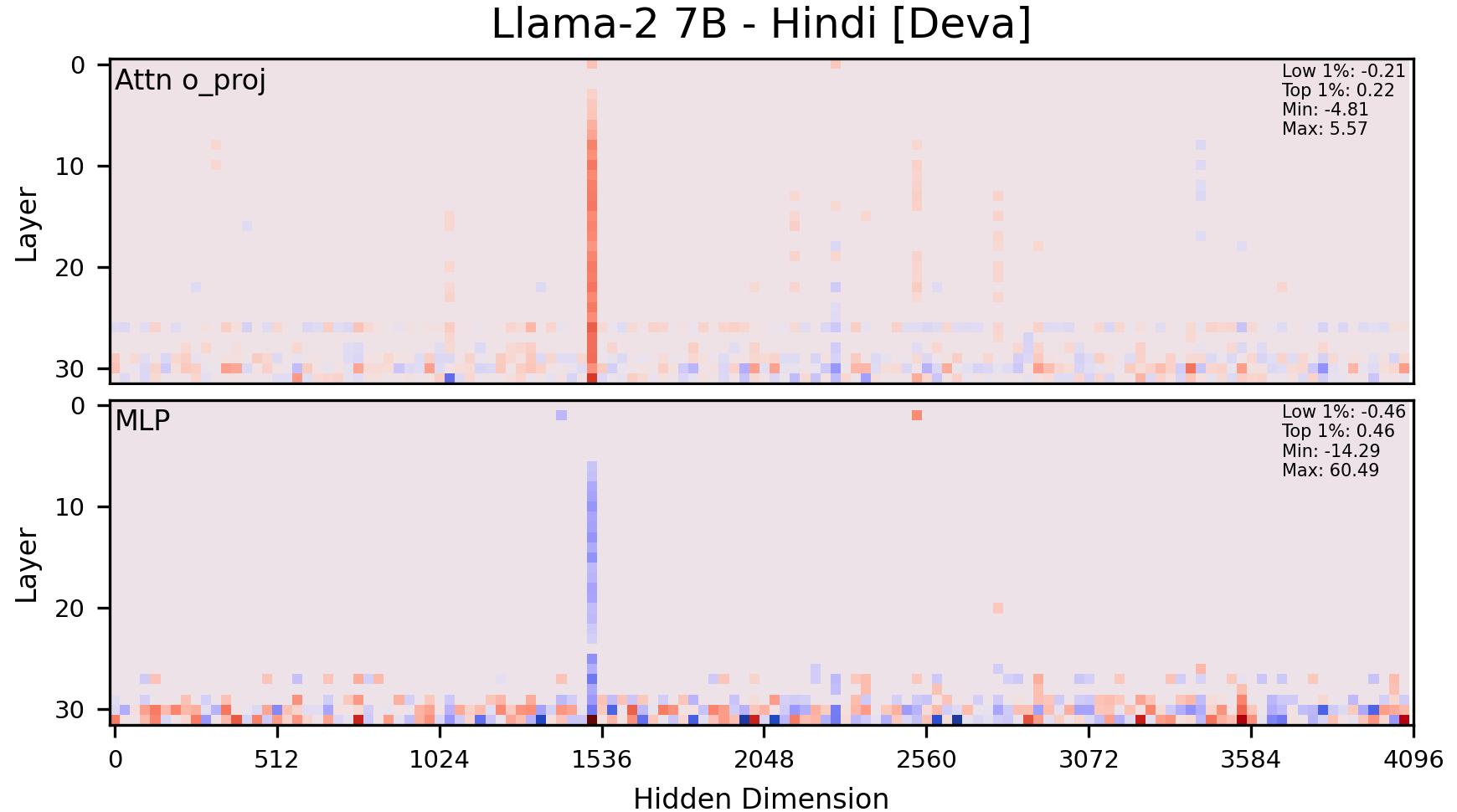} %
    \end{minipage}
    \vskip -0.3in
\end{figure}

\begin{figure}[H]
    \centering
    \begin{minipage}{0.33\textwidth}
        \centering
        \includegraphics[width=1.0\textwidth]{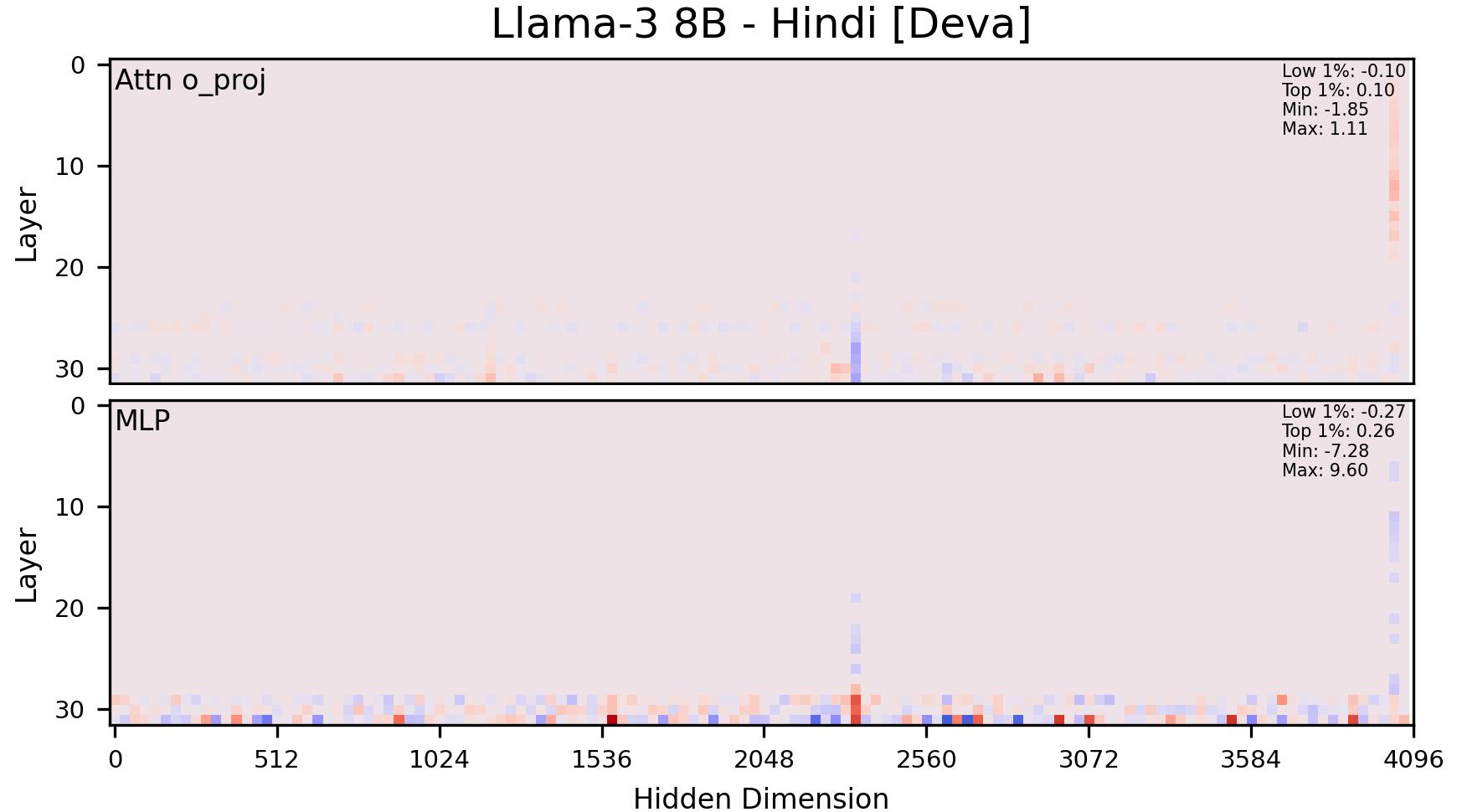} %
    \end{minipage}\hfill
    \begin{minipage}{0.33\textwidth}
        \centering
        \includegraphics[width=1.0\textwidth]{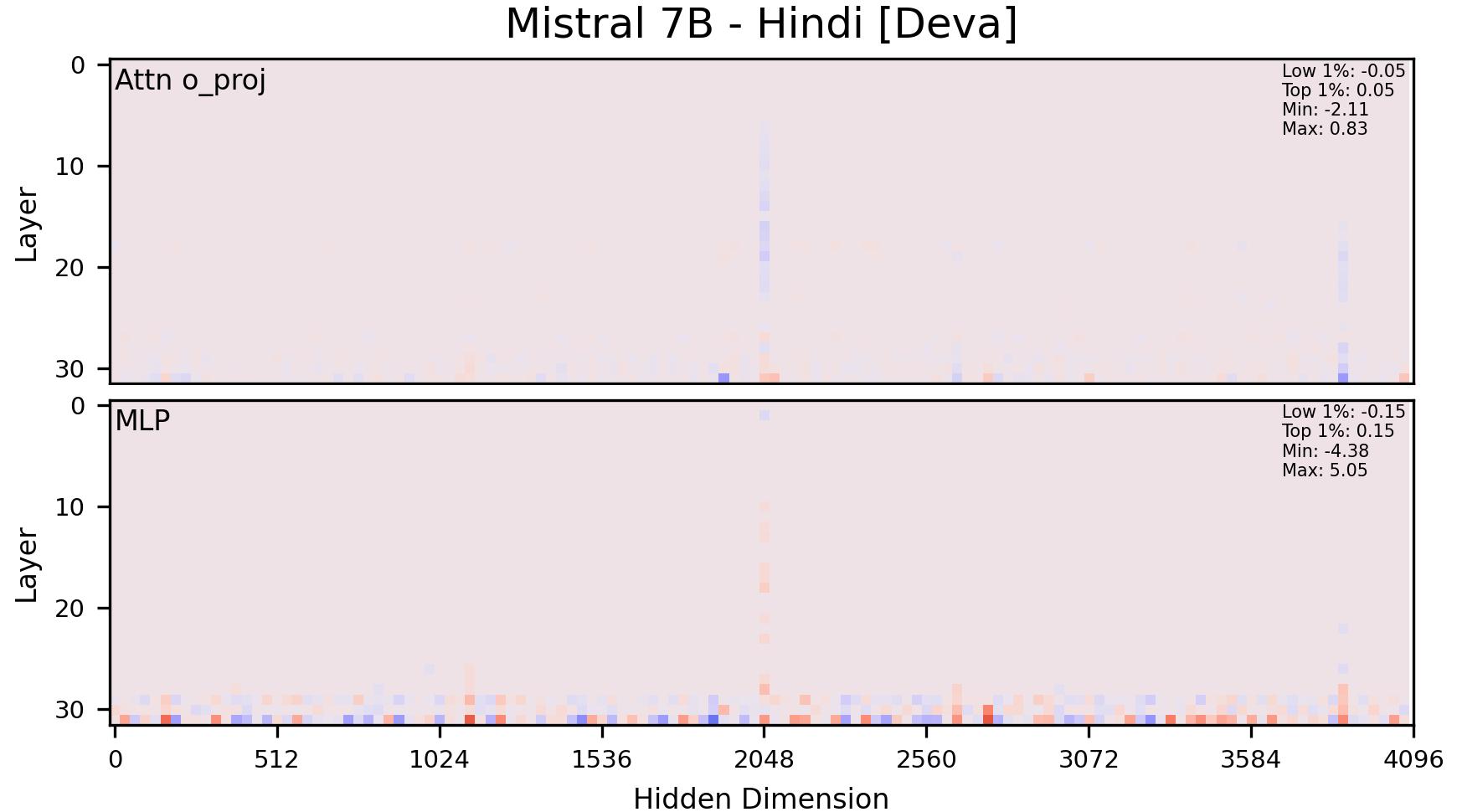} %
    \end{minipage}
    \begin{minipage}{0.33\textwidth}
        \centering
        \includegraphics[width=1.0\textwidth]{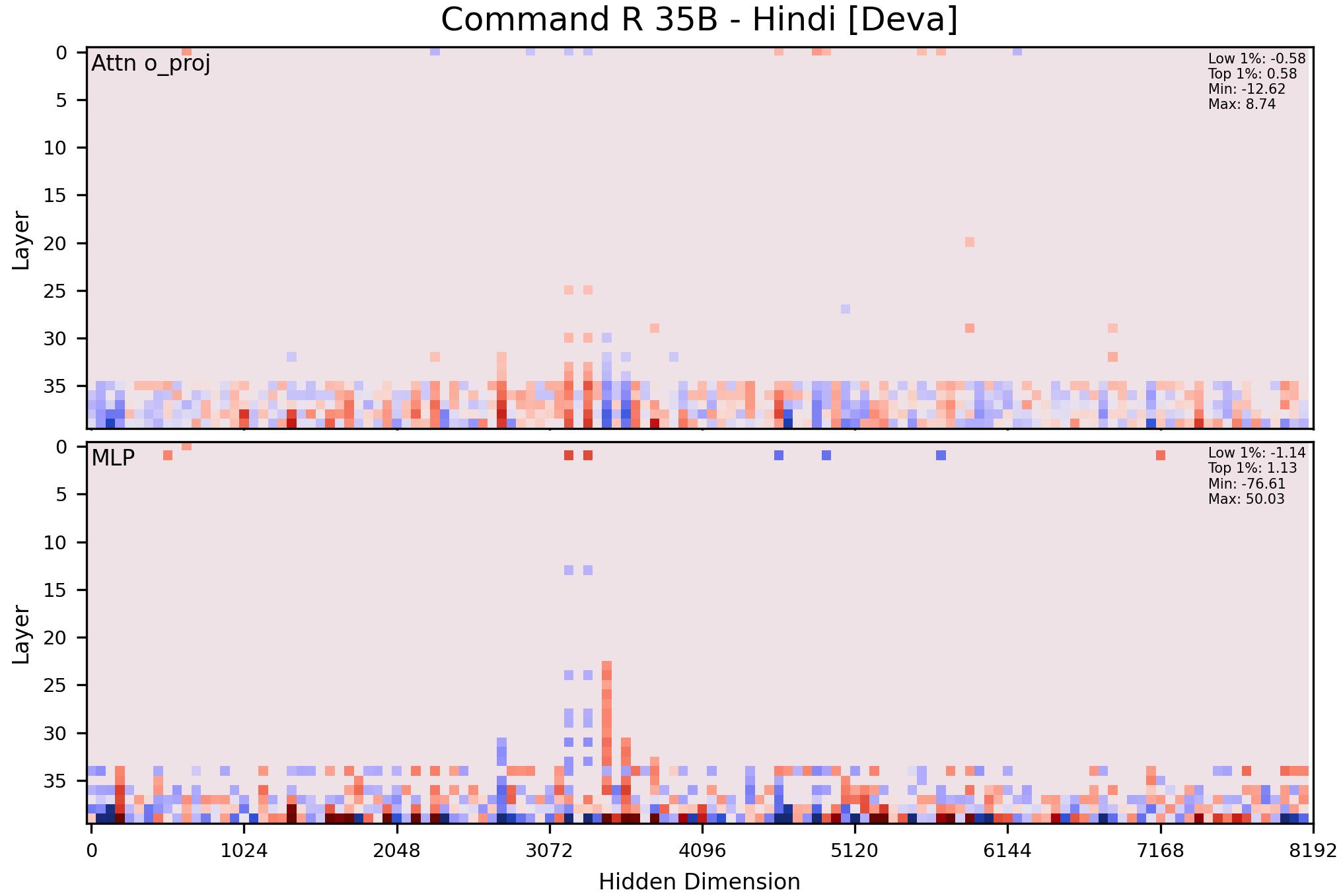} %
    \end{minipage}
\end{figure}

\begin{figure}[H]
    \centering
    \begin{minipage}{0.33\textwidth}
        \centering
        \includegraphics[width=1.0\textwidth]{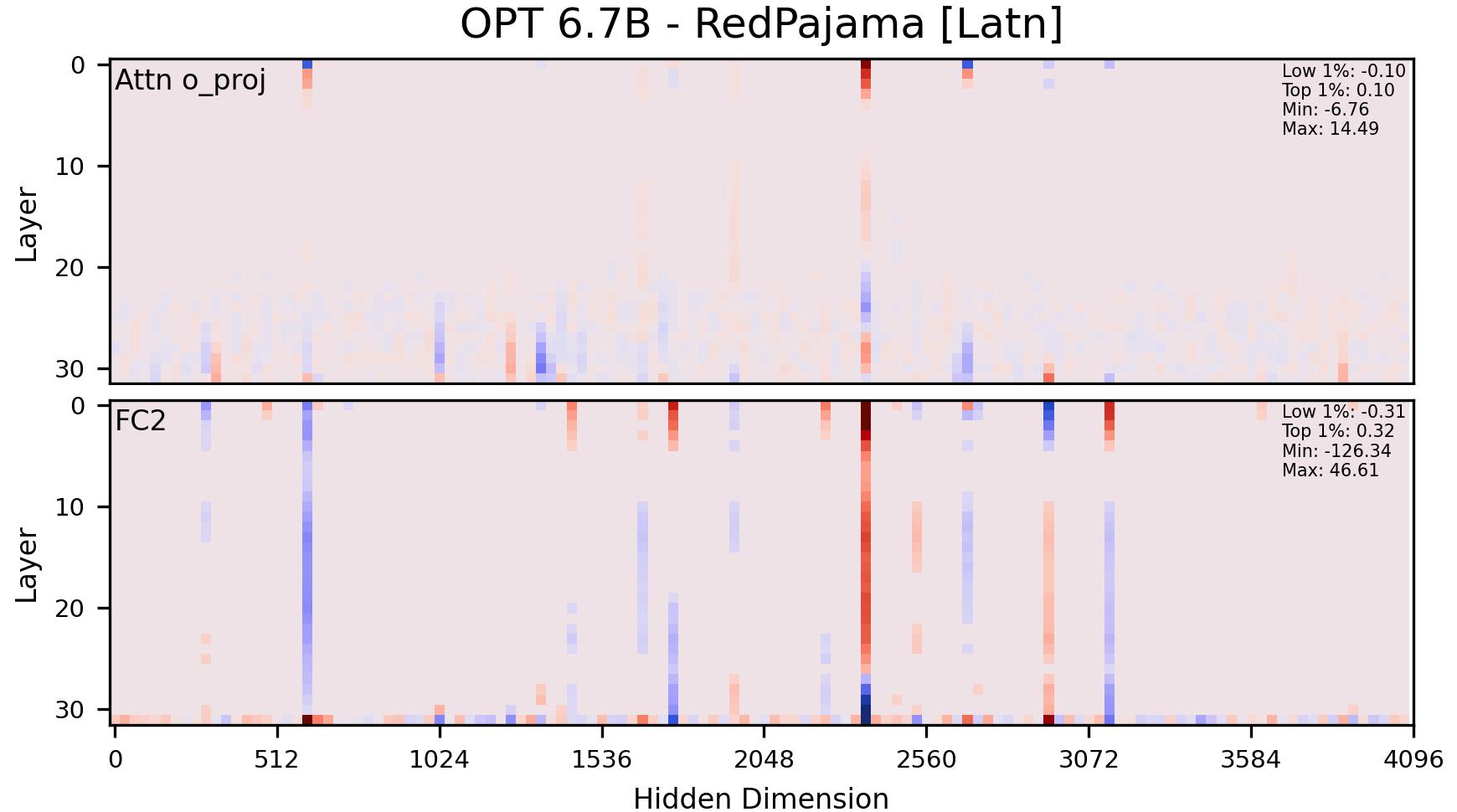} %
    \end{minipage}\hfill
    \begin{minipage}{0.33\textwidth}
        \centering
        \includegraphics[width=1.0\textwidth]{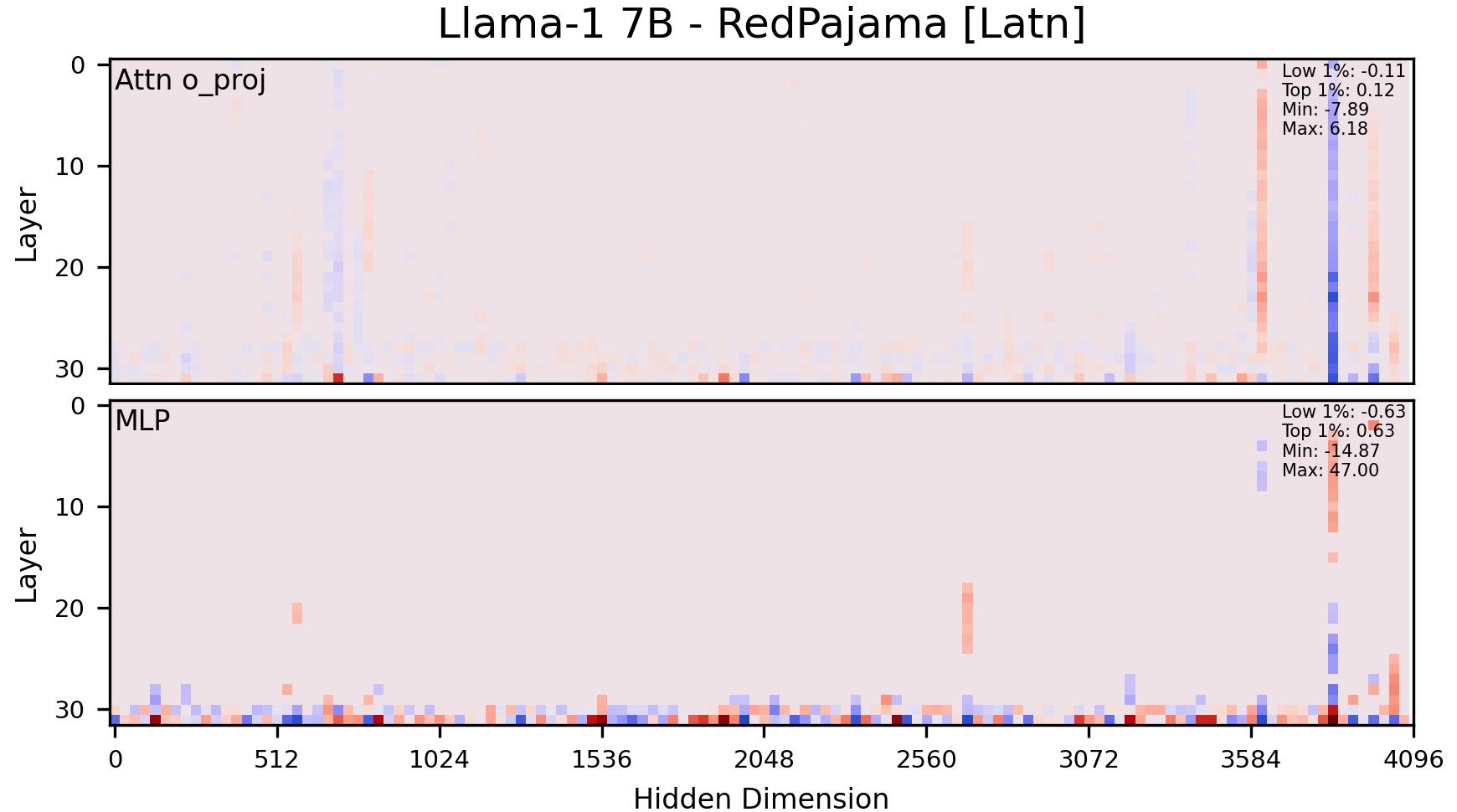} %
    \end{minipage}
    \begin{minipage}{0.33\textwidth}
        \centering
        \includegraphics[width=1.0\textwidth]{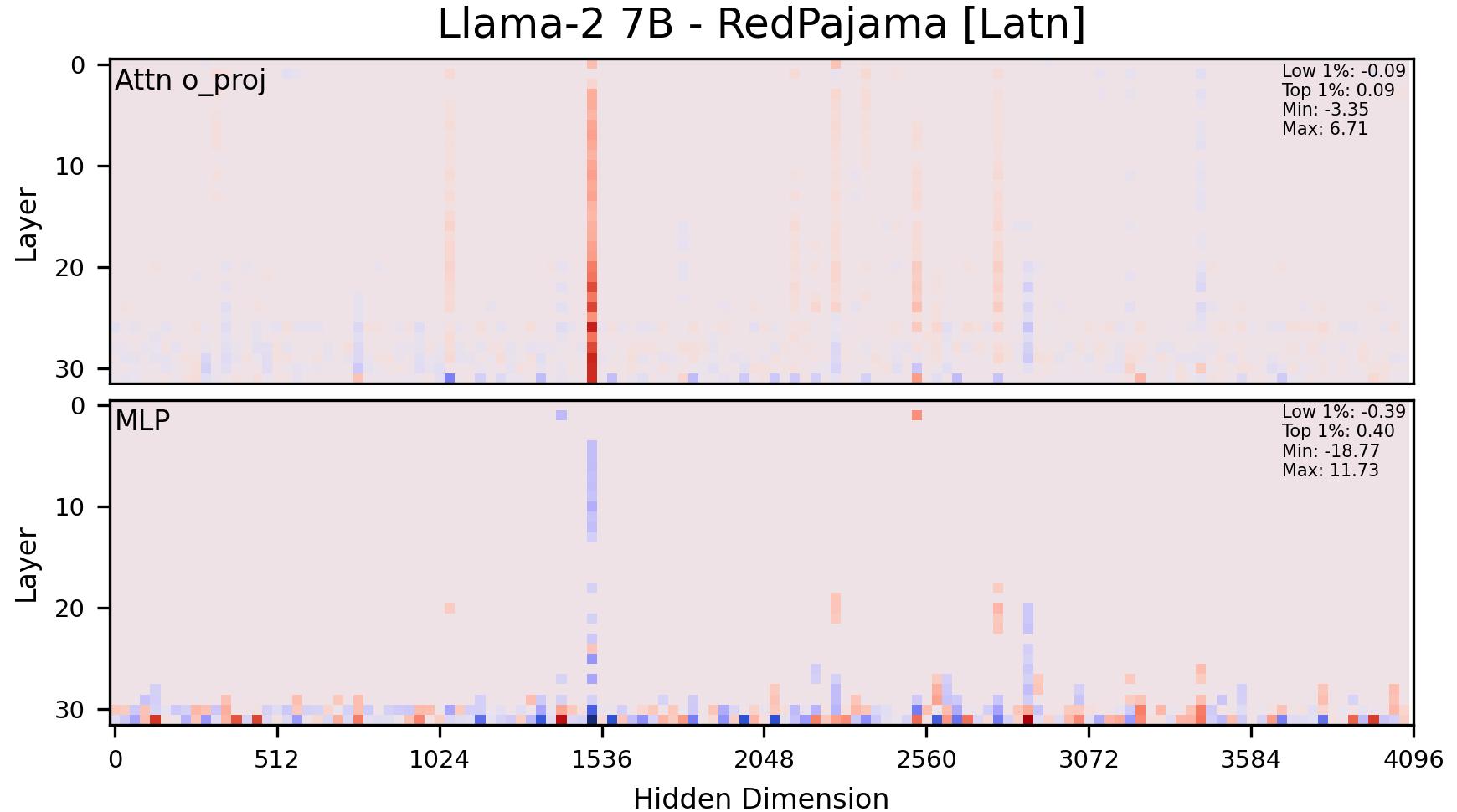} %
    \end{minipage}
    \vskip -0.3in
\end{figure}

\begin{figure}[H]
    \centering
    \begin{minipage}{0.33\textwidth}
        \centering
        \includegraphics[width=1.0\textwidth]{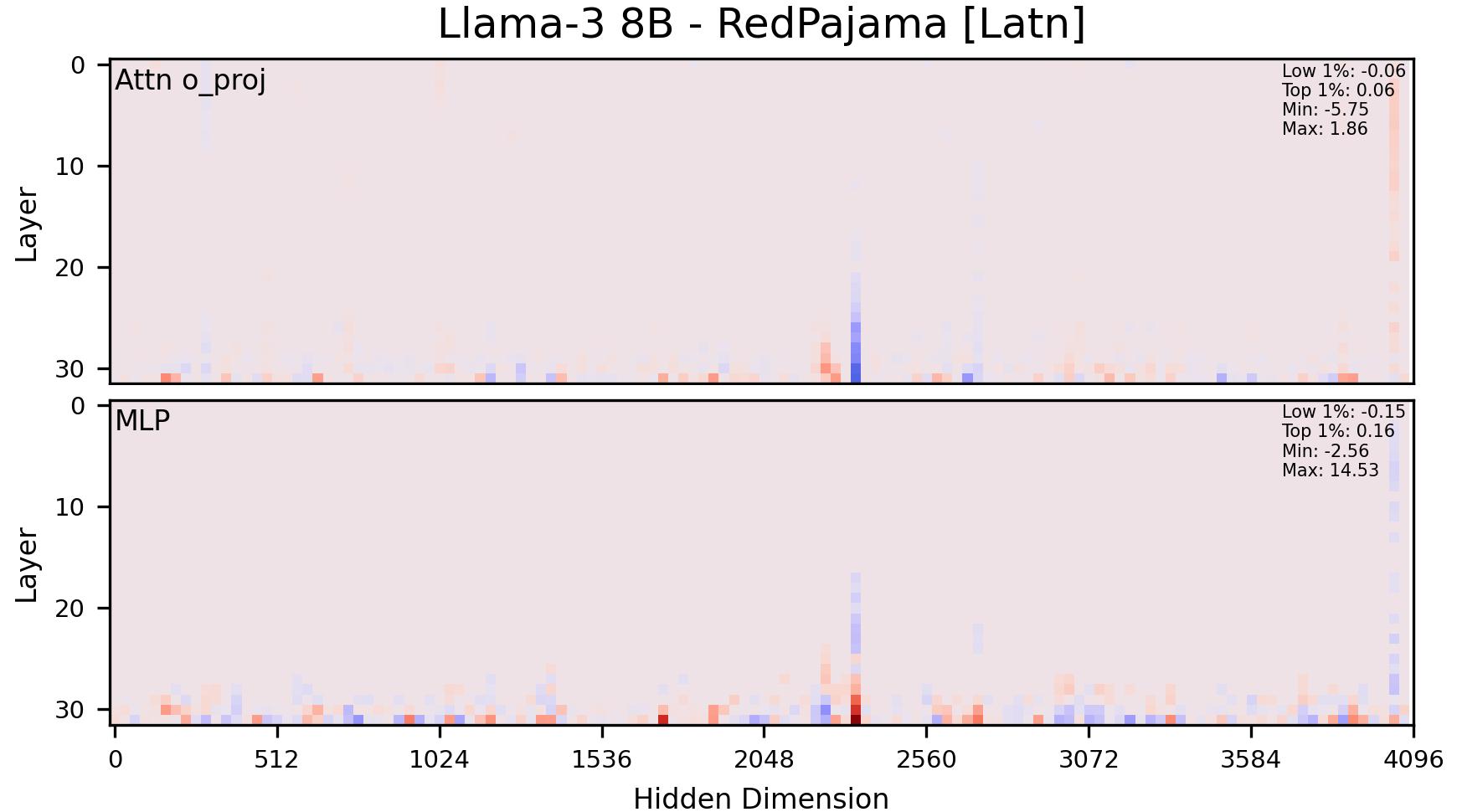} %
    \end{minipage}\hfill
    \begin{minipage}{0.33\textwidth}
        \centering
        \includegraphics[width=1.0\textwidth]{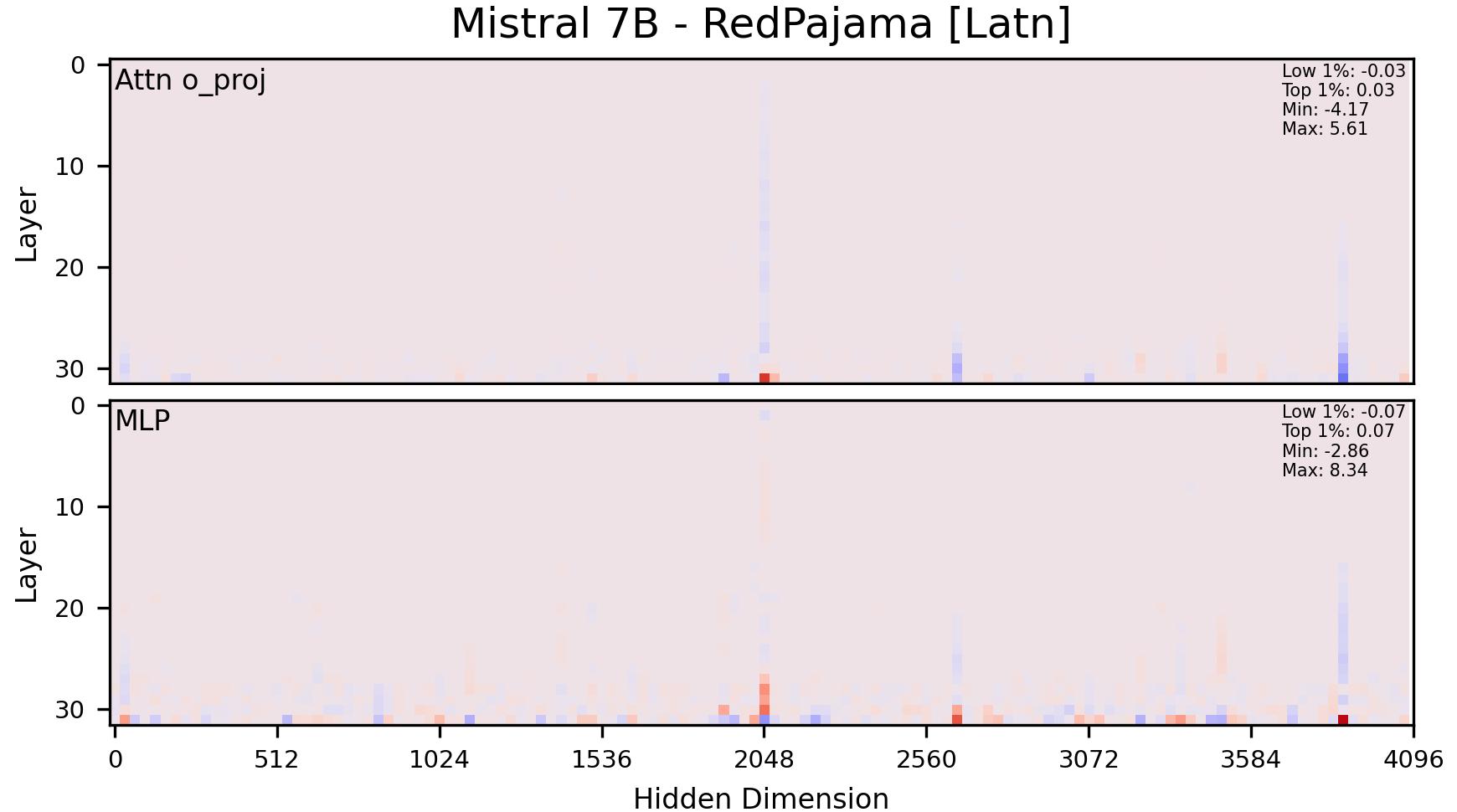} %
    \end{minipage}
    \begin{minipage}{0.33\textwidth}
        \centering
        \includegraphics[width=1.0\textwidth]{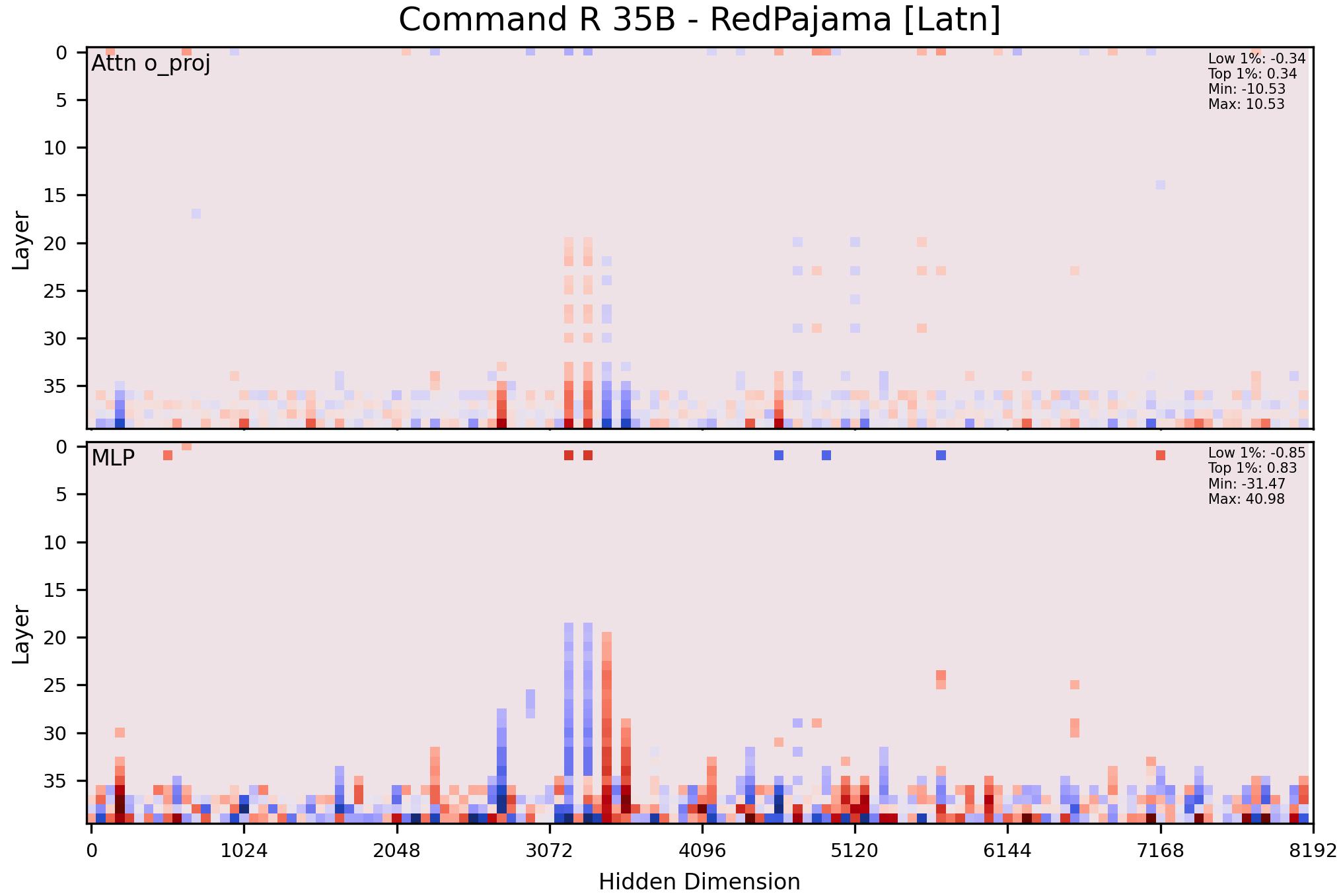} %
    \end{minipage}
\end{figure}

\begin{figure}[H]
    \centering
    \begin{minipage}{0.33\textwidth}
        \centering
        \includegraphics[width=1.0\textwidth]{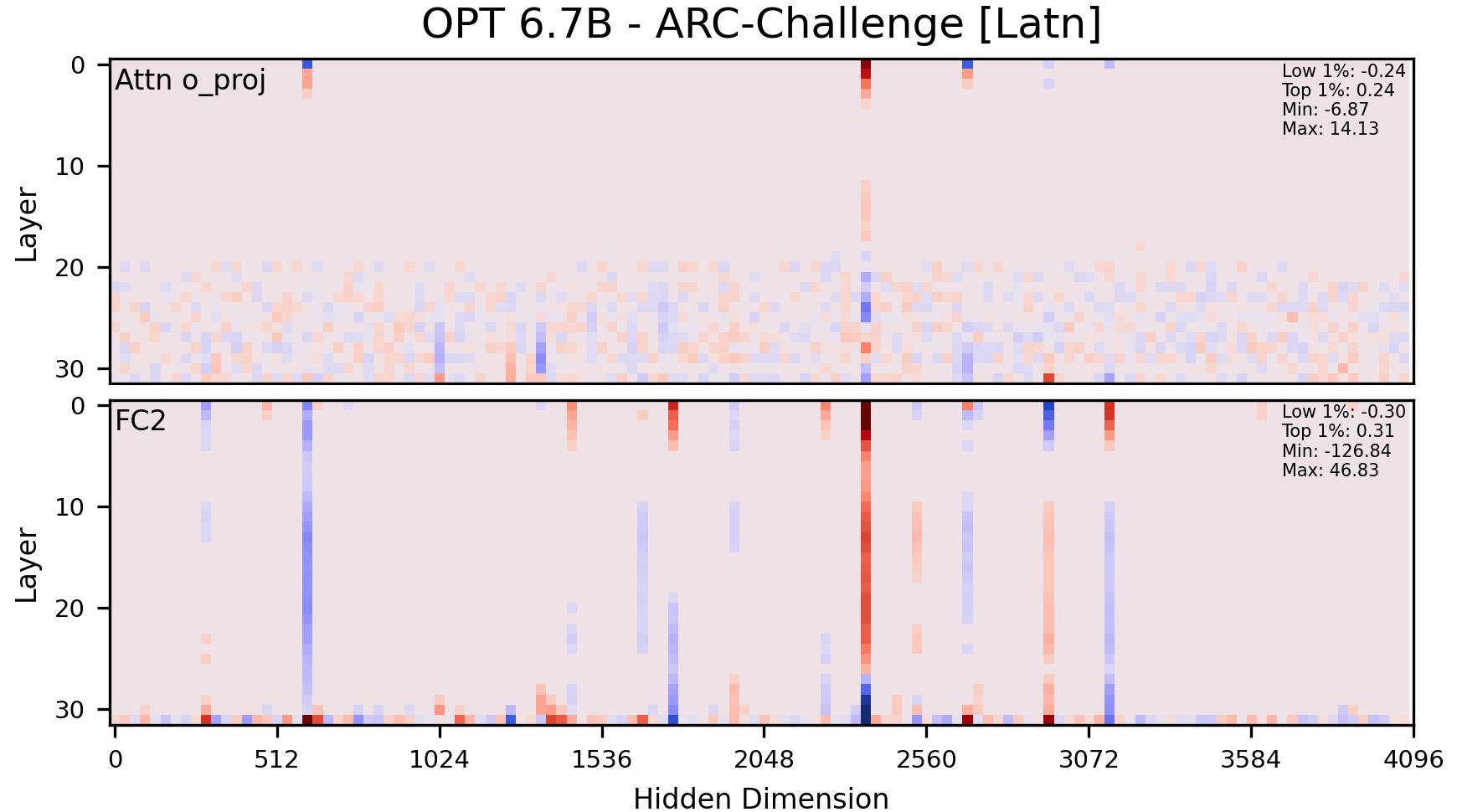} %
    \end{minipage}\hfill
    \begin{minipage}{0.33\textwidth}
        \centering
        \includegraphics[width=1.0\textwidth]{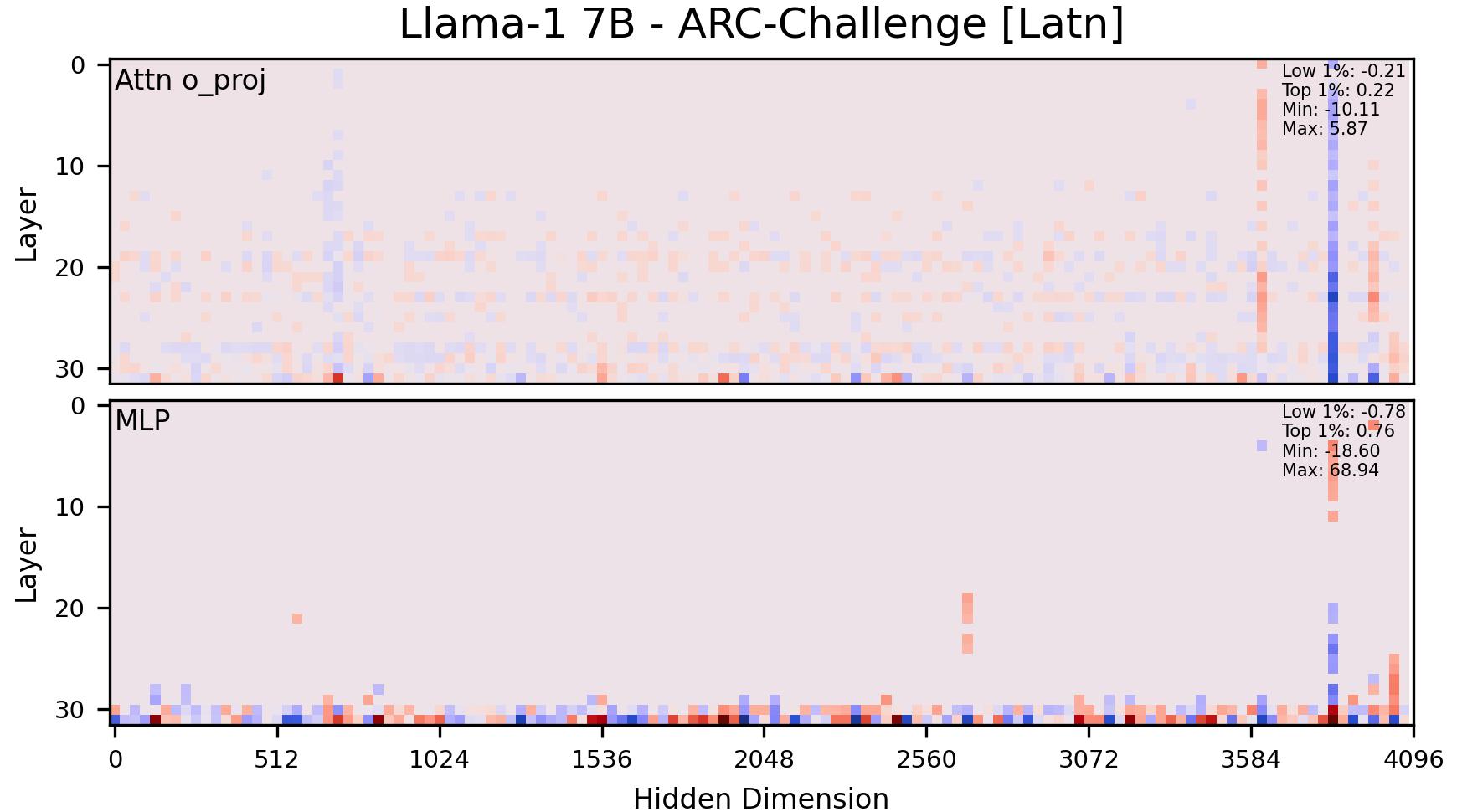} %
    \end{minipage}
    \begin{minipage}{0.33\textwidth}
        \centering
        \includegraphics[width=1.0\textwidth]{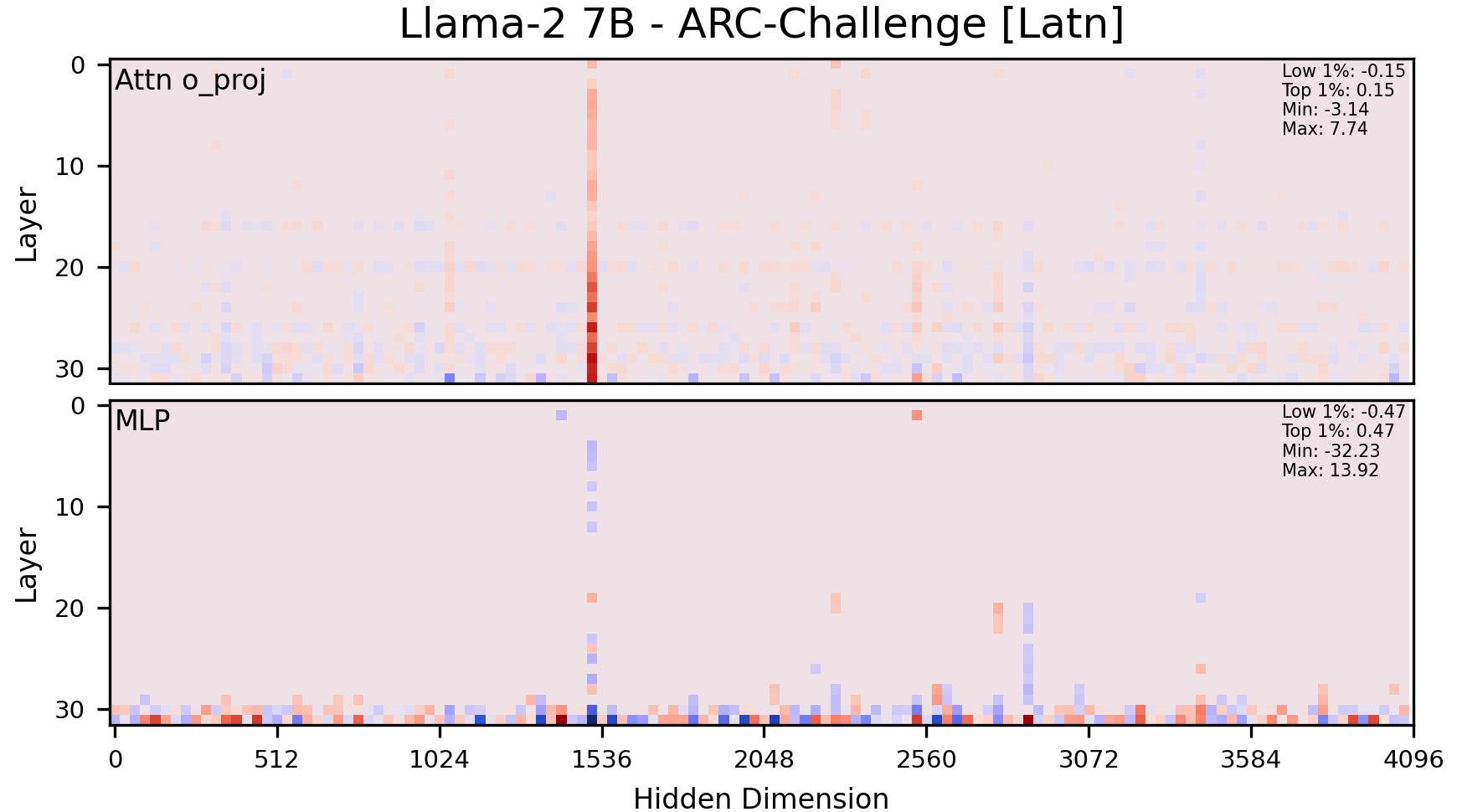} %
    \end{minipage}
    \vskip -0.3in
\end{figure}

\begin{figure}[H]
    \centering
    \begin{minipage}{0.33\textwidth}
        \centering
        \includegraphics[width=1.0\textwidth]{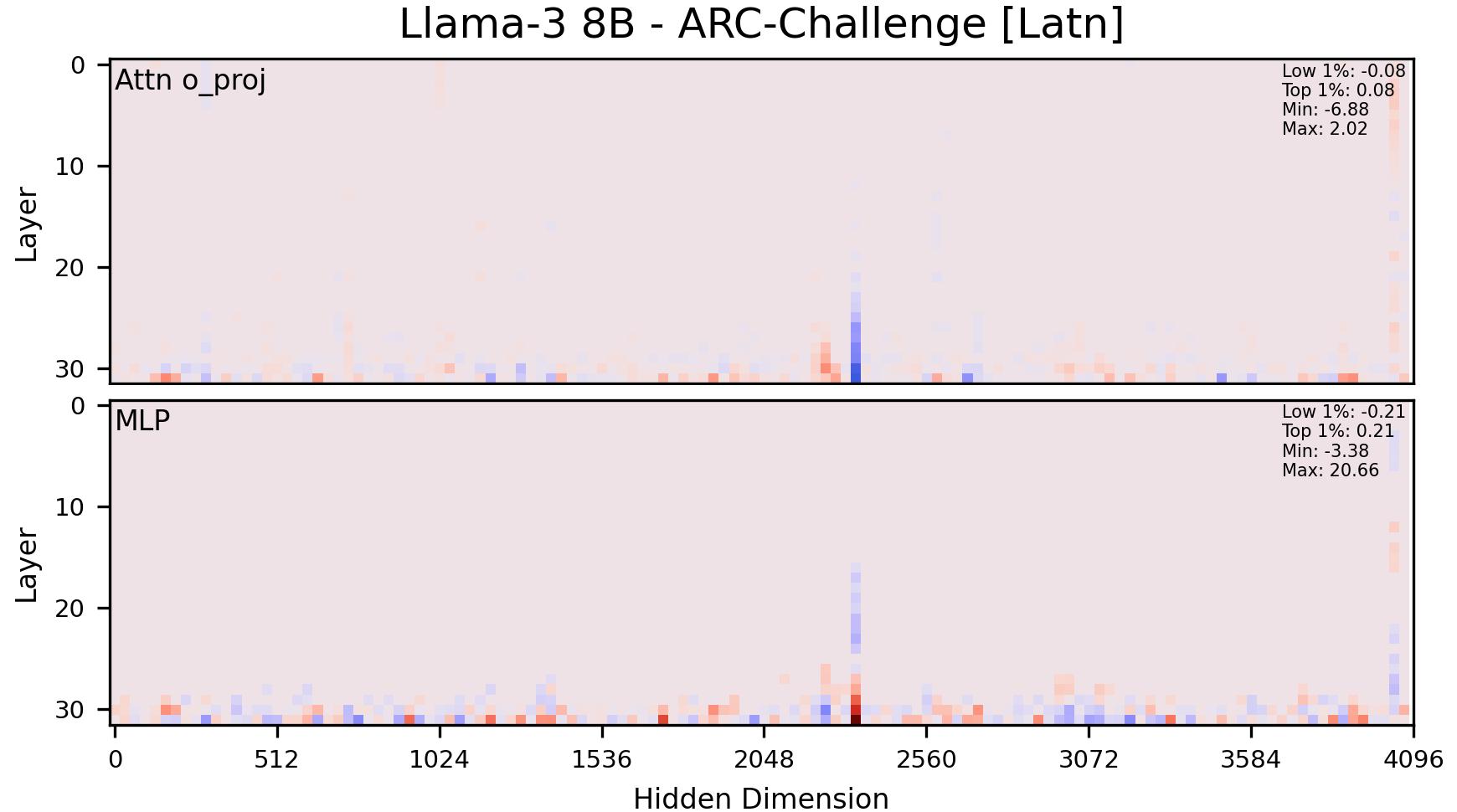} %
    \end{minipage}\hfill
    \begin{minipage}{0.33\textwidth}
        \centering
        \includegraphics[width=1.0\textwidth]{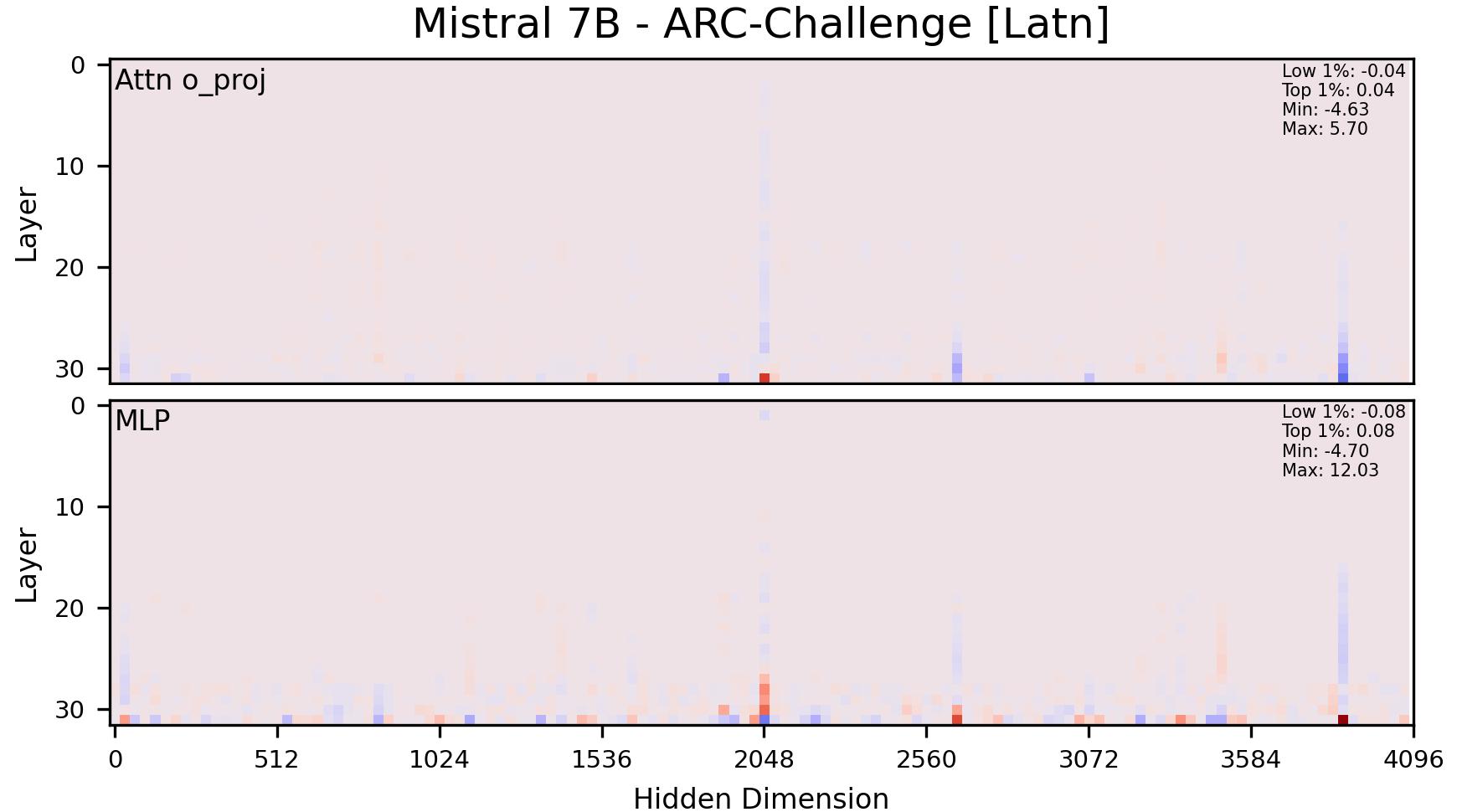} %
    \end{minipage}
    \begin{minipage}{0.33\textwidth}
        \centering
        \includegraphics[width=1.0\textwidth]{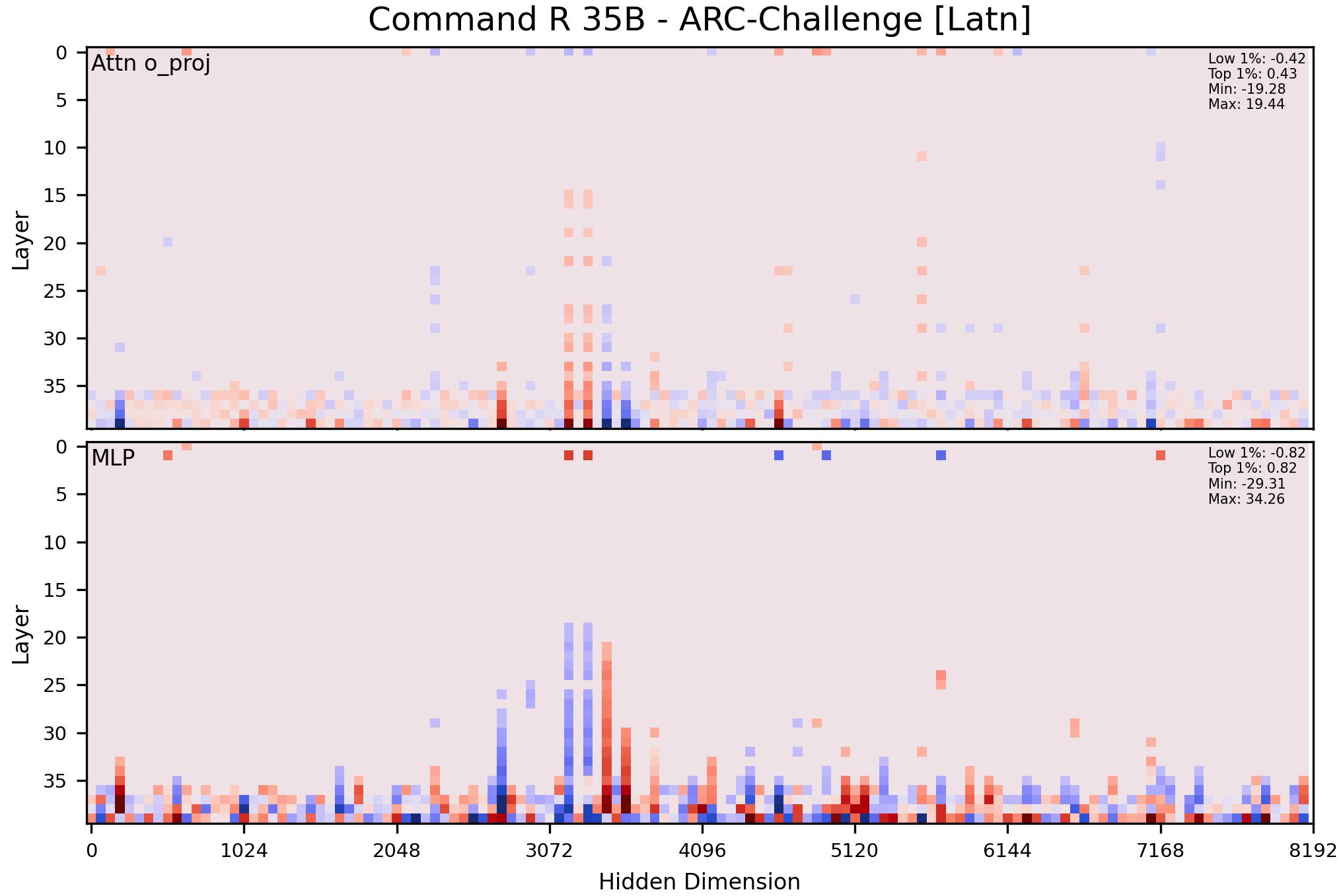} %
    \end{minipage}
\end{figure}

\begin{figure}[H]
    \centering
    \begin{minipage}{0.33\textwidth}
        \centering
        \includegraphics[width=1.0\textwidth]{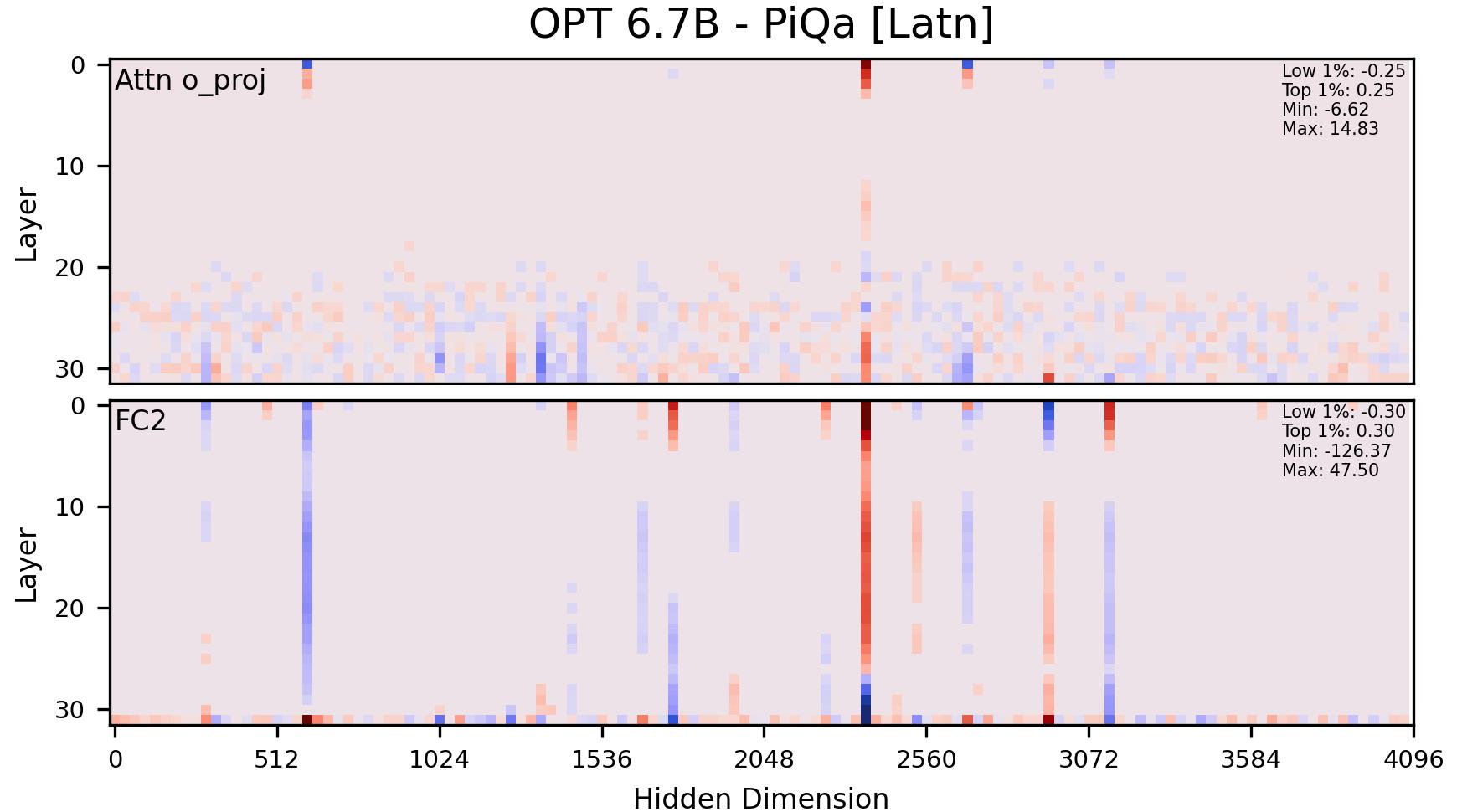} %
    \end{minipage}\hfill
    \begin{minipage}{0.33\textwidth}
        \centering
        \includegraphics[width=1.0\textwidth]{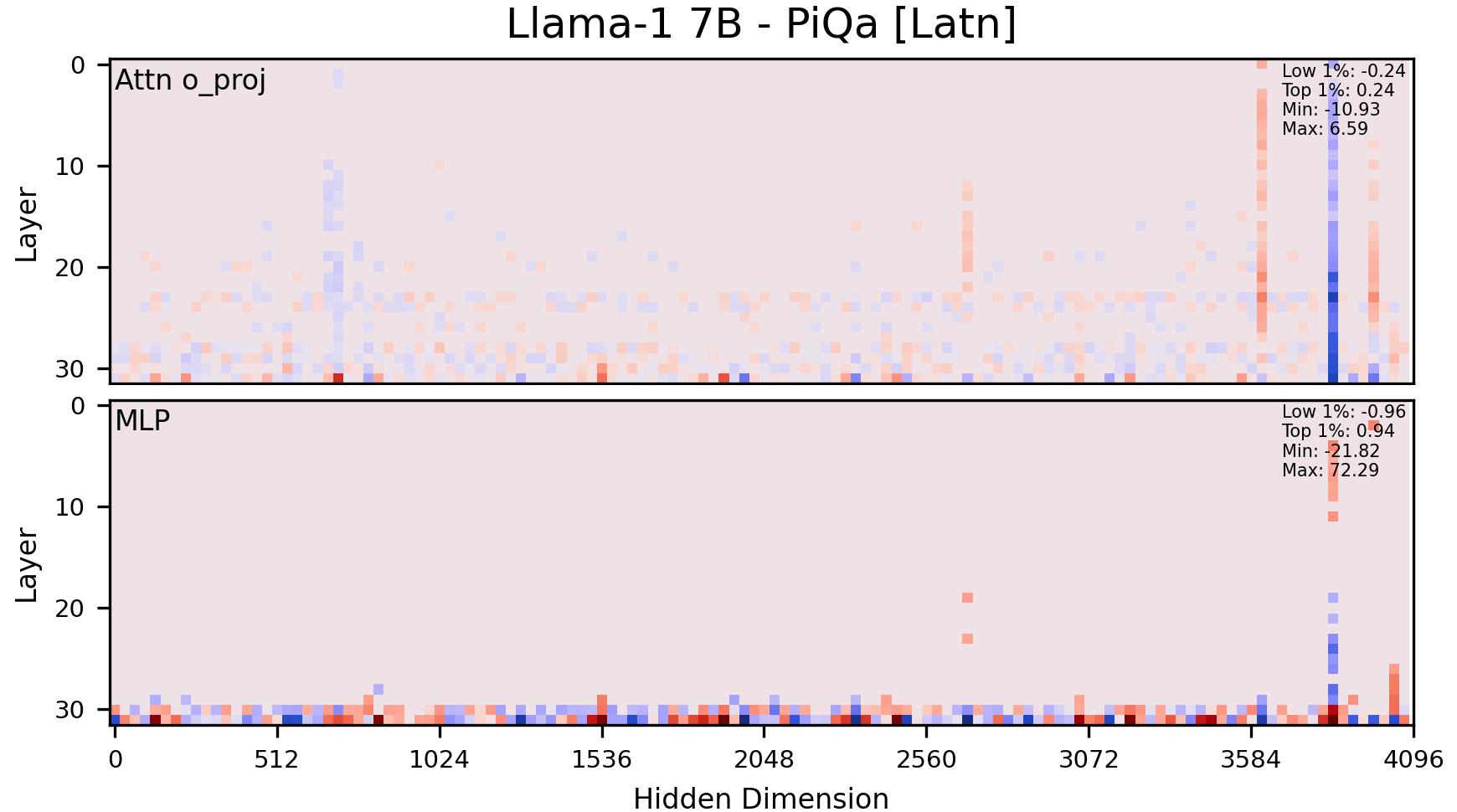} %
    \end{minipage}
    \begin{minipage}{0.33\textwidth}
        \centering
        \includegraphics[width=1.0\textwidth]{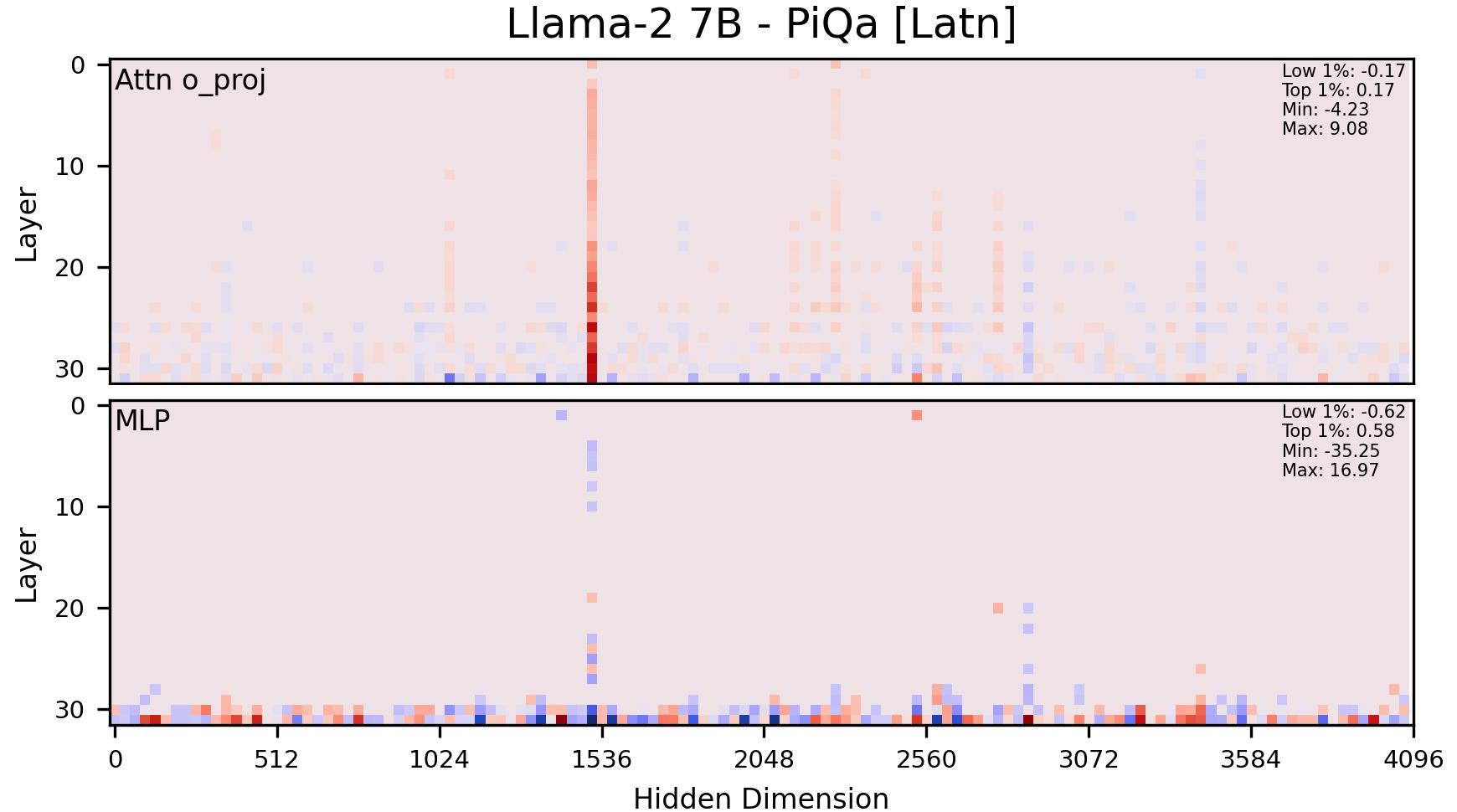} %
    \end{minipage}
    \vskip -0.3in
\end{figure}
\begin{figure}[H]
    \centering
    \begin{minipage}{0.33\textwidth}
        \centering
        \includegraphics[width=1.0\textwidth]{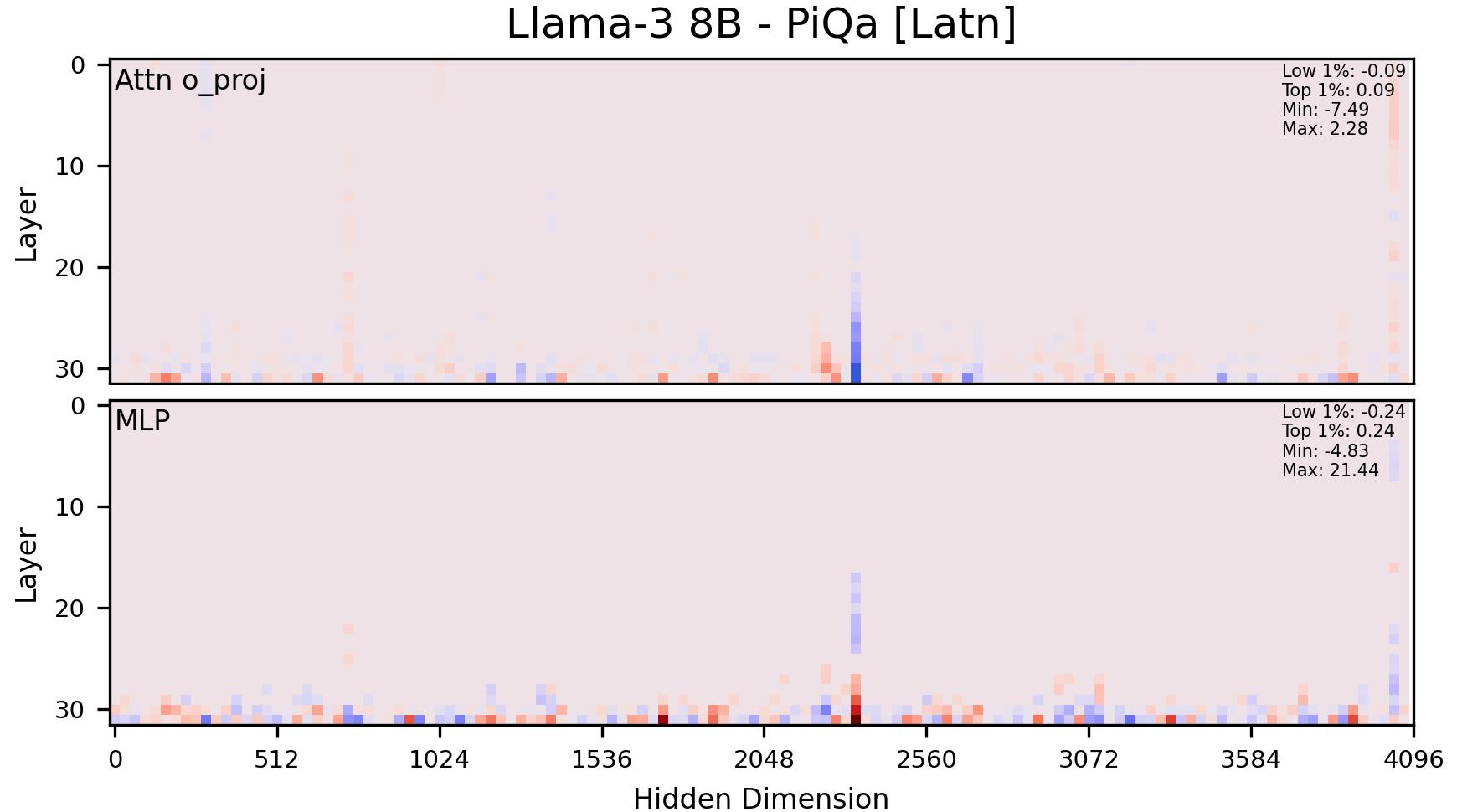} %
    \end{minipage}\hfill
    \begin{minipage}{0.33\textwidth}
        \centering
        \includegraphics[width=1.0\textwidth]{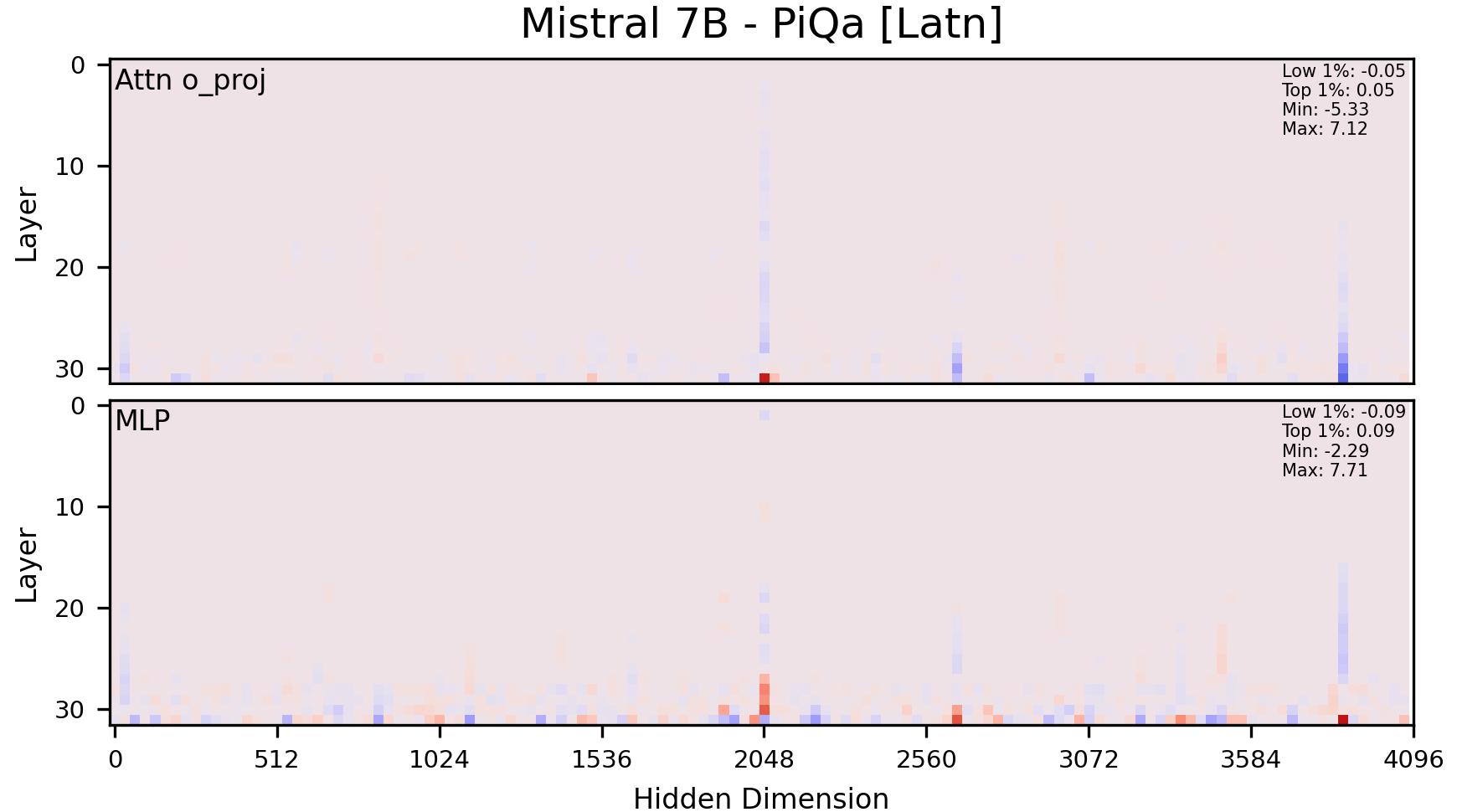} %
    \end{minipage}
    \begin{minipage}{0.33\textwidth}
        \centering
        \includegraphics[width=1.0\textwidth]{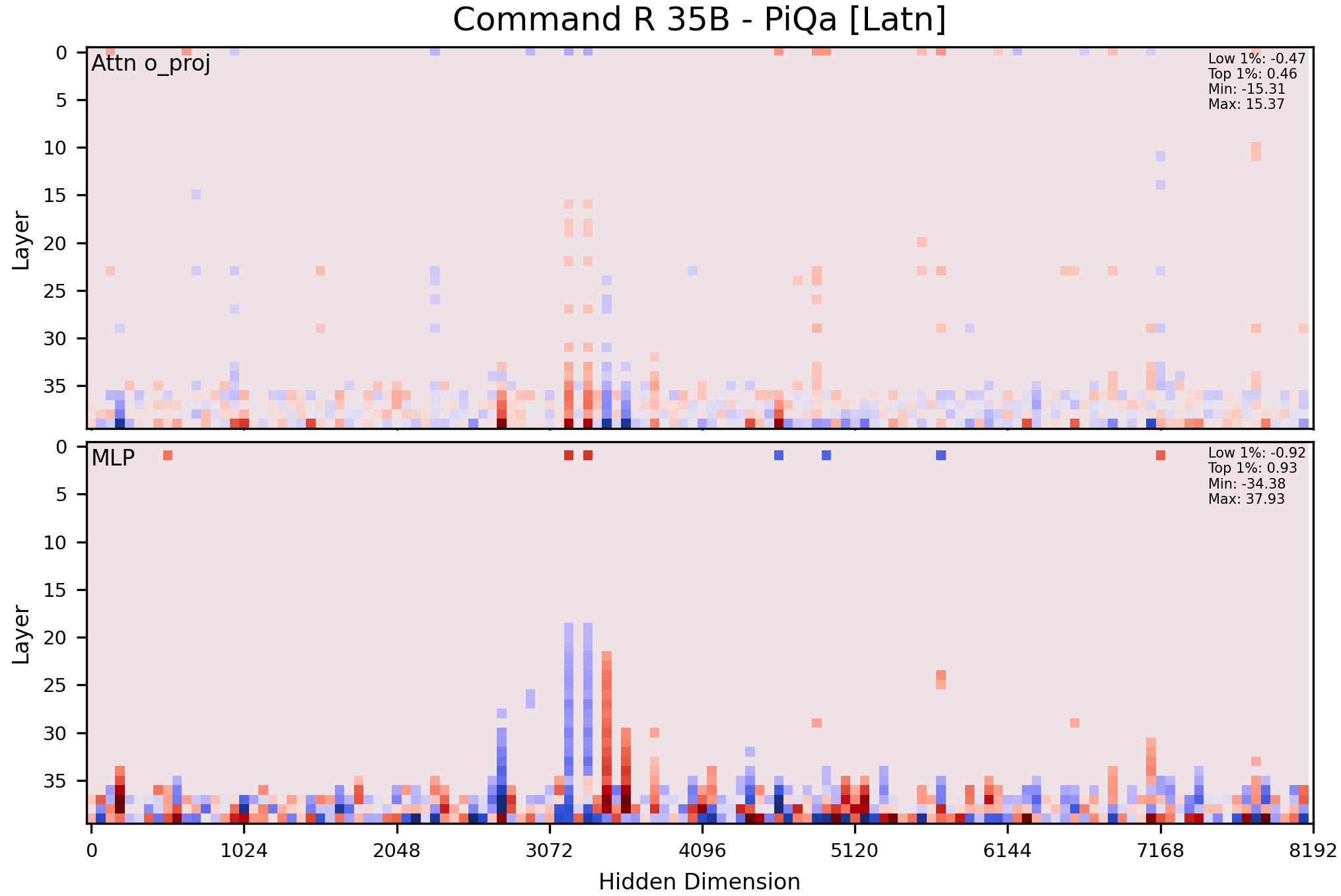} %
    \end{minipage}
\end{figure}

\begin{figure}[H]
    \centering
    \begin{minipage}{0.33\textwidth}
        \centering
        \includegraphics[width=1.0\textwidth]{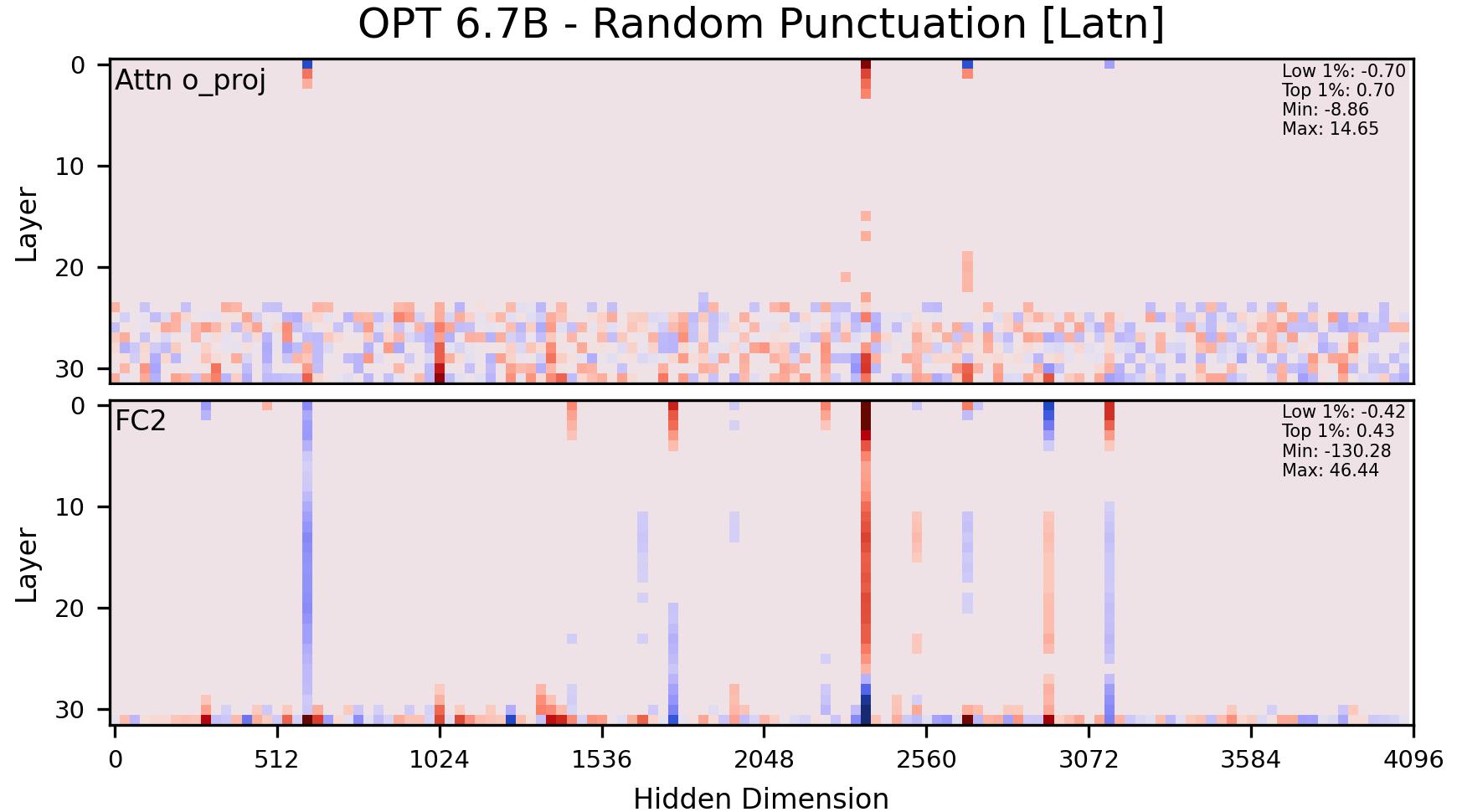} %
    \end{minipage}\hfill
    \begin{minipage}{0.33\textwidth}
        \centering
        \includegraphics[width=1.0\textwidth]{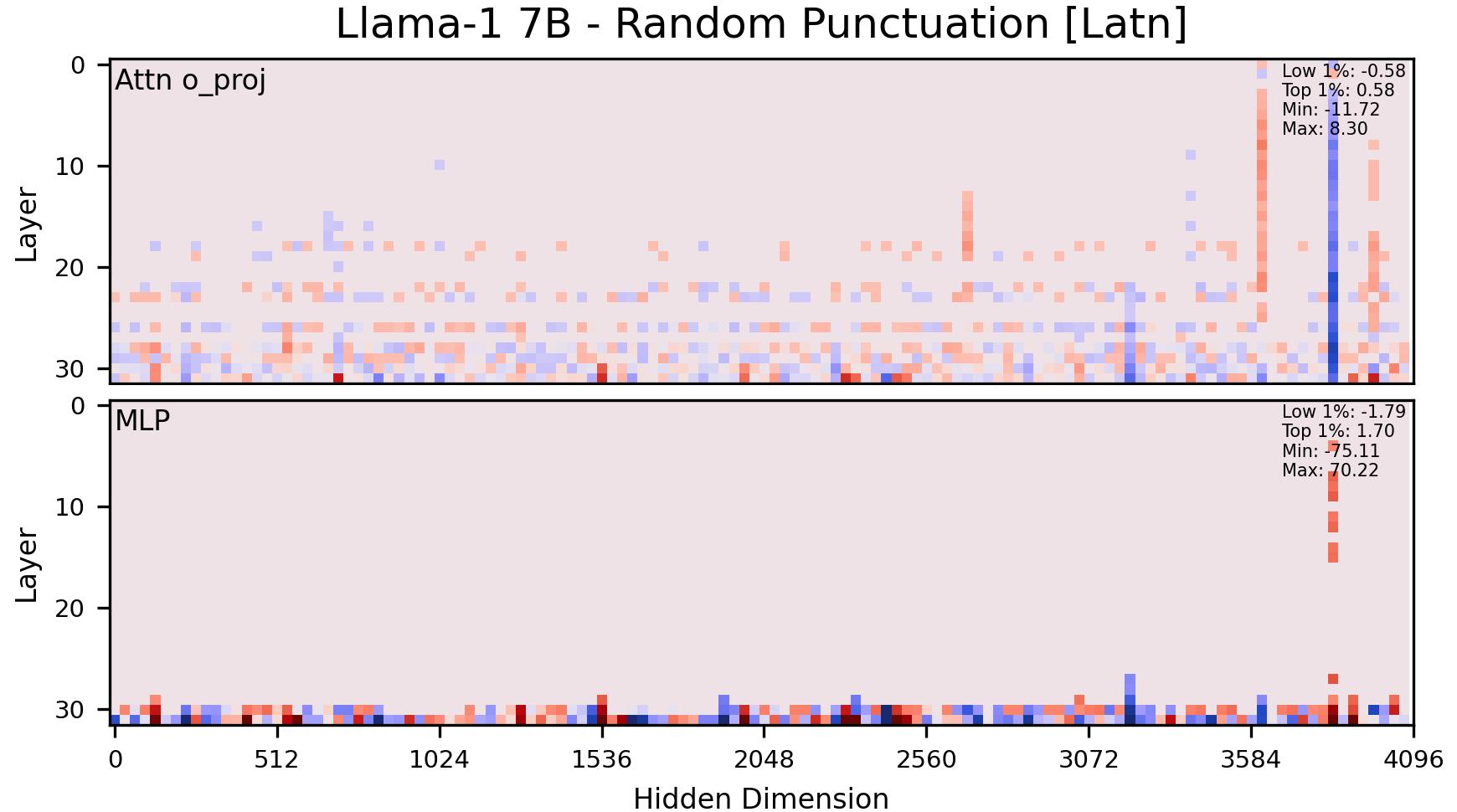} %
    \end{minipage}
    \begin{minipage}{0.33\textwidth}
        \centering
        \includegraphics[width=1.0\textwidth]{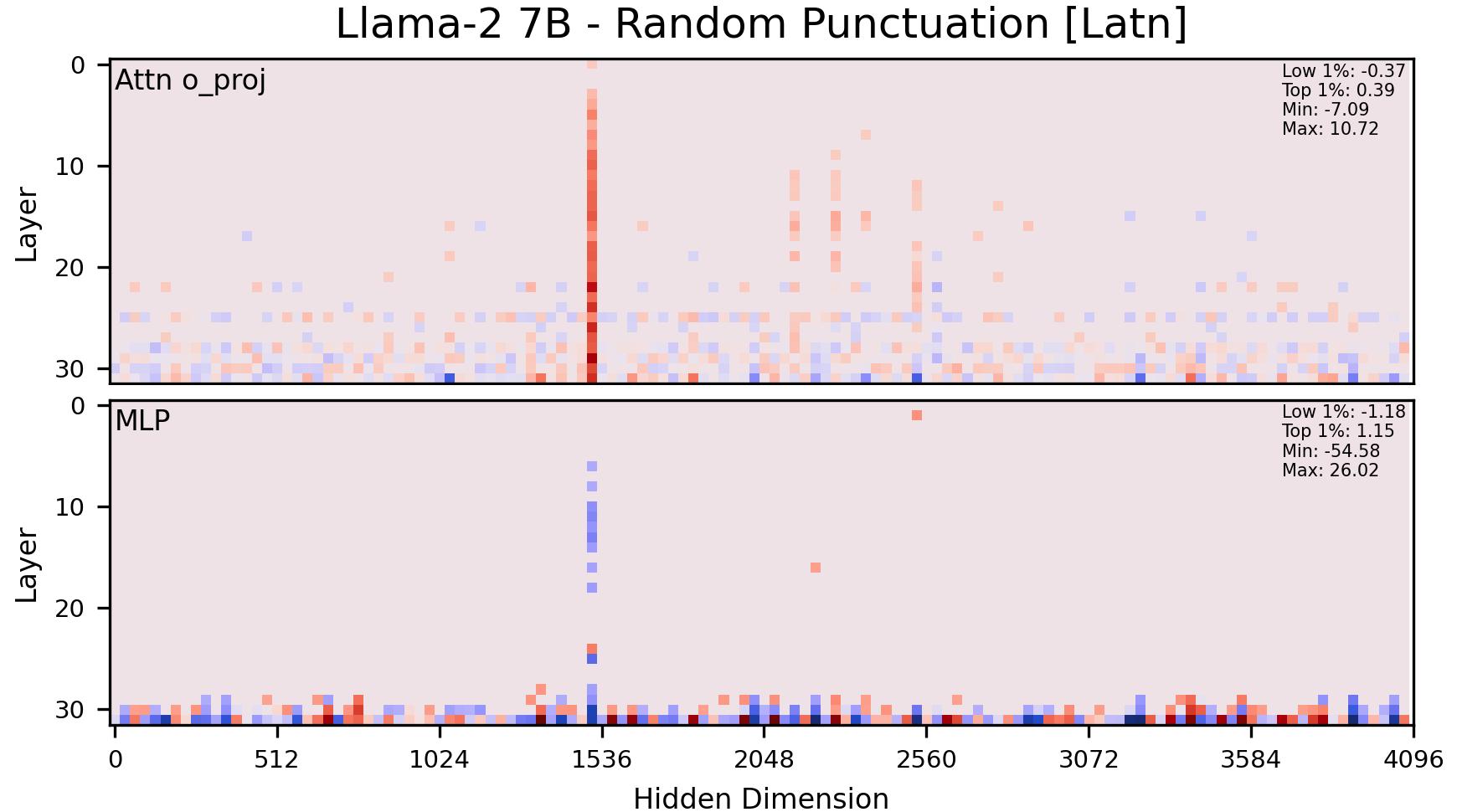} %
    \end{minipage}
    \vskip -0.3in
\end{figure}
\begin{figure}[H]
    \centering
    \begin{minipage}{0.33\textwidth}
        \centering
        \includegraphics[width=1.0\textwidth]{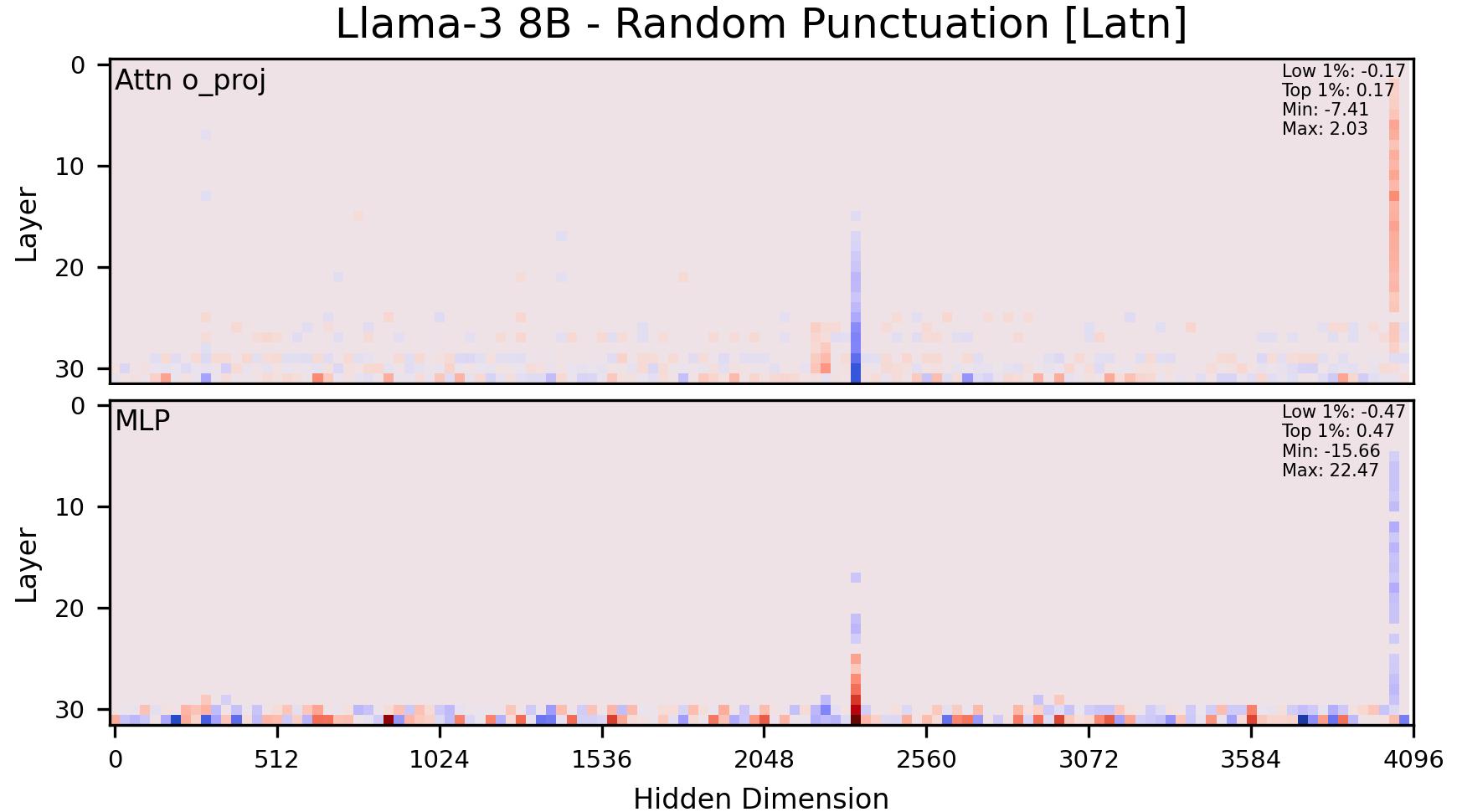} %
    \end{minipage}\hfill
    \begin{minipage}{0.33\textwidth}
        \centering
        \includegraphics[width=1.0\textwidth]{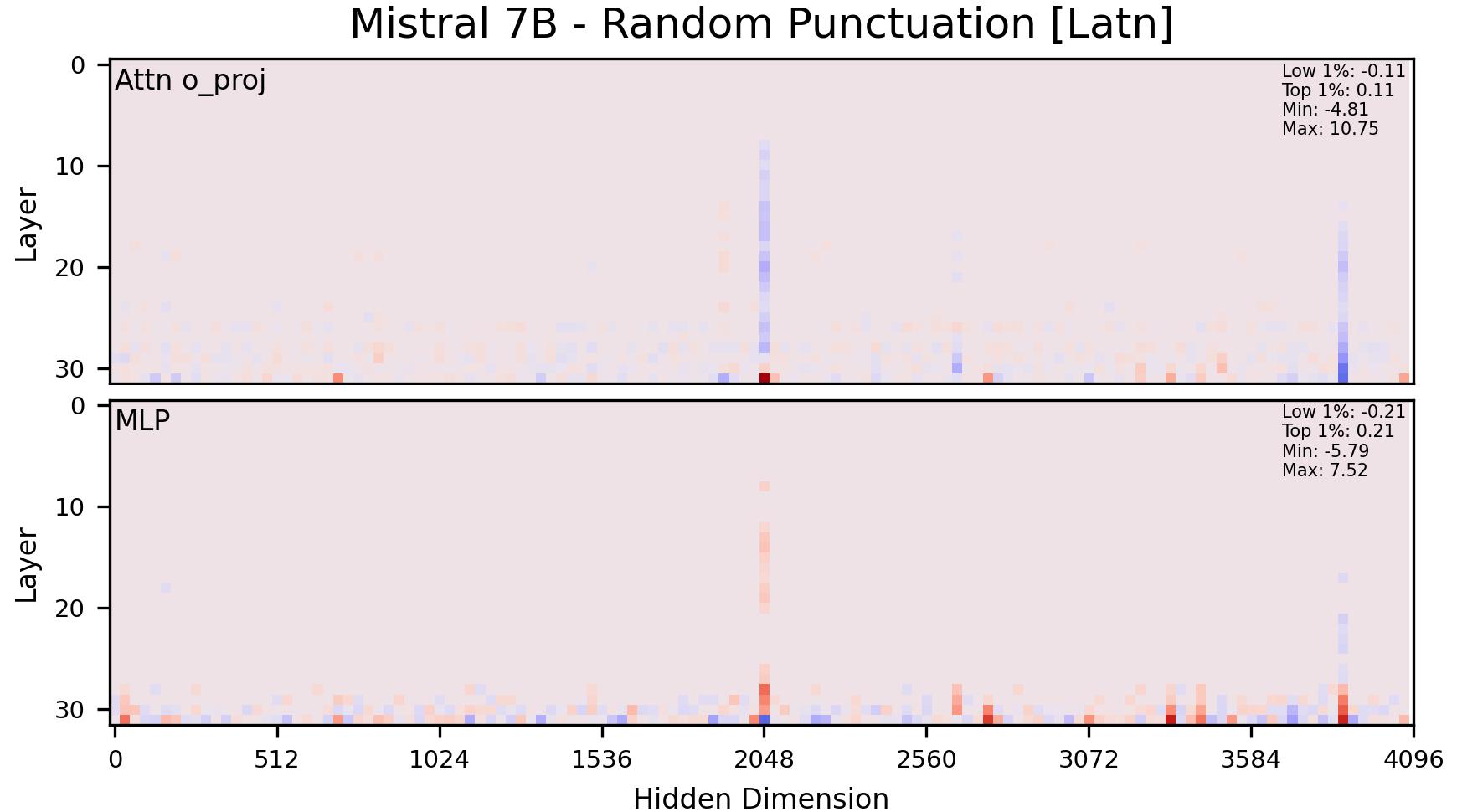} %
    \end{minipage}
    \begin{minipage}{0.33\textwidth}
        \centering
        \includegraphics[width=1.0\textwidth]{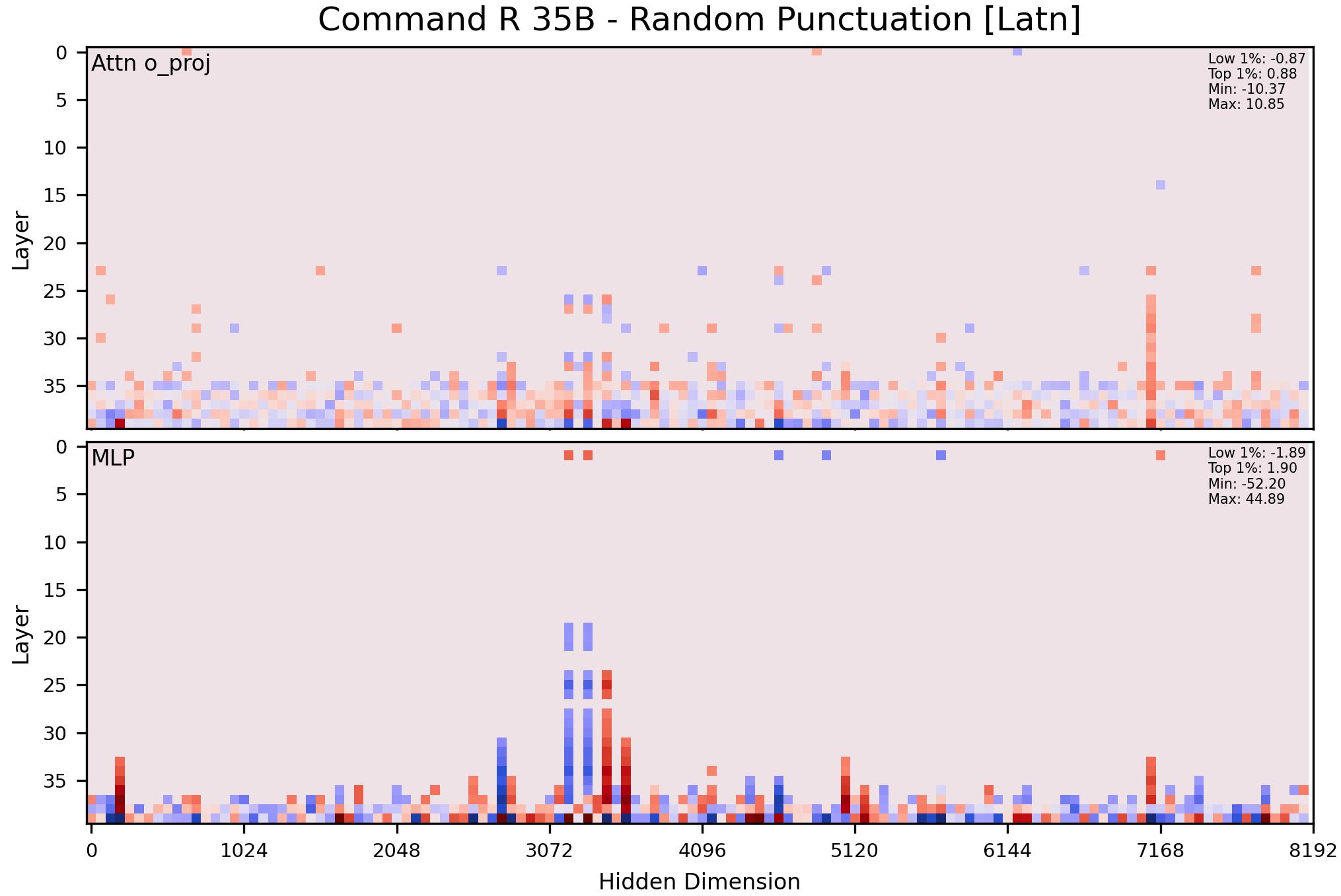} %
    \end{minipage}
\end{figure}

%% file: core/E-appendix.tex
\section{Activation Distributions Plots} \label{all_cropped_histogramsogram_distributions}
\begin{figure}[H]
   \begin{minipage}{0.25\textwidth}
     \centering
     \includegraphics[width=1\linewidth]{images/all_cropped_histograms/eng_Latn_attn_hist.png}
   \end{minipage}\hfill
   \begin{minipage}{0.25\textwidth}
     \centering
     \includegraphics[width=1\linewidth]{images/all_cropped_histograms/eng_Latn_mlp_hist.png}
   \end{minipage}\hfill
   \begin{minipage}{0.25\textwidth}
     \centering
     \includegraphics[width=1\linewidth]{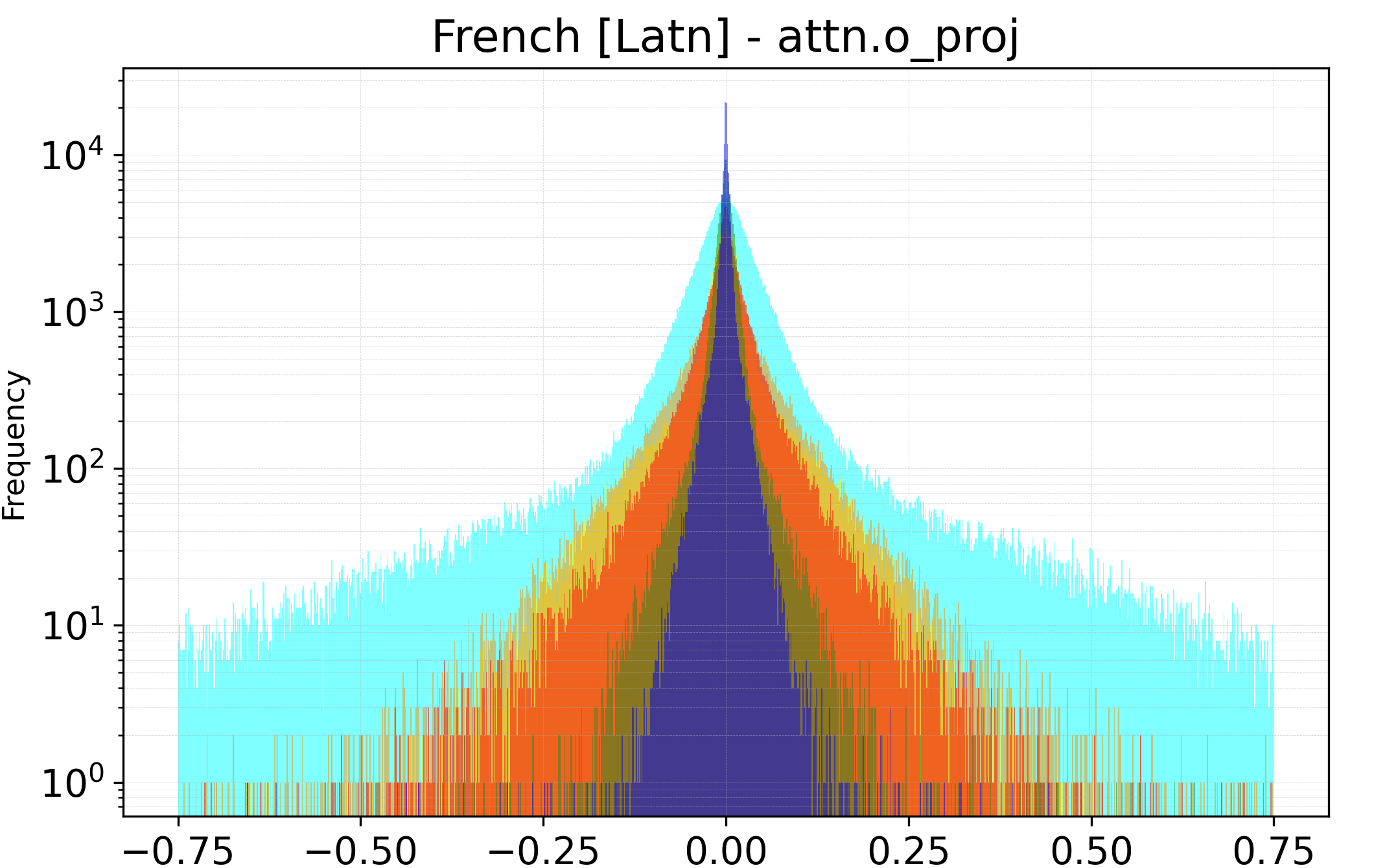}
   \end{minipage}\hfill
   \begin{minipage}{0.25\textwidth}
     \centering
     \includegraphics[width=1\linewidth]{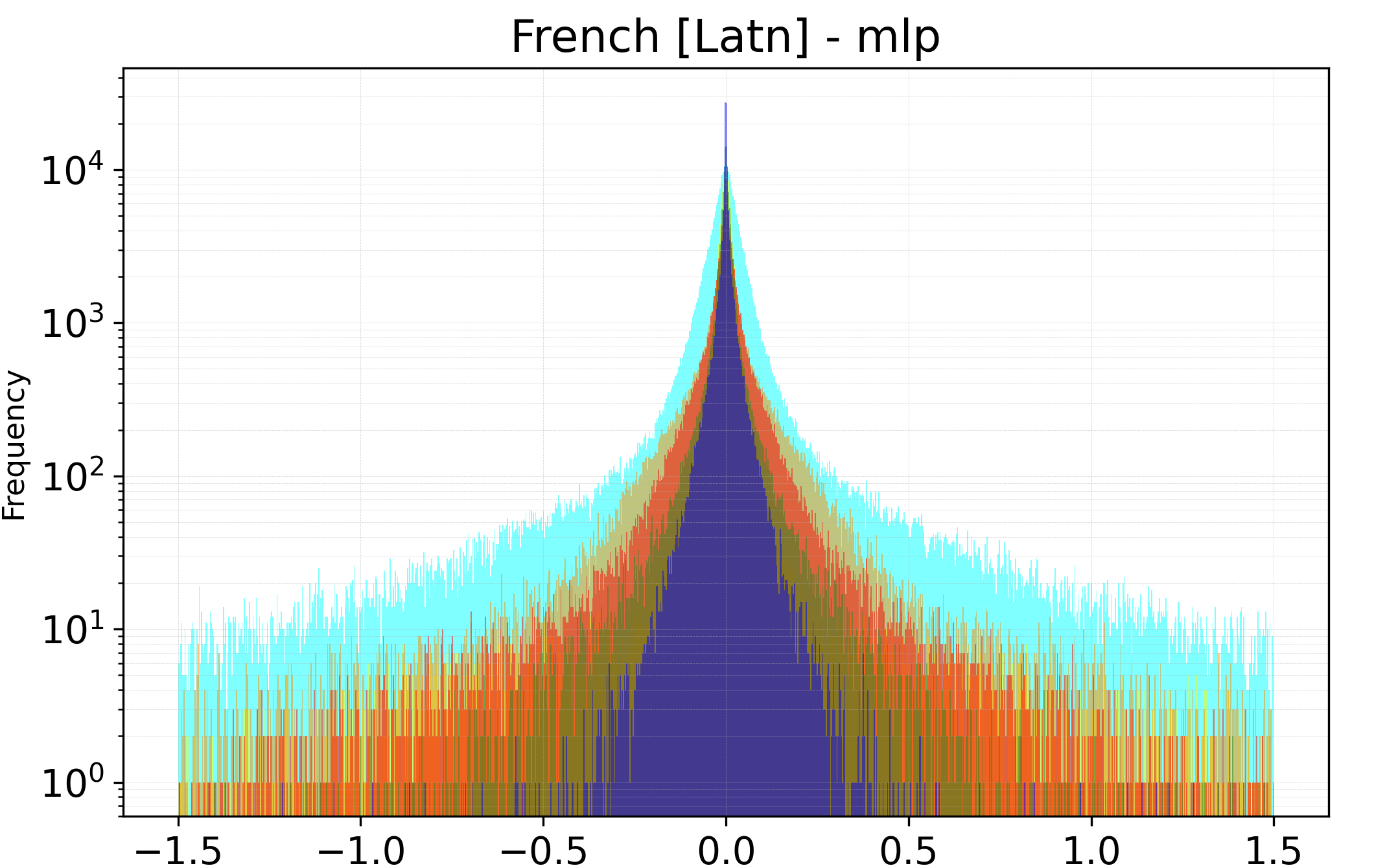}
   \end{minipage}
   \centerline{\includegraphics[width=1.0\columnwidth]{images/models.png}}
\end{figure}

\begin{figure}[H]
   \begin{minipage}{0.25\textwidth}
     \centering
     \includegraphics[width=1\linewidth]{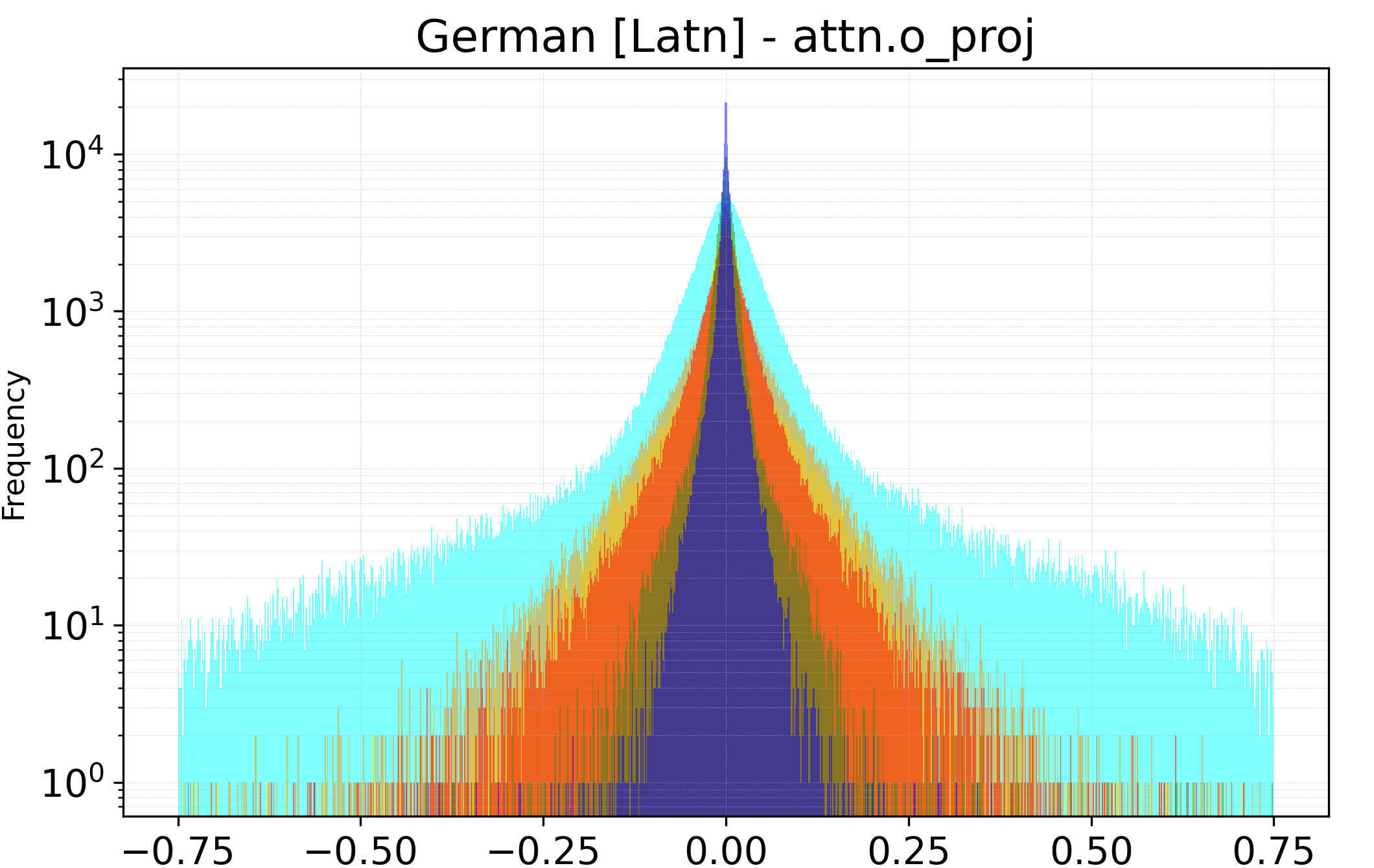}
   \end{minipage}\hfill
   \begin{minipage}{0.25\textwidth}
     \centering
     \includegraphics[width=1\linewidth]{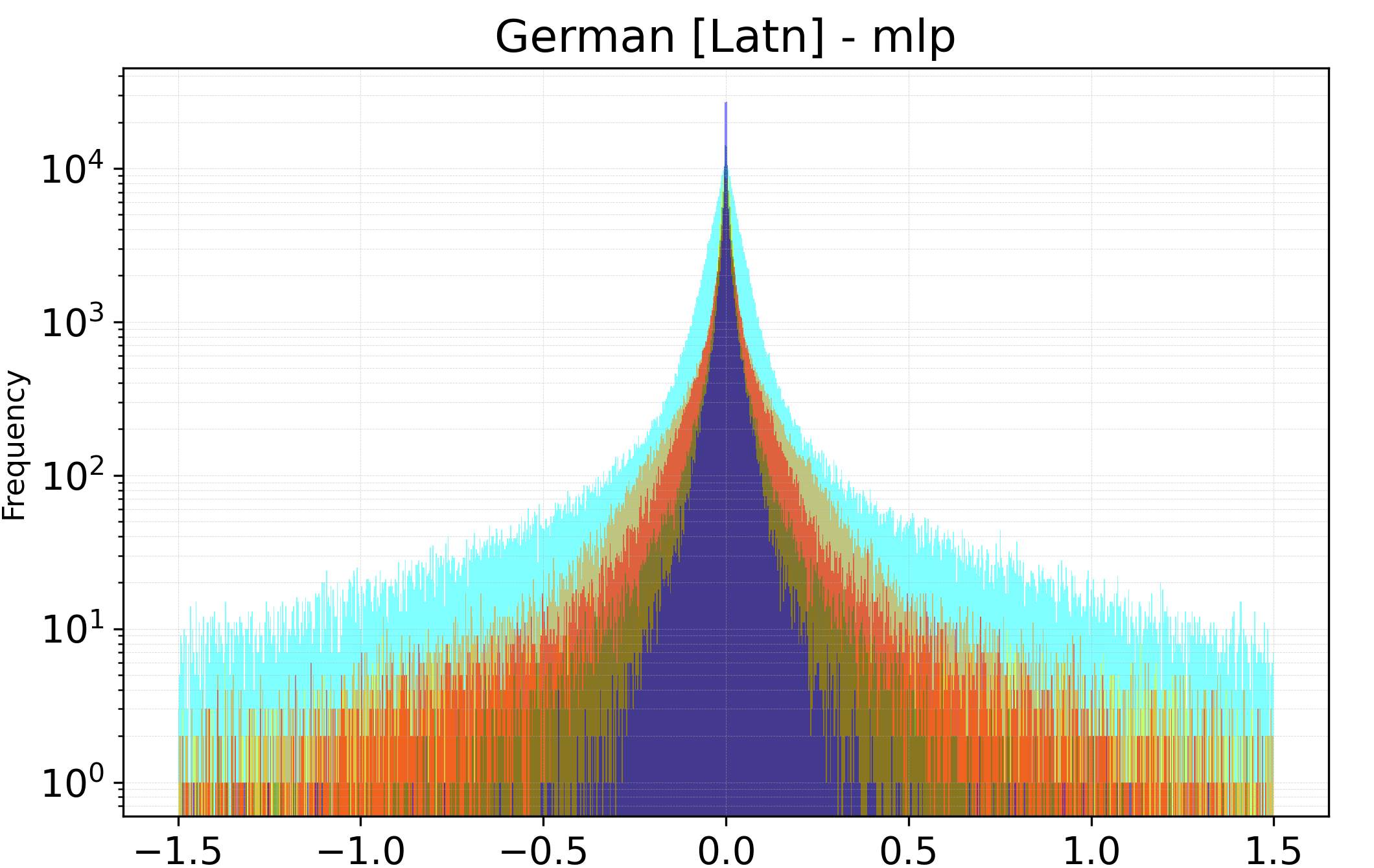}
   \end{minipage}\hfill
   \begin{minipage}{0.25\textwidth}
     \centering
     \includegraphics[width=1\linewidth]{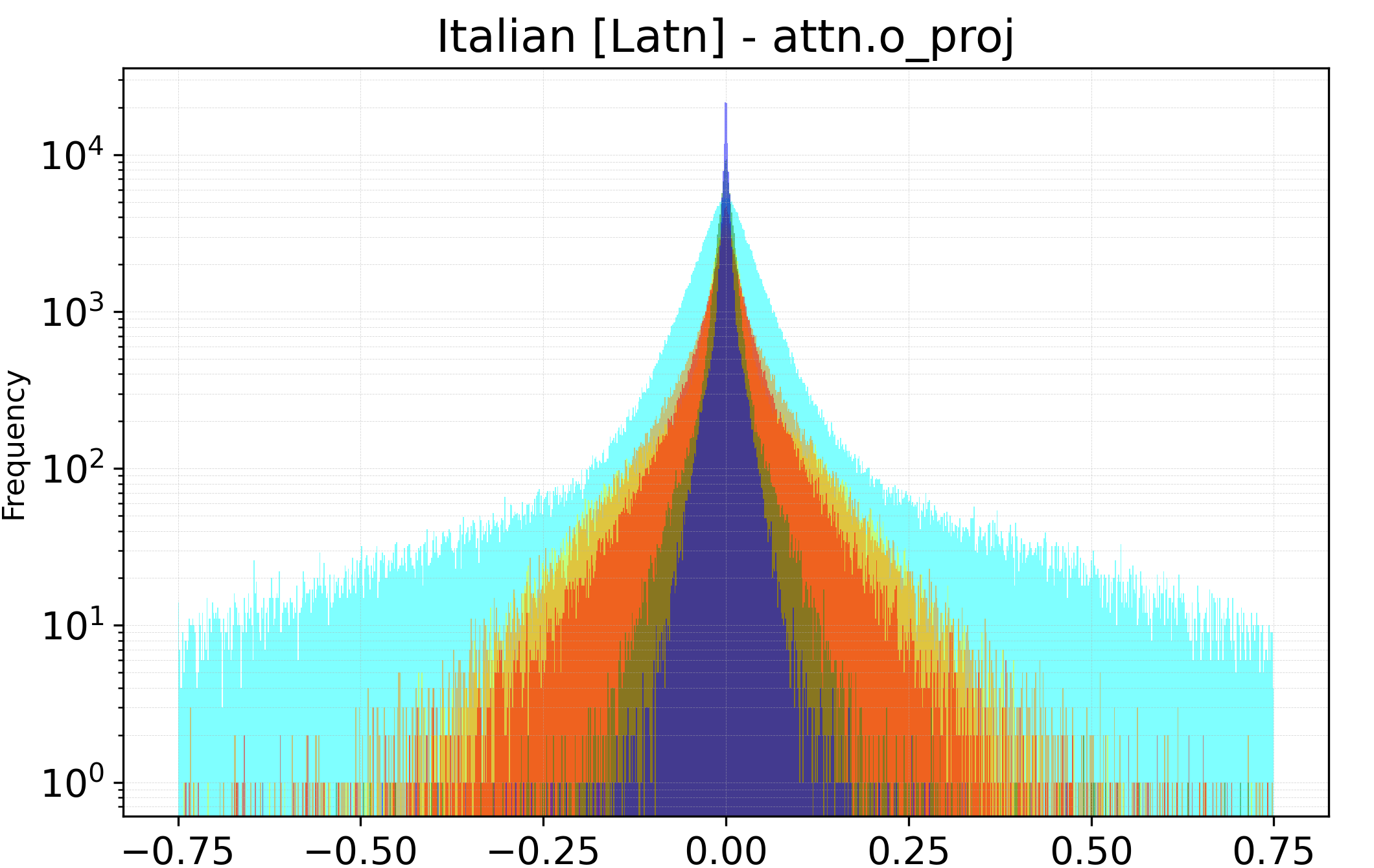}
   \end{minipage}\hfill
   \begin{minipage}{0.25\textwidth}
     \centering
     \includegraphics[width=1\linewidth]{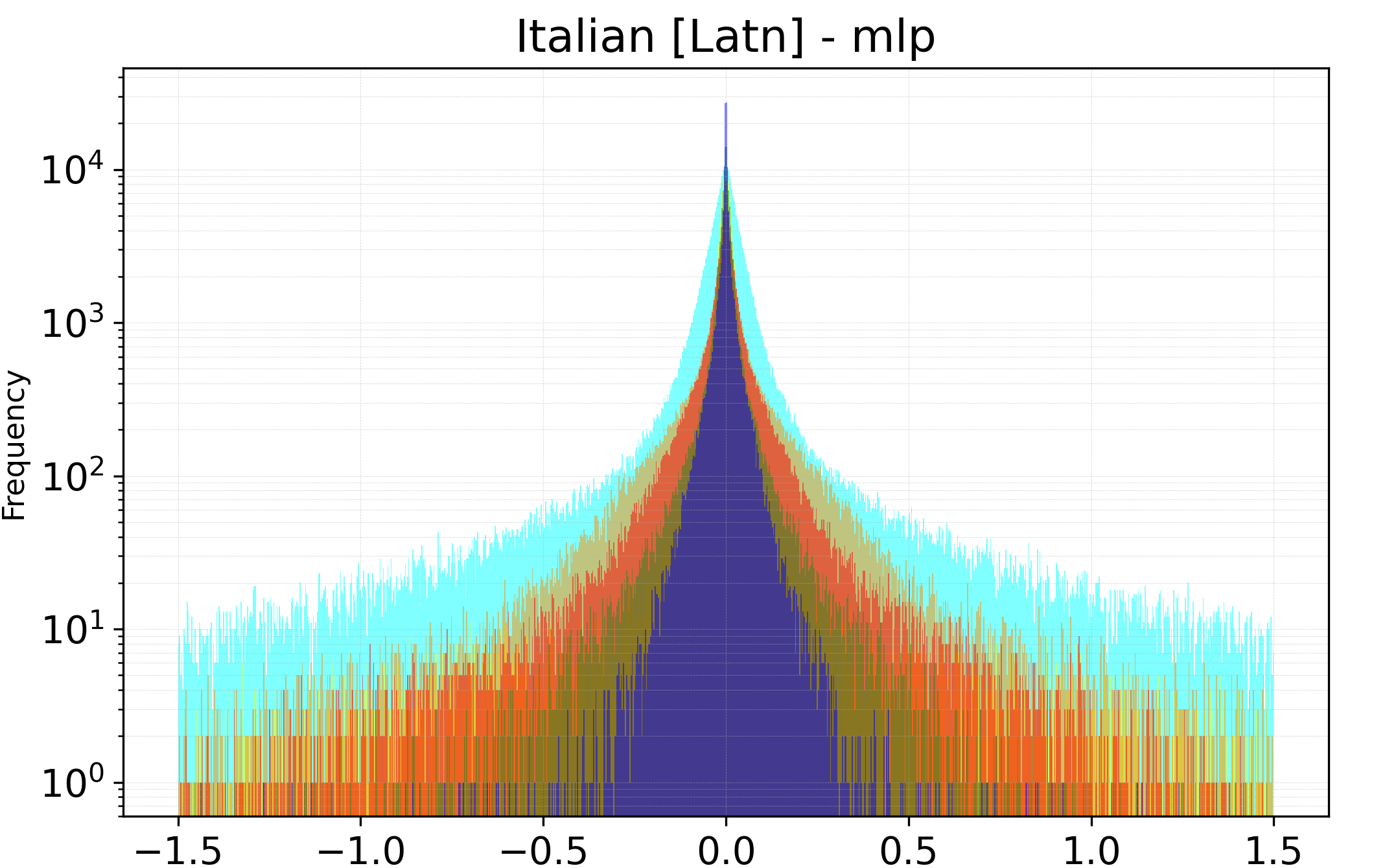}
   \end{minipage}
   \centerline{\includegraphics[width=1.0\columnwidth]{images/models.png}}
\end{figure}

\begin{figure}[H]
   \begin{minipage}{0.25\textwidth}
     \centering
     \includegraphics[width=1\linewidth]{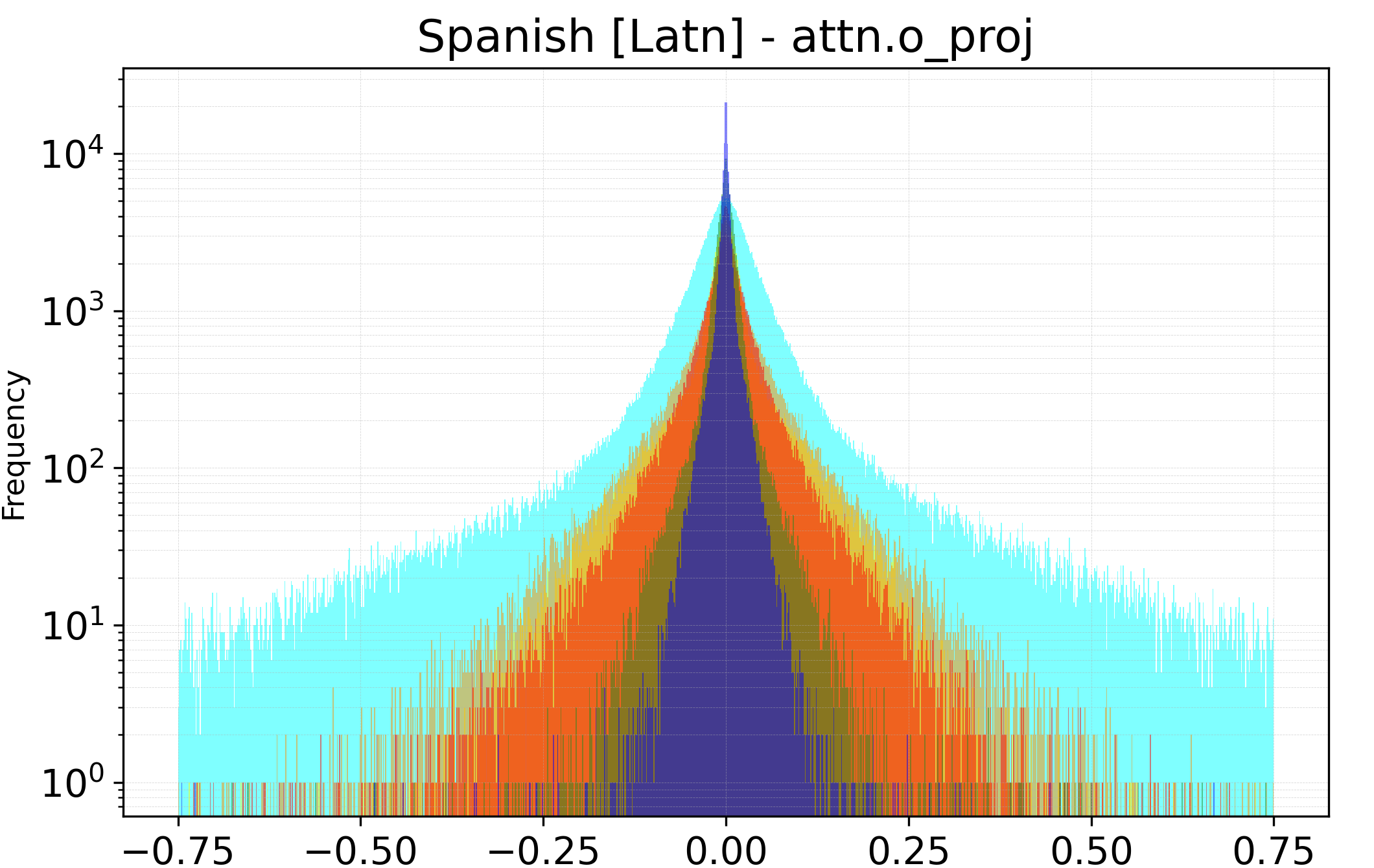}
   \end{minipage}\hfill
   \begin{minipage}{0.25\textwidth}
     \centering
     \includegraphics[width=1\linewidth]{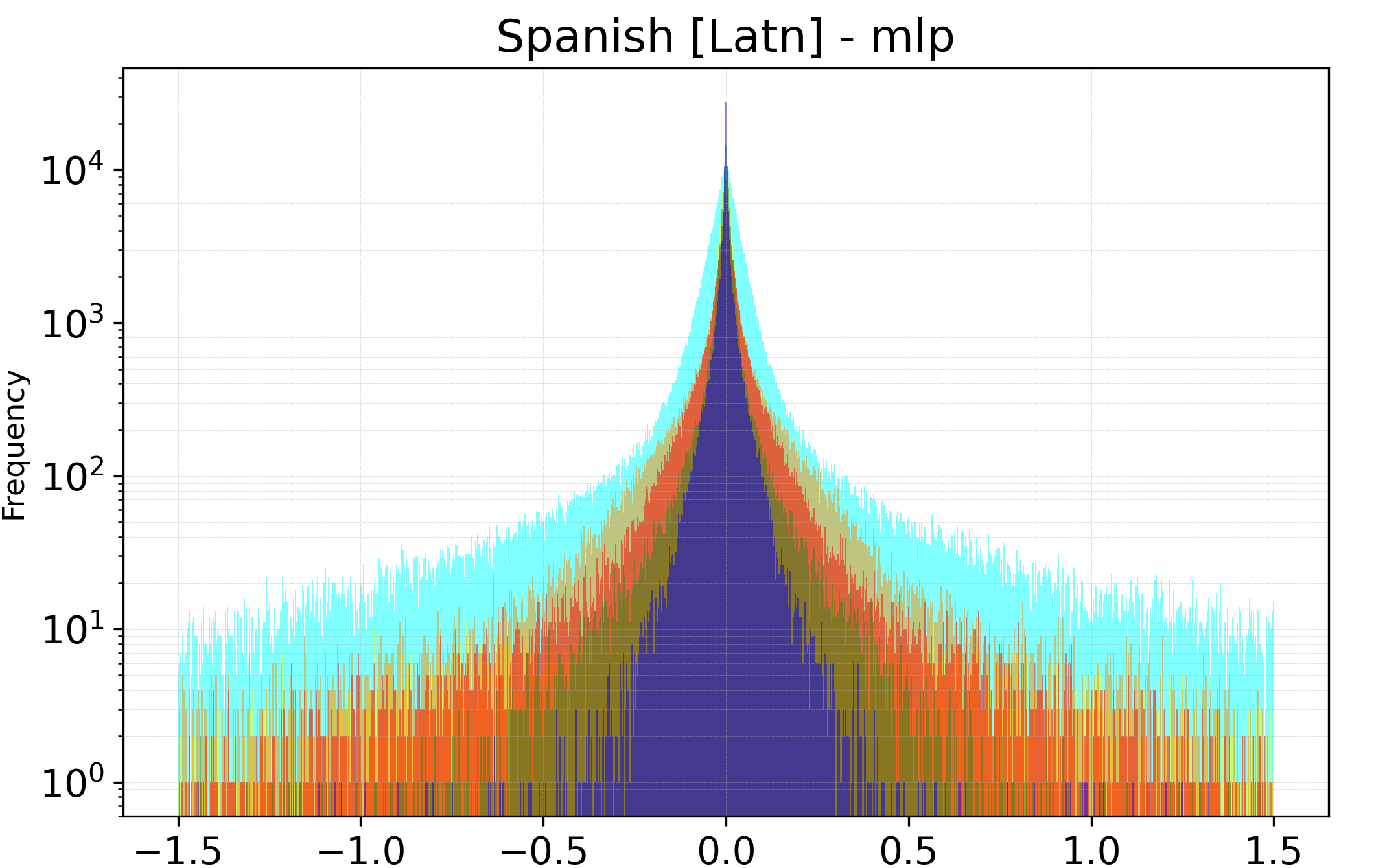}
   \end{minipage}\hfill
   \begin{minipage}{0.25\textwidth}
     \centering
     \includegraphics[width=1\linewidth]{images/all_cropped_histograms/cmn_Hans_attn_hist.png}
   \end{minipage}\hfill
   \begin{minipage}{0.25\textwidth}
     \centering
     \includegraphics[width=1\linewidth]{images/all_cropped_histograms/cmn_Hans_mlp_hist.png}
   \end{minipage}
   \centerline{\includegraphics[width=1.0\columnwidth]{images/models.png}}
\end{figure}

\begin{figure}[H]
   \begin{minipage}{0.25\textwidth}
     \centering
     \includegraphics[width=1\linewidth]{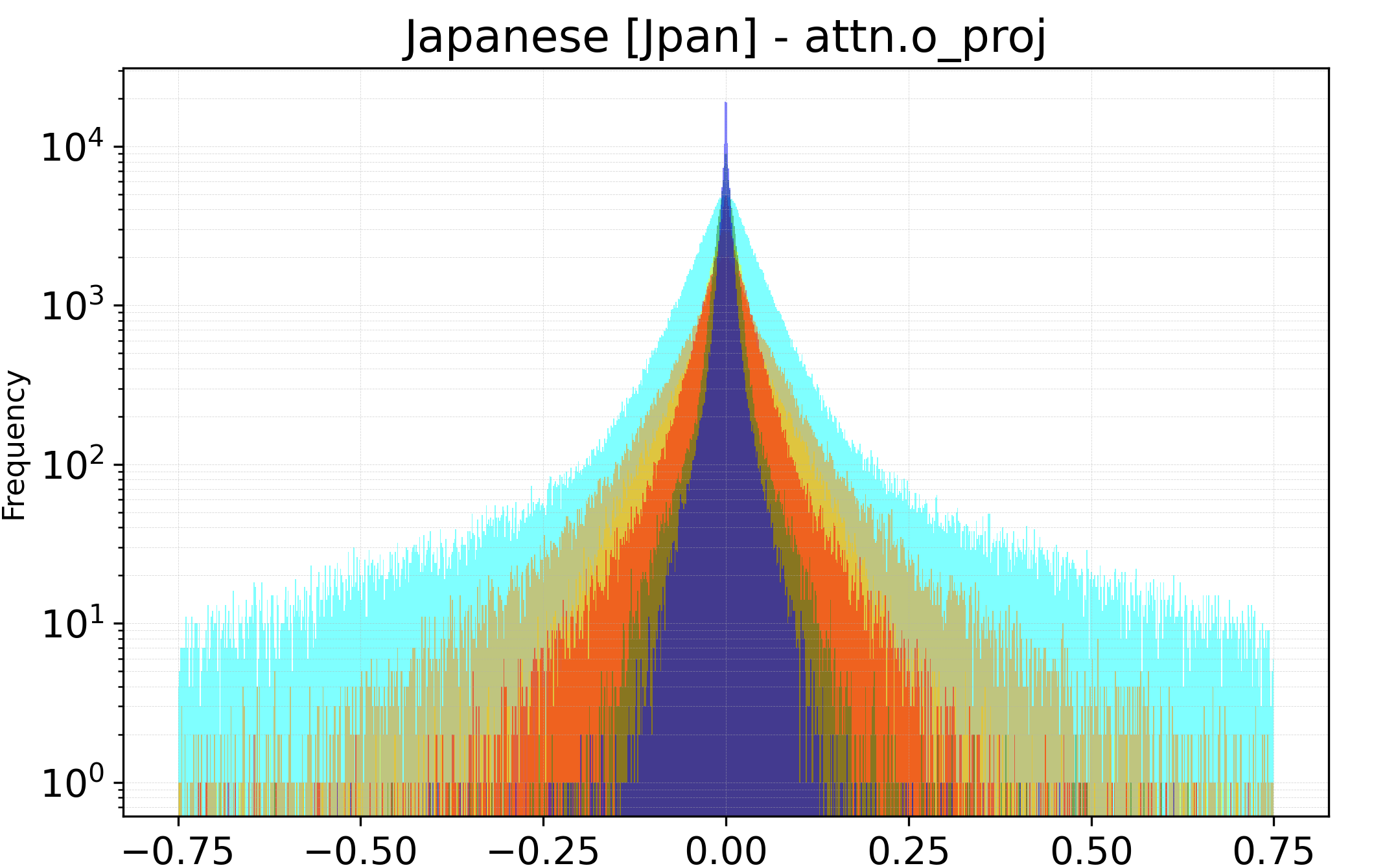}
   \end{minipage}\hfill
   \begin{minipage}{0.25\textwidth}
     \centering
     \includegraphics[width=1\linewidth]{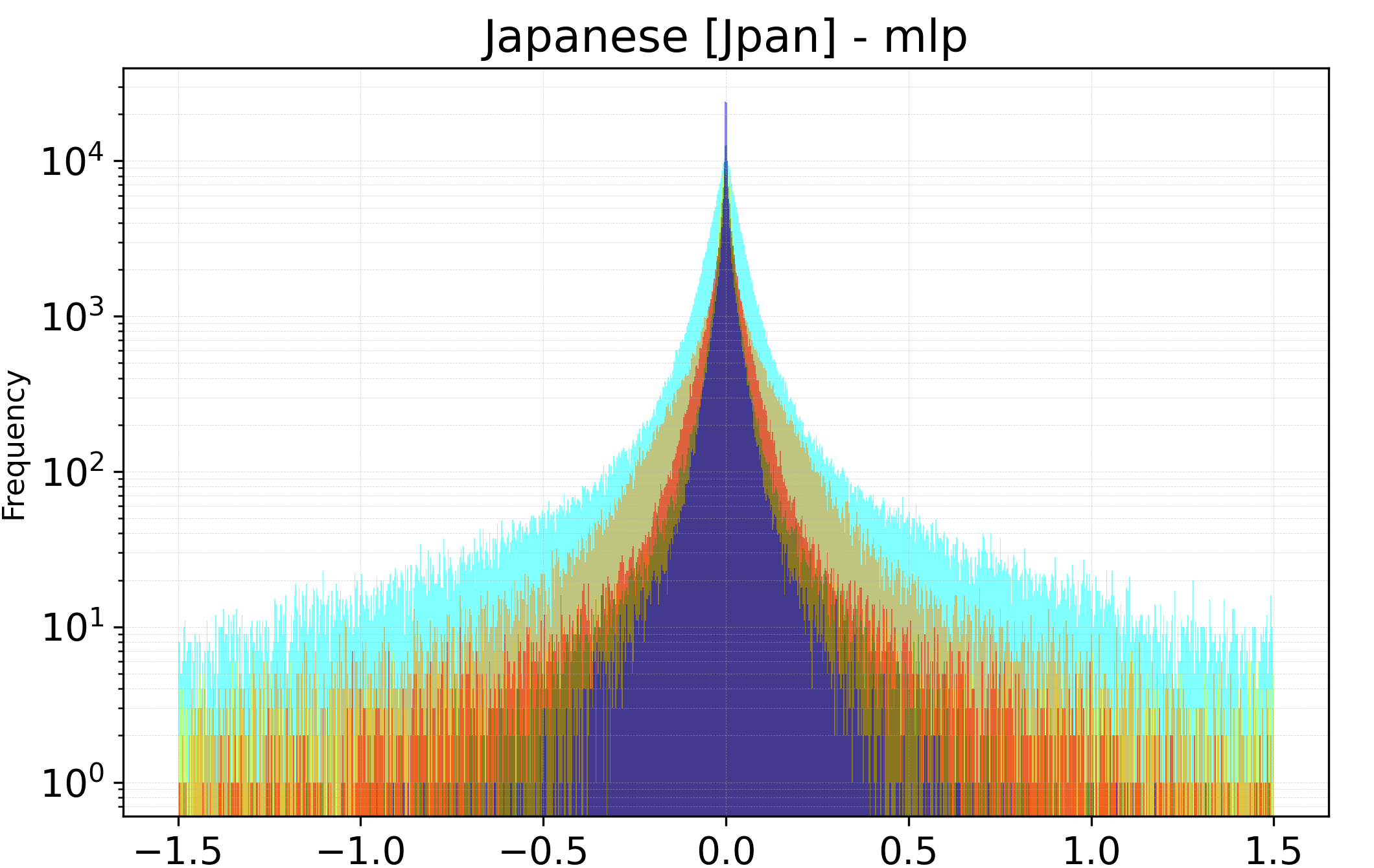}
   \end{minipage}\hfill
   \begin{minipage}{0.25\textwidth}
     \centering
     \includegraphics[width=1\linewidth]{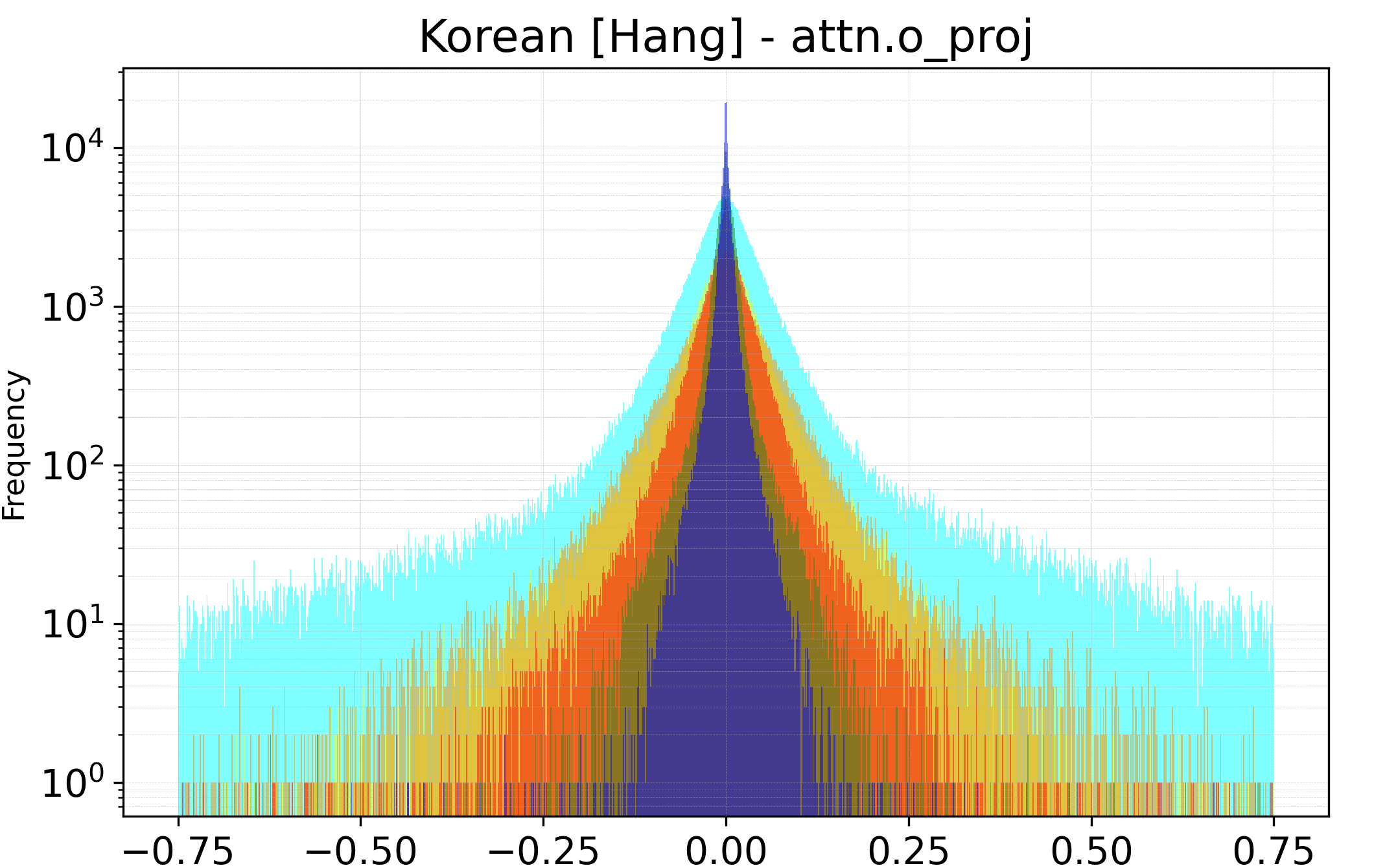}
   \end{minipage}\hfill
   \begin{minipage}{0.25\textwidth}
     \centering
     \includegraphics[width=1\linewidth]{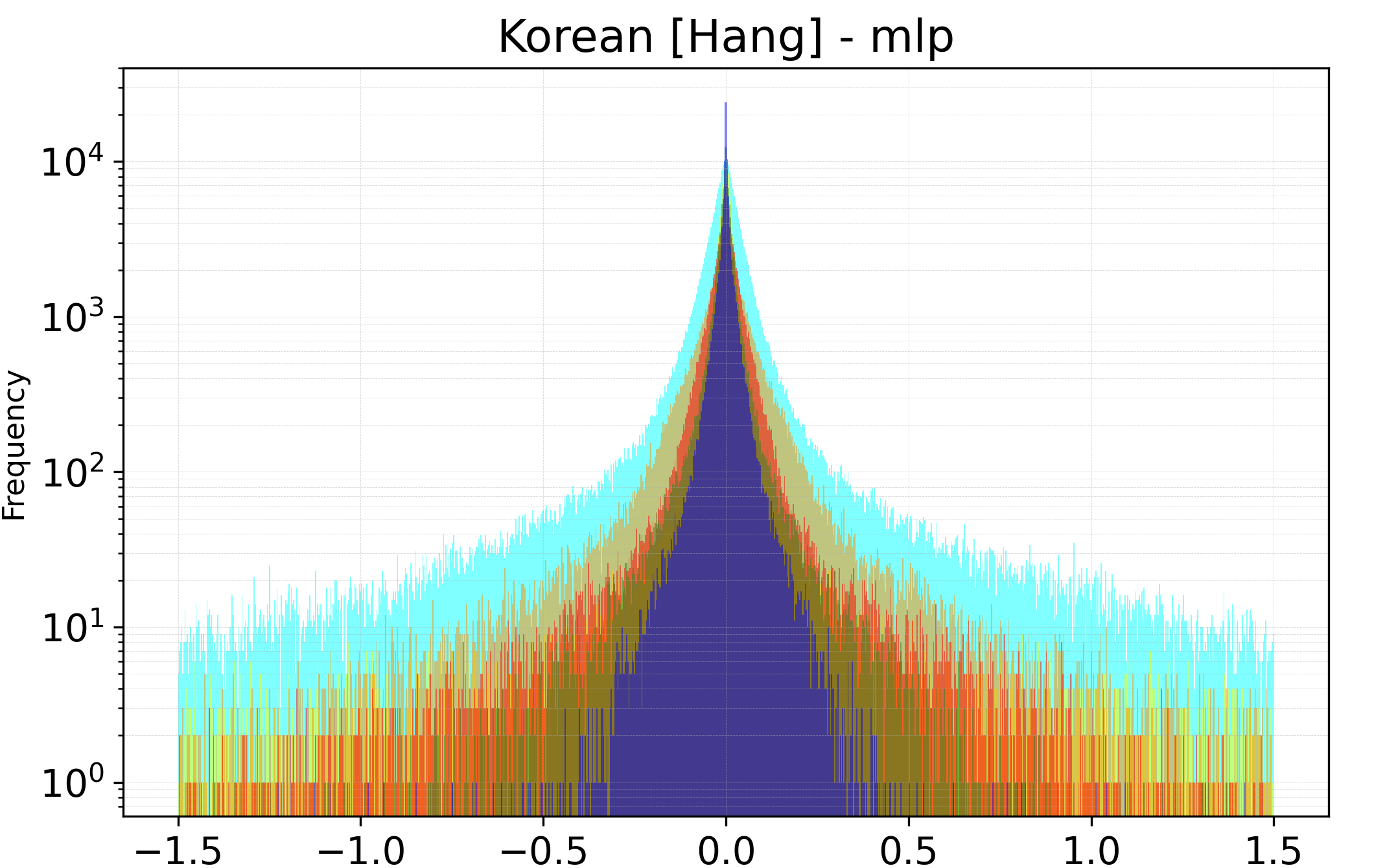}
   \end{minipage}
   \centerline{\includegraphics[width=1.0\columnwidth]{images/models.png}}
\end{figure}

\begin{figure}[H]
   \begin{minipage}{0.25\textwidth}
     \centering
     \includegraphics[width=1\linewidth]{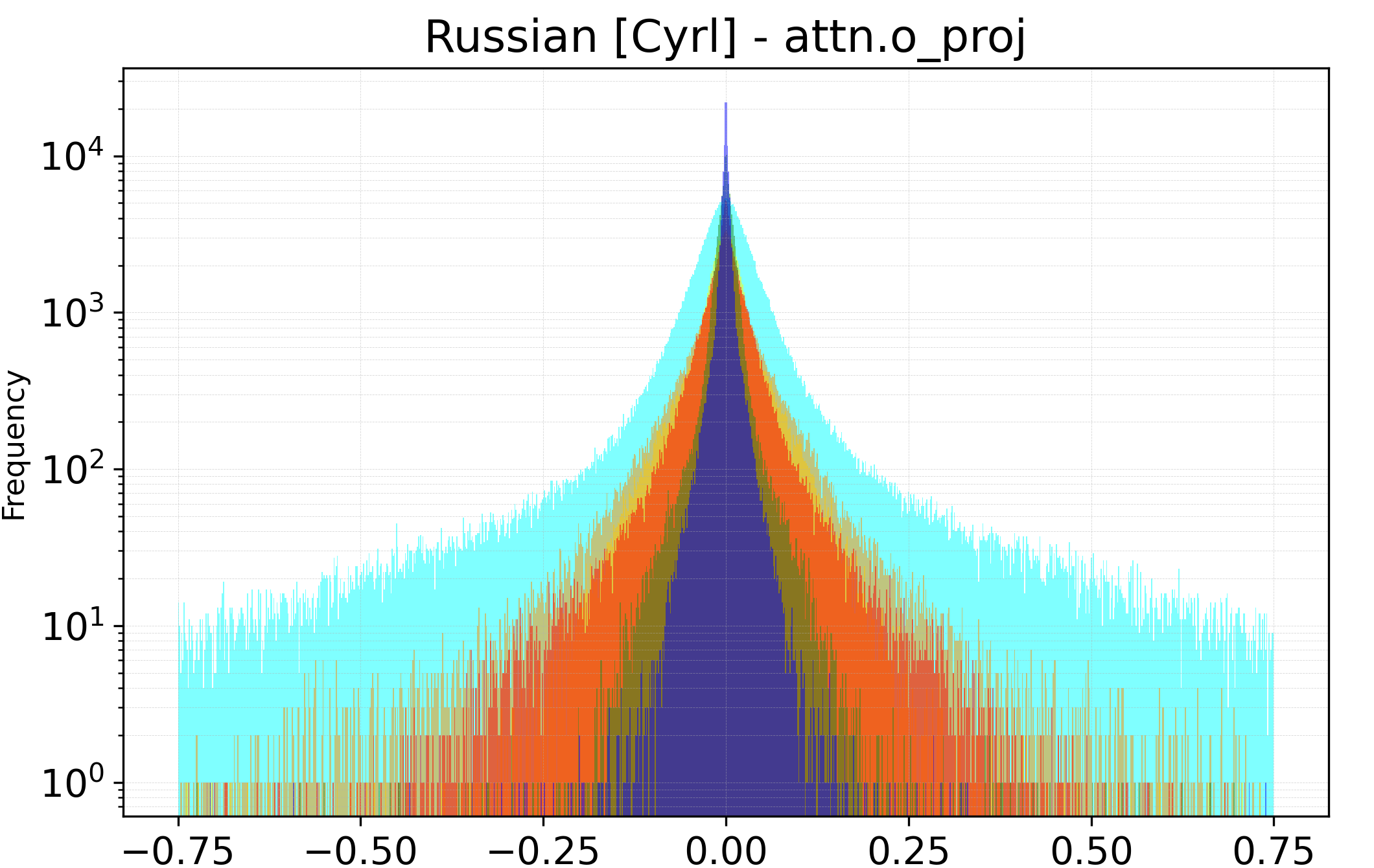}
   \end{minipage}\hfill
   \begin{minipage}{0.25\textwidth}
     \centering
     \includegraphics[width=1\linewidth]{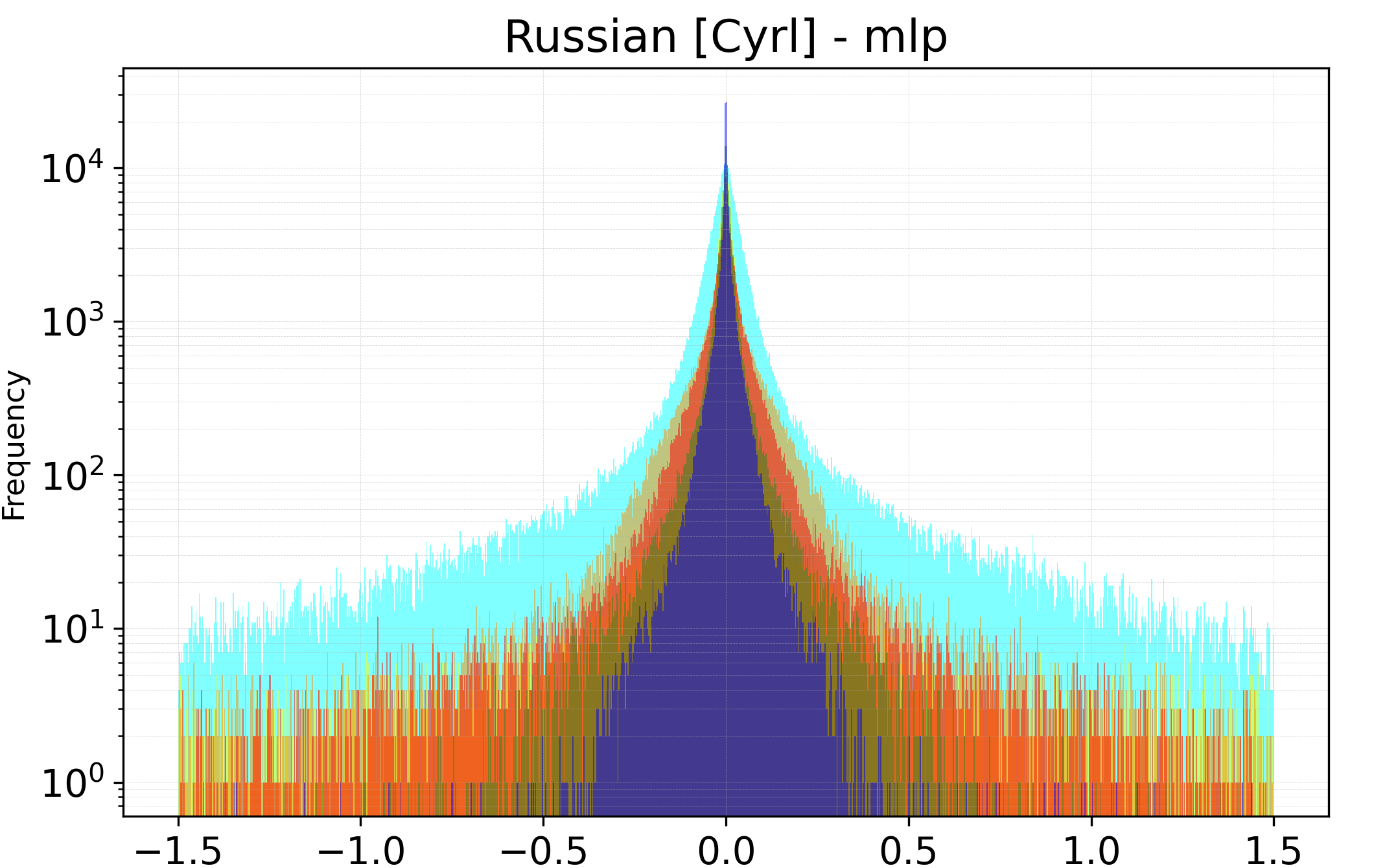}
   \end{minipage}\hfill
   \begin{minipage}{0.25\textwidth}
     \centering
     \includegraphics[width=1\linewidth]{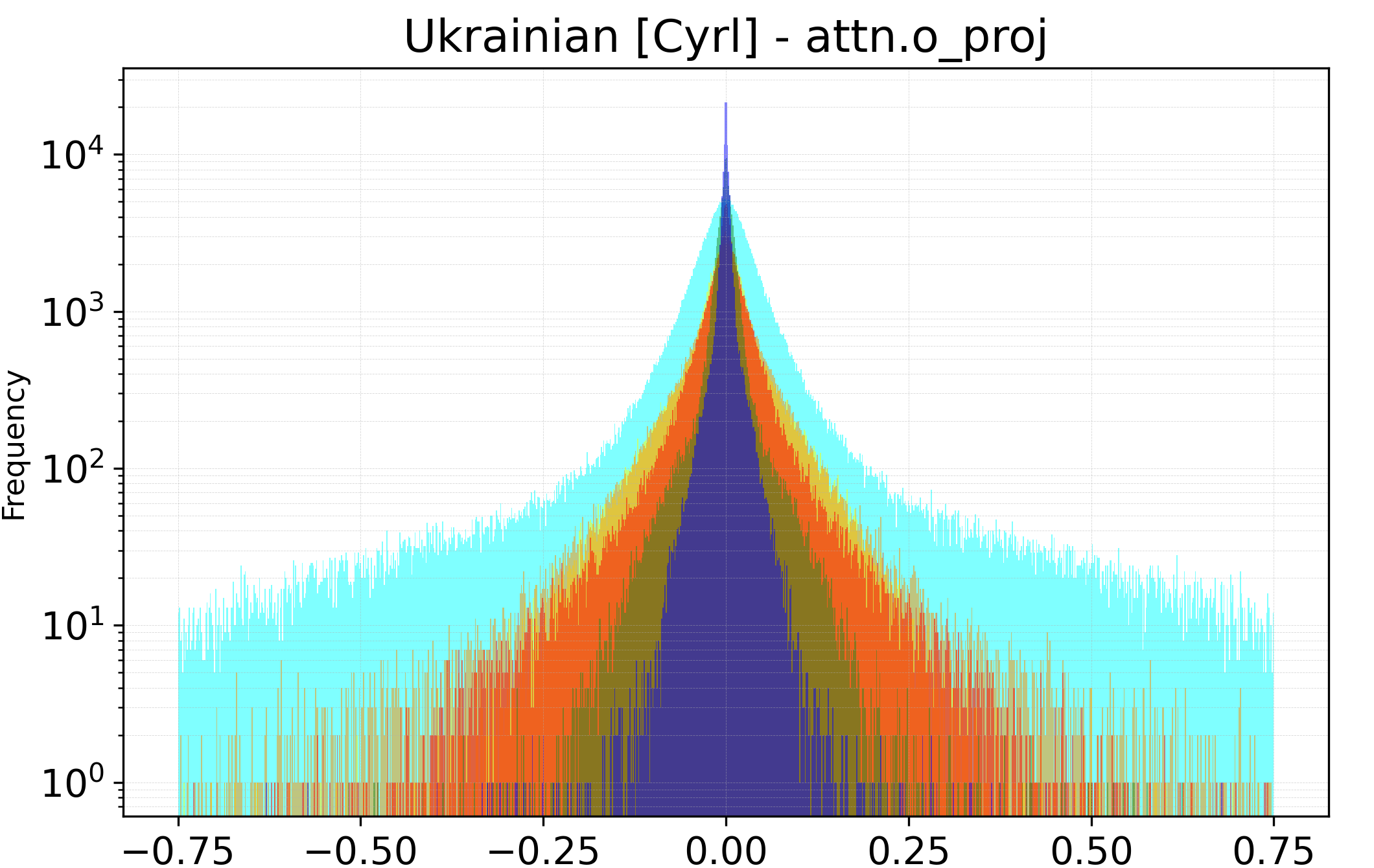}
   \end{minipage}\hfill
   \begin{minipage}{0.25\textwidth}
     \centering
     \includegraphics[width=1\linewidth]{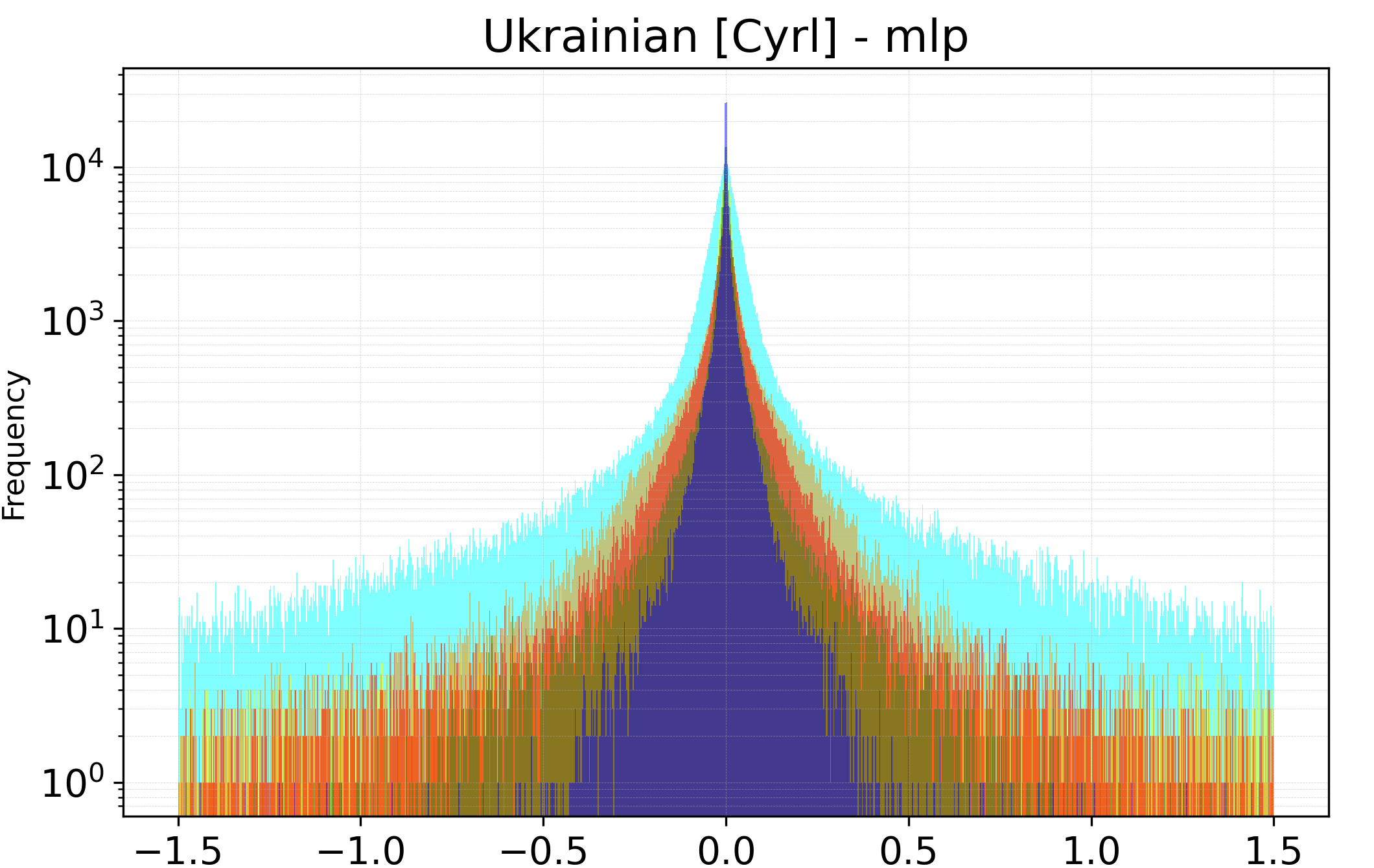}
   \end{minipage}\hfill
   \centerline{\includegraphics[width=1.0\columnwidth]{images/models.png}}
\end{figure}

\begin{figure}[H]
   \begin{minipage}{0.25\textwidth}
     \centering
     \includegraphics[width=1\linewidth]{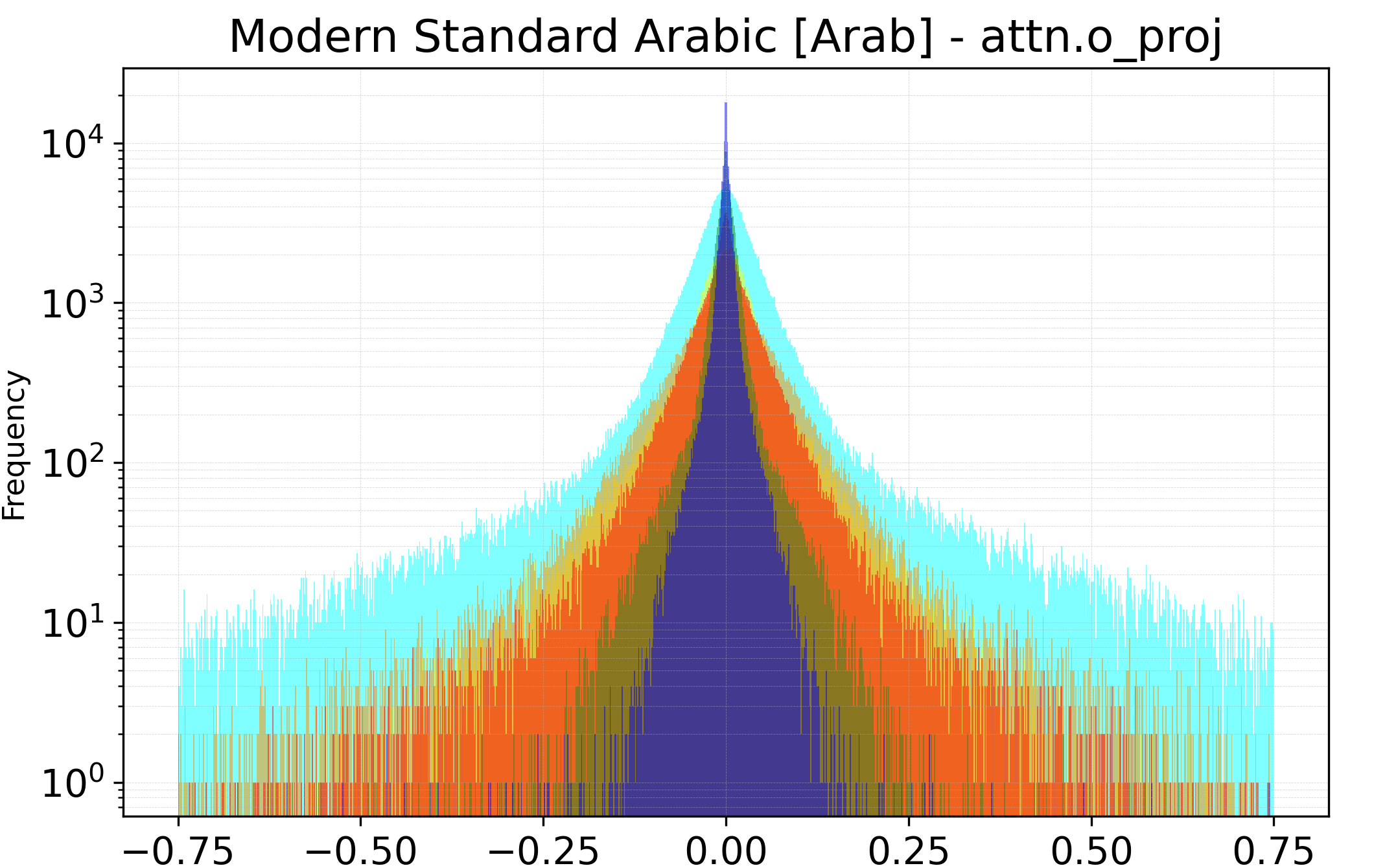}
   \end{minipage}\hfill
   \begin{minipage}{0.25\textwidth}
     \centering
     \includegraphics[width=1\linewidth]{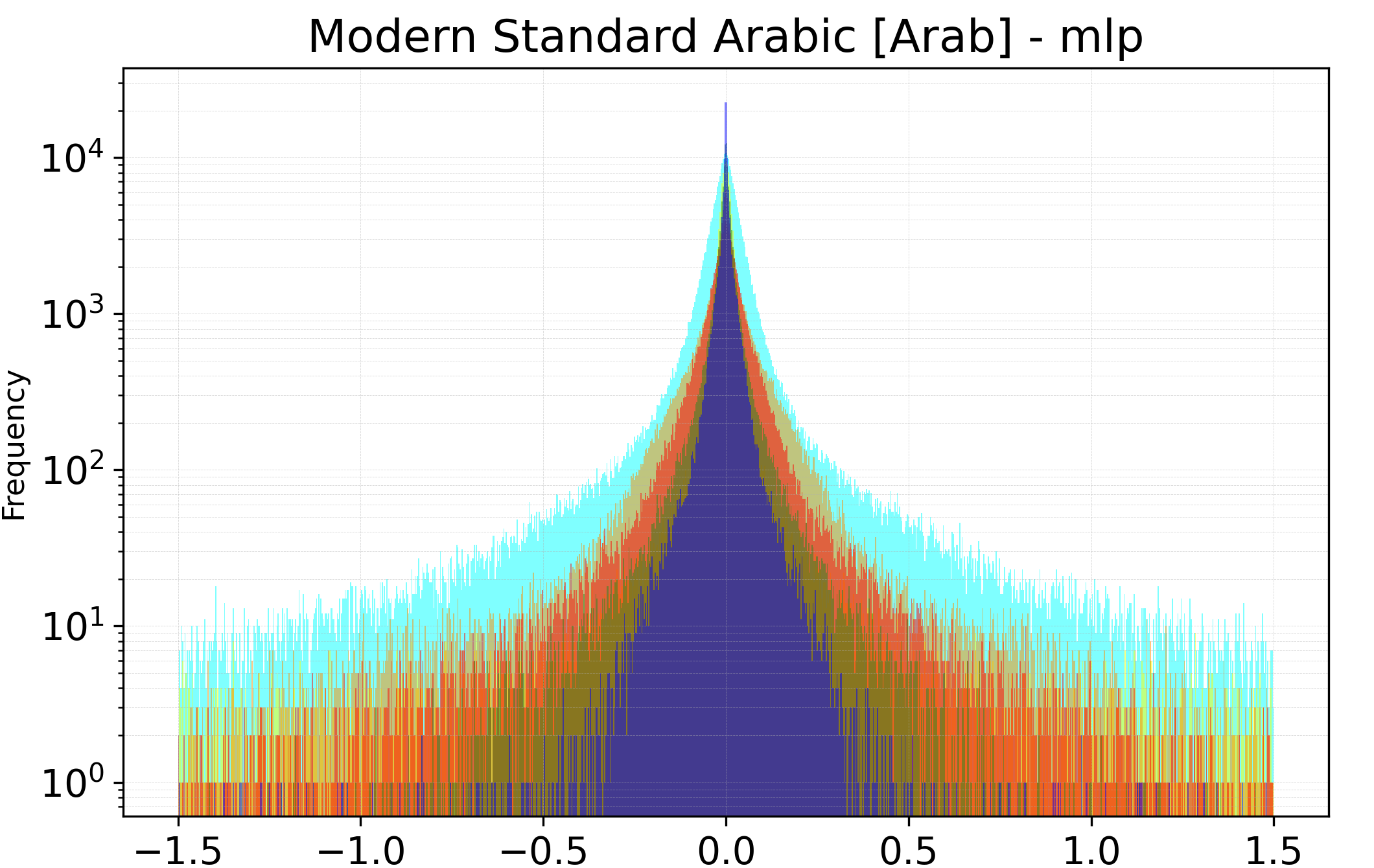}
   \end{minipage}\hfill
   \begin{minipage}{0.25\textwidth}
     \centering
     \includegraphics[width=1\linewidth]{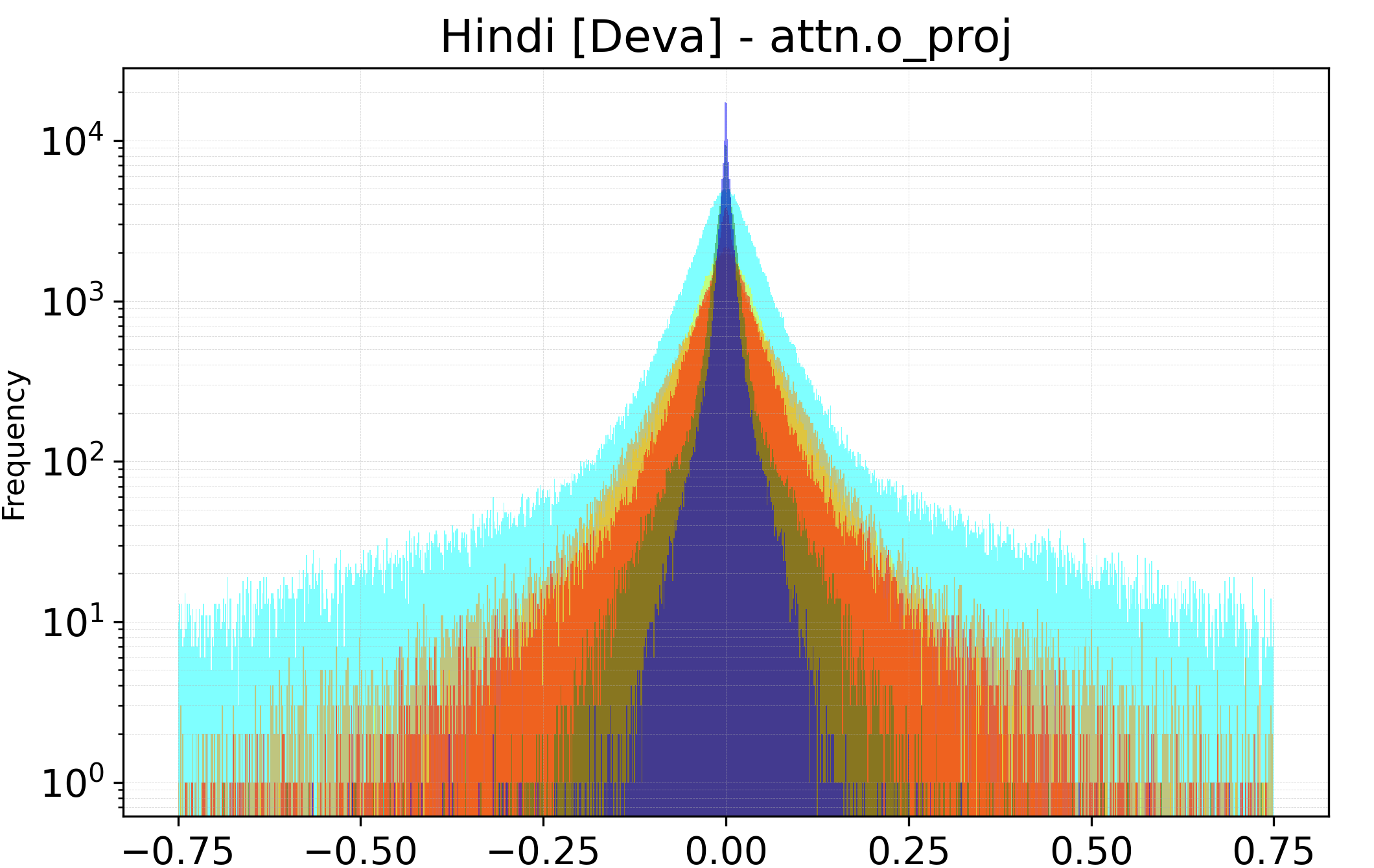}
   \end{minipage}\hfill
   \begin{minipage}{0.25\textwidth}
     \centering
     \includegraphics[width=1\linewidth]{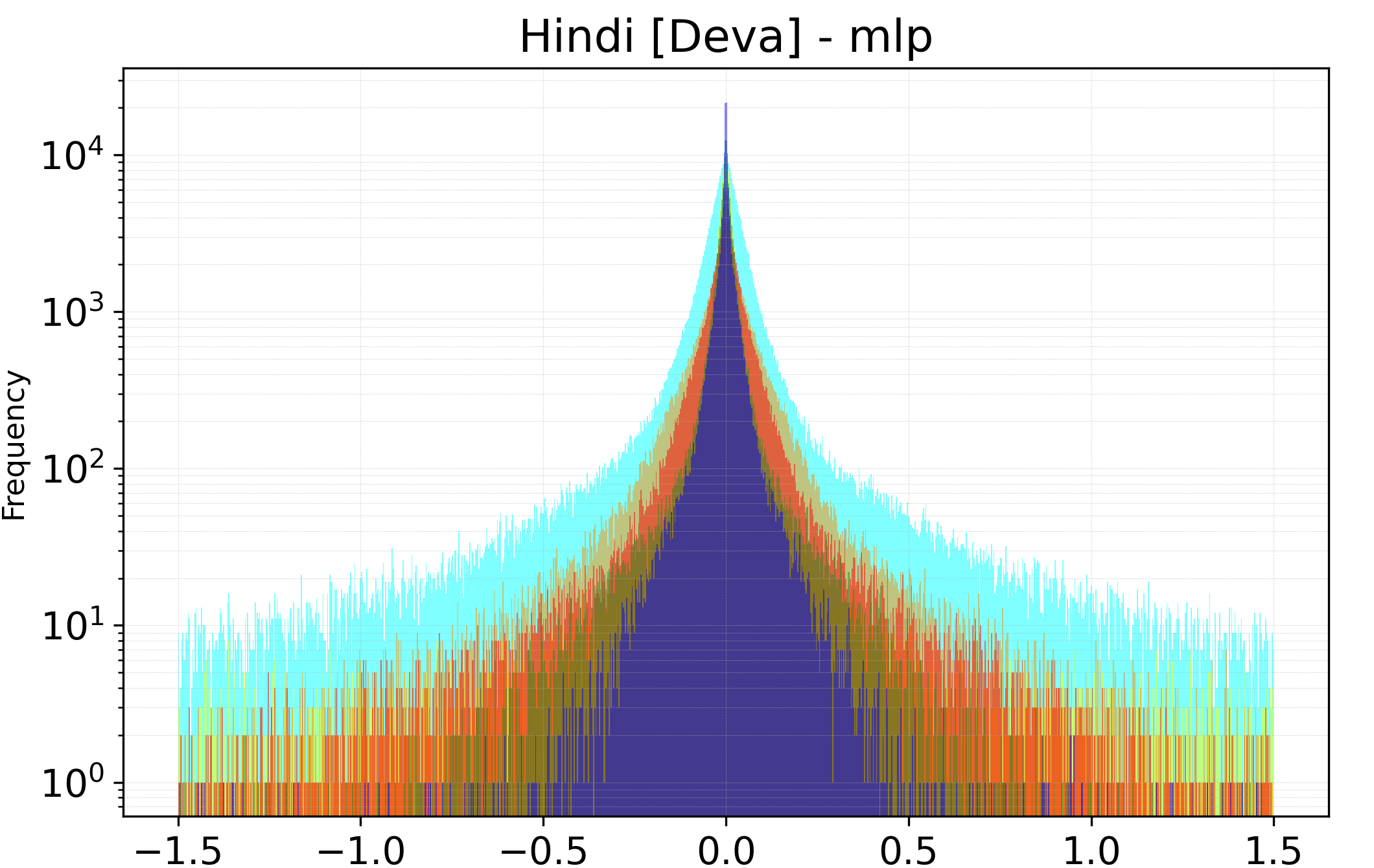}
   \end{minipage}\hfill
   \centerline{\includegraphics[width=1.0\columnwidth]{images/models.png}}
\end{figure}

\begin{figure}[H]
   \begin{minipage}{0.25\textwidth}
     \centering
     \includegraphics[width=1\linewidth]{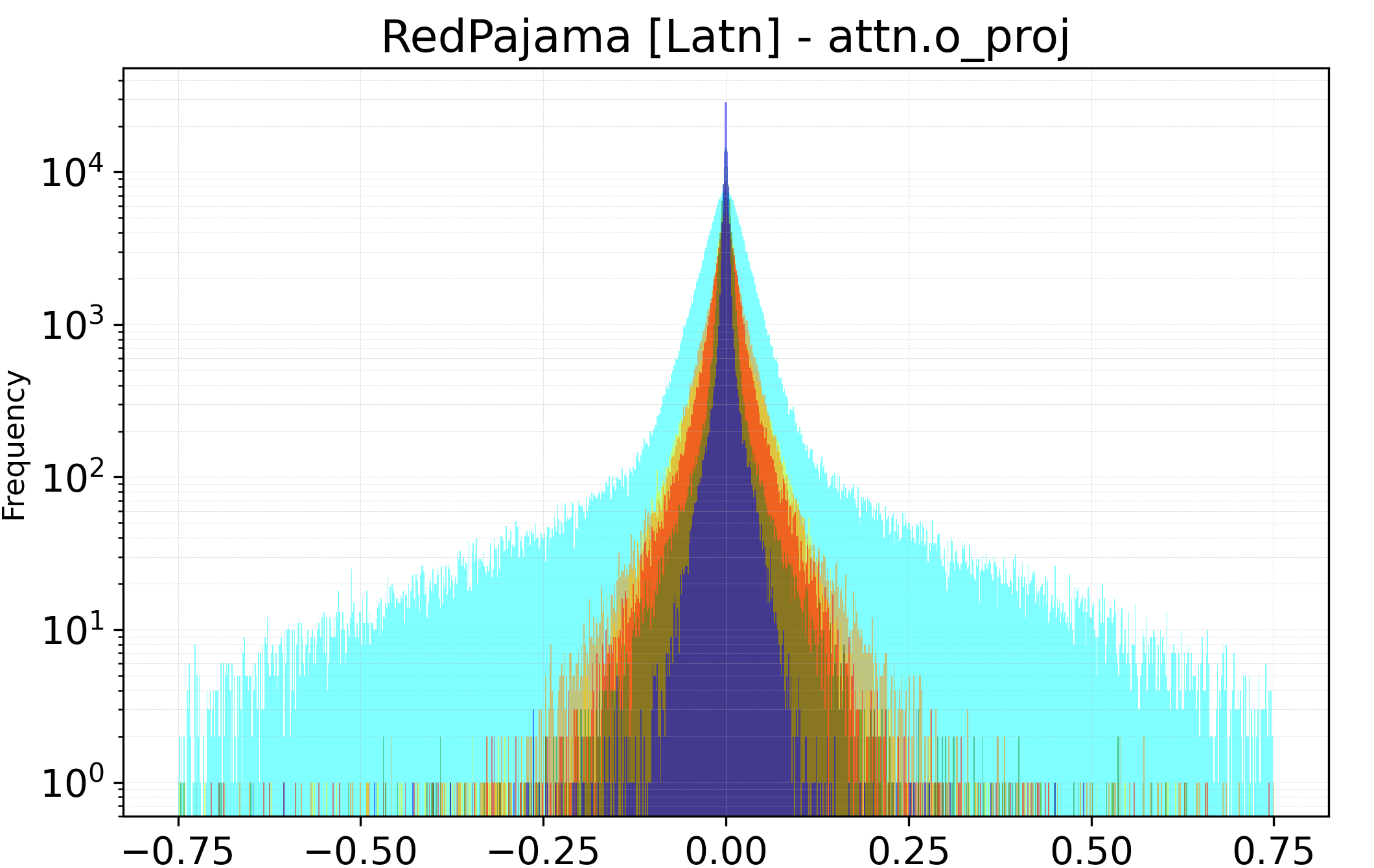}
   \end{minipage}\hfill
   \begin{minipage}{0.25\textwidth}
     \centering
     \includegraphics[width=1\linewidth]{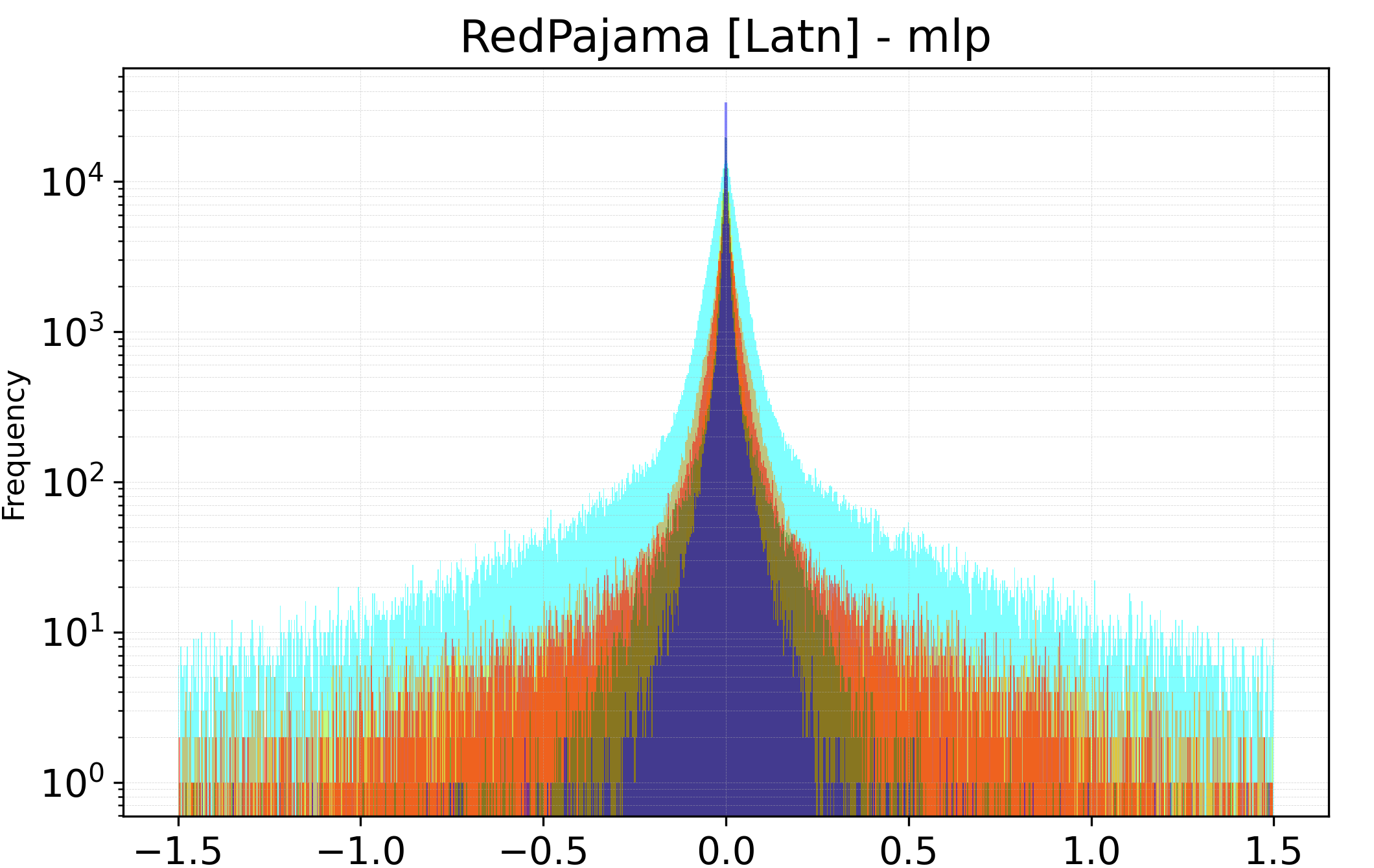}
   \end{minipage}\hfill
   \begin{minipage}{0.25\textwidth}
     \centering
     \includegraphics[width=1\linewidth]{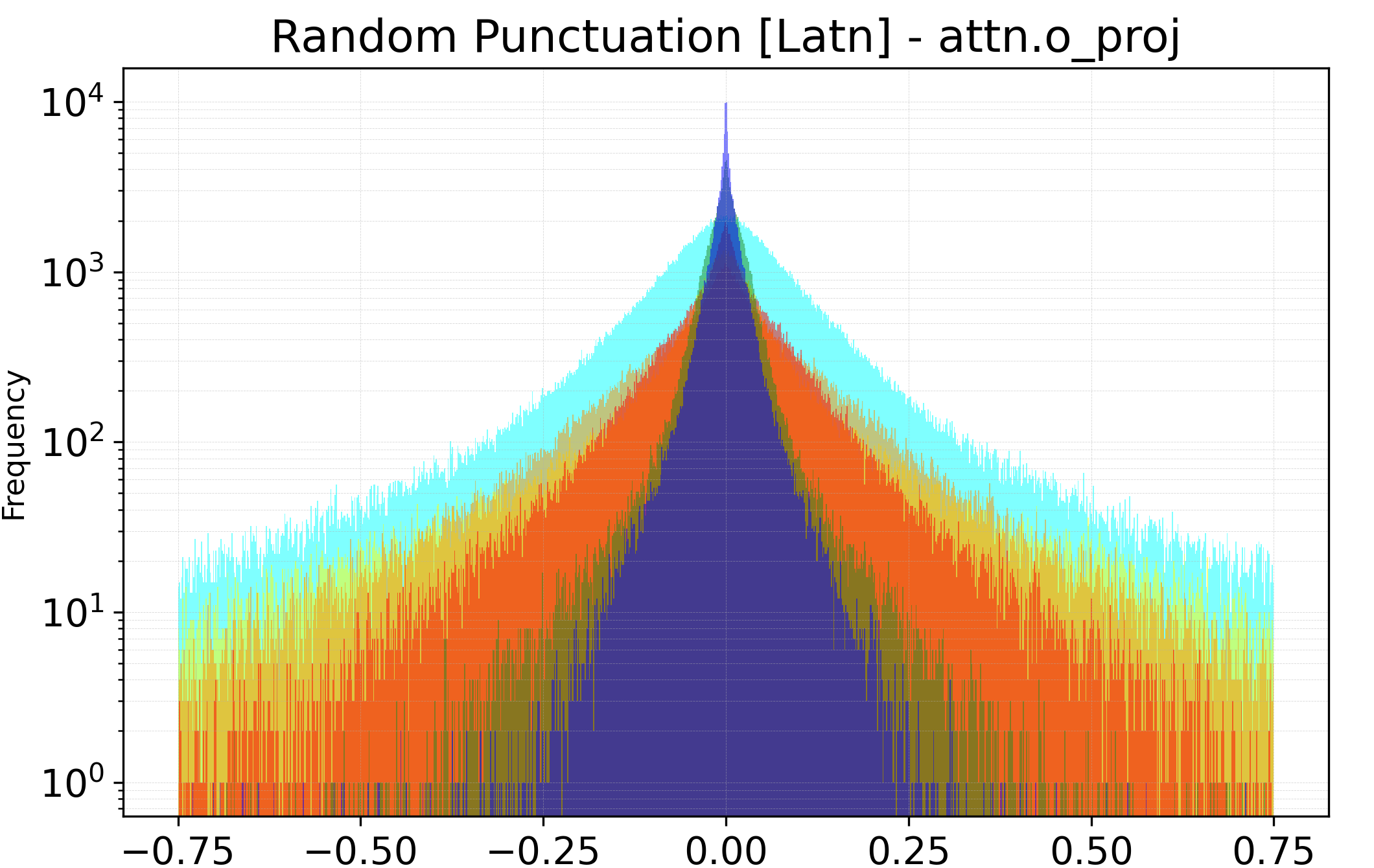}
   \end{minipage}\hfill
   \begin{minipage}{0.25\textwidth}
     \centering
     \includegraphics[width=1\linewidth]{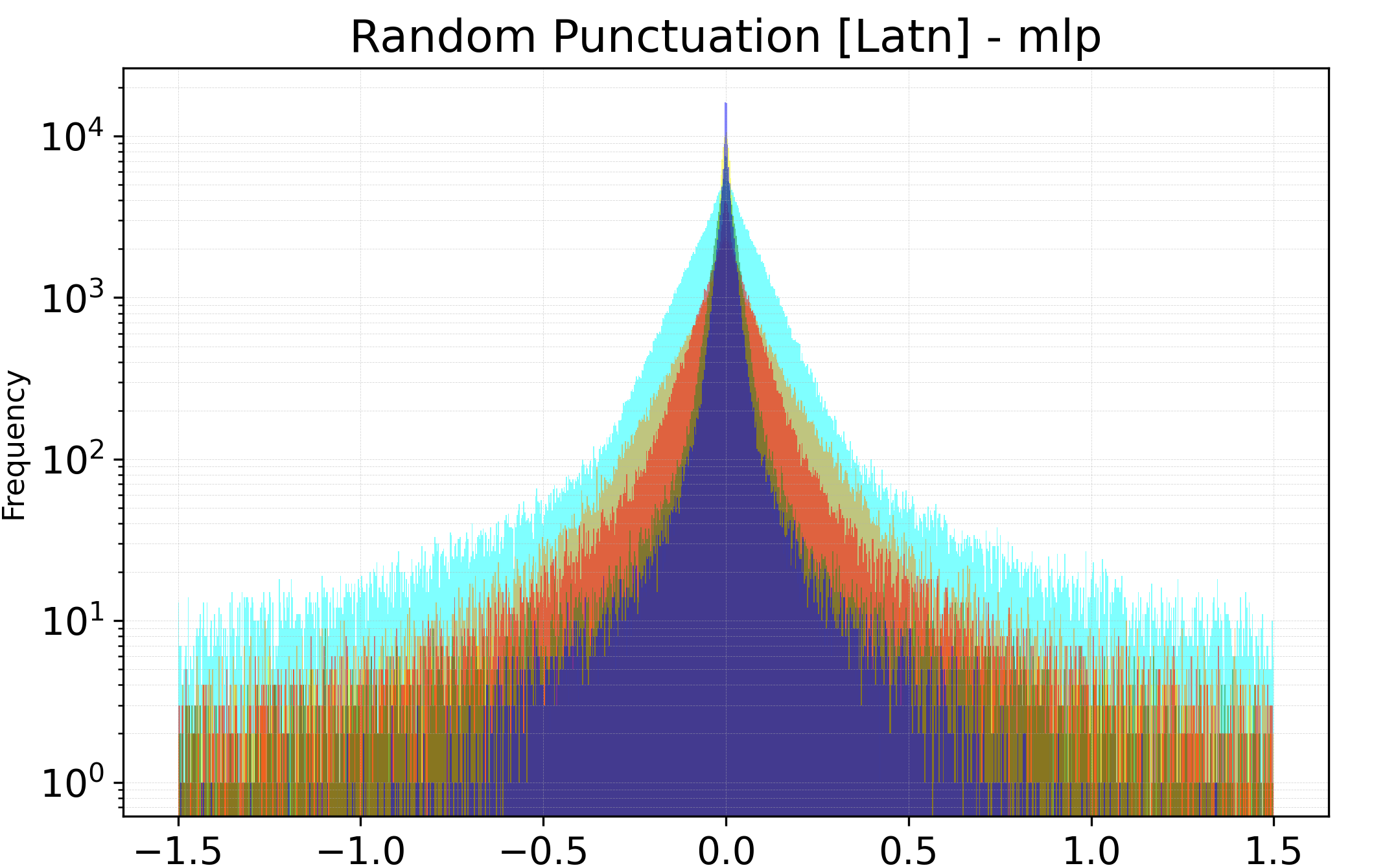}
   \end{minipage}
   \centerline{\includegraphics[width=1.0\columnwidth]{images/models.png}}
\end{figure}

\begin{figure}[H]
   \begin{minipage}{0.25\textwidth}
     \centering
     \includegraphics[width=1\linewidth]{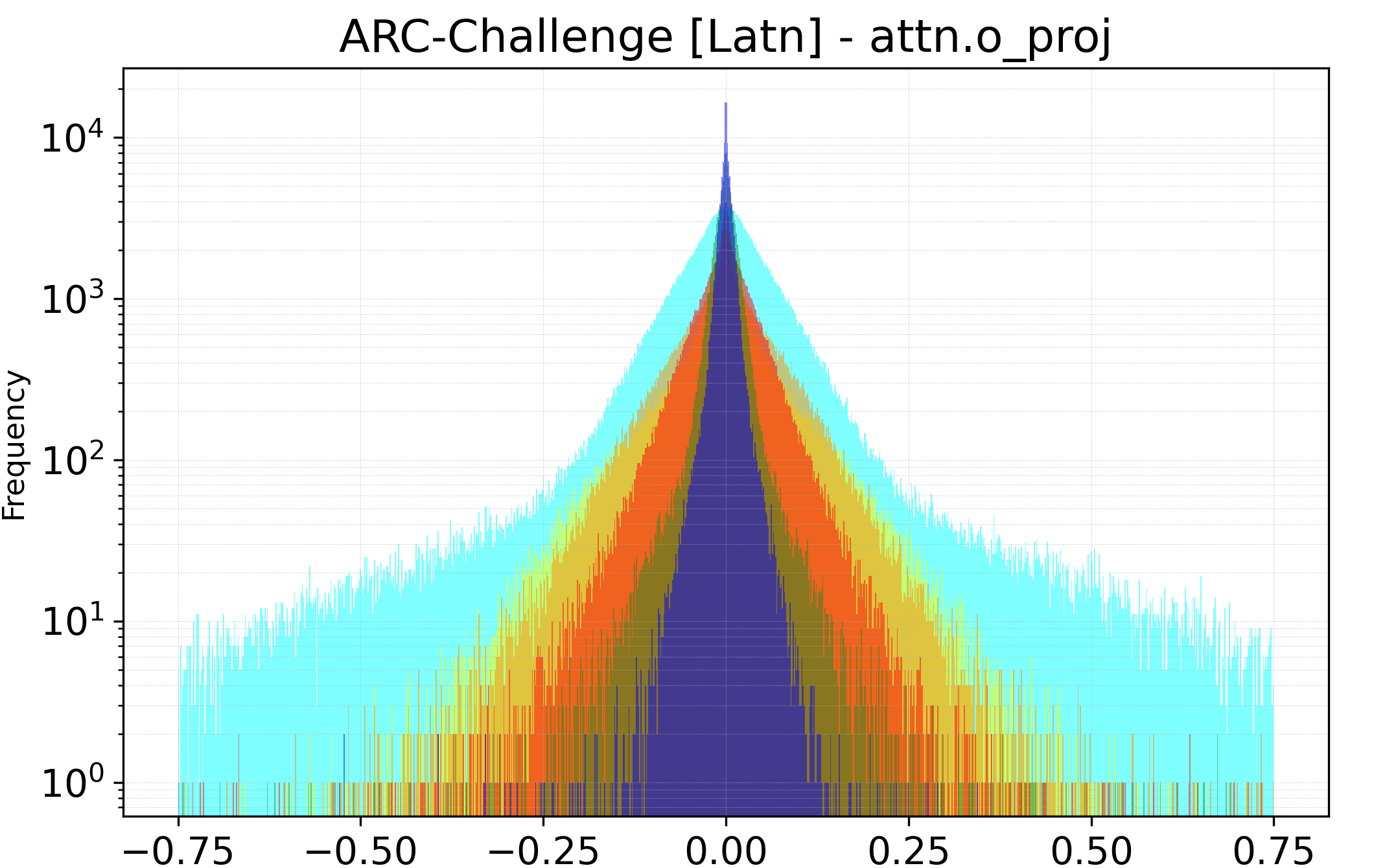}
   \end{minipage}\hfill
   \begin{minipage}{0.25\textwidth}
     \centering
     \includegraphics[width=1\linewidth]{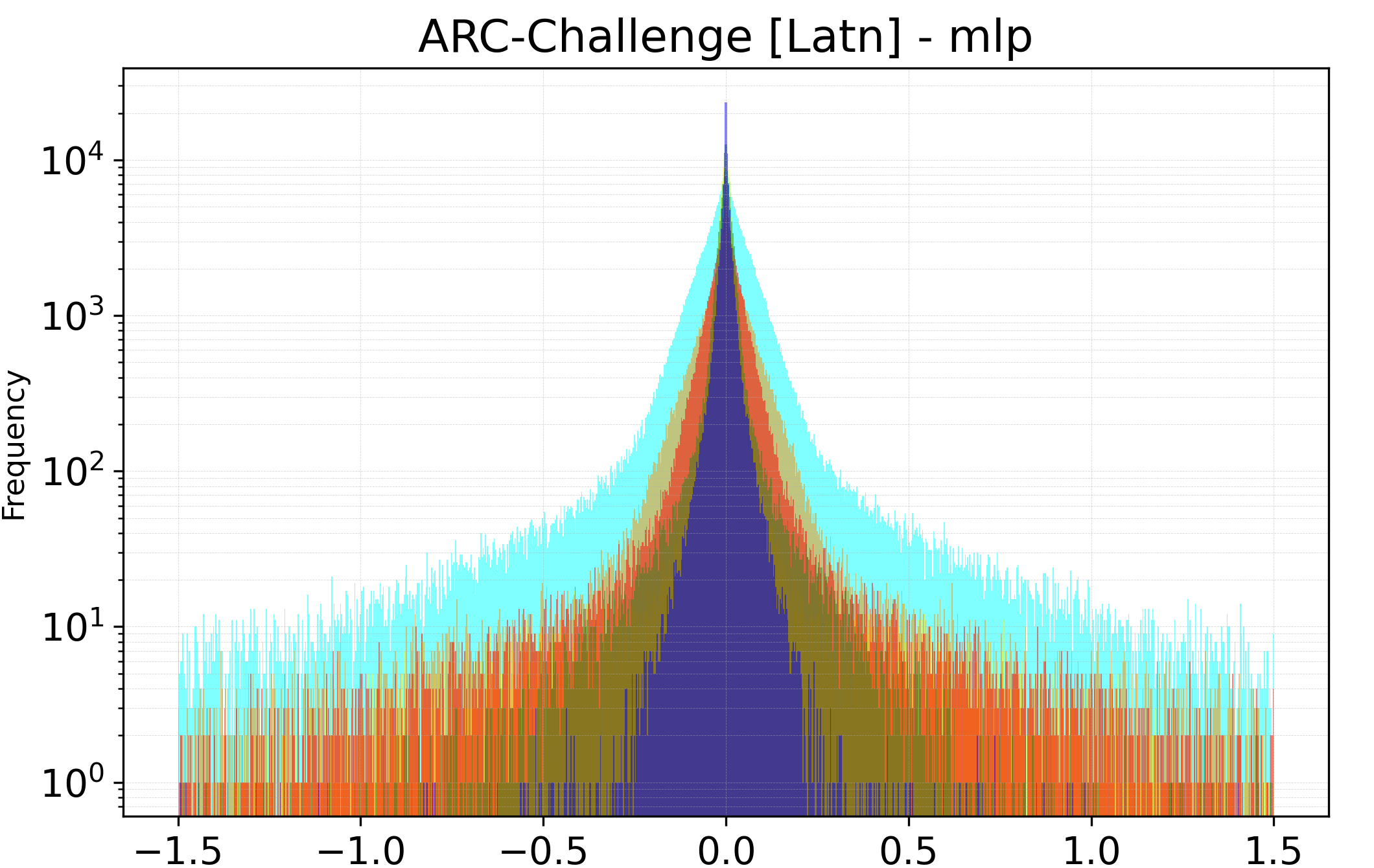}
   \end{minipage}\hfill
   \begin{minipage}{0.25\textwidth}
     \centering
     \includegraphics[width=1\linewidth]{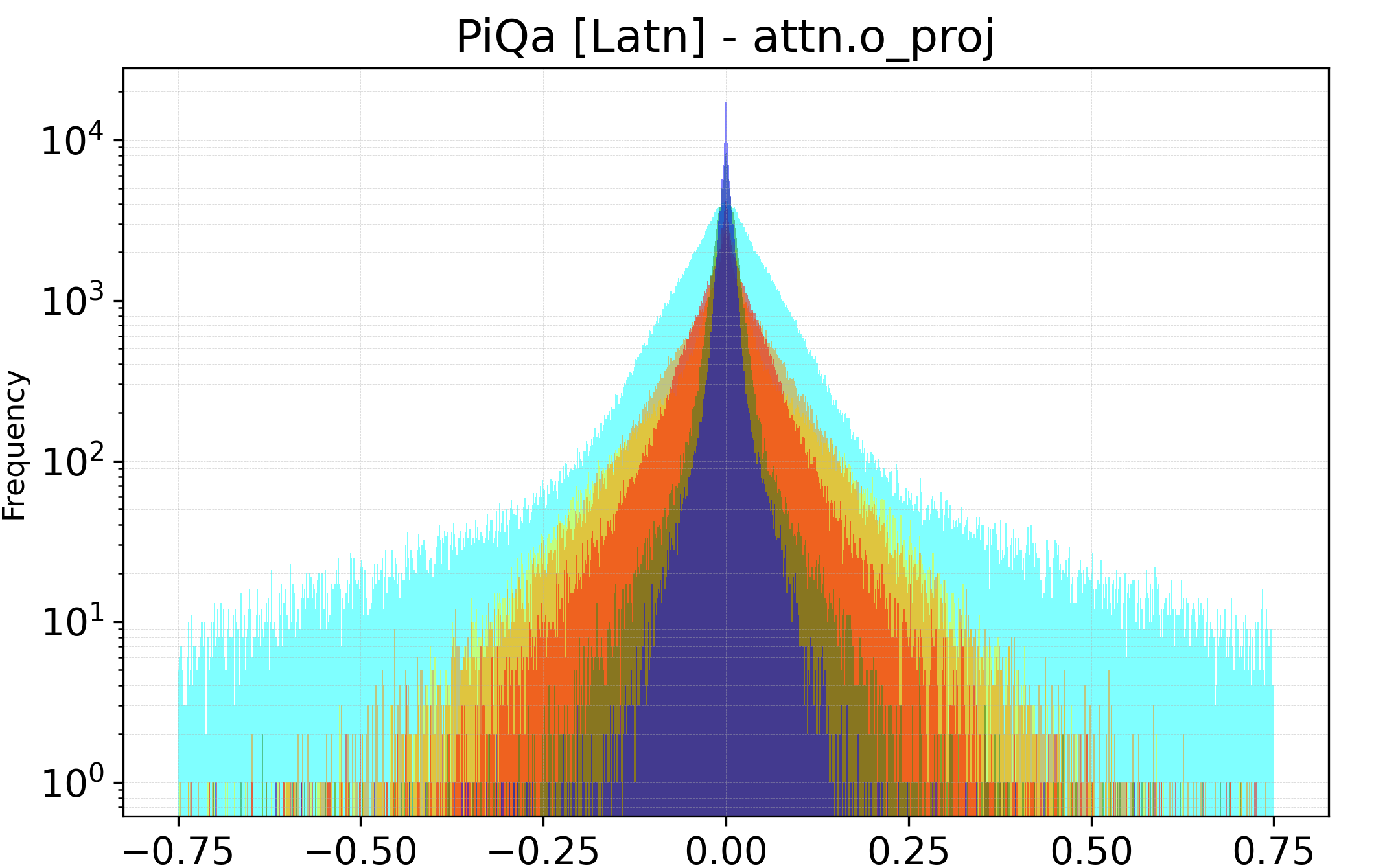}
   \end{minipage}\hfill
   \begin{minipage}{0.25\textwidth}
     \centering
     \includegraphics[width=1\linewidth]{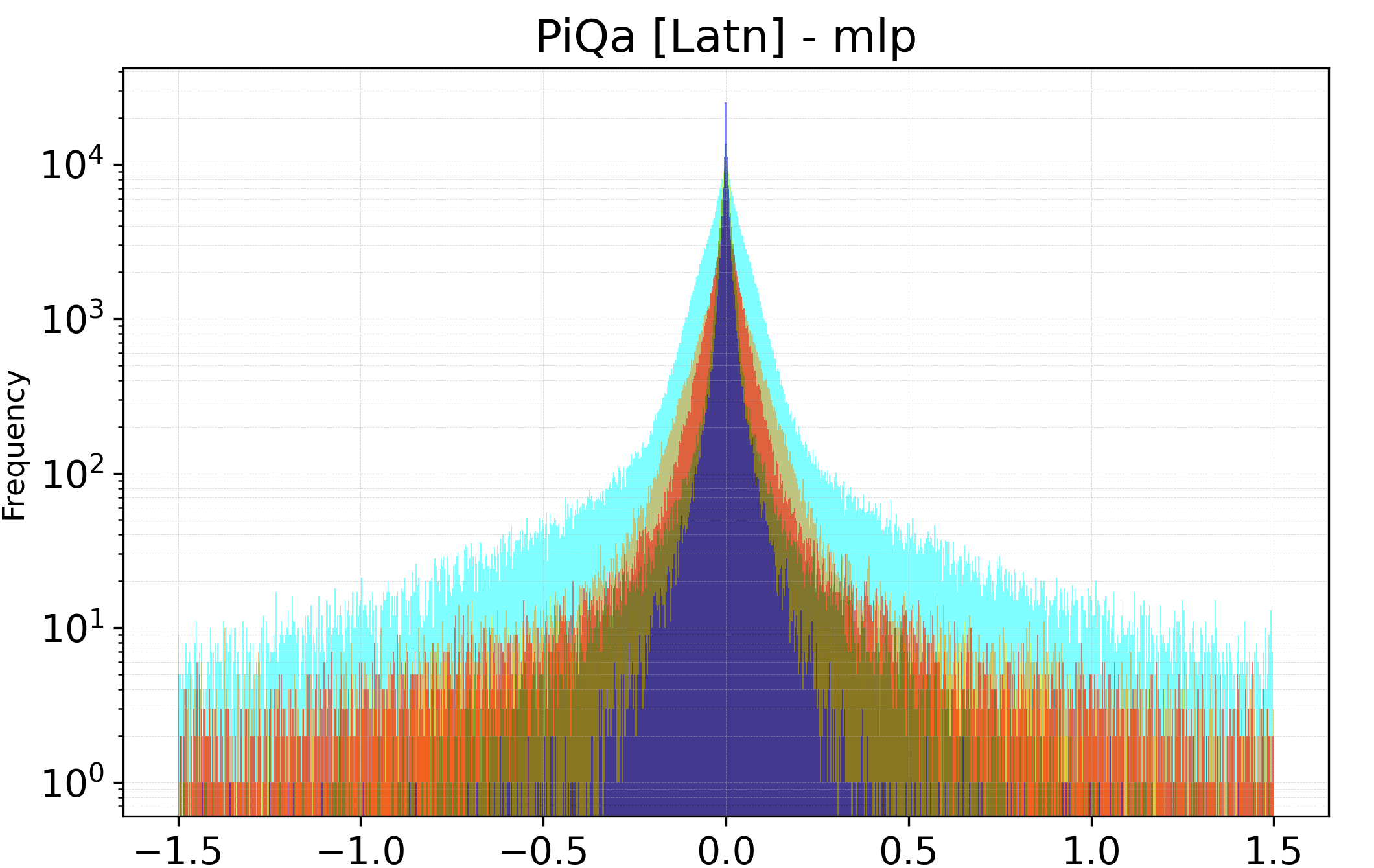}
   \end{minipage}
   \centerline{\includegraphics[width=1.0\columnwidth]{images/models.png}}
\end{figure}

%% file: main.bbl
\begin{thebibliography}{37}
\providecommand{\natexlab}[1]{#1}
\providecommand{\url}[1]{\texttt{#1}}
\expandafter\ifx\csname urlstyle\endcsname\relax
  \providecommand{\doi}[1]{doi: #1}\else
  \providecommand{\doi}{doi: \begingroup \urlstyle{rm}\Url}\fi

\bibitem[Ahmadian et~al.(2023)Ahmadian, Dash, Chen, Venkitesh, Gou, Blunsom,
  {\"U}st{\"u}n, and Hooker]{ahmadian2023intriguing}
Arash Ahmadian, Saurabh Dash, Hongyu Chen, Bharat Venkitesh, Stephen Gou, Phil
  Blunsom, Ahmet {\"U}st{\"u}n, and Sara Hooker.
\newblock Intriguing properties of quantization at scale.
\newblock \emph{arXiv preprint arXiv:2305.19268}, 2023.

\bibitem[AI@Meta(2024)]{llama3modelcard}
AI@Meta.
\newblock Llama 3 model card.
\newblock 2024.
\newblock URL
  \url{https://github.com/meta-llama/llama3/blob/main/MODEL_CARD.md}.

\bibitem[Bisk et~al.(2020)Bisk, Zellers, Gao, Choi, et~al.]{bisk2020piqa}
Yonatan Bisk, Rowan Zellers, Jianfeng Gao, Yejin Choi, et~al.
\newblock Piqa: Reasoning about physical commonsense in natural language.
\newblock In \emph{Proceedings of the AAAI conference on artificial
  intelligence}, volume~34, pages 7432--7439, 2020.

\bibitem[C4AI(2024)]{commandr}
C4AI.
\newblock Model card for c4ai command-r.
\newblock 2024.
\newblock URL \url{https://huggingface.co/CohereForAI/c4ai-command-r-v01}.

\bibitem[Chee et~al.(2024)Chee, Cai, Kuleshov, and De~Sa]{chee2024quip}
Jerry Chee, Yaohui Cai, Volodymyr Kuleshov, and Christopher~M De~Sa.
\newblock Quip: 2-bit quantization of large language models with guarantees.
\newblock \emph{Advances in Neural Information Processing Systems}, 36, 2024.

\bibitem[Chowdhery et~al.(2023)Chowdhery, Narang, Devlin, Bosma, Mishra,
  Roberts, Barham, Chung, Sutton, Gehrmann, et~al.]{chowdhery2023palm}
Aakanksha Chowdhery, Sharan Narang, Jacob Devlin, Maarten Bosma, Gaurav Mishra,
  Adam Roberts, Paul Barham, Hyung~Won Chung, Charles Sutton, Sebastian
  Gehrmann, et~al.
\newblock Palm: Scaling language modeling with pathways.
\newblock \emph{Journal of Machine Learning Research}, 24\penalty0
  (240):\penalty0 1--113, 2023.

\bibitem[Clark et~al.(2018)Clark, Cowhey, Etzioni, Khot, Sabharwal, Schoenick,
  and Tafjord]{clark2018think}
Peter Clark, Isaac Cowhey, Oren Etzioni, Tushar Khot, Ashish Sabharwal, Carissa
  Schoenick, and Oyvind Tafjord.
\newblock Think you have solved question answering? try arc, the ai2 reasoning
  challenge.
\newblock \emph{arXiv preprint arXiv:1803.05457}, 2018.

\bibitem[Computer(2023)]{together2023redpajama}
Together Computer.
\newblock Redpajama: an open dataset for training large language models, 2023.
\newblock URL \url{https://github.com/togethercomputer/RedPajama-Data}.

\bibitem[Costa-juss{\`a} et~al.(2022)Costa-juss{\`a}, Cross, {\c{C}}elebi,
  Elbayad, Heafield, Heffernan, Kalbassi, Lam, Licht, Maillard,
  et~al.]{costa2022no}
Marta~R Costa-juss{\`a}, James Cross, Onur {\c{C}}elebi, Maha Elbayad, Kenneth
  Heafield, Kevin Heffernan, Elahe Kalbassi, Janice Lam, Daniel Licht, Jean
  Maillard, et~al.
\newblock No language left behind: Scaling human-centered machine translation.
\newblock \emph{arXiv preprint arXiv:2207.04672}, 2022.

\bibitem[Dettmers and Zettlemoyer(2023)]{dettmers2023case}
Tim Dettmers and Luke Zettlemoyer.
\newblock The case for 4-bit precision: k-bit inference scaling laws.
\newblock In \emph{International Conference on Machine Learning}, pages
  7750--7774. PMLR, 2023.

\bibitem[Dettmers et~al.(2022)Dettmers, Lewis, Belkada, and
  Zettlemoyer]{dettmers2022llm}
Tim Dettmers, Mike Lewis, Younes Belkada, and Luke Zettlemoyer.
\newblock Llm. int8 (): 8-bit matrix multiplication for transformers at scale.
\newblock \emph{arXiv preprint arXiv:2208.07339}, 2022.

\bibitem[Dettmers et~al.(2023{\natexlab{a}})Dettmers, Pagnoni, Holtzman, and
  Zettlemoyer]{dettmers2023qlora}
Tim Dettmers, Artidoro Pagnoni, Ari Holtzman, and Luke Zettlemoyer.
\newblock Qlora: Efficient finetuning of quantized llms.
\newblock \emph{arXiv preprint arXiv:2305.14314}, 2023{\natexlab{a}}.

\bibitem[Dettmers et~al.(2023{\natexlab{b}})Dettmers, Svirschevski, Egiazarian,
  Kuznedelev, Frantar, Ashkboos, Borzunov, Hoefler, and
  Alistarh]{dettmers2023spqr}
Tim Dettmers, Ruslan Svirschevski, Vage Egiazarian, Denis Kuznedelev, Elias
  Frantar, Saleh Ashkboos, Alexander Borzunov, Torsten Hoefler, and Dan
  Alistarh.
\newblock Spqr: A sparse-quantized representation for near-lossless llm weight
  compression.
\newblock \emph{arXiv preprint arXiv:2306.03078}, 2023{\natexlab{b}}.

\bibitem[Doumbouya et~al.(2023)Doumbouya, Diané, Cissé, Diané, Sow,
  Doumbouya, Bangoura, Bayo, Condé, Diané, Piech, and Manning]{mt4nko-23}
Moussa Doumbouya, Baba~Mamadi Diané, Solo~Farabado Cissé, Djibrila Diané,
  Abdoulaye Sow, Séré~Moussa Doumbouya, Daouda Bangoura, Fodé~Moriba Bayo,
  Ibrahima Sory~2. Condé, Kalo~Mory Diané, Chris Piech, and Christopher
  Manning.
\newblock Machine translation for nko: Tools, corpora, and baseline results.
\newblock In \emph{Proceedings of the Eighth Conference on Machine
  Translation}, pages 312--343, Singapore, 2023. Association for Computational
  Linguistics.
\newblock URL \url{https://aclanthology.org/2023.wmt-1.34}.

\bibitem[Frantar et~al.(2022)Frantar, Ashkboos, Hoefler, and
  Alistarh]{frantar2022gptq}
Elias Frantar, Saleh Ashkboos, Torsten Hoefler, and Dan Alistarh.
\newblock Gptq: Accurate post-training quantization for generative pre-trained
  transformers.
\newblock \emph{arXiv preprint arXiv:2210.17323}, 2022.

\bibitem[Gala et~al.(2023)Gala, Chitale, AK, Doddapaneni, Gumma, Kumar, Nawale,
  Sujatha, Puduppully, Raghavan, Kumar, Khapra, Dabre, and
  Kunchukuttan]{indictrans2-23}
Jay Gala, Pranjal~A. Chitale, Raghavan AK, Sumanth Doddapaneni, Varun Gumma,
  Aswanth Kumar, Janki Nawale, Anupama Sujatha, Ratish Puduppully, Vivek
  Raghavan, Pratyush Kumar, Mitesh~M. Khapra, Raj Dabre, and Anoop
  Kunchukuttan.
\newblock Indictrans2: Towards high-quality and accessible machine translation
  models for all 22 scheduled indian languages.
\newblock 2023.

\bibitem[Goyal et~al.(2022)Goyal, Gao, Chaudhary, Chen, Wenzek, Ju, Krishnan,
  Ranzato, Guzmán, and Fan]{flores101-22}
Naman Goyal, Cynthia Gao, Vishrav Chaudhary, Peng-Jen Chen, Guillaume Wenzek,
  Da~Ju, Sanjana Krishnan, Marc’Aurelio Ranzato, Francisco Guzmán, and
  Angela Fan.
\newblock The {F}lores-101 evaluation benchmark for low-resource and
  multilingual machine translation.
\newblock \emph{Transactions of the Association for Computational Linguistics},
  10, 2022.

\bibitem[Guzmán et~al.(2019)Guzmán, Chen, Ott, Pino, Lample, Koehn,
  Chaudhary, and Ranzato]{flores1-19}
Francisco Guzmán, Peng-Jen Chen, Myle Ott, Juan Pino, Guillaume Lample,
  Philipp Koehn, Vishrav Chaudhary, and Marc’Aurelio Ranzato.
\newblock The {FLORES} evaluation datasets for low-resource machine
  translation: {N}epali{--}{E}nglish and {S}inhala{--}{E}nglish.
\newblock In \emph{Proceedings of the 2019 Conference on Empirical Methods in
  Natural Language Processing and the 9th International Joint Conference on
  Natural Language Processing (EMNLP-IJCNLP)}, pages 6098--6111, Hong Kong,
  China, 2019. Association for Computational Linguistics.
\newblock URL \url{https://aclanthology.org/D19-1632}.

\bibitem[Hoffmann et~al.(2022)Hoffmann, Borgeaud, Mensch, Buchatskaya, Cai,
  Rutherford, Casas, Hendricks, Welbl, Clark, et~al.]{hoffmann2022training}
Jordan Hoffmann, Sebastian Borgeaud, Arthur Mensch, Elena Buchatskaya, Trevor
  Cai, Eliza Rutherford, Diego de~Las Casas, Lisa~Anne Hendricks, Johannes
  Welbl, Aidan Clark, et~al.
\newblock Training compute-optimal large language models.
\newblock \emph{arXiv preprint arXiv:2203.15556}, 2022.

\bibitem[Jiang et~al.(2023)Jiang, Sablayrolles, Mensch, Bamford, Chaplot,
  Casas, Bressand, Lengyel, Lample, Saulnier, et~al.]{jiang2023mistral}
Albert~Q Jiang, Alexandre Sablayrolles, Arthur Mensch, Chris Bamford,
  Devendra~Singh Chaplot, Diego de~las Casas, Florian Bressand, Gianna Lengyel,
  Guillaume Lample, Lucile Saulnier, et~al.
\newblock Mistral 7b.
\newblock \emph{arXiv preprint arXiv:2310.06825}, 2023.

\bibitem[Kalamkar et~al.(2019)Kalamkar, Mudigere, Mellempudi, Das, Banerjee,
  Avancha, Vooturi, Jammalamadaka, Huang, Yuen, et~al.]{kalamkar2019study}
Dhiraj Kalamkar, Dheevatsa Mudigere, Naveen Mellempudi, Dipankar Das, Kunal
  Banerjee, Sasikanth Avancha, Dharma~Teja Vooturi, Nataraj Jammalamadaka,
  Jianyu Huang, Hector Yuen, et~al.
\newblock A study of bfloat16 for deep learning training.
\newblock \emph{arXiv preprint arXiv:1905.12322}, 2019.

\bibitem[Kaplan et~al.(2020)Kaplan, McCandlish, Henighan, Brown, Chess, Child,
  Gray, Radford, Wu, and Amodei]{kaplan2020scaling}
Jared Kaplan, Sam McCandlish, Tom Henighan, Tom~B Brown, Benjamin Chess, Rewon
  Child, Scott Gray, Alec Radford, Jeffrey Wu, and Dario Amodei.
\newblock Scaling laws for neural language models.
\newblock \emph{arXiv preprint arXiv:2001.08361}, 2020.

\bibitem[Kim et~al.(2023)Kim, Hooper, Gholami, Dong, Li, Shen, Mahoney, and
  Keutzer]{kim2023squeezellm}
Sehoon Kim, Coleman Hooper, Amir Gholami, Zhen Dong, Xiuyu Li, Sheng Shen,
  Michael~W Mahoney, and Kurt Keutzer.
\newblock Squeezellm: Dense-and-sparse quantization.
\newblock \emph{arXiv preprint arXiv:2306.07629}, 2023.

\bibitem[Li et~al.(2023)Li, Bubeck, Eldan, Del~Giorno, Gunasekar, and
  Lee]{li2023textbooks}
Yuanzhi Li, S{\'e}bastien Bubeck, Ronen Eldan, Allie Del~Giorno, Suriya
  Gunasekar, and Yin~Tat Lee.
\newblock Textbooks are all you need ii: phi-1.5 technical report.
\newblock \emph{arXiv preprint arXiv:2309.05463}, 2023.

\bibitem[Lin et~al.(2023)Lin, Tang, Tang, Yang, Dang, and Han]{lin2023awq}
Ji~Lin, Jiaming Tang, Haotian Tang, Shang Yang, Xingyu Dang, and Song Han.
\newblock Awq: Activation-aware weight quantization for llm compression and
  acceleration.
\newblock \emph{arXiv preprint arXiv:2306.00978}, 2023.

\bibitem[Merity et~al.(2016)Merity, Xiong, Bradbury, and
  Socher]{merity2016pointer}
Stephen Merity, Caiming Xiong, James Bradbury, and Richard Socher.
\newblock Pointer sentinel mixture models.
\newblock \emph{arXiv preprint arXiv:1609.07843}, 2016.

\bibitem[Metaseq(2022)]{Metaseq2022issue}
Metaseq.
\newblock Metaseq github issue, 2022.
\newblock URL \url{https://github.com/facebookresearch/metaseq/issues/213}.

\bibitem[Rajbhandari et~al.(2021)Rajbhandari, Ruwase, Rasley, Smith, and
  He]{rajbhandari2021zero}
Samyam Rajbhandari, Olatunji Ruwase, Jeff Rasley, Shaden Smith, and Yuxiong He.
\newblock Zero-infinity: Breaking the gpu memory wall for extreme scale deep
  learning.
\newblock In \emph{Proceedings of the International Conference for High
  Performance Computing, Networking, Storage and Analysis}, pages 1--14, 2021.

\bibitem[Sakaguchi et~al.(2021)Sakaguchi, Bras, Bhagavatula, and
  Choi]{sakaguchi2021winogrande}
Keisuke Sakaguchi, Ronan~Le Bras, Chandra Bhagavatula, and Yejin Choi.
\newblock Winogrande: An adversarial winograd schema challenge at scale.
\newblock \emph{Communications of the ACM}, 64\penalty0 (9):\penalty0 99--106,
  2021.

\bibitem[Touvron et~al.(2023{\natexlab{a}})Touvron, Lavril, Izacard, Martinet,
  Lachaux, Lacroix, Rozi{\`e}re, Goyal, Hambro, Azhar,
  et~al.]{touvron2023Llama}
Hugo Touvron, Thibaut Lavril, Gautier Izacard, Xavier Martinet, Marie-Anne
  Lachaux, Timoth{\'e}e Lacroix, Baptiste Rozi{\`e}re, Naman Goyal, Eric
  Hambro, Faisal Azhar, et~al.
\newblock Llama: Open and efficient foundation language models.
\newblock \emph{arXiv preprint arXiv:2302.13971}, 2023{\natexlab{a}}.

\bibitem[Touvron et~al.(2023{\natexlab{b}})Touvron, Martin, Stone, Albert,
  Almahairi, Babaei, Bashlykov, Batra, Bhargava, Bhosale,
  et~al.]{touvron2023Llama2}
Hugo Touvron, Louis Martin, Kevin Stone, Peter Albert, Amjad Almahairi, Yasmine
  Babaei, Nikolay Bashlykov, Soumya Batra, Prajjwal Bhargava, Shruti Bhosale,
  et~al.
\newblock Llama 2: Open foundation and fine-tuned chat models.
\newblock \emph{arXiv preprint arXiv:2307.09288}, 2023{\natexlab{b}}.

\bibitem[Tseng et~al.(2024)Tseng, Chee, Sun, Kuleshov, and
  De~Sa]{tseng2024quip}
Albert Tseng, Jerry Chee, Qingyao Sun, Volodymyr Kuleshov, and Christopher
  De~Sa.
\newblock Quip\#: Even better llm quantization with hadamard incoherence and
  lattice codebooks.
\newblock \emph{arXiv preprint arXiv:2402.04396}, 2024.

\bibitem[Wei et~al.(2022)Wei, Zhang, Zhang, Gong, Zhang, Zhang, Yu, and
  Liu]{wei2022outlier}
Xiuying Wei, Yunchen Zhang, Xiangguo Zhang, Ruihao Gong, Shanghang Zhang,
  Qi~Zhang, Fengwei Yu, and Xianglong Liu.
\newblock Outlier suppression: Pushing the limit of low-bit transformer
  language models.
\newblock \emph{Advances in Neural Information Processing Systems},
  35:\penalty0 17402--17414, 2022.

\bibitem[Williams and Aletras(2023)]{williams2023does}
Miles Williams and Nikolaos Aletras.
\newblock How does calibration data affect the post-training pruning and
  quantization of large language models?
\newblock \emph{arXiv preprint arXiv:2311.09755}, 2023.

\bibitem[Xiao et~al.(2023)Xiao, Lin, Seznec, Wu, Demouth, and
  Han]{xiao2023smoothquant}
Guangxuan Xiao, Ji~Lin, Mickael Seznec, Hao Wu, Julien Demouth, and Song Han.
\newblock Smoothquant: Accurate and efficient post-training quantization for
  large language models.
\newblock In \emph{International Conference on Machine Learning}, pages
  38087--38099. PMLR, 2023.

\bibitem[Zhang et~al.(2022)Zhang, Roller, Goyal, Artetxe, Chen, Chen, Dewan,
  Diab, Li, Lin, et~al.]{zhang2022opt}
Susan Zhang, Stephen Roller, Naman Goyal, Mikel Artetxe, Moya Chen, Shuohui
  Chen, Christopher Dewan, Mona Diab, Xian Li, Xi~Victoria Lin, et~al.
\newblock Opt: Open pre-trained transformer language models.
\newblock \emph{arXiv preprint arXiv:2205.01068}, 2022.

\bibitem[Zhu et~al.(2023)Zhu, Li, Liu, Ma, and Wang]{zhu2023survey}
Xunyu Zhu, Jian Li, Yong Liu, Can Ma, and Weiping Wang.
\newblock A survey on model compression for large language models.
\newblock \emph{arXiv preprint arXiv:2308.07633}, 2023.

\end{thebibliography}
